\documentclass[10pt,twoside,dutch,english]{book}                

\usepackage{bbm}
\usepackage{floatrow}
\usepackage[plain]{algorithm}

\usepackage{algpseudocode}
\DeclareMathAlphabet\mathbfcal{OMS}{cmsy}{b}{n}

\usepackage{lscape}
\usepackage[T1]{fontenc}
\usepackage{amsmath}
\usepackage{amssymb}
\usepackage{acronym}
\usepackage[english=usenglishmax]{hyphsubst}
\usepackage[usenames, dvipsnames, table]{xcolor}

\usepackage{mathpazo} 

\usepackage{quotchap}   
\usepackage{fancyhdr}
\usepackage[main=english,dutch,greek]{babel}
\usepackage[paperheight=240mm,paperwidth=160mm, verbose, asymmetric, centering, total={160mm, 240mm}]{geometry}  

\usepackage[hang,up]{caption}     
\usepackage{cite}
\usepackage{ifsym}

\usepackage{verbatim}   
\usepackage{float}      
\usepackage{multirow}   
\usepackage{nccmath}

\ifx\pdftexversion\undefined
\usepackage[dvips]{graphicx}
\else
\usepackage[pdftex]{graphicx}
\fi
\usepackage{url}        
\usepackage{expdlist}   
\usepackage{hhline}     
\usepackage{afterpage}  
\usepackage{marvosym}   
\usepackage{ifthen}     
\usepackage{soul}
\usepackage{xspace}
\usepackage{rotating}
\usepackage{tablefootnote}
\usepackage{array}
\newcolumntype{P}[1]{>{\centering\arraybackslash}p{#1}}

\usepackage{pifont}

\usepackage{tabularx} 
\usepackage{bigstrut}
\usepackage{framed}
\usepackage{tikz-dependency}
\usepackage{tikz-qtree}
\usepackage{longtable} 
\usepackage{hyphenat}
\usepackage{epstopdf}
\usepackage{hyperref} 
\usepackage{tikz,pgfplots}
\usepackage{lineno,hyperref}
\usepackage[inline,shortlabels]{enumitem}
\usepackage{amsthm}
\theoremstyle{plain}
\theoremstyle{definition}
\usepackage{makecell}
\usepackage{kbordermatrix}%
 
\usepackage{chngcntr} 
 
\usepackage{listings}
\usepackage[T1]{fontenc} 

\newcolumntype{L}{>{\raggedright\arraybackslash}X} 
\newcolumntype{R}{>{\raggedleft\arraybackslash}X} 
\newcolumntype{C}{>{\centering\arraybackslash}X} 

\definecolor{light-gray}{gray}{0.8}

\usepackage{pdfpages}
\usepackage{nicefrac}
\usepackage{booktabs}
\usepackage[warn]{textcomp}
\usepackage{textgreek}
\usepackage{eurosym}
\usepackage{mathtools}
\usepackage{adjustbox}
\usepackage{setspace}
\usepackage[normalem]{ulem}
\usepackage{cellspace}
\usepackage{tabu}
\usepackage{alltt}
\usepackage{tabularx,booktabs}
\usepackage{siunitx}

\lstdefinestyle{base}{
  moredelim=**[is][\color{red}]{@}{@} , 
  moredelim=**[is][\color{OliveGreen}]{!}{!},
  moredelim=**[is][\color{blue}]{?}{?},
  moredelim=[is][\bfseries]{[*}{*]},
  moredelim=**[is][\color{purple}]{)}{)},
  moredelim=**[is][\color{orange}]{(}{(}
}

\usepackage{cleveref}
\usepackage{stmaryrd}
\usepackage{diagbox} 

\newcommand{\equref}[1]{Eq.~(\ref{#1})}

\DeclarePairedDelimiterX{\probarg}[1]{(}{)}{%
  \ifnum\currentgrouptype=16 \else\begingroup\fi
  \activatebar#1
  \ifnum\currentgrouptype=16 \else\endgroup\fi
}
\newcommand{\innermid}{\nonscript\;\delimsize\vert\nonscript\;}
\newcommand{\activatebar}{%
  \begingroup\lccode`\~=`\|
  \lowercase{\endgroup\let~}\innermid 
  \mathcode`|=\string"8000
}
\newcommand{\cmark}{\ding{51}}%
\newcommand{\xmark}{\ding{55}}%

\usepackage{color}

\newcommand{\oldtext}[1]{}

\newcommand{\Fone}{{F$_{1}$}}

\newcommand{\revklim}[1]{#1}

\usepackage[textsize=small]{todonotes}

\newcommand{\oldphdtext}[1]{}

\newcommand{\datasetname}{DWIE}
\newcommand{\ouraida}{AIDA$^{+}$}
\newcommand{\propformat}[1]{\texttt{#1}}
\newcommand{\ourdataset}{TempEL}

\usepackage{subcaption}

\newcommand{\appendixB}{\renewcommand{\thechapter}{B}\chapter}

\makeatletter
     \renewcommand*\l@figure{\@dottedtocline{1}{1em}{3.2em}}
\makeatother
\makeatletter
     \renewcommand*\l@table{\@dottedtocline{1}{1em}{3.2em}}
\makeatother

\usepackage{hyperref}

\newcounter{Hsection}

\newcounter{Hsubsection}

\makeatletter
\def\Url@twoslashes{\mathchar`\/\@ifnextchar/{\kern-.2em}{}}
\g@addto@macro\UrlSpecials{\do\/{\Url@twoslashes}}
\makeatother

\newcommand{\etal}{et al.\ }  

\newcommand{\eg}{e.g., }
\newcommand{\ie}{i.e., }
\newcommand{\figref}[1]{Fig.~\ref{#1}}    
\newcommand{\Figref}[1]{Figure~\ref{#1}}  
\newcommand{\tabref}[1]{Table~\ref{#1}}
\newcommand{\Tabref}[1]{Table~\ref{#1}}
\newcommand{\secref}[1]{Section~\ref{#1}}
\newcommand{\chapref}[1]{\iflanguage{dutch}{Hoofdstuk}{Chapter}~\ref{#1}}
\newcommand{\appendixref}[1]{\iflanguage{dutch}{Bijlage}{Appendix}~\ref{#1}}
 
\newcommand\LSTM{\mbox{LSTM}} 
\newcommand\LSTMs{{\LSTM}s} 
\newcommand\biLSTM{Bi\LSTM}
\newcommand\biLSTMs{Bi{\LSTM}s} 
\newcommand\TreeLSTM{Tree-\LSTM}
\newcommand\TreeLSTMs{Tree-{\LSTM}s} 
\newcommand\BLSTM{B-\LSTM}
\newcommand\TLSTM{T-\LSTM}
\newcommand\NTLSTM{NT-\LSTM}

\newcommand\MaxILP{ILP Coverage}
\newcommand\TopILP{ILP Naive}

\definecolor{wordcolor}{HTML}{36a9e0}
\definecolor{embeddingcolor}{HTML}{f9b233}
\definecolor{biLSTMcolor}{HTML}{e84e1b}
\definecolor{attentioncolor}{HTML}{a2195b}
\definecolor{hiddencolor}{HTML}{008d36}

\definecolor{darkspringgreen}{rgb}{0.09, 0.45, 0.27}
\colorlet{positive0001}{darkspringgreen!75}
\colorlet{positive001}{darkspringgreen!50}
\colorlet{positive01}{darkspringgreen!25}

\definecolor{deepcarmine}{rgb}{0.66, 0.13, 0.24}

\colorlet{negative0001}{deepcarmine!75}
\colorlet{negative001}{deepcarmine!50}
\colorlet{negative01}{deepcarmine!25}

\usepackage{tikz}
\usepackage{pgfplots}
\pgfplotsset{compat=1.3}
\usepackage{filecontents}
\usetikzlibrary{calc}
\usetikzlibrary{arrows}
\usetikzlibrary{patterns}
\usetikzlibrary{plotmarks}
\usetikzlibrary{intersections} 
\usetikzlibrary{decorations.pathmorphing}
\usetikzlibrary{decorations.pathreplacing}
\usetikzlibrary{decorations.markings,arrows} 
\usetikzlibrary{decorations.shapes}
\usetikzlibrary{fit}
\usetikzlibrary{backgrounds}
\usetikzlibrary{trees}
\usetikzlibrary{matrix}
\usetikzlibrary{arrows.meta}

\newcommand{\Size}{1.7cm}

\tikzset{Square/.style={
    inner sep=0pt, 
    minimum size=\Size,
    }
}

\newcommand{\equsrefrange}[2]{\eqs~\ref{#1}--\ref{#2}}

\fancypagestyle{plain}{
\fancyhf{}

}

\renewcommand{\chaptermark}[1]{\markboth{#1}{}}

\newcommand\oddpageleftmark{}
\newcommand\evenpagerightmark{}

\def\PARstart#1#2{\begingroup\def\par{\endgraf\endgroup\lineskiplimit=0pt}
    \setbox2=\hbox{\uppercase{#2} }\newdimen\tmpht \tmpht \ht2
    \advance\tmpht by \baselineskip\font\hhuge=cmr10 at \tmpht
    \setbox1=\hbox{{\hhuge #1}}
    \count7=\tmpht \count8=\ht1\divide\count8 by 1000 \divide\count7 by\count8
    \tmpht=.001\tmpht\multiply\tmpht by \count7\font\hhuge=cmr10 at \tmpht
    \setbox1=\hbox{{\hhuge #1}} \noindent \hangindent1.05\wd1
    \hangafter=-2 {\hskip-\hangindent \lower1\ht1\hbox{\raise1.0\ht2\copy1}%
    \kern-0\wd1}\copy2\lineskiplimit=-1000pt}

\AtBeginDocument{%

   \renewcommand{\bibname}{References}%
}

\newcommand\PhDauthor{Klim Zaporojets}

\newcommand\PhDEnglishTitle{Neural Approaches to Sequence Labeling for Information Extraction}

\newboolean{PhDpromotorMultiple}
\setboolean{PhDpromotorMultiple}{true} 

\newboolean{IWTscholarship}
\setboolean{IWTscholarship}{false} 
\newboolean{FWOscholarship}
\setboolean{FWOscholarship}{false} 

\hypersetup{											
 		pdftitle=\PhDEnglishTitle,    
    pdfauthor=\PhDauthor,     
    pdfsubject={PhD dissertation, Ghent University, Belgium},   
    colorlinks,%
    citecolor=black,%
    filecolor=black,%
    linkcolor=black,%
    urlcolor=black
}

\restylefloat{figure}
\restylefloat{table}
\begin{document}
\graphicspath{{fig/}}
\floatstyle{plaintop}
\floatstyle{ruled}
\floatsetup[figure]{capposition=bottom}
\floatsetup[table]{capposition=bottom}


\includepdf[pages=-]{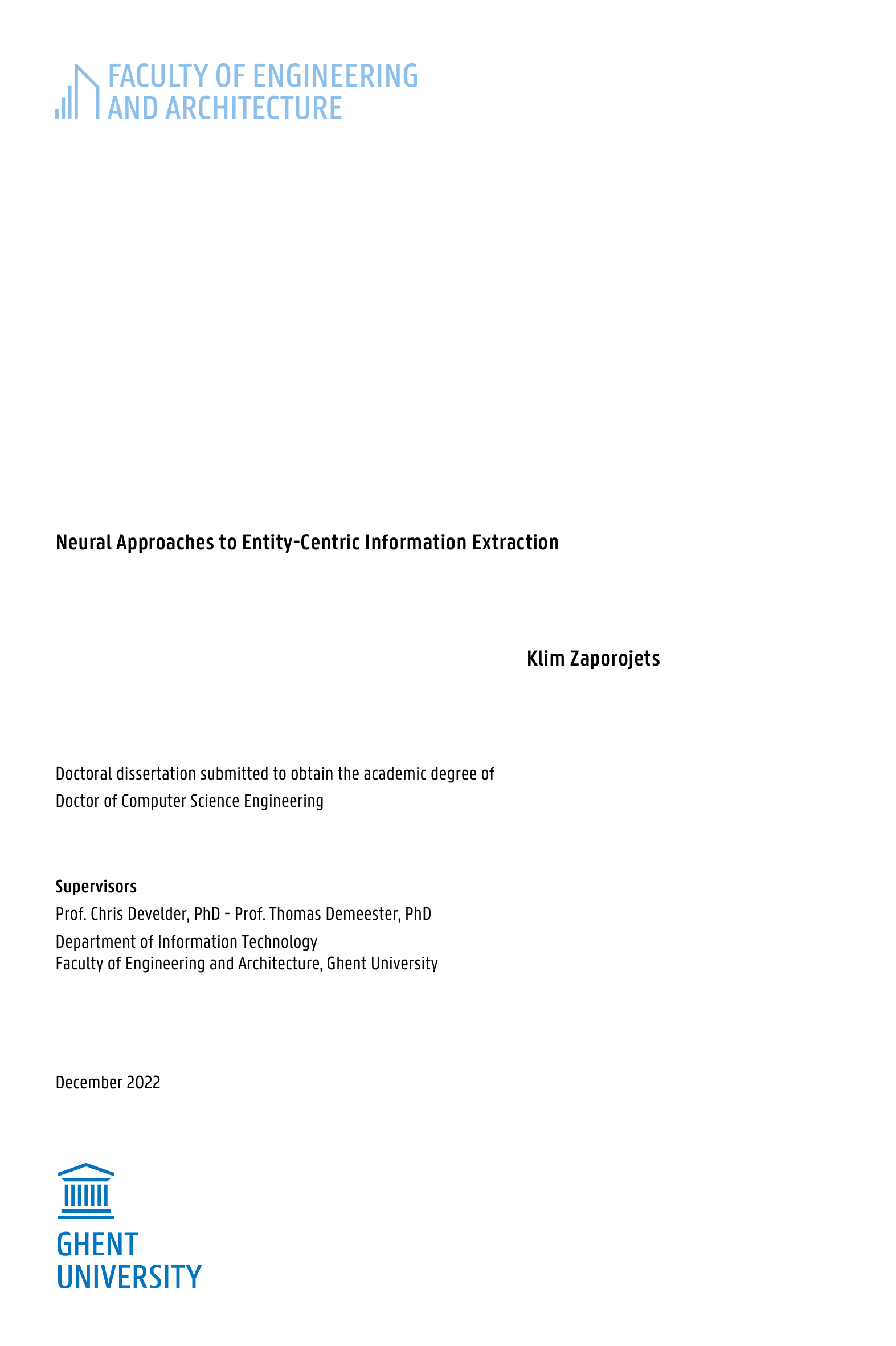}
\clearpage{\pagestyle{empty}\cleardoublepage}
\thispagestyle{empty}




 

\clearpage{\pagestyle{empty}\cleardoublepage}


\selectlanguage{dutch}
\frontmatter
\chapter{Acknowledgements}
First and foremost, I am extremely grateful to my supervisors Prof. Chris Develder and Prof. Thomas Demeester for giving me the great opportunity to pursue my PhD studies and for their continued support and encouragement. For their patience, assertive advice, and for being on my side and not giving up on me even in the hardest moments throughout my PhD journey, specially during the initial years when I was struggling with rejections and getting the first work published. 
Finally, I also want to thank my advisors and the IDLab and UGent administration in general for allowing me to pursue my doctoral studies despite working remotely and spending much of the time with my family in Denmark. 

I would also like to express special gratitude to ir. Johannes Deleu for all the hours and hours of brainstorming and discussion sessions together. I am sure that without his guidance, this thesis would not have been possible. In fact, it would not be an overstatement to acknowledge him as the initial generator of the main ideas and conceptualizations this thesis is based upon. I also would like to thank him for sharing the initial code of his models, so I could experiment and extend it for each of the different projects I was involved in. 

My special gratitude goes to the members of my PhD thesis
committee, Prof. Filip De Turck, Prof. Veronique Hoste, Prof. Isabelle Augenstein, Prof. Yvan Saeys, Prof. Tom Dhaene, dr. Pasquale Minervini for dedicating time
and effort to read my thesis and to provide constructive feedback.

I would like to express my deepest appreciation to all of the current and former members of our group. I feel extremely lucky and blessed to have been part of such an amazing team: Dr. Lucas Sterckx, Dr. Giannis Bekoulis, Dr. Nasrin Sadeghianpourhamami, Dr. Matthias Strobbe, Dr. Fr\'{e}deric Godin, Dr. Cedric De Boom, Semere Kiros Bitew, Amir Hadifar, Manu Lahariya, Maarten De Raedt, François Remy, Sofie Labat, Yiwei Jiang, Ruben Janssens, Jens-Joris Decorte, Karel D'Oosterlinck, Paloma Rabaey and C\'{e}dric Goemaere.

I am also extremely grateful to all the collaborators I worked with at Ghent University as well as at other institutions. I would like to particularly acknowledge all the members of the CopeNLU team at University of Copenhagen, where I interned during the Spring 2022. Specially I would like to express enormous gratitude to Prof. Isabelle Augenstein and dr. Lucie-Aim\'{e}e Kaffee, whose guidance was critical to get our temporal entity linking paper accepted to NeurIPS 2022 conference. 

I would also like to recognize a huge effort of all the master students I worked with. Specially, I am deeply indebted to MSc Severine Verlinden, whose hard work and dedication has resulted in the publication at the top ACL conference of the work on injecting external knowledge to text, presented in one of the chapters of this thesis. I also want to express a huge gratitude to MSc Vincent Schelstraete and MSc Nick Peeters.

Finally, I can not end this section without getting too personal and acknowledging my family for all the love and support during my PhD! I could not have undertaken this journey without the support of my wife and kids. Despite living in another country and having to spend many days, weeks or even months without me, they unconditionally encouraged me to pursue my studies. I would also like to thank from the bottom of my heart to my parents-in-law Chris and Ria, who always made me feel welcome in their house, and are now also giving refuge to my aunt and cousin who escaped the war in Ukraine. Finally, I am extremely grateful to my parents and my family in Argentina who, despite living in another continent, have supported and motivated me throughout all of my PhD.

\begin{flushright}{\emph{Ghent, Fall 2022\\
Klim}}
\end{flushright}

\selectlanguage{english}

\clearpage{\pagestyle{empty}\cleardoublepage}


\renewcommand{\contentsname}{Table of Contents} 
\pdfbookmark{Table of Contents}{bookmark:toc}

\tableofcontents
\clearpage{\pagestyle{empty}\cleardoublepage}

\pdfbookmark{List of Figures}{bookmark:lof}
\listoffigures
\clearpage{\pagestyle{empty}\cleardoublepage}

\pdfbookmark{List of Tables}{bookmark:lot}
\listoftables
\clearpage{\pagestyle{empty}\cleardoublepage}

\pdfbookmark{List of Acronyms}{bookmark:loa}

\clearpage      
\thispagestyle{empty}   
\mbox{}         
\clearpage{\pagestyle{empty}\cleardoublepage}   

\renewcommand{\bibname}{References}     

\selectlanguage{dutch}  
\renewcommand\evenpagerightmark{{\scshape\small Samenvatting}}
\renewcommand\oddpageleftmark{{\scshape\small Summary in Dutch}}

\hyphenation{tech-no-lo-gische toe-pas-sing het-geen be-paalt uit-voe-ren in-he-rent be-las-ting ont-wik-ke-ling mo-ge-lijk-heid schie-ten ver-bin-ding toe-pas-sing-en in-fra-struc-tuur con-fi-gu-re-ren be-per-ken-de com-pu-ter-vi-sie-tech-nie-ken twee-de op-los-sing}

\chapter[Samenvatting]{Samenvatting\\-- Summary in Dutch --}

Artifici\"{e}le intelligentie (AI) heeft een enorme impact op ons dagelijks leven met toepassingen zoals stemassistenten, gezichtsherkenning, chatbots, autonoom rijdende auto's, enz. Natural Language Processing (NLP) is een disciplineoverschrijdende AI en linguïstiek die zich toelegt op het bestuderen van het begrip van de tekst. Dit is een zeer uitdagend gebied vanwege de ongestructureerde aard van de taal met veel dubbelzinnige en hoekgevallen. Zinnen met meerdere betekenissen, zoals 'De geit is klaar om te eten', zijn bijvoorbeeld extreem moeilijk te interpreteren voor computers (en zelfs mensen) zonder aanvullende contextuele kennis. Toch is er de afgelopen jaren een snelle vooruitgang geboekt op het gebied van NLP met zeer nuttige toepassingen zoals automatische tekstvertaling, gespreksagenten, nepnieuws en detectie van haatspraak in onder andere sociale media.

In dit proefschrift behandelen we een zeer specifiek gebied van NLP dat het begrip van \textit{entities} in tekst aanpakt. Het concept van \textit{entity} is erg dubbelzinnig en kan voor verschillende interpretaties vatbaar zijn, afhankelijk van een specifieke toepassing en studiegebied. Het meest klassieke gebruik in Natural Language Processing is om te verwijzen naar \textit{named entity's} om echte of fictieve objecten aan te duiden die worden weergegeven met eigennamen. Typische voorbeelden zijn organisaties (\eg ``Google'', ``Gent University''), mensen (\eg ``Lionel Messi'', ``Joe Biden''), karakters (\eg ``Batman'', ``Superman''), locaties (\eg ``Gent'', ``Denemarken''), oa. Deze entiteitsaanduidingen in tekst worden gebruikt om te worden verbonden met een rijkere Knowledge Base (KB) zoals Wikipedia. Het gebruik van deze KB's is nuttig om aanvullende informatie te verkrijgen en de applicaties te voorzien van extra kennis die nodig is om de tekst te begrijpen.
De lezer kan een meer gedetailleerde inleiding op het onderwerp van dit proefschrift vinden in ~\chapref{chap:introduction}. Daar geven we een overzicht van de literatuur en introduceren we alle noodzakelijke concepten om het gepresenteerde werk beter te begrijpen.

We beginnen dit proefschrift~(\chapref{chap:dwie}) met een radicaal andere, \textit{entity-centric} kijk op de informatie in tekst. We stellen dat, in plaats van individuele vermeldingen in tekst te gebruiken om hun betekenis te begrijpen, we applicaties moeten bouwen die zouden werken in termen van entiteitsconcepten. Deze entiteitgestuurde benadering houdt in dat alle vermeldingen die naar dezelfde entiteit verwijzen (\eg ``Gent'') in coreferentiecluster worden gegroepeerd en de rest van de taken (\eg. relatieextractie, entiteitskoppeling, etc.) op clusterniveau worden uitgevoerd. Deze benadering heeft het voordeel dat de informatie van alle entiteitsvermeldingen die naar één enkele entiteit in het document verwijzen, wordt benut. Als gevolg hiervan vereist de entiteitsgerichte benadering een weergave op documentniveau van de tekst. Helaas heeft de NLP-gemeenschap geen evaluatie- en trainingsbronnen (dwz datasets) geproduceerd die deze focus op documentniveau zouden hebben voor meerdere taken tegelijk. We pakken deze onderzoekskloof aan door een DWIE-dataset (Deustche Welle Information Extraction) te introduceren waarin we vier verschillende taken op entiteitsniveau annoteren: coreferentieresolutie, entiteitskoppeling, relatie-extractie en benoemde entiteitherkenning. We laten verder zien hoe deze taken elkaar aanvullen in een gezamenlijk informatie-extractiemodel.

In het volgende hoofdstuk van dit proefschrift (\chapref{chap:coreflinker}), presenteren we een meer gedetailleerd model over hoe de entiteitsgerichte benadering kan worden gebruikt voor de taak \textit{entity linking}. De \textit{entity linking} bestaat uit het toewijzen van het anker \textit{mentions} in tekst aan doel \textit{entities} die beschrijven
ze in een Knowledge Base (KB) (\eg Wikipedia). In ons werk laten we zien dat deze taak kan worden verbeterd door te overwegen entiteitskoppeling uit te voeren op het coreferentieclusterniveau in plaats van op elk van de vermeldingen afzonderlijk. Door deze aanpak te volgen, is ons gezamenlijke model in staat om de informatie van alle kernvermeldingen tegelijk te gebruiken bij het kiezen van de kandidaat-entiteit. Als gevolg hiervan leidt dit tot consistentere voorspellingen tussen vermeldingen die naar hetzelfde concept verwijzen, met name een verbetering van de prestaties op hoekgevallen die bestaan uit impopulaire vermeldingen.

Ons volgende idee wordt beschreven in~\chapref{chap:injecting_knowledge} van dit proefschrift. Daar hanteren we een iets andere benadering met entiteiten: in plaats van puur tekstuele informatie te gebruiken om informatie-extractietaken op te lossen, zoals relatie-extractie, bestuderen we ook hoe de informatie van entiteiten uit Knowledge Base kan worden geïntegreerd. We bereiken een aanzienlijke verbetering van alle geëvalueerde taken door informatie te injecteren van zowel Wikipedia als Wikidata KB's. Bovendien, terwijl de taken die we aanpakken zijn geannoteerd en gedefinieerd op \textit{named entity}-niveau, is de informatie die we in onze tekst injecteren afkomstig van alle bestaande entiteiten die zijn gedefinieerd in de geteste KB's. We vinden dat deze techniek zonder toezicht nog steeds de entiteiten kan detecteren die relevanter zijn voor een bepaalde tekst.

Ten slotte wordt de laatste entiteitgerelateerde bijdrage van dit proefschrift beschreven in \chapref{chap:temporal_el}. Daar gaan we nog een stap verder en analyseren we de evolutie van de entiteiten vanuit een tijdsperspectief. Om dit te bereiken, creëren we een nieuwe dataset die bestaat uit 10 jaarlijkse snapshots van Wikipedia-entiteiten van 2013 tot 2022. We bestuderen verder hoe de taak \textit{entity linking} wordt beïnvloed door \begin{enumerate*}[(i)]
    \item wijzigingen van bestaande entiteiten in de tijd, en
    \item creatie van nieuwe opkomende entiteiten.
\end{enumerate*}. Verder beperken we onze analyse niet tot het domein van \textit{named bodies}, maar nemen we alle bestaande entiteiten en concepten op die in Wikipedia zijn gedefinieerd. Onze analyse toont een voortdurende afname van de prestaties in de tijd, wat aangeeft dat de entiteiten uit latere versies van Wikipedia moeilijker te ondubbelzinnig zijn dan entiteiten uit eerdere versies. Bovendien laten we zien dat de prestatiedaling vooral scherp is voor entiteiten die aanvullende nieuwe kennis nodig hebben (\eg nieuwe entiteiten met betrekking tot de COVID-19-pandemie) waarvoor het model niet vooraf is getraind.

Daarnaast omvat dit proefschrift ander onderzoekswerk dat is gepubliceerd in vooraanstaande tijdschriften en conferenties die geen verband houden met het centrale onderwerp van dit proefschrift. Daarom stellen we in appendix \chapref{chap:mwp} voor om terugkerende neurale netwerken te gebruiken om de structuur op vergelijkingsbomen na te bootsen om wiskundige wereldproblemen op te lossen. Met onze aanpak laten we een aanzienlijke verbetering zien. Verder beschrijven we in appendix \chapref{chap:clpsych} onze bijdrage aan de gedeelde taak van CLPsych 2018, waarbij we competitieve resultaten behalen met behulp van een ensemble bestaande uit meerdere modellen om depressie en angst te voorspellen in tekstuele enquêtes.
\clearpage{\pagestyle{empty}\cleardoublepage}   
\selectlanguage{english}
\renewcommand\evenpagerightmark{{\scshape\small Summary}}
\renewcommand\oddpageleftmark{{\scshape\small Summary}}

\hyphenation{}

\chapter[Summary]{Summary}

Artificial Intelligence (AI) has huge impact on our daily lives with applications such as voice assistants, facial recognition, chatbots, autonomously driving cars, etc. Natural Language Processing (NLP) is a cross-discipline of AI and Linguistics, dedicated to study the understanding of the text. This is a very challenging area due to unstructured nature of the language, with many ambiguous and corner cases. For example, sentences with multiple meanings such as ``The goat is ready to eat.'' are extremely hard to interpret for computers (and even for humans) without additional contextual knowledge. Yet, in recent years we have witnessed a rapid progress in the field of NLP with highly useful applications such as automatic text translation, conversational agents, fake news and hate speech detection in social media, among others. 

In this thesis, we address a very specific area of NLP which tackles the understanding of \textit{entities} in text. The concept of \textit{entity} is very ambiguous and can be subject to different interpretations depending on a specific application and area of study. The most classical use in Natural Language Processing is to refer to \textit{named entities} denoting real-world or fictitious objects that are represented with proper names. Typical examples include organizations (\eg ``Google'', ``Ghent University''), people (\eg ``Lionel Messi'', ``Joe Biden''), fictional characters (\eg ``Batman'', ``Superman''), locations (\eg ``Ghent'', ``Denmark''), among others. These entity denotations in text are used to be connected to a richer Knowledge Bases (KB) such as Wikipedia. The use of these KBs is beneficial to get additional information and provide the applications with extra knowledge needed to understand the text.
The reader can find a more detailed introduction to the topic of this thesis in~\chapref{chap:introduction}. There, we give an overview of the literature and introduce all the necessary concepts to better understand the presented work. 

We start this thesis in \chapref{chap:dwie} with proposing a radically different, \textit{entity-centric} view on the information in text. We argue that, instead of using individual mentions in text to understand their meaning, we need to build applications that would operate in terms of entity concepts. This entity-centric approach involves grouping all the mentions referring to the same entity (\eg ``Ghent'') in coreference cluster and perform the rest of the tasks (\eg relation extraction, entity linking, etc.) on a cluster level. This approach has the advantage of leveraging the information across all the entity mentions referring to a single entity in the document at once. As a consequence, the entity-centric approach requires a document-level view on the text. Unfortunately, the NLP community has produced no evaluation nor training resources (\ie datasets) that would have this document-level focus for multiple information extraction tasks. We tackle this research gap by introducing DWIE (Deustche Welle Information Extraction) dataset in which we annotate four different tasks on entity level: coreference resolution, entity linking, relation extraction, and named entity recognition. We further demonstrate 
the interdependence of these tasks
in a joint information extraction model. 

In the following \chapref{chap:coreflinker}, 
we present a more detailed model on how the entity-centric approach can be used for \textit{entity linking} task. The \textit{entity linking} consists in mapping the anchor \textit{mentions} in text to target \textit{entities} that describe
them in a Knowledge Base (KB) (\eg Wikipedia). In our work, we showcase that this task can be improved by considering performing entity linking on the coreference cluster level instead of on each of the mentions individually. By adopting this approach, our joint model is able to use the information of all the coreferent mentions at once when choosing the candidate entity. As a result, this leads to more consistent predictions among mentions referring to the same concept, especially boosting the performance on corner cases consisting of unpopular mentions. 

Our next idea is described in~\chapref{chap:injecting_knowledge} of this thesis. There, we adopt a slightly different approach involving entities: instead of using purely textual information to solve information extraction tasks such as relation extraction, we also study how the information of entities from a Knowledge Base can be integrated. We achieve significant improvement on all of the evaluated tasks by injecting information both from Wikipedia, as well as from Wikidata KBs. Furthermore, while the tasks we are tackling are annotated and defined on \textit{named entity} level, the information we inject in our text comes from all the existing entities defined in the experimented KBs. We find that this unsupervised technique is still able to detect the entities that are more relevant for a particular text. 

Finally, the last entity-related contribution of this thesis is described in \chapref{chap:temporal_el}. There, we go one step further and analyze the evolution of the entities from temporal perspective. In order to achieve this, we create a new dataset which consists of 10 yearly snapshots of Wikipedia entities from 2013 until 2022. We further study how \textit{entity linking} task is affected by \begin{enumerate*}[(i)]
    \item changes of existing entities in time, and 
    \item creation of new emerging entities
\end{enumerate*}. Furthermore, we do not restrict our analysis to the realm of \textit{named entities}, but incorporate all existing entities and concepts defined in Wikipedia. Our analysis showcases a continual decrease of performance over time, indicating that the entities from later versions of Wikipedia are harder to disambiguate than entities from earlier versions. Additionally, we demonstrate that this decrease of performance 
is exacerbated on
entities requiring additional new knowledge (\eg new entities related to COVID-19 pandemic) for which the model was not pre-trained. 

Additionally, this thesis includes other research work published in top-tier journals and conferences not related to the central topic of this thesis. Thus, in \appendixref{chap:mwp} we propose to use recursive neural networks to mimic the structure of equation trees to solve mathematical word problems. We showcase a significant improvement using our approach. Furthermore, in \appendixref{chap:clpsych} we describe our contribution to the CLPsych 2018 shared task where we achieve competitive results using an ensemble consisting of multiple models to predict depression and anxiety in textual surveys. 

\clearpage{\pagestyle{empty}\cleardoublepage}   

\mainmatter     
\selectlanguage{english}
\renewcommand*{\thesection}{\thechapter.\arabic{section}}

\newcommand\fdtsvrightmarktmp{{\scshape\small Chapter }}
\renewcommand\evenpagerightmark{{\scshape\small\chaptername\ \thechapter}}
\renewcommand\oddpageleftmark{{\scshape\small\leftmark}}

\baselineskip 13.0pt

\graphicspath{{klim_ch_intro/figures/}}

\hyphenation{}

\chapter[Introduction]{Introduction}
\label{chap:introduction}

\renewcommand\evenpagerightmark{{\scshape\small Chapter \arabic{chapter}}}
\renewcommand\oddpageleftmark{{\scshape\small Introduction}}

\renewcommand{\bibname}{References}
\newcommand{\revklimintro}[1]{{#1}}
\newcommand{\revklimtohide}[1]{\textcolor{RoyalBlue}{}}


\begin{flushright}
\end{flushright}

Natural Language Processing (NLP) has recently gained a lot of attention in society. Formally, NLP is a subfield of Linguistics and Artificial Intelligence (AI) concerned with automatic processing of textual data by computers. It spans a wide range of research areas with high societal impact. For example, research in \textit{information extraction} (IE) \cite{martinez2020information,wang2018clinical,zaporojets2021dwie,corcoglioniti2016knowledge} 
\revklimintro{on extracting the most relevant information from textual documents,}
supports the construction of
robust search engines such as Google that allow millions of people to find useful information on internet. Research in \textit{text classification} allows to automatically divide incoming e-mail in different categories, identify fraudulent profiles in e-commerce sites, detect hate speech in social platforms such as Facebook, etc. The latest advances in \textit{conversational agents} \cite{hussain2019survey}, allow to assist people in all variety of daily tasks such as getting medical assistance \cite{laranjo2018conversational}, online shopping \cite{bavaresco2020conversational} and cooking \cite{jiang2020recipe,jiang2022cookdial}.  
Development in \textit{psycholinguistics} enables models to accurately predict the chances a person may suffer from depression \cite{zaporojets2018predicting,trifan2020understanding}, anxiety \cite{jacobson2021deep} \revklimintro{or even the inclination to commit suicide \cite{tadesse2019detection,bitew2019predicting}}.
Recent developments in \textit{fake news} \cite{thorne2018fever,aly2021feverous} and \textit{stance} \cite{augenstein2016stance,riedel2017simple} detection 
facilitate
users to find trustworthy and unbiased information online. This list of NLP areas with applications is far from exhaustive, but should give the reader a good idea of the breadth and 
high impact of the research in NLP. 

In this thesis, we focus on \textit{information extraction} (IE), the sub-field of NLP that studies the extraction of structured information from unstructured text. 
Concretely, we study how the information about the \textit{entities} described in text can be extracted and used to solve a number of IE tasks. 
One can think of entities as concepts that can represent any physical or abstract object. The description of these entities is usually collected in encyclopedias such as Wikipedia,\footnote{\url{https://www.wikipedia.org/}} Fandom,\footnote{\url{https://www.fandom.com/}} DBPedia,\footnote{\url{https://www.dbpedia.org/}} etc. 
Our works in \cite{zaporojets2021dwie} (\chapref{chap:dwie}) and \cite{zaporojets2021consistent} (\chapref{chap:coreflinker}) deal with a specific type of entities denominated \textit{named entities} that include names of people (\eg ``Joe Biden'', ``Lionel Messi'', ``Galileo Galilei''), places (\eg ``Belgium''), organizations (\eg ``Google'', ``Microsoft'', ``Ghent University''), etc. As a rule of thumb, named entities are usually written with the first letter in uppercase (\ie are proper nouns). In the first part of our work we explore the importance of thinking in terms of entities that transcends the written entity \textit{mentions} (\ie words that refer to a specific entity) in text (Chapters \ref{chap:dwie} and \ref{chap:coreflinker}). Furthermore, in \cite{verlinden2021injecting} (\chapref{chap:injecting_knowledge}), we focus on exploring how additional entity information can enrich other IE tasks such as coreference resolution (\secref{ch_intro:sec:coreference_resolution}), named entity recognition (\secref{ch_intro:sec:named_entity_recognition}), and relation extraction (\secref{ch_intro:sec:relation_extraction}).
Finally, in \cite{zaporojets2022tempel} (\chapref{chap:temporal_el}) we propose a fundamentally different, \textit{evolutionary} view on the entity linking (\secref{ch_intro:sec:entity_linking}) task. There, we introduce a new TempEL dataset which consists of entity linking annotations grouped in 10 yearly snapshots. Our experimental results showcase a continual temporal decrease in performance of the EL task: the biggest drop is observed for new entities that require additional world knowledge, non-existing during the pre-training phase of the models. 
 
In this introductory chapter we describe \begin{enumerate*} [(i)]
    \item the various information extraction tasks this thesis is about (\secref{ch_intro:sec:ie_tasks}),
    \item the learning approaches used to train the models that are relevant to this thesis (Sections \ref{ch_intro:sec:ml_approaches}--\ref{ch_intro:sec:learning_in_nlp_tasks}), and 
    \item a highlight of the main contributions of our work (\secref{ch_intro:sec:research_contributions}) with the list of produced publications (\secref{ch_intro:sec:publications}).
\end{enumerate*}



\section{Information extraction tasks}
\label{ch_intro:sec:ie_tasks}
In this thesis, we focus on \textit{information extraction} (IE), which includes tasks used to extract structured information from unstructured text. This structured information is a result of solving multiple diverse tasks on a given piece of text \cite{jurafsky2000speech,martinez2020information,niklaus2018survey,nasar2018information,grishman2015information,sarawagi2008information}, many of which go beyond the scope of the current work. 
%
In this thesis, we focus on the tasks that allow to identify entities described in text as well as the semantic relations between these entities (see entity-centric approach in \secref{ch_intro:sec:entity_linking} for details). The first task consists in identifying and classifying the named mentions such as ``Meghan Markle`` in the example of our DWIE dataset in \figref{ch_intro:fig:entity_centric}. This task is called \textit{named entity recognition} (NER) \cite{nadeau2007survey,yadav2018survey,li2020survey}, and is further described in \secref{ch_intro:sec:named_entity_recognition}.  Once the mentions have been identified, we proceed to group them in clusters, each one referring to one specific entity. For example, the mentions ``Meghan'' and ``Meghan Markle'' in \figref{ch_intro:fig:entity_centric} are clustered together since they both refer to the same person. This task is known as \textit{coreference resolution} \cite{sukthanker2020anaphora,stylianou2021neural}, and is further described in \secref{ch_intro:sec:coreference_resolution}. We further identify semantic relations between the clusters, such as the relation \textit{spouse\_of} between the clusters representing \textit{Meghan} and \textit{Harry}. This task is known as \textit{relation extraction} \cite{pawar2017relation,kumar2017survey}, described in \secref{ch_intro:sec:relation_extraction}. Finally, we connect the mention clusters to the respective entities in encyclopedias (formally known as Knowledge Bases or KB) such as Wikipedia. This task is known as \textit{entity linking} and is further detailed in \secref{ch_intro:sec:entity_linking}.
The remainder of this subsection describes
the main characteristics as well as the main datasets used to evaluate the performance of the models for each of these tasks. 
\subsection{Named entity recognition}
\label{ch_intro:sec:named_entity_recognition}
\begin{figure}[t]
\centering
\includegraphics[width=1.0\columnwidth]{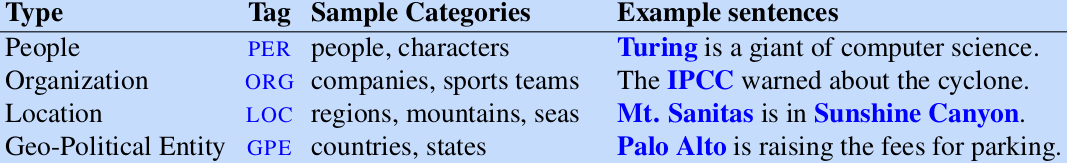}
\captionsetup{singlelinecheck=off}
\caption[Example of named entities]{Example of named entities. Source: \cite{jurafsky2018speech}.}  
\label{ch_intro:fig:ner_intro}
\end{figure}

The task of named entity recognition (NER) consists in 
finding
and classifying named entity mentions in text. A \textit{named entity} mention is a proper name referring to real world objects such as countries, organizations, universities, etc. Named entities tend to be written with the first letter upper-cased (see \figref{ch_intro:fig:ner_intro}). This definition is commonly extended to include mentions denoting dates, times and numerical expressions (\eg prices). The most 
widely
used dataset to evaluate NER is CoNLL-2003 \cite{sang2003introduction}, and consists of 35,089 annotated named entity mentions splitted in four entity types: person (PER), location (LOC), organization (ORG), and miscellaneous (MISC). Other 
extensively
used NER datasets are WNUT 2017 \cite{derczynski2017results}, Ontonotes v5 \cite{weischedel2013ontonotes} and Few-NERD \cite{ding2021few}, to mention a few. 
\subsection{Coreference resolution}
\label{ch_intro:sec:coreference_resolution}
\begin{figure}[t]
\centering
\includegraphics[width=0.7\columnwidth, trim={1.0cm 16.5cm 10.5cm 1.0cm},clip]{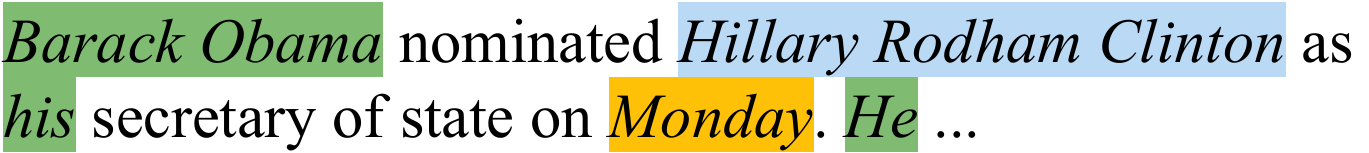}
\captionsetup{singlelinecheck=off}
\caption[Example of coreference resolution]{Example of coreference resolution task composed of three coreference mention clusters, depicted with different colors. Thus, the mentions ``he'' and ``He'' are coreferent with ``Barack Obama''. The rest of the mentions (``Hillary Rodham Clinton'' and ``Monday'') form singleton clusters, each composed of a single mention. Adapted from \cite{rahman2009supervised}.}
\label{ch_intro:fig:coref_intro}
\end{figure}
The coreference resolution task consists in detecting references to the same entity in a text. These references are \textit{mentions} as illustrated in the example of \figref{ch_intro:fig:coref_intro}, where the mention ``He'' is coreferent with ``Barack Obama''. The main dataset to measure the coreference resolution performance is CoNLL-2012 \cite{pradhan2012conll} which is part of the OntoNotes corpus \cite{weischedel2013ontonotes}. It consists of a set of articles coming from newswire, magazines, broadcast news and conversations, web data, and conversational speech domains. 
Other datasets in coreference resolution include Task-1 of SemEval 2010 \cite{recasens2010semeval} 
and GAP \cite{webster2018mind}, the latter being a gender-balanced coreference dataset consisting of ambiguous pronouns that have to be resolved to the correct coreferent name.

\subsection{Relation extraction}
\label{ch_intro:sec:relation_extraction}
\begin{figure}[t]
\centering
\includegraphics[width=.7\columnwidth]{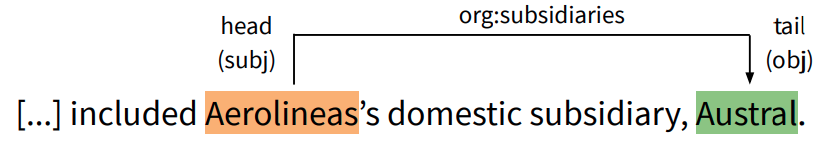}
\captionsetup{singlelinecheck=off}
\caption[Example of relation extraction]{Example of relation extraction from TACRED \cite{zhang2017position} dataset. Source: \cite{alt2020tacred}.}  
\label{ch_intro:fig:rel_intro}
\end{figure}
The task of relation extraction consists in identifying semantically meaningful relations between two mentions 
in text.  \Figref{ch_intro:fig:rel_intro} illustrates an example from the TACRED dataset \cite{zhang2017position} of the relation \texttt{org:subsidiaries} between the mentions ``Aerolineas'' and ``Austral''. This relation denotes that ``Austral'' (\textit{object} or \textit{tail} of the relation) is a subsidiary of (\textit{relation type}) ``Aerolineas'' (\textit{head} or \textit{subject} of the relation). The relation types are 
defined upfront
and vary from dataset to dataset. For example, the BC5CDR \cite{li2016biocreative, wei2015overview} dataset contains only a single relation type (indicating whether a disease is caused by a particular chemical), while DocRED \cite{yao2019docred} contains 96 distinct Wikipedia-derived relation types. Other datasets used in relation extraction include ACE 2004 \cite{doddington2004automatic}, ACE 2005 \cite{walker2006ace}, CoNLL04 \cite{roth2004linear}, SemEval 2010 - Task 8 \cite{hendrickx2010semeval}, SciERC \cite{luan2018multi} and TAC-KBP \cite{ji2010overview,ji2015overview,ji2017overview}. 

\newgeometry{margin=2.5cm}
\begin{landscape}
\begin{figure}[t]
\centering
\includegraphics[width=0.9\columnwidth]{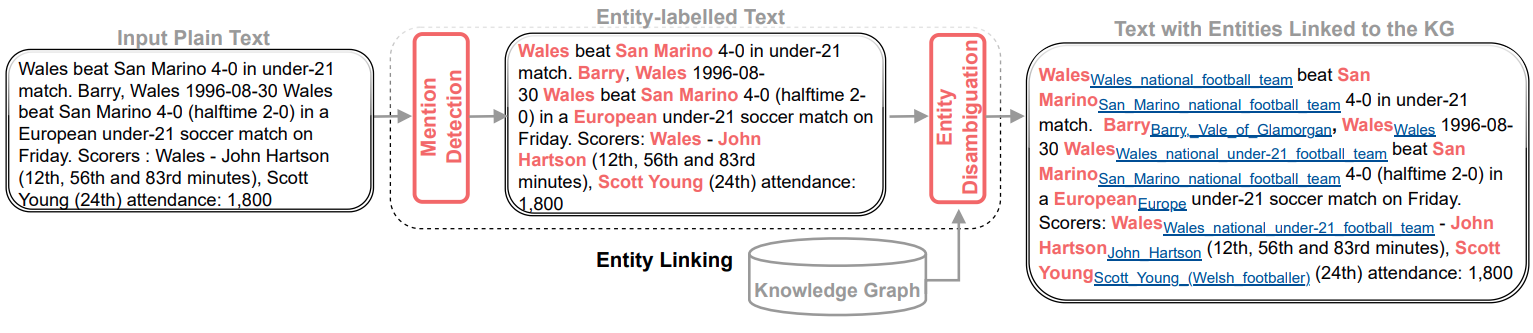}
\captionsetup{singlelinecheck=off}
\caption[Example of entity linking task]{Example of entity linking task, which involves two steps: \begin{enumerate*}[(i)]
    \item mention detection, and
    \item entity disambiguation
\end{enumerate*}. Source: \cite{sevgili2020neural}.}  
\label{ch_intro:fig:linking_intro}
\end{figure}
\ \\
\ \\ 
\end{landscape}
\restoregeometry
\clearpage
\subsection{Entity linking}
\label{ch_intro:sec:entity_linking}
Entity linking (EL) task consists in mapping a particular mention in text to an entry in a knowledge base (\eg Wikipedia) which defines the concept (entity) denoted by the mention. 
Some works \cite{ganea2017deep, kolitsas2018end, sevgili2020neural, zhang2021entqa, zaporojets2021consistent} classify entity linking in two subtasks: mention detection (MD) and entity disambiguation (ED) (see \figref{ch_intro:fig:linking_intro}), and refer to both of these tasks together 
as \textit{entity linking}. Yet, most 
mainstream works \cite{rao2013entity, wu2019zero, logeswaran2019zero, onoe2020fine, raiman2022deeptype}, 
do not 
make this distinction and use the concept of \textit{entity linking} as a synonym of entity disambiguation. 
We go more in detail on these two different settings in \secref{ch_intro:sec:learning_in_nlp_tasks} of this chapter, where we describe the difference between \textit{single-task} and \textit{joint} models. 
Most of the datasets in entity linking use Wikipedia (\ie wikification; \cite{mihalcea2007wikify}) as the target Knowledge Base to which all the entity mentions are disambiguated. 

Most current state-of-the-art EL models \cite{yamada2020global, orr2020bootleg, de2020autoregressive, zhang2021entqa, de2021highly} report on 
datasets from predominantly the news domain such as AIDA \cite{hoffart2011robust}, 
KORE50~\cite{hoffart2011robust}, AQUAINT \cite{milne2008learning}, ACE 2004, MSNBC \cite{ratinov2011local}, N$^3$ \cite{roder2014n3}, 
VoxEL\cite{rosales2018voxel}, and TAC-KBP 2010-2015 \cite{ji2010overview,ji2015overview}. Other frequently used datasets include the web-based IITB \cite{kulkarni2009collective} and OKE 15/16 \cite{nuzzolese2015open}, as well as the tweet-based Derczynski \cite{derczynski2015analysis}. Additionally, larger yet automatically annotated datasets such as WNED-WIKI and WNED-CWEB \cite{guo2018robust} 
have been also widely adopted. 
Finally, 
a number of
resources such as the domain-specific biomedical MedMentions \cite{mohan2018medmentions}, the zero-shot ZeShEL \cite{logeswaran2019zero}, and the multi-task DWIE \cite{zaporojets2021dwie} (see \chapref{chap:dwie}) and \ouraida \cite{zaporojets2021consistent} datasets have been recently introduced. 
Many of the mentioned datasets are further covered by entity linking evaluation frameworks such as GERBIL \cite{usbeck2015gerbil,roder2018gerbil} and KILT \cite{petroni2020kilt} that provide a common interface to evaluate the models. 

\section{Machine learning methodology}
\label{ch_intro:sec:ml_approaches}
In this section we will provide an introductory description of machine learning methodologies to solve the tasks explained in 
\secref{ch_intro:sec:ie_tasks}. Due to 
the
sheer amount of architectures to solve the described tasks, 
we 
limit 
our discussion
to the ones that are relevant to the main contributions of this thesis. 
First, we describe the \textit{span-based information extraction} (\secref{ch_intro:sec:span_based_el}) approach used in the models presented in Chapters \ref{chap:dwie}-\ref{chap:injecting_knowledge} to detect candidate mention spans in the text. Next, we describe the backbone of the graph propagation algorithm to transfer the local contextual information between these mention spans (see \secref{ch_intro:sec:graph_propagation_mechanisms}). The use of such graph algorithm allows to boost the performance of entity linking tasks by efficiently exchanging information between the mentions spread across the document in \chapref{chap:dwie}. In \secref{ch_intro:sec:dense_passage_retrieval}  we describe a novel technique used to efficiently retrieve similar documents given a query commonly known as \textit{dense passage retrieval (DPR)}. We use this method in \chapref{chap:temporal_el} of this thesis to efficiently retrieve candidate entities given the mention context when tackling temporal entity linking task. Recent work \cite{ayoola2022refined,wu2019zero,zhang2021entqa}, evidences that such fast DPR algorithm is the key backbone component to achieve state of the art results in entity linking task, reason why we use it as a baseline in the \chapref{chap:temporal_el} of this thesis. Finally, in \secref{ch_intro:sec:external_knowledge_sources} we describe two main external knowledge sources, each one containing millions of entities that we use to link to the mentions in text.
The first one represents textual-based knowledge bases such as Wikipedia, where each entity is described using plain text. For example, the description of the entity referring to \textit{Meghan Markle} is located at the following Wikipedia url: \url{https://en.wikipedia.org/wiki/Meghan,_Duchess_of_Sussex}. The second knowledge source is a structured knowledge graph (KG). Concretely, in this thesis we use Wikidata as such structured knowledge repository which consists of entities, their attributes, and relations between them. 
The distinction between these two knowledge sources is particularly crucial in the context of  \chapref{chap:injecting_knowledge}, where we combine representations of both of them to obtain major boost in performance as compared to using each of these sources separately. 
\subsection{Span-based information extraction}
\label{ch_intro:sec:span_based_el}
\begin{figure}[t]
\centering
\includegraphics[width=1.0\columnwidth]{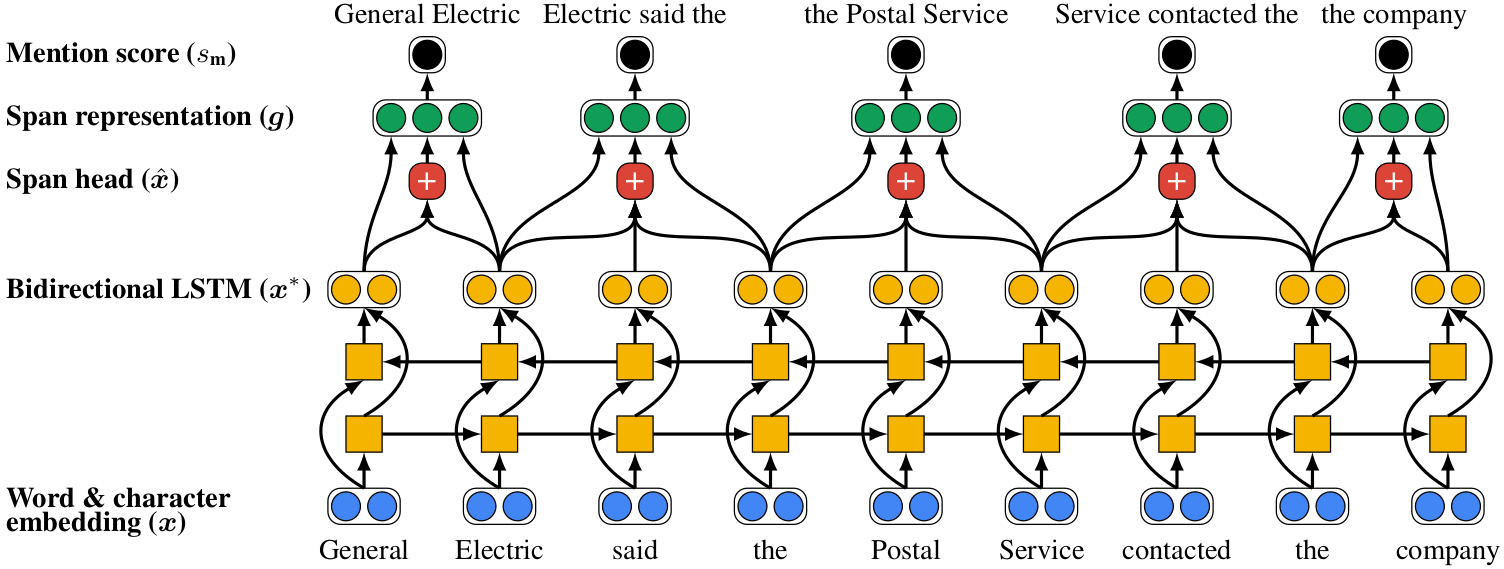}
\captionsetup{singlelinecheck=off}
\caption[Illustration of span-based approach in Coreference Resolution task]{Illustration of span-based approach in Coreference Resolution task. Source: \cite{lee2017end}.}  
\label{ch_intro:fig:spans}
\end{figure}
The \textit{span-based} information-extraction approach has been popularized by the work of \cite{lee2017end}. The authors propose, instead of using traditional sequential models such as the ones based on Conditional Random Fields \cite{mccallum2003early,sutton2012introduction,bekoulis2018joint}, to work on all possible token spans of up to a specified length.\footnote{The length of 10 is enough to cover more than 99\% of the mentions in most of the current information extraction datasets.}
Such a span-driven approach has a number advantages compared to the sequential models. 
First, 
they allow to straightforwardly model the loss function for the information extraction tasks, backpropagating towards the pruner and the internal LSTM weights. This is because the loss is directly calculated on candidate mention spans and not on intermediate LSTM states as in sequential models. 
Furthermore, the textual spans provide a natural component to perform further graph propagation as will be explained in \secref{ch_intro:sec:graph_propagation_mechanisms}. 
Finally, our observations suggest that the span-based models allow to recover more mentions of texts, specially the longer ones consisting of multiple tokens.
\Figref{ch_intro:fig:spans} illustrates how the span representation is calculated by concatenating left and right BiLSTM-based token representations. 
The total number of spans per document of maximum width $L$ can be calculated as: 
\begin{equation}
|S| = \sum\limits_{k=1}^{w_{\mathrm{max}}} |T| - k + 1 \;={w_{\mathrm{max}}}\left(|T| - \frac{w_{\mathrm{max}}-1}{2}\right)\label{ch_intro:eq:spans},
\end{equation}
where $T$ represents the total number of tokens in the document, and $w_{\mathrm{max}}$ is the maximum span width.
In order to avoid memory overflow, the resulting $\vert S \vert$ spans are pruned using a \textit{pruner} component to a manageable fraction of all the tokens in the document. Next, a separate model for a particular IE task is used independently or jointly with the pruner model to predict. The success of span-based approaches \cite{lee2017end, lee2018higher, luan2019general,luan2018multi,wadden2019entity,he2018jointly} has also been demonstrated in BERT based models. Thus, \cite{joshi2020spanbert} introduce SpanBERT, a BERT model pre-trained directly on spans instead of tokens (as is the case of BERT) in text. This model has been successfully used as the backbone to achieve state-of-the-art results in numerous information extraction tasks \cite{xu2020revealing,xu2022modeling,wu2020corefqa}. We use span-based models in our work described in Chapters \ref{chap:dwie}, \ref{chap:coreflinker} and \ref{chap:injecting_knowledge} of this thesis. Furthermore, we use SpanBERT as the pre-trained model in the architecture described in \chapref{chap:coreflinker}. 
\subsection{Graph propagation mechanisms}
\label{ch_intro:sec:graph_propagation_mechanisms}
The span representations obtained from the tokens (see \figref{ch_intro:fig:spans}) can be formalized as follows: 
\begin{equation}
\textbf{g}^0_i = [\textbf{e}_{l}; \textbf{e}_{r}; \boldsymbol{\psi}_{r-l}] \label{ch_intro:eq:spans}
\end{equation}
Where $\textbf{g}^0_i$ is the representation for span $s_i$, ranging from token $l$ to token $r$, by concatenating their respective BiLSTM states $\textbf{e}_{l}$ and $\textbf{e}_{r}$ with an embedding $\boldsymbol{\psi}_{r-l}$ for the span width $w_i = r-l$. 

These representations only depend on the underlying BiLSTM states which are inefficient in retaining the context information located further than 50 tokens away \cite{khandelwal2018sharp}. 
Recent work tackled this problem by using \textit{graph propagation techniques} \cite{wu2020comprehensive,zhou2020graph} in span-based models. These techniques are also referred to with the term of \textit{higher order inference (HOI)} \cite{xu2020revealing,lee2018higher}, and consist in iteratively propagating contextual information between spans. More formally, the propagation operation can be defined as follows: 
\begin{align}
    \textbf{g}_i^{t+1} &= \textbf{f}^t_x(s_i) \odot \textbf{g}^t_i + \left(1 - \textbf{f}^t_x(s_i)\right) \odot \textbf{u}^t_x(s_i),
    \label{ch_intro:eq:update_spans2}
\end{align}
where the $n$-dimentional vector $\textbf{f}^t_x(s_i)$, 
can be interpreted as 
a gating vector that acts as a switch between the current span representations $\textbf{g}^t_i \in \mathbb{R}^n$, and the update span vector $\textbf{u}^t_x(s_i) \in \mathbb{R}^n$. The various graph propagation methods differ in how $\textbf{u}^t_x(s_i)$ is calculated. In our work on entity-centric joint information extraction described in \chapref{chap:dwie}, we introduce our own task-independent attention-based graph propagation technique (\texttt{AttProp}). 

Finally, recently there has been an extensive study evaluating the gains of using HOI on BERT-based (as opposed to BiLSTM-based as described above) coreference resolution models \cite{xu2020revealing}. The authors conclude that, while the coreference resolution models experience additional gains with the incorporation of HOI techniques, it is minimal compared to the gains when using HOI on top of ELMo \cite{peters2018deep} or LSTM \cite{hochreiter1997long} token representation techniques. 
\subsection{Dense passage retrieval}
\label{ch_intro:sec:dense_passage_retrieval}
\begin{figure}[t]
\centering
\includegraphics[width=0.7\columnwidth]{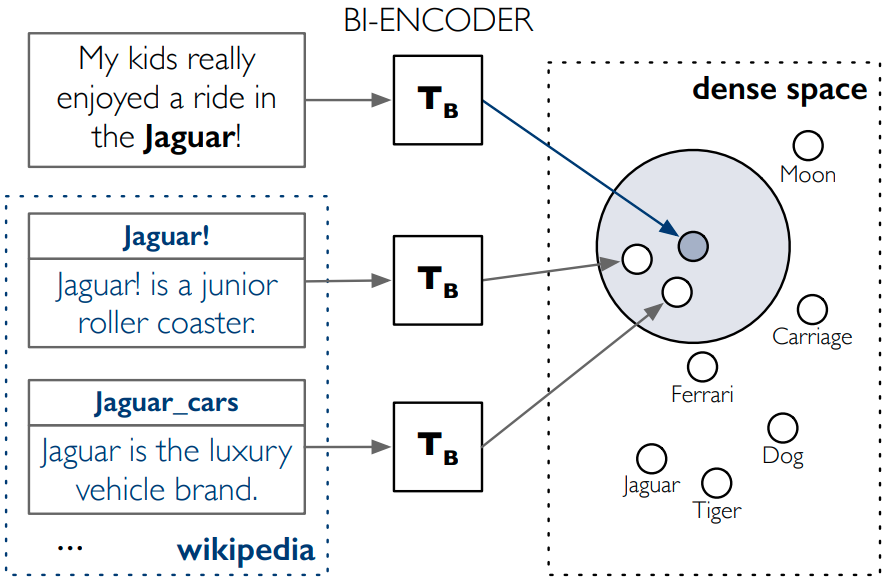}
\captionsetup{singlelinecheck=off}
\caption[Illustration of dense passage retrieval (DPR) applied on entity linking task.]{Illustration of Dense passage retrieval (DPR) applied on entity linking task. Source: \cite{wu2019zero}.}
\label{ch_intro:fig:biencoder}
\end{figure}
The dense passage retrieval (DPR) concept was introduced by \cite{karpukhin2020dense} and is used in our work on temporal entity linking (\chapref{chap:temporal_el}). Concretely, DPR consists in using \textit{dense representations} to match a \textit{query} text with \textit{passages}. The example in \figref{ch_intro:fig:biencoder} showcases the mechanism behind DPR for the entity linking task. Concretely, a set of Wikipedia pages (``Jaguar!'', ``Jaguar\_cars'') are encoded in the same dense space as the mentions (``Jaguar'' on top) linked to them. Next, a dot product operation is used to match a query (entity mention with the context) to the Wikipedia entity descriptions in a single dense space. More formally, the similarity operation is defined as follows: 
\begin{equation}
    \mathrm{sim}(q,p) = E_Q(q)^T E_P(p)
\end{equation}
Where $E_Q$ module encodes the text $q$ in a fixed $d$-dimensional vector. Similarly, $E_P$ module encodes the candidate passage $p$ into another $d$-dimensional vector. 
Most commonly, $E_Q$ and $E_P$ are BERT-based pre-trained encoders, fine-tuned on a specific task (\eg entity linking). 

\subsection{External knowledge sources}
\label{ch_intro:sec:external_knowledge_sources}
The entities are defined in multiple types of knowledge bases (KBs). A well-known knowledge base is Wikipedia. Each of the entities is described in the corresponding Wikipedia page. For example, the entity \textit{Ghent University} is described on the Wikipedia page \url{https://en.wikipedia.org/wiki/Ghent_University}. Yet, 
KBs 
such as Wikipedia provide only a textual description of entities. Inherently, computers are not adapted to interpret information in this unstructured textual format. As a result, the research community has recently shown a growing interest in representing the information in a structured manner by means of knowledge graphs (KGs) \cite{ji2021survey,gutierrez2021knowledge,yan2018retrospective}. In a 
KG, 
a particular node representing an entity is connected to other nodes as well as associated with certain attributes that describe it. One of the most well-known Knowledge Graphs is Wikidata\cite{vrandevcic2014wikidata}.\footnote{\url{https://www.wikidata.org/}} This knowledge graph interconnects millions of entities\footnote{At the moment of current writing, Wikidata contains more than 97 million entities.} using edges of different types. For example, the entity \textit{Ghent University} is connected to the entity \textit{Belgium} by the edge of type \textit{country}, and with the node \textit{William I of the Netherlands} by the edge of type \textit{founded by}. In our work described in \chapref{chap:injecting_knowledge} we rely on both the textual Wikipedia 
KB
as well as the structured Wikidata 
KG 
to obtain robust entity representations that are used to inject additional knowledge in information extraction models. 

\section{Entity-centric approach}
\label{ch_intro:sec:entity_centric}
\begin{figure}[t]
\centering
\includegraphics[width=1.0\columnwidth, trim={0.3cm 3.2cm 1.5cm 2.0cm},clip]{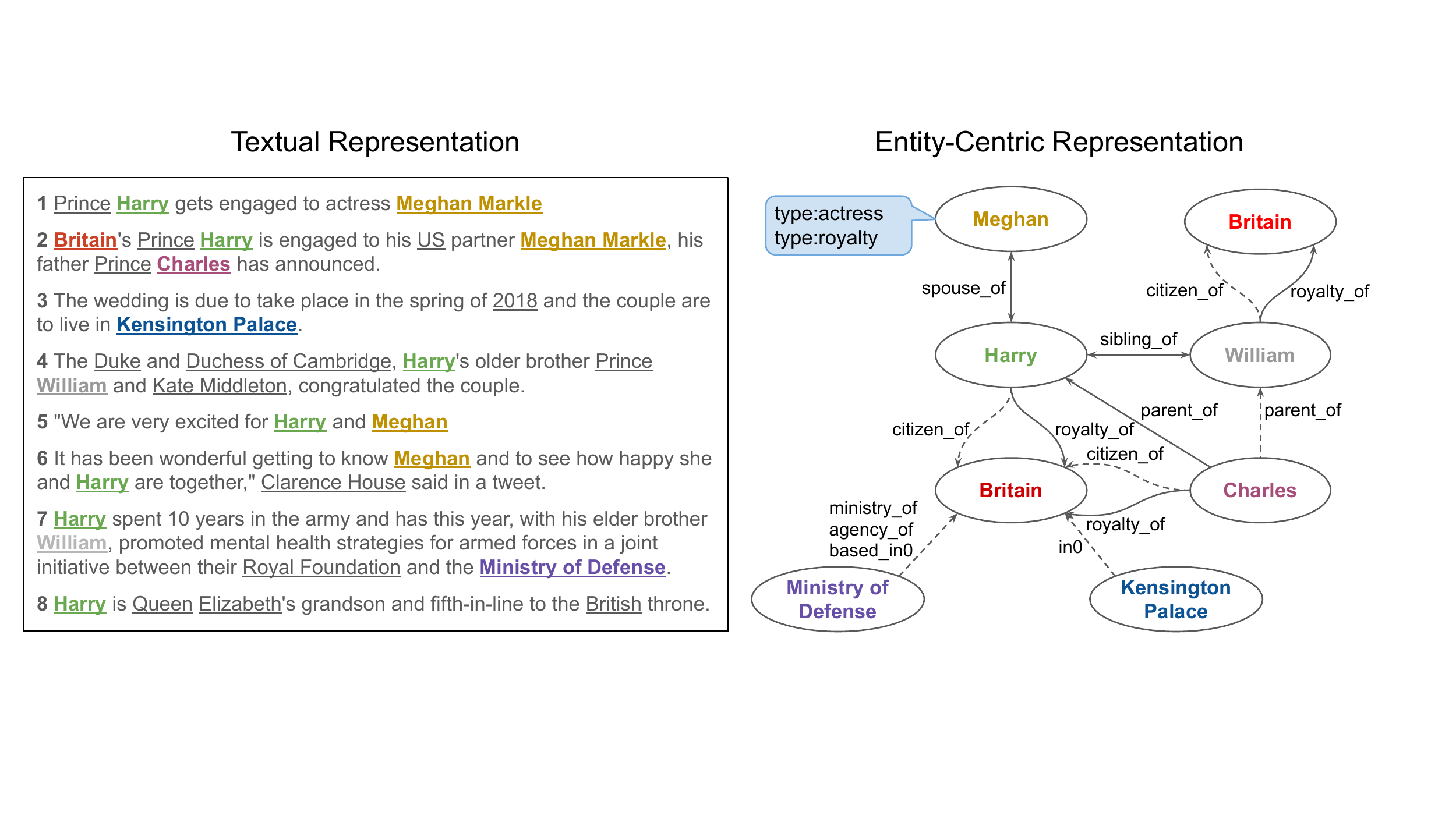}
\captionsetup{singlelinecheck=off}
\caption[Illustrative example of entity-centric approach.]{An example taken from the \datasetname~dataset (see \chapref{chap:dwie}) to illustrate the \textit{entity-centric} approach. Each of the entity mentions in the full \textit{document text} of the left is identified and clustered into entities represented in the graph in the right part of the figure. 
The color of the mentions in text indicates the entity they represent. 
The remaining annotations 
such as entity and relation types as well as entity linking (not shown in the figure),
are performed on entity level (\ie are \textit{entity-centric}). 
}
\label{ch_intro:fig:entity_centric}
\end{figure}
The term \textit{entity-centric} is widely 
used
and is key to understand some of the main contributions of 
this thesis. 
The main goal of entity-centric approaches is
to encourage to develop
of models that reason in
terms of entities (concepts) instead of on individual entity mentions in text. 
One specific type of entities 
we work
with in Chapters 
\ref{chap:dwie}--\ref{chap:injecting_knowledge} 
are named entities, 
which include 
all entities denoted with proper names (\eg names of people, companies, countries, etc.).
\Figref{ch_intro:fig:entity_centric} illustrates an example from the entity-centric \datasetname~dataset, which will be introduced formally in \chapref{chap:dwie}. The left part of the figure depicts the text that is annotated. 
The right part of the figure illustrates the corresponding structured entity-centric representation where each of the 
information extraction (IE) 
annotations are done on entity level. 
Each such entity may have multiple coreferent entity mentions in the text.
This approach allows to summarize the information of the whole document (\eg entity and relation types) in a single graph. Furthermore, the document-level perspective of entity-centric annotations also enables to extract information that is not explicitly mentioned in text, but rather can be deduced from the content of the document (represented by dashed arrows in \figref{ch_intro:fig:entity_centric}). 
This contrasts with mention-driven \cite{doddington2004automatic,walker2006ace,ellis2014overview,ellis2015overview,ji2010overview,ji2015overview,ji2017overview,song2015light} annotations that rely on specific and explicit textual triggers. In this thesis, we are interested in the following entity-centric annotations: 
\begin{enumerate}
    \item \textbf{Coreferent entity mentions}: the coreferent entity mentions are grouped in entity mention \textit{clusters}, each one representing a single entity (right part of \figref{ch_intro:fig:entity_centric}). This clustering of mentions is addressed by \textit{coreference resolution} IE task (see \secref{ch_intro:sec:coreference_resolution}), and is tackled by models introduced in Chapters 
    \ref{chap:dwie}--\ref{chap:injecting_knowledge} 
    of this thesis.
    \item \textbf{Relation types}: indicate semantically meaningful relations between identified entities. Two entities can be connected by multiple relation types 
    (\ie the relation annotations are \textit{multilabel}), 
    such as \textit{based\_in} and \textit{ministry\_of} between entities \textit{Ministry of Defense} and \textit{Britain} in \figref{ch_intro:fig:entity_centric}. The prediction of these types is addressed by the \textit{relation extraction} IE task (see \secref{ch_intro:sec:relation_extraction}), and is tackled by models introduced in Chapters \ref{chap:dwie} and \ref{chap:injecting_knowledge} of this thesis.
    \item \textbf{Entity types}: describe the main characteristics of a particular entity. 
    Similar to relation types, 
    the entity types are multilabel (\ie a particular entity can be associated with multiple entity types). For example, in \figref{ch_intro:fig:entity_centric} the entity \textit{Meghan} is described with entity types \textit{type:actress} and \textit{type:royalty}. The predictions of entity types is addressed by the named entity recognition (NER) IE task (see \secref{ch_intro:sec:named_entity_recognition}), and is tackled by models introduced in Chapters \ref{chap:dwie} and \ref{chap:injecting_knowledge}.
    \item \textbf{Entity linking} (not shown in the \figref{ch_intro:fig:entity_centric}): each of the entities are connected (linked) to the respective entity in the Wikipedia Knowledge Base. For example, the entity Meghan of the right part of \figref{ch_intro:fig:entity_centric} is linked
    to the page \emph{\href{https://en.wikipedia.org/wiki/Meghan,_Duchess_of_Sussex}{Meghan,\_Duchess\_of\_Sussex}} in Wikipedia. The prediction of these links is addressed by the \textit{entity linking} IE task (see \secref{ch_intro:sec:entity_linking}), and is tackled by models introduced in \chapref{chap:coreflinker} using an entity-centric approach, and in \chapref{chap:temporal_el} in a more traditional mention-driven setting (\ie predicting entity links for each of the mentions separately). 
\end{enumerate}
\section{Learning in NLP tasks}
\label{ch_intro:sec:learning_in_nlp_tasks}
\begin{figure}[!t]
\centering
\includegraphics[width=0.95\columnwidth, trim={2.0cm 2.5cm 2.3cm 2.5cm},clip]{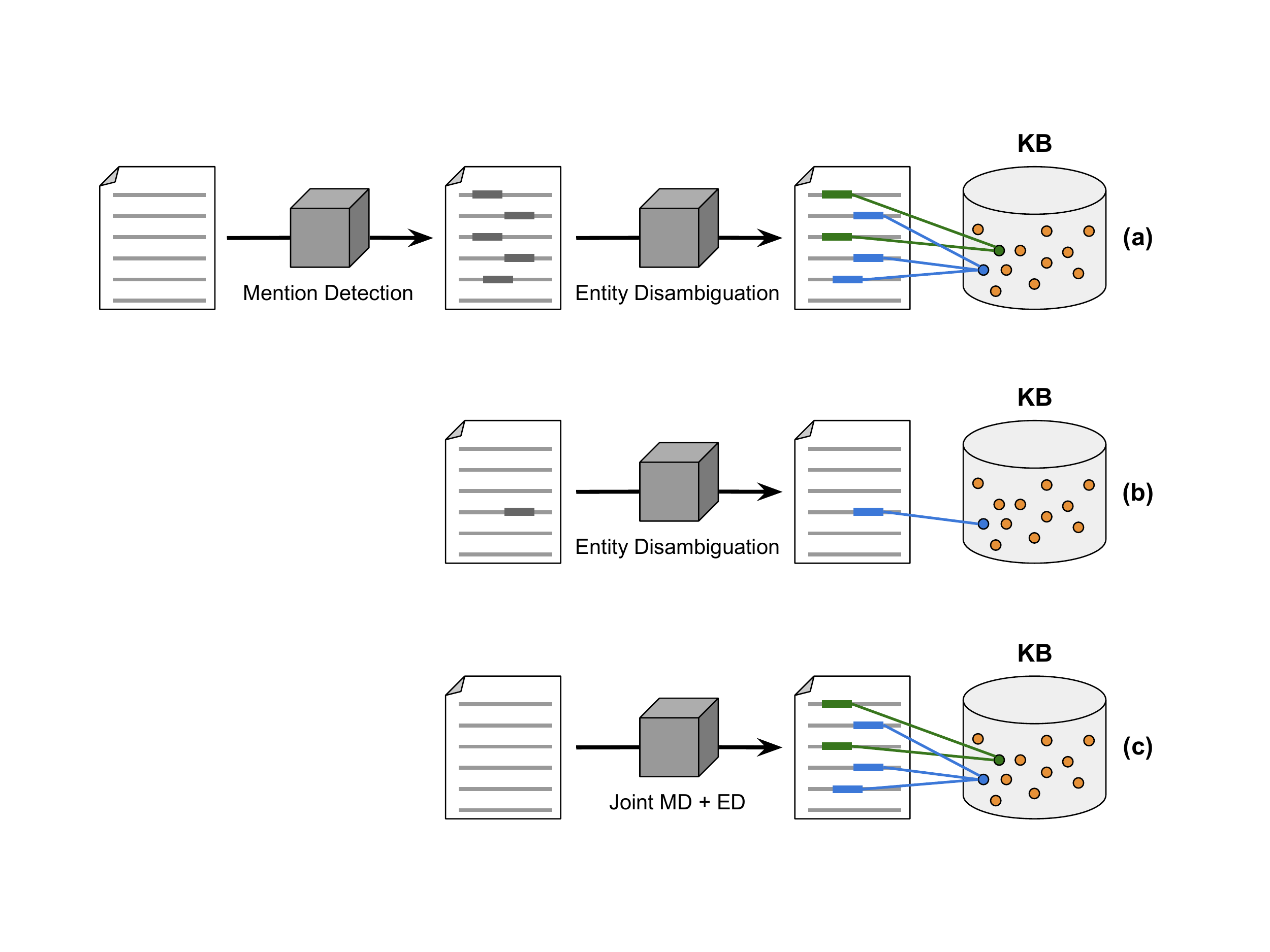}
\captionsetup{singlelinecheck=off}
\caption[Single task and joint setups illustration]{Figure showcasing an example of \textit{entity detection} and \textit{entity disambiguation} information extraction (IE) tasks in order to illustrate three IE learning setups: \begin{enumerate*}[(a)]
    \item\label{ch_intro:fig:nlp_tasks:pipeline} pipelined setup where separate 
    models
    are used sequentially for each of the tasks in order to produce the desired output given a \textit{plain document} (\ie without annotations) as input,
    \item\label{ch_intro:fig:nlp_tasks:single}single task setup, where the ground truth annotations (\ie mention(s) to disambiguate in the showcased example) are already given as input to the model,
    \item\label{ch_intro:fig:nlp_tasks:joint}\textit{end-to-end} joint setup where a single model is trained and evaluated jointly on multiple tasks necessary to produce the desired output given a \textit{plain document} as input.
\end{enumerate*} 
}
\label{ch_intro:fig:nlp_tasks}
\end{figure}
In this section we will describe three approaches to solve information extraction (IE) tasks relevant to our thesis. We will use the example in \figref{ch_intro:fig:nlp_tasks}, which depicts two subtasks necessary to fully solve the entity-linking task (\ie with unannotated documents given as input) (see also \secref{ch_intro:sec:entity_linking}). The first subtask is \textit{mention detection} (MD), 
to detect all the mentions in text to be linked to a Knowledge Base. In most 
datasets \cite{zaporojets2021consistent, zaporojets2021dwie,hoffart2011robust,hoffart2012kore,roder2014n3}, these mentions are limited to named entity mentions (\ie proper nouns). The second subtask is \textit{entity disambiguation} (ED), whose goal is to link each of the detected mentions by MD to entities in the Knowledge Base (KB). Both of these subtasks are needed to solve entity linking (EL) task 
starting from an unannotated plain document input (\ie setups depicted in \figref{ch_intro:fig:nlp_tasks}\ref{ch_intro:fig:nlp_tasks:pipeline} and \figref{ch_intro:fig:nlp_tasks}\ref{ch_intro:fig:nlp_tasks:joint}).

\subsection{Single task learning}
Traditionally, IE architectures tackling multiple (sub-)tasks were solved using a pipelined approach as depicted in the example in \figref{ch_intro:fig:nlp_tasks}\ref{ch_intro:fig:nlp_tasks:pipeline}. For 
instance, 
\cite{ceccarelli2013dexter,van2013learning,piccinno2014tagme,hoffart2011robust} propose an entity linking pipeline that first performs mention detection and then entity disambiguation. 
While straightforward to implement, this setup suffers from at least two drawbacks: \begin{enumerate*}[(i)]
    \item sequential error accumulation of models executed in the pipeline, and 
    \item inability to leverage possible inter-relations between tasks (\eg knowing that a particular textual span can be linked to a KB can help the mention detection component). 
\end{enumerate*}

\Figref{ch_intro:fig:nlp_tasks}\ref{ch_intro:fig:nlp_tasks:single} depicts another single task learning setting. 
A set of ground truth annotations (\eg a selection of ground truth mentions in text) are given to the model as input, and it only has to perform a specific task (\ie \textit{entity disambiguation} in  \figref{ch_intro:fig:nlp_tasks}\ref{ch_intro:fig:nlp_tasks:single}). While not solving the entity linking task completely, this approach has at least two advantages: \begin{enumerate*}[(i)]
    \item it 
    alleviates the 
    study of the performance of a single component responsible for a particular task, and 
    \item it opens the possibility to filter specific ground truth annotations to be fed to the input model
\end{enumerate*}. This latter point is exploited in recent entity 
disambiguation 
works \cite{logeswaran2019zero,provatorova2020named,eshel2017named, onoe2020fine} to test the performance of state-of-the-art entity linking models on more challenging entity mentions (\eg entity mentions linked to unpopular entities). Yet, this may also present the disadvantage of not reflecting the performance in a \textit{real-world} scenario, which is dominated by trivial mentions (\eg mentions linked to most popular entities) as demonstrated by \cite{guo2018robust}.
In \chapref{chap:temporal_el}, we use this setup to create temporally evolving \textit{entity disambiguation} dataset where we filter out the trivial mentions (\eg mentions whose surface form is the same as that of the title of linked entity) in order to focus on most challenging cases.
\subsection{Joint learning}
\label{ch_intro:sec:multi_task}
More recently, there has been a shift towards creating end-to-end models able to be trained and evaluated on multiple tasks jointly. These models share a single neural net architecture, and typically a single loss function is used during their training. The main advantage of such models is 
an easier deployment
since they do not require coupling of multiple components in a pipeline (see \figref{ch_intro:fig:nlp_tasks}\ref{ch_intro:fig:nlp_tasks:pipeline}). Consequently, they 
do not suffer from
error propagation from one model to the next one, characteristic of the pipelined 
approach.
For instance, recent 
entity linking architectures
\cite{kolitsas2018end, de2020autoregressive, broscheit2019investigating, zaporojets2021consistent, zhang2021entqa,ayoola2022refined}
model 
both the mention detection (MD) and entity disambiguation (ED) tasks jointly as illustrated \figref{ch_intro:fig:nlp_tasks}\ref{ch_intro:fig:nlp_tasks:joint}. This contrasts with entity linking models focusing exclusively on the ED task \cite{wu2019zero, logeswaran2019zero, barba2022extend, yamada2020global, raiman2022deeptype, orr2020bootleg, rao2013entity, onoe2020fine}, taking as input the ground truth mentions as illustrated in \figref{ch_intro:fig:nlp_tasks}\ref{ch_intro:fig:nlp_tasks:joint}. 
Additionally, related work \cite{martins2019joint, bekoulis2018joint, bekoulis2018adversarial, zaporojets2021consistent, zaporojets2021dwie,ruder2019latent}
has also shown an improvement in performance when combining multiple IE tasks in a joint model. This is explained by the interdependence between tasks, where the information from one task can benefit other task(s). For example, knowing 
the type of a particular entity mention (NER task) can help the model to restrict the entities this mention can be potentially linked to in a KB (EL task) \cite{martins2019joint}. 

From a more detailed perspective, we distinguish between \textit{multi-task} and \textit{joint} learning. The \textit{multi-task} learning is characterized by using one or more related tasks from single or separate datasets to train the model. Normally these tasks share some commonalities which act as regularizers to the shared weights of neural model \cite{collobert2011natural} or label spaces \cite{augenstein2018multi}. 
A typical example are language models such as BERT \cite{joshi2019bert} that are pre-trained on predicting tokens, which has a big impact on a huge number of natural language processing tasks such as coreference resolution \cite{joshi2020spanbert}, question answering \cite{yasunaga2021qa}, relation extraction \cite{zhang2021document}, etc. 
In this thesis, we use purely multi-tasking approach in Chapters \ref{chap:dwie} and \ref{chap:injecting_knowledge} by combining the summing the losses of each of the tasks to obtain the final loss. Empirically, this setup has positive effect with a boost in performance for relation extraction task in \chapref{chap:dwie}. 
Furthermore, we use the term of \textit{joint} to refer to multi-task models strictly trained on a single dataset, such as the models in Chapters \ref{chap:dwie} and \ref{chap:injecting_knowledge}. We also include in the category of \textit{joint} architectures that not necessarily share the neural network weights, but rather connect different tasks in structured way. This is the example of the joint model introduced in \chapref{chap:coreflinker} where we connect the coreference and entity linking tasks in a single structured task using a single loss function. 

In this thesis,
we use \textit{joint} learning to 
perform
cluster-level predictions 
inherent
to the entity-centric approach (see \secref{ch_intro:sec:entity_centric}) on datasets such as  DWIE (\chapref{chap:dwie}) and AIDA+ (\chapref{chap:coreflinker}). Concretely, in \chapref{chap:dwie} we experiment with a joint loss which consists of a sum of losses for each of the tasks the model is trained on. We discover that this setup, besides allowing to make cluster-level predictions using a single end-to-end model, also leads to a significant improvement in performance on the relation extraction task. This indicates that in the proposed joint model, the relation extraction task benefits from the information 
contained in
the rest of the tasks such as NER and coreference resolution (coref). In \chapref{chap:coreflinker}, we go one step further and 
frame entity linking (EL) and 
coref
tasks as a single structured task. This allows us to have a single loss term for both of the tasks. We show that this joint approach leads to a general improvement of both 
coref and EL
tasks of up to +5\% F1-score. Furthermore, our joint model is also able to solve corner cases of EL task with an improvement of up to +50\%
in 
accuracy compared to the standalone EL model. 

\section{Temporal outline of the research}
The content of the current PhD thesis with the main focus on entity-centric information extraction was not evident at the beginning of the PhD (almost five years ago). Chronologically, our first work \cite{zaporojets2018predicting} is described in \appendixref{chap:clpsych}, and tackles the shared task in CLPsych 2018 workshop. The proposed architecture achieves competitive results in predicting the various metrics used to measure depression and anxiety based on the content of textual surveys. Next, towards the end of 2018, our interest shifted to investigate the limits and shortcomings of machine learning 
models 
to reason on the facts described in Wikipedia entities. Concretely, we focused on studying the results of then state-of-the-art models used in recently introduced FEVER \cite{thorne2018fever} shared task challenge. This challenge consists in verification of textual claims using entity descriptions from Wikipedia. Our analysis of the results showed the difficulty of all the models when reasoning with sentences involving numbers. This inspired us to pursue the work \cite{zaporojets2021solving} where we propose an innovative structural approach to solve arithmetic word problems which involve multiple numbers, given a textual description of a problem (see \appendixref{chap:mwp}). 

In parallel, during 2018 and 2019, we started annotating DWIE (Deutsche Welle corpus for Information Extraction) dataset as part of CPN project\footnote{\url{https://www.projectcpn.eu/}}.
This resulted in our work \cite{zaporojets2021dwie} described in \chapref{chap:dwie}. 
The main challenge we faced when working on this project was to design an architecture that would exploit the complementary information across the mentions referring to the same entity (entity-centric approach) in the document. In our baseline model presented in \chapref{chap:dwie} we succeed in harnessing this interrelated cross-mention information passing by applying graph propagation techniques, which resulted in a significant boost in performance.  
The ideas for our next two works described in Chapters \ref{chap:coreflinker}--\ref{chap:injecting_knowledge} were conceived almost simultaneously in 2020. The first idea \cite{zaporojets2021consistent} described in \chapref{chap:coreflinker} consists in framing entity linking and coreference resolution tasks in a single structural architecture with a single loss function. Conversely, the idea \cite{verlinden2021injecting} presented in \chapref{chap:injecting_knowledge} intends to exploit the entity information in external Knowledge Bases to use it jointly with the textual mentions entities to boost the performance of various information extraction tasks. 

Finally, the work \cite{zaporojets2022tempel} described in \chapref{chap:temporal_el} is a result of a joint effort between CopeNLU\footnote{\url{https://www.copenlu.com/}} and T2K\footnote{\url{https://ugentt2k.github.io/}} research groups and was done during the research visit to University of Copenhagen under the supervision of prof. Isabelle Augenstein in spring of 2022. The idea of working with dynamically evolving entity linking task originated and was shaped during numerous online brainstorm meetings in the course of the winter of 2021-2022. Our main goal was to tackle what, based on our experience when working on Chapters~\ref{chap:dwie}--\ref{chap:injecting_knowledge}, we thought was one of the main limitations of the currently available entity linking datasets: they are bound to entities and mentions created at a specific point in time. As a result, these datasets are unable to measure the effect of temporal evolution on entity linking task. Our first plan to tackle this problem was to create a dataset by splitting the news documents in our DWIE dataset (\chapref{chap:dwie}) according to their publication year, linking the mentions to the corresponding Wikipedia version. 
Yet, this would have produced a very low number of annotated mentions in each of the temporal snapshots, with a high popularity bias (\ie dominated by most popular entities). As a consequence, we decided to make use of a much larger human-annotated corpus, namely Wikipedia, in order to produce a more challenging large-scale temporally evolving entity-linking dataset: TempEL, described in \chapref{chap:temporal_el}. 
%
\section{Research contributions}
\label{ch_intro:sec:research_contributions}
\begin{table}
\small
    \centering
    \begin{tabular}{c p{4cm} p{5cm} }
\toprule
\textbf{Chapter}&\textbf{Task}& \textbf{Contribution}  \\
\toprule

\ref{chap:dwie}& {Multi-task information extraction.} & Define multi-tasking IE dataset and baselines. \\
\ref{chap:coreflinker}& {Joint coreference and entity linking.} & Joint coreference and entity linking architecture that conceives both tasks in a single structured representation. \\
\ref{chap:injecting_knowledge}& {Named entity recognition, coreference resolution and relation extraction.} & New neural architecture that combines textual spans with entity representations from external Knowledge Bases.  \\
\ref{chap:temporal_el}& {Entity linking.} & New dataset that allows to track the temporal evolution of entities in Wikipedia and the impact of temporal shift on entity linking task. \\
\bottomrule
    \end{tabular}
    \caption{Overview of the contributions presented in this thesis.}
    \label{ch_intro:tab:chap_contributions}
\end{table}

In this section, we present an overview of the main research contributions of this thesis. We organize the addressed research problems in chapters, each one tackling clearly defined research questions. \tabref{ch_intro:tab:chap_contributions} summarizes the information extraction tasks and contributions of each of the chapters. In Chapters \ref{chap:dwie}--\ref{chap:injecting_knowledge} we explore how an entity-centric approach can further boost the performance of information extraction tasks compared to baseline mention-centric architectures, with special focus on multi-task joint models (\secref{ch_intro:sec:multi_task}). Conversely, in Chapter \ref{chap:temporal_el} we explore the 
performance of EL solutions for entities as they evolve over larger timeframes
Below, we provide a summary of the contributions of each of the chapters: 
\begin{itemize}
    \item In Chapter \ref{chap:dwie} we propose a radically different, \textit{entity-centric} view on the information in text. We argue that, instead of using individual mentions in text to understand their meaning, we need to build applications that would operate in terms of entity concepts. This entity-centric approach involves grouping all the mentions referring to the same entity (\eg ``Ghent'') in a single coreference cluster and perform the rest of the tasks (\eg relation extraction, entity linking, etc.) on this cluster level. Our approach has the advantage of leveraging the information across all the entity mentions referring to a single entity in the document at once. As a consequence, the entity-centric approach requires a document-level view on the text. Yet, the NLP community has produced no evaluation and training resources (\ie datasets) that would have this document-level focus for multiple tasks at once. We tackle this research gap by introducing DWIE (Deustche Welle Information Extraction) dataset in which we annotate four different tasks on entity level: coreference resolution, entity linking, relation extraction, and named entity recognition. We further demonstrate how these tasks complement each other in a joint information extraction model.
    \item In \chapref{chap:coreflinker}, 
    we
    develop an entity-centric architecture 
    to make 
    entity linking predictions directly on 
    the entity cluster level instead of on each of the entity mentions separately. 
    To do so, 
    we frame the coreference (coref) and entity linking (EL) tasks as a single structured task. This contrasts with previous attempts to join coref+EL tasks \cite{hajishirzi2013joint,dutta2015c3el,angell2021clustering}, where 
    both of the 
    models are trained separately and additional logic is required to merge the predictions of coref and EL tasks. 
    Concretely, in this chapter we contribute with: \begin{enumerate*}[(i)]
    \item 2 architectures 
    for joint entity linking (EL) and corefence resolution, 
    \item an extended AIDA dataset \cite{hoffart2011robust}, adding new annotations of linked and NIL coreference clusters,
    \item experimental analysis on 2 datasets where our joint coref+EL models achieve up to +5\% F1-score on both tasks compared to standalone models. We also show up to +50\% in accuracy
    for hard cases of EL where entity mentions lack the correct entity in their candidate list. 

\end{enumerate*}
    \item In \chapref{chap:injecting_knowledge}, we explore how the entity knowledge contained in external Knowledge Bases (KBs) can be 
    injected
    in text to further enhance the performance of three information extraction (IE) tasks: named entity recognition, coreference resolution, and relation extraction. Furthermore, we analyze what KB representation is more beneficial for these IE tasks: either \textit{KB-graph} trained on Wikidata, or \textit{KB-text} trained directly on Wikipedia. We particularly contribute with 
\begin{enumerate*}[(i)]
\item a first span-based end-to-end architecture incorporating KB knowledge in a joint entity-centric setting, exploiting unsupervised entity linking (EL) to select KB entity candidates,
\item exploration of prior- and attention-based mechanisms to combine the EL candidate representations into the model,
\item assessment of the complementarity of KB-graph and KB-text representations, and
\item consistent gains of up to 5\% F1-score when incorporating KB knowledge in 3 document-level IE tasks evaluated on 2 different datasets. 
\end{enumerate*}
\item In \chapref{chap:temporal_el} we propose a fundamentally different, \textit{evolutionary} view on the entity linking (see \secref{ch_intro:sec:entity_linking}) task. There, we introduce a new, \ourdataset~dataset which consists of Wikipedia entity linking annotations grouped in 10 yearly snapshots. We further contribute with a study of how \textit{entity linking} task is affected by \begin{enumerate*}[(i)]
    \item changes of existing entities in time, and 
    \item creation of new emerging entities
\end{enumerate*}. Our experimental results showcase a continual temporal decrease in performance of the EL task, with the biggest drop for new entities that require additional world knowledge non-existing during the pre-training phase of the models. 

\end{itemize}
Additionally, this thesis appendices' include other research work published in top-tier journals and conferences not related to the central contribution summarized in \tabref{ch_intro:tab:chap_contributions}. Thus, in \cite{zaporojets2021solving} (see \appendixref{chap:mwp}) we propose to use recursive neural networks to mimic the structure of equation trees to solve mathematical word problems. We showcase a significant improvement using our approach over baselines. Furthermore, in \cite{zaporojets2018predicting} (see \appendixref{chap:clpsych}) we describe our contribution to CLPsych 2018 shared task where we achieve competitive results using an ensemble consisting of multiple models to predict depression and anxiety in textual surveys. 
\newpage
\section{Publications}
\label{ch_intro:sec:publications}
The research 
output
obtained during this PhD 
has been published in scientific journals and presented at a series of international conferences and workshops. The following list provides an overview of these publications.

\renewcommand{\labelenumi}{[\arabic{enumi}]}  

\subsection[Publications in international journals\\(listed in the Science Citation Index)]{Publications in international journals\\
(listed in the Science Citation Index\footnote{The publications listed are recognized as `A1 publications', according to the following definition used by Ghent University: \textit{``A1 publications are articles listed in the Science Citation Index, the Social Science Citation Index or the Arts and Humanities Citation Index of the ISI Web of Science, restricted to contributions listed as article, review, letter, note or proceedings paper.''}}%
)}

\begin{enumerate}
	\item[\cite{zaporojets2021dwie}] 
\begin{sloppypar}
		\textbf{K.~Zaporojets}, J.~Deleu, T.~Demeester,  and C.~Develder,
		\textit{{DWIE}: An entity-centric dataset for multi-task document-level information extraction}.
		Information Processing \& Management.   
		58: 102563, 2021. (\textbf{acceptance rate:}  11\%)
\end{sloppypar}
	\item[\cite{zaporojets2021solving}] 
\begin{sloppypar}
		\textbf{K.~Zaporojets}, G.~Bekoulis, J.~Deleu, T.~Demeester,  and C.~Develder,
		\textit{{S}olving Arithmetic Word Problems by Scoring Equations with Recursive Neural Networks}.
		Expert Systems with Applications.   
		174: 114704, 2021. (\textbf{acceptance rate:}  12\%)
\end{sloppypar}

\end{enumerate}

\subsection[Publications in international conferences]{Publications in international conferences}
\begin{enumerate}[I]
 \setcounter{enumi}{2}
\item[\cite{zaporojets2022tempel}] 
\begin{sloppypar}
		\textbf{K.~Zaporojets}, LA.~Kaffee, J. Deleu, T.~Demeester, C.~Develder, I.~Augenstein,
		\textit{TempEL: A Dataset to Evaluate Temporal Effect on Entity Linking Task}.
		2022 Conference on Neural Information Processing Systems Datasets and Benchmarks Track: NeurIPS, 2022. 
\end{sloppypar}
\item[\cite{zaporojets2021towards}]
\begin{sloppypar}
		\textbf{K.~Zaporojets}, J.~Deleu, Y.~Jiang, T.~Demeester,  and C.~Develder,
		\textit{{T}owards Consistent Document-level Entity Linking: Joint Models for Entity Linking and Coreference Resolution}.
		2022 Conference on the Association for Computational Linguistics: ACL, 2022. pp. 778-784. \textbf{acceptance rate:} 25.2\%
\end{sloppypar} 
\item[\cite{verlinden2021injecting}]
\begin{sloppypar}
		\textbf{S.~Verlinden}$^\mathrm{*}$, \textbf{K.~Zaporojets}$^\mathrm{*}$, J.~Deleu, T.~Demeester,  and C.~Develder,
		\textit{Injecting Knowledge Base Information into End-to-End Joint Entity and Relation Extraction and Coreference Resolution}.
	    Findings of the Association for Computational Linguistics: ACL-IJCNLP, 2021. pp. 1952-1957. \textbf{acceptance rate:} 34.9\%
		 
		 * Equal contribution
\end{sloppypar}
\item[\cite{zaporojets2018predicting}]
\begin{sloppypar}
		\textbf{K.~Zaporojets}, L, Sterckx, J. Deleu, T.~Demeester, C.~Develder,
		\textit{Predicting psychological health from childhood essays: the UGent-IDLab CLPsych 2018 shared task system.}
		5th Annual Workshop on Computer Linguistics and Clinical Psychology (CLPsych 2018) at NAACL-HLT 2018. pp. 119-125.
\end{sloppypar}
\end{enumerate}

\subsection{Publications in international journals and conferences (not included in this thesis)}
\begin{enumerate}[I]
 \setcounter{enumi}{6}

\item[\cite{bitew2019predicting}]
\begin{sloppypar}
		S.~Bitew, G.~Bekoulis, J.~Deleu, L.~Sterckx, \textbf{K.~Zaporojets}, T.~Demeester, and C.~Develder,
		\textit{{P}redicting Suicide Risk from Online Postings in Reddit --
The UGent-IDLab submission to the CLPysch 2019 Shared Task A}.
		6th Ann. Workshop on Computational Linguistics and Clinical Psychology (CLPsych 2019) at NAACL-HLT,   
		  2019. pp. 158-161.
\end{sloppypar}
\item[\cite{jiang2020recipe}]
\begin{sloppypar}
		Y.~Jiang, \textbf{K.~Zaporojets}, J.~Deleu, T.~Demeester, and C.~Develder,
		\textit{{R}ecipe instruction semantics corpus ({RISeC}): resolving semantic structure and zero anaphora in recipes}.
		2020 Conference of the Asia-Pacific Chapter of the Association Computational Linguistics and 10th International Joint Conference on Natural Language Processing: AACL-IJCNLP 2020. pp. 821-826. 
\end{sloppypar}
\item[\cite{jiang2022cookdial}]
\begin{sloppypar}
		Y.~Jiang, \textbf{K.~Zaporojets}, J.~Deleu, T.~Demeester, and C.~Develder,
		\textit{{C}ook{D}ial: A dataset for task-oriented dialogs grounded in procedural documents}.
		Applied Intelligence Journal 2022. pp. 1-19. 
\end{sloppypar}
%
%
\end{enumerate}




\renewcommand{\labelenumi}{\arabic{enumi}.} 

\clearpage

\renewcommand*{\thesection}{\thechapter.\arabic{section}}       



\clearpage{\pagestyle{empty}\cleardoublepage}

\graphicspath{{klim_ch_dwie/figures/}}

\newcommand*\linenomathpatch[1]{%
  \cspreto{#1}{\linenomath}%
  \cspreto{#1*}{\linenomath}%
  \cspreto{end#1}{\endlinenomath}%
  \cspreto{end#1*}{\endlinenomath}%
}
\linenomathpatch{equation}
\linenomathpatch{gather}
\linenomathpatch{align}

\sisetup{output-exponent-marker=\ensuremath{\mathrm{e}}}


\usetikzlibrary{calc}
\pgfplotsset{/pgfplots/error bars/error bar style={thick}}  
\pgfplotsset{compat = 1.3}

\definecolor{darkspringgreen}{rgb}{0.09, 0.45, 0.27}
\colorlet{presentattr}{darkspringgreen!40}
\colorlet{ourdataset}{darkspringgreen!30}

\definecolor{deepcarmine}{rgb}{0.66, 0.13, 0.24}
\colorlet{absentattr}{deepcarmine!40}

\definecolor{Gainsboro}{rgb}{0.86,0.86,0.86}
\colorlet{graycell}{Gainsboro}

\algnewcommand{\LeftComment}[1]{\Statex \(\triangleright\) #1}

\newcommand{\cf}{cf.\ }
\newcommand{\vs}{vs.\ }

\newcommand{\topicname}[1]{\textbf{#1}}
\newcommand{\entityname}[1]{\textbf{#1}}

\newcommand{\figsref}[2]{Figs.~\ref{#1}--\ref{#2}} 
\newcommand{\Figsref}[2]{Figures~\ref{#1}--\ref{#2}}    
\newcommand{\Algref}[1]{Algorithm~\ref{#1}}
\newcommand{\eqsref}[2]{Eq.~(\ref{#1})--(\ref{#2})}
\newcommand{\appref}[1]{\ref{#1}} 

\newcommand{\relation}[3]{$\textsf{#1}\langle\textit{#2},$ $\textit{#3}\rangle$}

\newcommand{\relationalign}[3]{\textsf{#1}\langle\textit{#2}, \textit{#3}\rangle}
\newcommand{\relationalignsingle}[2]{\textsf{#1}\langle\textit{#2}\rangle}

\newcommand{\boldpartitle}[1]{\noindent \textbf{#1} ---}

\urlstyle{sf}
\makeatletter
\let\UrlSpecialsOld\UrlSpecials
\def\UrlSpecials{\UrlSpecialsOld\do\/{\Url@slash}\do\_{\Url@underscore}}%
\def\Url@slash{\@ifnextchar/{\kern-.11em\mathchar47\kern-.2em}%
    {\kern-.0em\mathchar47\kern-.08em\penalty\UrlBigBreakPenalty}}
\def\Url@underscore{\nfss@text{\leavevmode \kern.06em\vbox{\hrule\@width.3em}}}
\makeatother

\widowpenalty100000
\clubpenalty100000

\hyphenation{}

\chapter[DWIE: an Entity-Centric Dataset for Multi-Task Document-Level Information Extraction
]{DWIE: an Entity-Centric Dataset for Multi-Task Document-Level Information Extraction}
\label{chap:dwie}

\renewcommand\evenpagerightmark{{\scshape\small Chapter \arabic{chapter}}}
\renewcommand\oddpageleftmark{{\scshape\small DWIE: an Entity-Centric Dataset}}

\renewcommand{\bibname}{References}

\begin{flushright}
\end{flushright}

\noindent 

\emph{In this chapter we introduce `Deutsche Welle corpus for Information Extraction', 
a newly created 
multi-task dataset that combines four main Information Extraction (IE) annotation subtasks:
\begin{enumerate*}[(i)]
\item Named Entity Recognition (NER),
\item Coreference Resolution,
\item Relation Extraction (RE), and
\item Entity Linking.
\end{enumerate*} Furthermore, we propose a radically different, \textit{entity-centric} view on the information in text. We argue that, instead of using individual mentions in text to understand their meaning, we need to build applications that would operate in terms of entity concepts. This approach has the advantage of leveraging the information across all the entity mentions referring to a single entity in the document at once. As a consequence, all the annotations on \datasetname~are done on coreference concept level. Each of the concepts can group one or more mentions that refer to the same entity in the Knowledge Base. We further demonstrate how these tasks complement each other in a joint information extraction model.}

\begin{center}
\par{$\star\star\star$}
\end{center}
\vspace{0.15in}

\par{\noindent\large{\textbf{K.~Zaporojets, J.~Deleu, C.~Develder and T.~Demeester}}}
\vspace{0.1in}
\par{\noindent\textbf{Information Processing \& Management, 2021.}}
\vspace{0.15in}

\par{\noindent\bf{Abstract}}
This paper presents \datasetname, the `Deutsche Welle corpus for Information Extraction', 
a newly created 
multi-task dataset that combines four main Information Extraction (IE) annotation subtasks:
\begin{enumerate*}[(i)]
\item Named Entity Recognition (NER),
\item Coreference Resolution,
\item Relation Extraction (RE), and
\item Entity Linking.
\end{enumerate*}
DWIE is conceived as an \emph{entity-centric} dataset that describes interactions and properties of conceptual entities on %
the level of the complete document.
This contrasts with currently dominant \emph{mention-driven} approaches 
that start from the detection and classification of 
named entity mentions in
individual sentences.
Further, {\datasetname} presented two main challenges when building and evaluating IE models for it. \textit{First}, the use of traditional 
mention-level 
evaluation metrics for NER and RE tasks on entity-centric \datasetname~dataset can result in measurements dominated by predictions on more frequently mentioned entities. We tackle this issue by proposing a new entity-driven metric that takes into account the number of mentions that compose each of the predicted and ground truth entities. 
\textit{Second}, the document-level multi-task annotations require the models to transfer information between entity mentions located in different parts of the document, as well as between different tasks, in a joint learning setting. To realize this, we propose to use graph-based neural message passing techniques between document-level mention spans. Our experiments show an improvement of up to 5.5 F$_\text{1}$ percentage points when incorporating neural graph propagation into our joint model. 
This demonstrates \datasetname's potential to stimulate further research 
in graph neural networks for  representation learning in multi-task IE. 
We make DWIE publicly available at \url{https://github.com/klimzaporojets/DWIE}.




\section{Introduction}
\label{ch_dwie:sec:introduction}
Information Extraction (IE) plays a fundamental role as a backbone component in many downstream applications. For example, 
an application
such as question answering
may be improved by relying on relation extraction (RE) \cite{yu2017improved,hu2019language}, coreference resolution \cite{gao2019interconnected,bhattacharjee2020investigating}, named entity recognition (NER) \cite{molla2006named,singh2018reinvent}, and entity linking (EL) 
\cite{chen2017reading,broscheit2019investigating} 
 components. 
This also holds for other applications such as
personalized news recommendation \cite{wang2018dkn,karimi2018news,wang2019multi}, 
fact checking \cite{thorne2018automated,zhang2020overview}, 
opinion mining \cite{sun2017review}, 
semantic search \cite{cifariello2019wiser}, and 
conversational agents \cite{roller2020recipes}. 
The last decade has shown a growing interest in IE datasets suitably annotated for
developing multi-task models where each of the tasks (\eg NER, RE, etc.) would benefit from the interaction with (an)other task(s) \cite{bekoulis2018joint,fei2020boundaries,lee2017end,lee2018higher,luan2019general}, to boost their performance.
However, the currently widely used IE datasets to build such multi-task models exhibit three major limitations. 
\textit{First}, the annotation schema adopted in most of 
these datasets
is mention-driven, focusing on annotating elements (\eg relations, entity types) that involve specific entity mentions explicitly mentioned in the text. 
This produces very localized annotations (\eg sentence-based relations between entity mentions) that do not reflect meaning that can be inferred on a more general 
document-level.
\textit{Second}, the number of annotated extraction tasks in most of the IE datasets is rather limited. Most of them focus on a single or at most a few different tasks. 
Furthermore, some other datasets, including the well-known TAC-KBPs \cite{ji2010overview,ellis2014overview,ji2015overview,ellis2015overview,ji2017overview}, use different non-overlapping corpora for each of the tracks that group a few related tasks. Consequently, current models addressing multiple IE tasks together often use multi-tasking (with different datasets per task) rather than really joint modeling approaches.
\textit{Finally}, the annotation of currently widely used IE datasets is driven by either relying on a priori defined annotation schemas \cite{doddington2004automatic, walker2006ace,song2015light,augenstein2017semeval,zhang2017position,hendrickx2010semeval} or on distantly supervised labeling techniques \cite{han2018fewrel,yao2019docred,riedel2010modeling,quirk2017distant,peng2017cross}. In consequence, the resulting annotations
are not necessarily representative of the actual information contained in the 
annotated 
corpus. 
\begin{figure}[t]
\centering
\includegraphics[width=1.0\columnwidth, trim={0cm 0cm 0cm 0cm},clip]{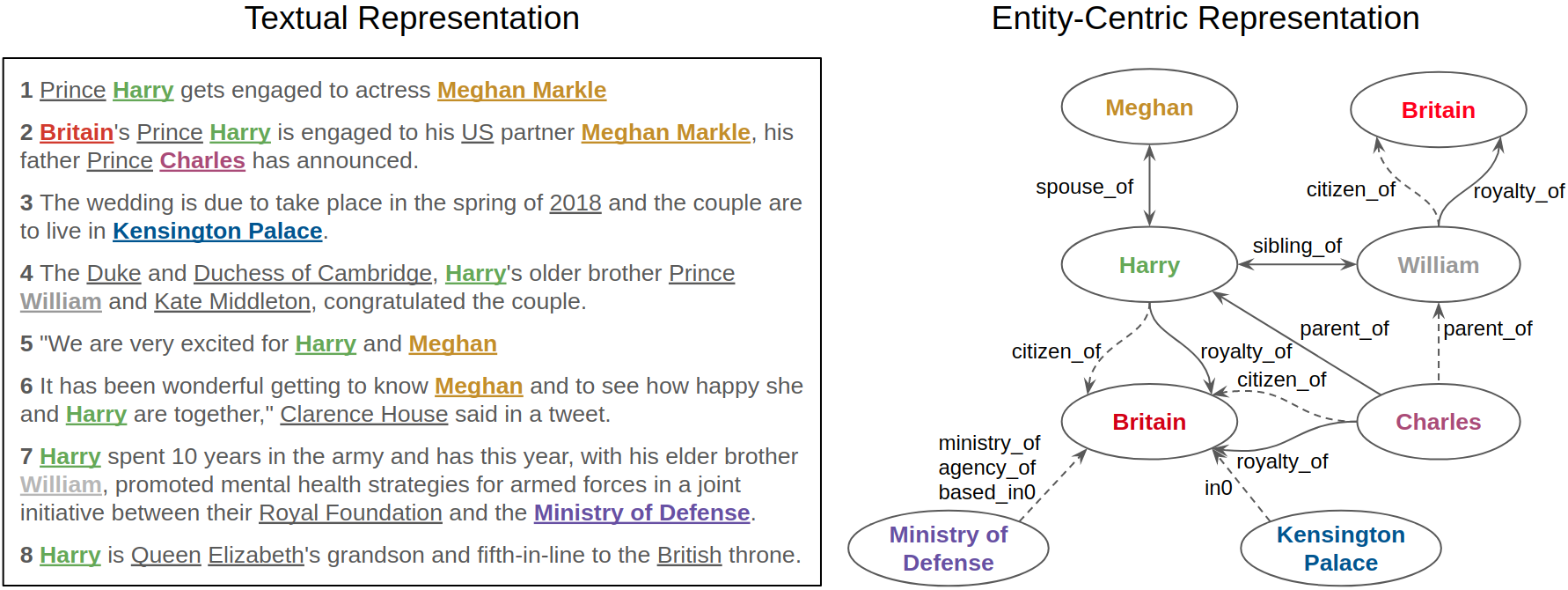}
\captionsetup{singlelinecheck=off}
\caption[Example from the \datasetname]{An example from the \datasetname~dataset with entity mentions underlined.
We show 8 of the 29 entities in the graph on the right.
It illustrates the relations that can be derived from the content of the article. The relations that are explicitly mentioned in the text (trigger-based) are depicted by solid arrows. Conversely, the relations that are implicit and/or need the whole document context (document-based) to be derived are represented by dashed arrows.}
\label{ch_dwie:fig:example}
\end{figure}

In this work, we tackle the aforementioned limitations 
of IE datasets 
by introducing a new 
dataset named DWIE. It consists of 802 general news articles in English, selected randomly from a corpus collected from Deutsche Welle\footnote{\url{https://www.dw.com}} between 2002 and 2018, as part of the CPN project.\footnote{\url{https://www.projectcpn.eu}} We focus on annotating four main IE tasks: \begin{enumerate*}[(i)]
\item Named Entity Recognition (NER),
\item Coreference Resolution,
\item Relation Extraction (RE), and
\item Entity Linking.\footnote{The linking is done to Wikipedia version 20181115.}
\end{enumerate*}
\Figref{ch_dwie:fig:example} 
shows an example snippet from the \datasetname~corpus.
We adopt an \textit{entity-centric} approach where all annotations (\ie for NER, RE and Entity Linking tasks) are made on the entity\footnote{Also referred to as \textit{entity cluster} or just \textit{cluster}.} level. Each of the entities is composed by the coreferenced entity mentions from the entire document (\eg the entity \textit{Meghan} in \figref{ch_dwie:fig:example} clusters the entity mentions ``Meghan Markle'' and ``Meghan'' across the whole document). 
This entity-centric approach contrasts with mention-driven annotations in widely used IE datasets \cite{doddington2004automatic,ji2015overview,song2015light,han2018fewrel,zhang2017position,hendrickx2010semeval,luan2018multi} where the annotation process is 
biased towards 
considering only local 
explicit textual evidence to annotate elements such as relations and entity types (\eg the relation \relation{spouse\_of}{Meghan}{Harry} that can be extracted from the 1st sentence in \figref{ch_dwie:fig:example}). Consequently, our {\datasetname} dataset paves the way for research on more complex document-level reasoning that goes beyond only the local textual context directly surrounding individual entity mentions. For example, consider the relation \relation{ministry\_of}{Ministry of Defense}{Britain} in \figref{ch_dwie:fig:example}: while the text of the document does not directly state such a relation, it can be deduced from a more general document-level entity-centric vision of the article, \ie combining the information involving the entities \textit{Ministry of Defense} and \textit{Harry} in sentence 7 with the one involving \textit{Britain} and \textit{Harry} in sentence 2.
Finally, the entity-centric approach provides entity linking annotations that are consistent across the document: by clustering mentions of the same entity, and then providing links to the Wikidata KB (or NIL if the entity does not appear there) for the whole cluster at once, we limit annotation errors or accidental inconsistencies (in the linking itself, but also in terms of NER labels).
To our knowledge, DWIE is the first dataset with this level of conceptual consistency over the considered information extraction tasks. We therefore expect that the dataset
will play a key role in advancing research exploring potential benefits of
\begin{enumerate*}[(i)]
    \item entity-level information extraction in terms of reducing potential inconsistent decisions (within EL across multiple mentions, as well as across multiple tasks), and
    \item using entity-centric information stored in a KB to complement the 
    otherwise exclusively
    text-dependent IE tasks such as NER, RE, and coreference resolution. 
\end{enumerate*}

Additionally, we use a bottom-up, data-driven annotation approach where we manually define our annotations (\eg in terms of the entity and relation types) to maximally reflect the information of the corpus at hand. 
Currently dominant datasets are driven by distant supervision and executed top-down, by which we mean that the selection of entity and relation types is a priori defined and limited in coverage (\ie the raw data potentially contains other types that thus remain un-annotated).
Conversely, we do not a priori limit the entity and relation types to annotate, but adopt a bottom-up approach driven by the data itself.
Our proposed bottom-up approach encompasses a three-pass annotation procedure where we use the first exploratory annotation pass to derive the main annotation types (annotation schema) from the corpus, and the next two passes to perform schema-driven annotations and refine them by carrying out an additional parallel annotation of the corpus for fixing errors inferred from inter-annotator inconsistencies.

Besides the dataset itself, we also contribute empirical modeling results 
to address the aforementioned IE tasks. 
Our goal is to study two important properties
that are inherent to \datasetname.
The first key property is the need for \textit{long-range contextual information sharing}
to make document-level predictions involving entities whose mentions are located in different parts of the document. 
The second key property 
involves the \textit{joint interaction between tasks} where the information obtained in one task can help to solve another task. For example, in \figref{ch_dwie:fig:example} knowing the types of entities (which involves NER and coreference tasks) \textit{Britain} and \textit{Kensington Palace} can boost the performance of the relation extraction task by limiting the number of possible relation types between these two entities (\eg \textsf{ministry\_of} but not \textsf{citizen\_of}).
In order to study the impact of these two phenomena inherent to our {\datasetname} dataset on the final results, we experiment with neural graph-based models \cite{li2015gated,xu2018powerful,wu2020comprehensive}. These models allow message passing between local contextual encodings, making it possible to measure the impact of local contextual information sharing both on a more general document level and across the tasks.
Furthermore, previous work already has shown the positive effect of using graph-based information passing techniques on single tasks \cite{lee2018higher, kantor2019coreference}, and between tasks \cite{luan2019general, wadden2019entity, fei2020boundaries, fu2019graphrel} on mention-driven datasets. We expand this work even further by extending these models to be used on the entity-centric, document-level {\datasetname} dataset. More specifically, we experiment with both single-task (\secref{ch_dwie:sec:single_task_models}) as well as joint (\secref{ch_dwie:sec:joint_training}) models to study the effect of contextual information propagation in 
single task and
joint settings.
Additionally, for the NER and RE tasks, we propose a new entity-centric evaluation metric that not only considers the predictions on separate entity mentions (as is  done in related IE datasets), but also accounts for the impact of the predictions on entity cluster level.

In summary, the main objective that we address in the current paper is to introduce an entity-centric multi-task IE dataset that covers different related tasks on a document level as well as provides a connection with external structured knowledge (through the entity linking task). Furthermore, we aim to explore how neural graph-based models can boost the performance 
by enabling local contextual information propagation across the document (single-task models) and between different tasks (joint models). The results presented in this paper  
suggest that, while challenging, \datasetname~opens up new possibilities of research in the domain of 
joint entity-centric information extraction methods.
The main contributions of our work are that: 
\begin{enumerate}[(1)]
    \item We construct a self-contained dataset (\secref{ch_dwie:sec:annotation}) with joint annotations for four basic information extraction tasks (NER, entity linking, coreference resolution, and 
    RE), 
    that provide entity-centric document-level annotations (as opposed to typical 
    mention-driven
    sentence-level annotations for, 
    \eg RE) connecting unstructured (text) and structured (KB) information sources. 
    \item We introduce a data-driven, bottom-up three-pass annotation approach complemented by context-based logical rules to build such dataset (\secref{ch_dwie:sec:annotation}).
    \item We propose a new evaluation metric for the NER and RE tasks (\secref{ch_dwie:sec:metrics}), in line with the entity-centric nature of \datasetname. 
    \item We extend the competitive graph-based neural IE model DyGIE \cite{luan2019general}  
    for the four IE tasks in {\datasetname}  
    (\secref{ch_dwie:sec:models}) and provide source code for NER, coreference resolution, and 
    RE. Furthermore, we introduce a new latent attention-driven \propformat{AttProp} graph propagation method and show its advantages in both single and joint model settings.
    The experimental results (\secref{ch_dwie:sec:results}) demonstrate the potential of such neural graph based models. 
\end{enumerate}

\section{Related work}
\label{ch_dwie:sec:related}

This section summarizes the overview of related datasets (\secref{ch_dwie:sec:related_datasets}), and explores the differences between our newly created {\datasetname} and other similar datasets widely used by the scientific community.
The main qualitative differences are presented in \Tabref{ch_dwie:tab:quali_compare}, while the quantitative comparison is provided in \Tabref{ch_dwie:tab:quanti_compare}. 
Next, we describe the current trends 
in IE to solve the tasks included in \datasetname, and compare them to our proposed approach (\secref{ch_dwie:sec:recent_advances}). Finally, we discuss currently used metrics to evaluate model performance on IE datasets and introduce some challenges in applying them to measuring the performance on \datasetname~(\secref{ch_dwie:sec:metrics_and_evaluation}). 

\subsection{Related datasets}
\label{ch_dwie:sec:related_datasets}

Most of IE datasets have focused on a single task, making it very challenging to develop systems that  jointly train for different annotation subtasks on a single corpus. 
Well-known single-task datasets include
\begin{enumerate*}[(i)]
    \item \emph{for NER:} CoNLL-2003 \cite{sang2003introduction} and WNUT 2017 \cite{derczynski2017results},
    \item \emph{for relation extraction:} Semeval-2010 T8 \cite{hendrickx2010semeval}, TACRED \cite{zhang2017position} and FewRel \cite{han2018fewrel},
    \item \emph{for entity linking:} IITB \cite{kulkarni2009collective}, CoNLL-YAGO \cite{hoffart2011robust}, and WikilinksNED \cite{eshel2017named}, and
    \item \emph{for coreference resolution:} CoNLL-2012 \cite{pradhan2012conll} and GAP \cite{webster2018mind}.
\end{enumerate*}
Conversely, in this work we propose a multi-task dataset as a single corpus annotated with different information extraction layers: named entities, mention clustering in entities (\ie coreference), relations between entity clusters of mentions, and entity linking.
We further complement our dataset with additional tasks such as document classification and keyword extraction. It is worth noting that our \textit{coreference} annotations differ from the widely adopted CoNLL-2012 \cite{pradhan2012conll} scheme in two aspects:
\begin{enumerate*}[(i)]
    \item we retain singleton entities composed by only one mention as a valid entity cluster, 
    \item we only cluster proper nouns, leaving out nominal and anaphoric expressions. 
\end{enumerate*}
{\renewcommand\baselinestretch{1}
\begin{table}[!t]
\centering
\caption[Qualitative comparison of the datasets]{Qualitative comparison of the datasets. We divide our comparison in five groups: \protect\begin{enumerate*}[(i)] 
	 \item \textit{Core Tasks} represent the main subtasks covered in \datasetname, 
	 \item \textit{Doc-Based} indicates whether different subtasks are annotated on the document-level,
	 \item \textit{Entity-Centric} indicates which annotations are done with respect to entity clusters (\cmark) as opposed to individual mentions (\xmark), 
	 \item \textit{Unaided} specifies whether the annotation process was completely manual (\cmark) or with some form of distant supervision (\xmark), and 
	 \item \textit{Open} indicates whether the dataset is freely available. 
	 \protect\end{enumerate*}} 
\resizebox{1.0\textwidth}{!}{
\begin{tabular}{l cccc cccc cccc cccc c}
\toprule
& \multicolumn{4}{c}{\textbf{Core Tasks}} & \multicolumn{4}{c}{\textbf{Doc-Based}} & \multicolumn{4}{c}{\textbf{Entity-Centric}} & \multicolumn{4}{c}{\textbf{Unaided}} & \\
\cmidrule(lr){2-5} \cmidrule(lr){6-9}\cmidrule(lr){10-13}\cmidrule(lr){14-17}
\multicolumn{1}{c}{\textbf{Dataset}} & \rotatebox{90}{\textbf{NER}} & \rotatebox{90}{\textbf{Coreference}} & \rotatebox{90}{\textbf{Relations}} & \rotatebox{90}{\textbf{Linking}} & \rotatebox{90}{\textbf{Coreference}} & \rotatebox{90}{\textbf{Relations}} & \rotatebox{90}{\textbf{Multi-label Rel}} & \rotatebox{90}{\textbf{Keywords}} & \rotatebox{90}{\textbf{Classification}} & \rotatebox{90}{\textbf{Multi-label Ent}} & \rotatebox{90}{\textbf{Relations}} & \rotatebox{90}{\textbf{Linking}} & \rotatebox{90}{\textbf{NER}} & \rotatebox{90}{\textbf{Coreference}} & \rotatebox{90}{\textbf{Relations}} & \rotatebox{90}{\textbf{Linking}} & \rotatebox{90}{\textbf{Open}}\\
\midrule
\textbf{\datasetname} & \cellcolor{presentattr}\cmark & \cellcolor{presentattr}\cmark & \cellcolor{presentattr}\cmark & \cellcolor{presentattr}\cmark & \cellcolor{presentattr}\cmark & \cellcolor{presentattr}\cmark & \cellcolor{presentattr}\cmark & \cellcolor{presentattr}\cmark & \cellcolor{presentattr}\cmark & \cellcolor{presentattr}\cmark & \cellcolor{presentattr}\cmark & \cellcolor{presentattr}\cmark & \cellcolor{presentattr}\cmark & \cellcolor{presentattr}\cmark & \cellcolor{presentattr}\cmark & \cellcolor{presentattr}\cmark & \cellcolor{presentattr}\cmark\\
TAC-KBP~\cite{ji2010overview,ji2015overview,ji2017overview} & \cellcolor{presentattr}\cmark & \cellcolor{presentattr}\cmark & \cellcolor{presentattr}\cmark & \cellcolor{presentattr}\cmark & \cellcolor{presentattr}\cmark & \cellcolor{absentattr}\xmark & \cellcolor{absentattr}\xmark & \cellcolor{absentattr}\xmark & \cellcolor{presentattr}\cmark & \cellcolor{absentattr}\xmark & \cellcolor{absentattr}\xmark & \cellcolor{presentattr}\cmark & \cellcolor{presentattr}\cmark & \cellcolor{presentattr}\cmark & \cellcolor{presentattr}\cmark & \cellcolor{presentattr}\cmark & \cellcolor{absentattr}\xmark\\
BC5CDR~\cite{li2016biocreative,wei2015overview} & \cellcolor{presentattr}\cmark & \cellcolor{presentattr}\cmark & \cellcolor{presentattr}\cmark & \cellcolor{presentattr}\cmark & \cellcolor{presentattr}\cmark & \cellcolor{presentattr}\cmark & \cellcolor{absentattr}\xmark & \cellcolor{absentattr}\xmark & \cellcolor{absentattr}\xmark & \cellcolor{absentattr}\xmark & \cellcolor{presentattr}\cmark & \cellcolor{absentattr}\xmark & \cellcolor{presentattr}\cmark & \cellcolor{presentattr}\cmark & \cellcolor{absentattr}\xmark & \cellcolor{presentattr}\cmark & \cellcolor{presentattr}\cmark\\
MUC-7~\cite{chinchor1998muc} & \cellcolor{presentattr}\cmark & \cellcolor{presentattr}\cmark & \cellcolor{presentattr}\cmark & \cellcolor{absentattr}\xmark & \cellcolor{presentattr}\cmark & \cellcolor{presentattr}\cmark & \cellcolor{absentattr}\xmark & \cellcolor{absentattr}\xmark & \cellcolor{presentattr}\cmark & \cellcolor{absentattr}\xmark & \cellcolor{presentattr}\cmark & \cellcolor{absentattr}\xmark & \cellcolor{presentattr}\cmark & \cellcolor{presentattr}\cmark & \cellcolor{presentattr}\cmark & \cellcolor{absentattr}\xmark & \cellcolor{absentattr}\xmark\\
SciERC~\cite{luan2018multi} & \cellcolor{presentattr}\cmark & \cellcolor{presentattr}\cmark & \cellcolor{presentattr}\cmark & \cellcolor{absentattr}\xmark & \cellcolor{presentattr}\cmark & \cellcolor{absentattr}\xmark & \cellcolor{absentattr}\xmark & \cellcolor{absentattr}\xmark & \cellcolor{presentattr}\cmark & \cellcolor{absentattr}\xmark & \cellcolor{absentattr}\xmark & \cellcolor{absentattr}\xmark & \cellcolor{presentattr}\cmark & \cellcolor{presentattr}\cmark & \cellcolor{presentattr}\cmark & \cellcolor{presentattr}\cmark & \cellcolor{presentattr}\cmark\\
DocRED~\cite{yao2019docred} & \cellcolor{presentattr}\cmark & \cellcolor{presentattr}\cmark & \cellcolor{presentattr}\cmark & \cellcolor{absentattr}\xmark & \cellcolor{presentattr}\cmark & \cellcolor{presentattr}\cmark & \cellcolor{presentattr}\cmark & \cellcolor{absentattr}\xmark & \cellcolor{presentattr}\cmark & \cellcolor{absentattr}\xmark & \cellcolor{presentattr}\cmark & \cellcolor{absentattr}\xmark & \cellcolor{absentattr}\xmark & \cellcolor{absentattr}\xmark & \cellcolor{absentattr}\xmark & \cellcolor{absentattr}\xmark & \cellcolor{presentattr}\cmark\\
Rich ERE~\cite{song2015light,aguilar2014comparison} & \cellcolor{presentattr}\cmark & \cellcolor{presentattr}\cmark & \cellcolor{presentattr}\cmark & \cellcolor{absentattr}\xmark & \cellcolor{presentattr}\cmark & \cellcolor{absentattr}\xmark & \cellcolor{absentattr}\xmark & \cellcolor{absentattr}\xmark & \cellcolor{presentattr}\cmark & \cellcolor{absentattr}\xmark & \cellcolor{absentattr}\xmark & \cellcolor{absentattr}\xmark & \cellcolor{presentattr}\cmark & \cellcolor{presentattr}\cmark & \cellcolor{presentattr}\cmark & \cellcolor{absentattr}\xmark & \cellcolor{absentattr}\xmark\\
ACE 2005~\cite{walker2006ace} & \cellcolor{presentattr}\cmark & \cellcolor{presentattr}\cmark & \cellcolor{presentattr}\cmark & \cellcolor{absentattr}\xmark & \cellcolor{absentattr}\xmark & \cellcolor{absentattr}\xmark & \cellcolor{absentattr}\xmark & \cellcolor{absentattr}\xmark & \cellcolor{presentattr}\cmark & \cellcolor{absentattr}\xmark & \cellcolor{absentattr}\xmark & \cellcolor{absentattr}\xmark & \cellcolor{presentattr}\cmark & \cellcolor{presentattr}\cmark & \cellcolor{presentattr}\cmark & \cellcolor{absentattr}\xmark & \cellcolor{absentattr}\xmark\\
OntoNotes 5.0~\cite{hovy2006ontonotes,weischedel2013ontonotes} & \cellcolor{presentattr}\cmark & \cellcolor{presentattr}\cmark & \cellcolor{absentattr}\xmark & \cellcolor{absentattr}\xmark & \cellcolor{absentattr}\xmark & \cellcolor{absentattr}\xmark & \cellcolor{absentattr}\xmark & \cellcolor{absentattr}\xmark & \cellcolor{absentattr}\xmark & \cellcolor{absentattr}\xmark & \cellcolor{absentattr}\xmark & \cellcolor{absentattr}\xmark & \cellcolor{presentattr}\cmark & \cellcolor{presentattr}\cmark & \cellcolor{presentattr}\cmark & \cellcolor{presentattr}\cmark & \cellcolor{presentattr}\cmark\\
ScienceIE~\cite{augenstein2017semeval} & \cellcolor{presentattr}\cmark & \cellcolor{absentattr}\xmark & \cellcolor{presentattr}\cmark & \cellcolor{absentattr}\xmark & \cellcolor{absentattr}\xmark & \cellcolor{presentattr}\cmark & \cellcolor{absentattr}\xmark & \cellcolor{presentattr}\cmark & \cellcolor{absentattr}\xmark & \cellcolor{absentattr}\xmark & \cellcolor{absentattr}\xmark & \cellcolor{absentattr}\xmark & \cellcolor{presentattr}\cmark & \cellcolor{absentattr}\xmark & \cellcolor{presentattr}\cmark & \cellcolor{absentattr}\xmark & \cellcolor{presentattr}\cmark\\
FewRel~\cite{han2018fewrel} & \cellcolor{presentattr}\cmark & \cellcolor{presentattr}\cmark & \cellcolor{presentattr}\cmark & \cellcolor{presentattr}\cmark & \cellcolor{absentattr}\xmark & \cellcolor{absentattr}\xmark & \cellcolor{absentattr}\xmark & \cellcolor{absentattr}\xmark & \cellcolor{absentattr}\xmark & \cellcolor{absentattr}\xmark & \cellcolor{presentattr}\cmark & \cellcolor{presentattr}\cmark & \cellcolor{absentattr}\xmark & \cellcolor{absentattr}\xmark & \cellcolor{absentattr}\xmark & \cellcolor{absentattr}\xmark & \cellcolor{presentattr}\cmark\\
GENIA~\cite{kim2003genia} & \cellcolor{presentattr}\cmark & \cellcolor{presentattr}\cmark & \cellcolor{presentattr}\cmark & \cellcolor{absentattr}\xmark & \cellcolor{presentattr}\cmark & \cellcolor{absentattr}\xmark & \cellcolor{absentattr}\xmark & \cellcolor{absentattr}\xmark & \cellcolor{absentattr}\xmark & \cellcolor{absentattr}\xmark & \cellcolor{absentattr}\xmark & \cellcolor{absentattr}\xmark & \cellcolor{presentattr}\cmark & \cellcolor{absentattr}\xmark & \cellcolor{absentattr}\xmark & \cellcolor{absentattr}\xmark & \cellcolor{presentattr}\cmark\\
AIDA CoNLL-YAGO~\cite{hoffart2011robust} & \cellcolor{presentattr}\cmark & \cellcolor{absentattr}\xmark & \cellcolor{absentattr}\xmark & \cellcolor{presentattr}\cmark & \cellcolor{absentattr}\xmark & \cellcolor{absentattr}\xmark & \cellcolor{absentattr}\xmark & \cellcolor{absentattr}\xmark & \cellcolor{absentattr}\xmark & \cellcolor{absentattr}\xmark & \cellcolor{absentattr}\xmark & \cellcolor{absentattr}\xmark & \cellcolor{presentattr}\cmark & \cellcolor{absentattr}\xmark & \cellcolor{absentattr}\xmark & \cellcolor{presentattr}\cmark & \cellcolor{presentattr}\cmark\\
SemEval 2010 T8~\cite{hendrickx2010semeval} & \cellcolor{presentattr}\cmark & \cellcolor{absentattr}\xmark & \cellcolor{presentattr}\cmark & \cellcolor{absentattr}\xmark & \cellcolor{absentattr}\xmark & \cellcolor{absentattr}\xmark & \cellcolor{absentattr}\xmark & \cellcolor{absentattr}\xmark & \cellcolor{absentattr}\xmark & \cellcolor{absentattr}\xmark & \cellcolor{absentattr}\xmark & \cellcolor{absentattr}\xmark & \cellcolor{presentattr}\cmark & \cellcolor{absentattr}\xmark & \cellcolor{presentattr}\cmark & \cellcolor{absentattr}\xmark & \cellcolor{presentattr}\cmark\\
NYT~\cite{riedel2010modeling} & \cellcolor{presentattr}\cmark & \cellcolor{absentattr}\xmark & \cellcolor{presentattr}\cmark & \cellcolor{presentattr}\cmark & \cellcolor{absentattr}\xmark & \cellcolor{absentattr}\xmark & \cellcolor{absentattr}\xmark & \cellcolor{absentattr}\xmark & \cellcolor{absentattr}\xmark & \cellcolor{absentattr}\xmark & \cellcolor{absentattr}\xmark & \cellcolor{absentattr}\xmark & \cellcolor{absentattr}\xmark & \cellcolor{absentattr}\xmark & \cellcolor{absentattr}\xmark & \cellcolor{absentattr}\xmark & \cellcolor{presentattr}\cmark\\
ACEtoWiki~\cite{bentivogli2010extending} & \cellcolor{absentattr}\xmark & \cellcolor{absentattr}\xmark & \cellcolor{absentattr}\xmark & \cellcolor{presentattr}\cmark & \cellcolor{absentattr}\xmark & \cellcolor{absentattr}\xmark & \cellcolor{absentattr}\xmark & \cellcolor{absentattr}\xmark & \cellcolor{absentattr}\xmark & \cellcolor{absentattr}\xmark & \cellcolor{absentattr}\xmark & \cellcolor{absentattr}\xmark & \cellcolor{absentattr}\xmark & \cellcolor{absentattr}\xmark & \cellcolor{absentattr}\xmark & \cellcolor{presentattr}\cmark & \cellcolor{presentattr}\cmark\\
WNUT 2017~\cite{derczynski2017results} & \cellcolor{presentattr}\cmark & \cellcolor{absentattr}\xmark & \cellcolor{absentattr}\xmark & \cellcolor{absentattr}\xmark & \cellcolor{absentattr}\xmark & \cellcolor{absentattr}\xmark & \cellcolor{absentattr}\xmark & \cellcolor{absentattr}\xmark & \cellcolor{absentattr}\xmark & \cellcolor{absentattr}\xmark & \cellcolor{absentattr}\xmark & \cellcolor{absentattr}\xmark & \cellcolor{presentattr}\cmark & \cellcolor{absentattr}\xmark & \cellcolor{absentattr}\xmark & \cellcolor{absentattr}\xmark & \cellcolor{presentattr}\cmark\\
CoNLL-2003~\cite{sang2003introduction} & \cellcolor{presentattr}\cmark & \cellcolor{absentattr}\xmark & \cellcolor{absentattr}\xmark & \cellcolor{absentattr}\xmark & \cellcolor{absentattr}\xmark & \cellcolor{absentattr}\xmark & \cellcolor{absentattr}\xmark & \cellcolor{absentattr}\xmark & \cellcolor{absentattr}\xmark & \cellcolor{absentattr}\xmark & \cellcolor{absentattr}\xmark & \cellcolor{absentattr}\xmark & \cellcolor{presentattr}\cmark & \cellcolor{absentattr}\xmark & \cellcolor{absentattr}\xmark & \cellcolor{absentattr}\xmark & \cellcolor{presentattr}\cmark\\
TACRED~\cite{zhang2017position} & \cellcolor{absentattr}\xmark & \cellcolor{absentattr}\xmark & \cellcolor{presentattr}\cmark & \cellcolor{absentattr}\xmark & \cellcolor{absentattr}\xmark & \cellcolor{absentattr}\xmark & \cellcolor{absentattr}\xmark & \cellcolor{absentattr}\xmark & \cellcolor{absentattr}\xmark & \cellcolor{absentattr}\xmark & \cellcolor{absentattr}\xmark & \cellcolor{absentattr}\xmark & \cellcolor{absentattr}\xmark & \cellcolor{absentattr}\xmark & \cellcolor{presentattr}\cmark & \cellcolor{absentattr}\xmark & \cellcolor{absentattr}\xmark\\
\bottomrule
\end{tabular}
}
\label{ch_dwie:tab:quali_compare}\end{table}}

Furthermore, most prominent efforts to produce jointly annotated datasets have focused on using a \emph{top-down} annotation approach. This method involves an a priori defined annotation schema that drives the process of selection and labeling of the corpus. 
The de facto datasets used in most of the joint learning baselines such as ACE 2005 
\cite{doddington2004automatic, walker2006ace}, 
TAC-KBPs \cite{ji2010overview,ellis2014overview,ji2015overview,ellis2015overview,ji2017overview} and Rich ERE \cite{song2015light} 
use this annotation approach. More specifically, during the creation of the ACE 2005 dataset, 
the annotators initially tagged candidate documents as ``good'' or ``bad'' depending on the estimated number and types of entities present in each one. In subsequent annotation stages, only ``good'' documents were fully annotated and included in the final dataset. Similarly, during the creation of the TAC-KBP 
datasets, the annotators focused on producing 
annotations evenly distributed among three entity types (PERs, ORGs, and GPEs) by annotating only the documents that contained a minimum number of entities related to event types. In the case of Rich ERE, 
the documents to tag were prioritized by the event trigger word density calculated per 1,000 tokens, thus focusing only on content with a high number of previously defined key event-related tokens. Furthermore, other IE-related datasets \cite{augenstein2017semeval,han2018fewrel,yao2019docred,zhang2017position,hendrickx2010semeval} use similar pre-filtering techniques in order to select the text to be annotated.
As a consequence, the corpus and annotations in these datasets tend to be biased and likely not representative of the language used in the different input domains. Conversely, we adopt a radically different \textit{bottom-up} approach where we derive the annotations (\eg entity classification types, relation types) from the data itself. 
This bottom-up data-driven procedure guarantees that the annotations in {\datasetname} are representative of the document corpus information and reflects the particularities of the language used in its journalistic domain. Furthermore, it better represents the properties that are inherently present in written corpora, \eg the long-tail distribution of different annotation types.
\setlength{\tabcolsep}{5pt}
\renewcommand{\arraystretch}{1.0}
{\renewcommand\baselinestretch{1}\begin{table}[t]
\centering
\caption[Numerical comparison of \datasetname~and well-known IE datasets]{Numerical comparison of \datasetname~and well-known IE datasets. Note that some datasets (including \datasetname) use an entity-centric approach, organizing entity mentions in entity clusters, and annotating entities, relations, and linking on the cluster level. Hence, we provide both mention-level as well as cluster-level (if a particular dataset supports it) statistics.}
\resizebox{1.0\textwidth}{!}{
\begin{tabular}{llllllllll} 
\toprule
& & \multicolumn{3}{c}{\textbf{Entities}} & \multicolumn{3}{c}{\textbf{Relations}} & \multicolumn{2}{c}{\textbf{Linking}} \\ 
\cmidrule(lr){3-5} \cmidrule(lr){6-8}\cmidrule(lr){9-10} 
Dataset & \# Tokens & \# Mentions & \# Entity & \# Entity & \# Relation & \# Relation & \# Relation & \# Mention & \# Cluster\\
 &  &  & clusters & types & mentions & clusters & types & KB Links & KB Links\\
\midrule
NYT & 5,765,332 & 1,388,982 & - & - & 142,823 & - & 52 & 1,388,982 & -\\
TACRED & 3,866,863 & - & - & - & 21,784 & - & 42 & - & -\\
TAC-KBP\tablefootnote{The EDL track only of TAC-KBP 2010.} & 3,053,336 & 6,495 & 3,750 & - & - & - & - & 3,818 & 2,094\\
OntoNotes 5.0 & 2,088,832 & 161,783 & 136,037 & - & - & - & - & - & -\\
FewRel\tablefootnote{Numbers based on publicly available train and development sets.} & 1,397,333 & 114,213 & 112,000 & - & 58,267 & 56,000 & 80 & 114,213 & 112,000\\
DocRED & 1,018,297 & 132,392 & 98,610 & 6 & 155,535 & 50,503 & 96 & -  & -\\
MUC-4 & 717,798 & 14,196 & - & 13 & - & - & - & - & -\\
GENIA & 554,346 & 56,743 & 10,728 & 5 & 2,337 & - & 2 & - & -\\
\cellcolor{ourdataset}{\datasetname} & \cellcolor{ourdataset}501,095 & \cellcolor{ourdataset}43,373 & \cellcolor{ourdataset}23,130 & \cellcolor{ourdataset}311 & \cellcolor{ourdataset}317,204 & \cellcolor{ourdataset}21,749 & \cellcolor{ourdataset}65 & \cellcolor{ourdataset}28,482 & \cellcolor{ourdataset}13,086\\
BC5CDR & 343,175 & 29,271 & 10,326 & 2 & 47,813 & 3,116 & 1 & 29,562 & 10,326\\
CoNLL-2003 & 301,418 & 35,089 & - & 4 & - & - & - & - & -\\
CoNLL-YAGO & 301,418 & 34,929 & - & - & - & - & - & 34,929 & -\\
ACE 2005 & 259,889 & 54,824 & 37,622 & 51 & 8,419 & 7,786 & 18 & - & -\\
ACEtoWiki & 259,889 & - & - & - & - & - & - & 16,310 & -\\
SEval 2010 T8 & 207,307 & 21,434 & - & - & 6,674 & - & 9 & - & -\\
ACE 2004 & 185,696 & 29,949 & 12,507 & 43 & 5,976 & 5,525 & 24 & - & -\\
WNUT 2017 & 101,857 & 3,890 & - & 6 & - & - & - & - & -\\
ScienceIE & 99,580 & 9,946 & 9,536 & 3 & 638 & - & 1 & - & -\\
SciERC & 65,334 & 8,094 & 1,015 & 6 & 2,687 & - & 7 & - & -\\
\bottomrule
\end{tabular}
}
\label{ch_dwie:tab:quanti_compare}
\end{table}}
\setlength{\tabcolsep}{5pt}

Finally, from the perspective of the necessary evidence to annotate a particular entity type or relation, we propose to make a distinction for the currently existing datasets between \textit{trigger-based} 
and 
\textit{document-based} 
annotations (see \textit{Doc-Based} comparison group in \Tabref{ch_dwie:tab:quali_compare}).  
The \textit{trigger-based} datasets require that a particular relation or entity 
type should only be annotated if it is supported by an explicit reference in a text. For example, in \figref{ch_dwie:fig:example} there is a concrete reference of the relation between ``Meghan'' and ``Harry'' in form of triggers such as ``gets engaged'' in sentence~1 and ``The wedding'' in sentence~2. Most of the traditionally used jointly annotated datasets such as ACE 2005 \cite{doddington2004automatic,walker2006ace}, TAC-KBPs \cite{ji2010overview,ellis2014overview,ji2015overview,ellis2015overview,ji2017overview} and Rich ERE \cite{song2015light}, as well as others, including FewRel \cite{han2018fewrel}, OntoNotes \cite{hovy2006ontonotes,weischedel2013ontonotes}, TACRED \cite{zhang2017position}, SemEval 2010 Task 8 \cite{hendrickx2010semeval} and SciERC \cite{luan2018multi}, are \emph{trigger-based}. The disadvantage of such an approach is that it only captures the most simple cases of relations and 
entity
types that are explicitly mentioned in the text. As a general rule, this also limits 
the datasets to cover only the relations between entity mentions (\ie the annotation process is mention-driven) that appear within a single or at most few adjacent sentences where the relation trigger occurs (see \figref{ch_dwie:fig:rel_span} in \secref{ch_dwie:sec:annotation} for a more detailed illustration of this phenomenon).
However, as we move to a broader \textit{document-based} interpretation, it is common to find relations that are not explicitly mentioned in text. Thus, in our example of \figref{ch_dwie:fig:example} the relation between ``Ministry of Defense'' and ``Britain'' is not explicitly indicated in the text. However, after reading the whole article we can infer relations such as \textsf{ministry\_of}, \textsf{agency\_of} and \textsf{based\_in} between these two entities. This document-level reasoning makes it essential to adopt an entity-centric approach (see \textit{Entity-Centric} comparison group in \Tabref{ch_dwie:tab:quali_compare}) where each \textit{entity}
comprises one or more entity \textit{mentions}, and the annotations (\ie \textit{relations}, \textit{entity tags} and \textit{entity linking} in \datasetname) are made on the entity level, thus abstracting from specific 
\textit{mention-driven} triggers. 

\subsection{Recent advances in information extraction}
\label{ch_dwie:sec:recent_advances}

In the last couple of years, 
the advances in joint modeling have been accompanied by an ever increasing interest in the use of graph-based neural networks \cite{li2015gated,xu2018powerful,wu2020comprehensive}. Initially, this approach has been applied to improve the performance of the single coreference resolution task by transferring document-level contextual information between coreferenced entity mention spans \cite{lee2018higher, kantor2019coreference}. Most recently, these graph propagation techniques have been successfully used in a joint setting \cite{luan2019general, wadden2019entity, fei2020boundaries, fu2019graphrel} 
by performing graph message passing updates between the shared spans across different tasks. However, while successful on mention-driven datasets such as ACE 2005 \cite{walker2006ace} and 
NYT \cite{riedel2010modeling}, 
as far as we are aware, the advantages of these techniques have not yet been investigated in an entity-centric document-level setting.  
We fill this gap by extending the neural graph-based model initially proposed by \cite{luan2019general} to be used on \datasetname~(see \secref{ch_dwie:sec:models}). More specifically, we explore the effect of performing document-level coreference (\propformat{CorefProp})~\cite{lee2018higher,luan2019general} and relation-driven (\propformat{RelProp})~\cite{luan2019general} graph message passing updates between the spans. Additionally, we introduce a new latent attention-based graph propagation method (\propformat{AttProp}) and compare it to 
previously proposed task-driven graph propagation methods (\propformat{CorefProp} and \propformat{RelProp}).  

\subsection{Metrics and evaluation}
\label{ch_dwie:sec:metrics_and_evaluation}

Current dominant IE systems consider \emph{mention-level} scoring of NER as well as 
RE
components when reporting on datasets such as CoNLL-2003 \cite{lample2016neural,chiu2016named,baevski2019cloze,akbik2019pooled,akbik2018contextual}, OntoNotes \cite{clark2018semi,chiu2016named, strubell2017fast}, ACE 2004 \cite{bekoulis2018adversarial,li2014incremental,zhang2017end}, ACE 2005 \cite{zhang2017end,luan2019general,fei2020boundaries}, TACRED \cite{zhang2017position,soares2019matching,zhang2018graph}, and SelEval 2010-Task 8 \cite{hu2020cross,peters2019knowledge, guo2019attention} among others.
In contrast, the \datasetname~dataset is entity-centric where all the annotations are done on the entity cluster level. Consequently, adopting a purely mention-based evaluation approach can lead to a dominance of the score by predictions on entities composed by many mentions as opposed to entities composed by only few ones. 
Conversely, a purely cluster-level evaluation would be overly strict, requiring correct prediction of relation/entity types as well as an exact match of the predicted entity clusters.
To tackle this problem, we propose a new scoring method that combines entity mention-level and cluster-level evaluation, while avoiding the pitfalls of either method alone (see \secref{ch_dwie:sec:metrics}).

\section{Annotation process}
\label{ch_dwie:sec:annotation}
\begin{table}
    \centering
    \caption[Examples of entity mentions in~\datasetname]{\textit{Descriptions} and \textit{Examples} (with entity mentions underlined) of each of the most granular entity classes in \datasetname~(\textit{ENTITY}, \textit{VALUE} and \textit{OTHER}) in the \textit{type} tag hierarchy. Additionally, for the type \textit{ENTITY}, we describe and give examples of each of its direct subtypes (\textit{location}, \textit{organization}, \textit{person}, \textit{misc}, \textit{event} and \textit{ethnicity}).}
    \label{ch_dwie:tab:top_entities_detail}

    \small
    \renewcommand{\arraystretch}{1.2}
    \begin{tabular}{lp{4.0cm}p{4.2cm}}
    \toprule
    \textbf{Entity Tag} &
    \multicolumn{1}{c}{\textbf{Description}} &
    \multicolumn{1}{c}{\textbf{Example}}
    \\
    \midrule
    ENTITY & 
    All nominal named entities. &
    ``\underline{UK} court rules \underline{WikiLeaks}' \underline{Assange} should be extradited to \underline{Sweden}''
    \\    
    \ \ \ \ location & 
    Entities referring to a particular geographical location. &
    ``\underline{Libya} is one of \underline{Germany}'s strongest trading partners in northern \underline{Africa}.''
    \\
    \ \ \ \ organization &
    Organizations such as companies, governmental organizations, etc. & 
    ``According to the report, \underline{Amazon} would pay the same level of royalty fees as \underline{Apple}.''
    \\ 
    \ \ \ \ person &
    Entities referring to people in general such as politicians, artists, sport players, etc. & 
    ``With \underline{Ramires} out, \underline{Drogba} could start as striker, with \underline{Torres} moving to the wing.''
    \\
    \ \ \ \ misc &
    Miscellaneous entity types such as names of work of arts, treaties, product names, etc.  & 
    ``According to the director's own words, \underline{The Post} is a `patriotic film'.''
    \\
    \ \ \ \ event &
    Events such as sport competitions, summits, etc. & 
    ``Last year's \underline{Champions League} final drew a crowd of just 14,303.''
    \\
    \ \ \ \ ethnicity &
    Entity type used to identify different ethnic groups. & 
    ``Attempt to assimilate \underline{Uyghurs} into dominant \underline{Han Chinese} culture.''
    \\
    VALUE &
    Values in general such as time, money, etc.  &
    ``It ended the \underline{2014} fiscal year \underline{45 million} \underline{euros} (\underline{\$51 million}) in the red.''
    \\
    OTHER & Includes the nominal variations of entity types (\eg 
    includes variations of country names such as ``German'', which is a variation of ``Germany''). 
     & ``\underline{Franco}-\underline{German} `war child' granted \underline{German} citizenship.'' \\ 
    \bottomrule
    \end{tabular}
\end{table}

\renewcommand\baselinestretch{1.0}\normalsize
In this work we introduce our \textit{bottom-up} data-driven annotation approach.
Our main goal is to get an annotation schema that reflects
the types of entities and relations that are effectively mentioned throughout the corpus
to maximally capture the information it contains. 
Therefore, we derive the annotation schema from the corpus itself, adopting three annotation passes that are detailed next:
\begin{enumerate*}[(i)]
  \item \textit{exploratory pass}, 
  \item \textit{schema-driven pass}, and 
  \item \textit{inter-annotator refinement}.
\end{enumerate*}
Each pass encompasses substeps to cover all IE subtasks:
\begin{enumerate*}[(i)]
\item mention annotation (\ie the entities and their types),
\item coreference resolution,
\item relation extraction on the entity level (\ie clustering all mentions referring to the same entity), and
\item entity linking (again, on the entity level, providing the same link for all clustered mentions).
\end{enumerate*}

\subsection{Exploratory pass} 
The first annotation pass aims to discover the annotation structure (\ie annotation schema) to be used on the corpus, in particular the types to use for named entity recognition (NER) and relation extraction (RE) tasks. Three annotators are involved in this step to provide annotations on the mention level: one expert annotator and two paid students. However, no parallel annotation is done and the role of the expert annotator is to annotate part of the corpus, as well as instructing and supervising the paid annotators. No a priori fixed schema is followed, but we ask the annotators to be as consistent as possible during the process. More specifically, the annotators are free to define their own entity and relation types for the NER and RE tasks that reflect the contents of the articles as long as they comply with the following generic guidelines: 
\begin{enumerate}
    \item \textbf{Named Entities:} any physical or abstract object (\eg ``Washington'', ``Jeff Davis'',``Nobel Prize'', ``Lisbon Treaty'', etc.) that can be denoted with a proper noun. Entities are usually upper-cased in the text, although values such as money and time can also be included. Use short and specific entity types (\eg person, organization, etc.) to classify entities, the types can be overlapping (a single entity can have multiple types). 
    \item \textbf{Relations:} identify meaningful relations between entities. The type of a relation should be specific and reflect the type of the connected entities as well as the semantic meaning of the relation. For example, instead of using a generic ``located in'' relation for entities located in a particular country, we can divide it in ``based in country'' for organizations that are based in a country, ``city located in country'' for cities located in the country, etc. The types of the relations should have short names, ideally not exceeding 15 characters. 
\end{enumerate}
By not constraining the annotation process to specific entity and relation types, we ensure that our annotations are representative of the actual information contained in the annotated corpus.

\begin{figure}[!t]
\centering
\includegraphics[width=1.0\columnwidth, trim={0cm 0cm 0cm 0cm},clip]{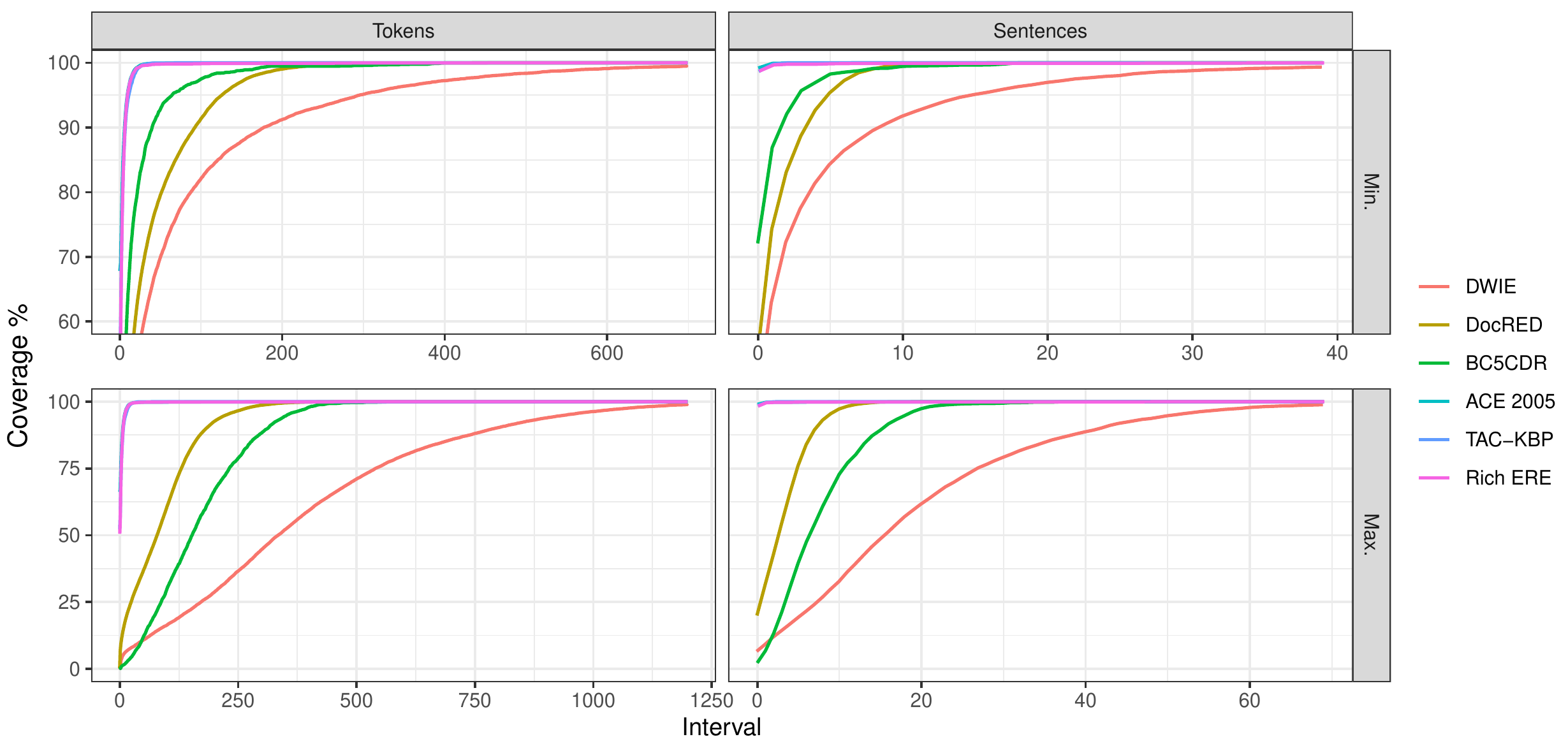}
\caption[Comparison of the coverage of relations of \datasetname]{Comparison of the coverage of the \% of relations with increasing  
interval in tokens
(left) and 
interval in sentences
(right). The graph at the top illustrates the relations coverage measuring the minimum distance between entities (closest mentions). Conversely, the graph at the bottom shows the coverage measuring the maximum distance between entity mentions.
In both 
graphs, we note that the distance between the related mentions in our dataset is higher than in other widely used datasets. }
\label{ch_dwie:fig:rel_span}
\end{figure}

\subsection{Schema-driven pass}
\label{ch_dwie:sec:schema_driven_pass}

\setlength{\tabcolsep}{5pt}
The main goal of this step is to create a consistent annotation schema for 
\begin{enumerate*}[(i)]
\item named entity types and
\item relation types
\end{enumerate*} based on the annotations made in the \textit{exploratory pass}.
As a first step, we identify the 
classification \textit{tags} to be assigned to \textbf{entities}. 
We divide these tags in five main categories: \textit{type}, \textit{topic}, \textit{iptc},
\textit{slot}, and \textit{gender} (see \tabref{ch_dwie:tab:main_entity_types}). 
Our \textit{type} tag is organized in a hierarchical structure (see \tabref{ch_dwie:tab:entities_stats} in  \appref{ch_dwie:app:dataset_insights}), making it easier to extend our annotations to more granular subtypes. \tabref{ch_dwie:tab:top_entities_detail} defines and provides examples 
of
each of the top \textit{type} tags in the entity type hierarchy (\textit{ENTITY}, \textit{VALUE} and \textit{OTHER}) as well as the direct subtypes of \textit{ENTITY}. The \textit{topic} tag allows to assign topics (\eg politics, culture, education, etc.)\ to the entities and it complements the \textit{type} tag (see \tabref{ch_dwie:tab:ner_labels_matrix}). The \textit{iptc} tag is used for the universally defined IPTC news categories based on a media taxonomy (\url{https://iptc.org/standards/subject-codes/}). The \textit{slot} tag is used for additional categorization that is transversal to different entity types. One example of this is the slot \textit{interviewee} that can be assigned to any person (entity of type \textit{person}) interviewed in a particular article.\footnote{Other possible slot values are: \textit{keyword}, \textit{head}, \textit{death}, \textit{interviewer} and \textit{expert}.} Finally, the \textit{gender} tag is used to indicate the gender of the entities that refer to people. 
By defining these multiple overlapping tag types, we 
realize 
that the entity classification is multi-label by nature and thus allows different complementary entity tags to be assigned to a particular entity.\footnote{The average number of labels per entity is 4.0 in our \datasetname~dataset.} 
This contrasts with prevailing single-label multi-class datasets such as ACE 2005~\cite{doddington2004automatic,walker2006ace}, TAC-KBPs~\cite{ellis2014overview,ji2015overview,ellis2015overview,ji2017overview}, Rich ERE~\cite{song2015light}, WNUT 2017~\cite{derczynski2017results} and CoNLL-2003~\cite{sang2003introduction}. 
\setlength{\tabcolsep}{5pt}
\renewcommand{\arraystretch}{1.0}
\renewcommand\baselinestretch{1}\small
\begin{table}[t]
    \centering
    \caption[Examples of relations in~\datasetname]{\textit{Descriptions} and \textit{Examples} of the top 5 most occurring relation types in \datasetname. The entity mentions involved in the relations are underlined.}
    \label{ch_dwie:tab:top_relations_detail}

    \small
    \renewcommand{\arraystretch}{1.5}
    \begin{tabular}{lp{4.0cm}p{4.2cm}}
    \toprule
    \renewcommand{\arraystretch}{1.0}
    \begin{tabular}{c@{}c@{}}\textbf{Relation} \\ \textbf{Type}\end{tabular}
    &
    \multicolumn{1}{c}{\textbf{Description}} &
    \multicolumn{1}{c}{\textbf{Example}}
    \\
    \midrule
    based\_in0 & 
    Relations between organizations and the countries they are based in, ex: \relation{based\_in0}{Uni of Cologne}{Germany} &
    ``Now he's back in \underline{Germany} carrying on with his cancer research at the \underline{University of Cologne}.''
    \\
    in0 &
    Relations between geographic locations and the countries they are located in, ex: \relation{in0}{Athens}{Greece} & 
    ``The murder of a left-wing activist in \underline{Athens} has shaken up \underline{Greece} and inspired a backlash.''
    \\ 
    citizen\_of &
    Relations between people and the country they are citizens of, ex: \relation{citizen\_of}{Guerrero}{Peru} & 
    ``Even as a teenager, \underline{Guerrero} played for the national side in his native \underline{Peru}.''
    \\
    based\_in0-x &
    Relations between organizations and the nominal variations of the countries they are based in, ex: \relation{based\_in0-x}{SPD}{German} & 
    ``\underline{SPD} denies `green light' for new \underline{German} government, but keeps options open''
    \\
    citizen\_of-x &
    Relations between people and the nominal variations of the countries they are citizens of, ex: \relation{citizen\_of-x}{Assange}{Australian} & 
    ``\underline{Australian} national \underline{Assange} said the accusations were politically motivated.''
    \\
    \bottomrule
    \end{tabular}
\end{table}
\renewcommand\baselinestretch{1.0}\normalsize

For our \textbf{relation} annotations, we focus on annotating relations between entities themselves (\cf\textit{document-based entity-centric} approach). 
Our adopted approach allows us to think concept-wise and come up not only with relations that are explicitly stated, but also those that can be implicitly inferred from the text. As a result, our dataset includes relations whose connected mentions are located further apart in the document. 
This can be seen in \figref{ch_dwie:fig:rel_span}, where we compare the \textit{minimum} 
(\textit{Min.}) 
 and \textit{maximum} 
 (\textit{Max.}) 
 distances between the mentions of the two entities connected by a relation for various mention-driven (Rich ERE\footnote{We use the Rich ERE dataset from the LDC2015E29 and LDC2015E68 catalogs.}, TAC-KBP\footnote{We use the TAK-KBP 2017 dataset from the LDC2017E54 and LDC2017E55 catalogs.}, and ACE 2005) and entity-centric (DocRED, BC5CDR, and the final version of our \datasetname~dataset) 
RE datasets. 
We note how other datasets that define the relation in terms of entities (BC5CDR and DocRED) require a higher number of token and sentence spans to cover all the relations in the respective dataset:
entity-centric relations very often involve mentions located in different sentences in the document that refer to those entities. This is not the case for 
mention-driven trigger-based
relations as in the TAC-KBP, Rich ERE and ACE 2005 datasets, where the annotation bias is towards finding explicitly mentioned relations, often involving entity mentions in a single sentence. 

Similarly as with entity tags, we organize our relation annotations using multi-label types (see 
\tabref{ch_dwie:tab:relations_multilabel_stats}
for details). \tabref{ch_dwie:tab:top_relations_detail} gives some examples from the \datasetname~corpus for the top 5 most occurring relation types (a detailed list can be consulted in \tabref{ch_dwie:tab:relations_stats}). For reasons of space, the examples only involve relations between entities whose mentions occur in a single sentence; for an example involving document-level relations we refer to \figref{ch_dwie:fig:example}.

Additionally, we define logical rules to automatically guarantee the consistency of the relations and their types. The following is an example, 
\begin{align}
\relationalign{based\_in2}{X}{Z} \land \relationalign{in0}{Z}{Y} \implies \relationalign{based\_in0}{X}{Y} \label{ch_dwie:eq:complex_logic}
\end{align}
reflecting the knowledge that if an organization \textit{X} is based in a city \textit{Z} (relation \textsf{based\_in2}), and that this city \textit{Z} is located in the country \textit{Y} (relation \textsf{in0}), the fact that company \textit{X} is also located in that country (relation \textsf{based\_in0}) is valid as well.
The goal of this step is mainly consistency of the annotations, but it implies that an effective predictor would need to perform some form of reasoning to correctly predict all relations in the dataset. A complete list of logical rules is provided in \appref{ch_dwie:app:rel_rules_list}. 

\subsection{Inter-annotator refinement} 
\label{ch_dwie:sec:interannotator}
\renewcommand{\thefootnote}{\fnsymbol{footnote}}  
{\renewcommand\baselinestretch{1}\begin{table}[t]
    \centering
\caption[The inter-annotation agreement Cohen's kappa scores]{The inter-annotation agreement Cohen's kappa scores for all the different annotation tasks \textit{before} and \textit{after} the dataset 
refinement
used to analyze and correct
the discrepancies between the parallel annotations.
}
\begin{minipage}{1.0\textwidth}
\renewcommand{\thefootnote}{\fnsymbol{footnote}}  
\renewcommand*\footnoterule{}
    \centering
    \begin{tabular}{lcc}
    \toprule
    {\textbf{Task}} & \textbf{Before Refinement} & \textbf{After Refinement} \\
    \midrule
    Named Entity & 0.8497 & 0.8703 \\
    \ \ \ Named Entity Detection & 0.9665 & 0.9673 \\ 
    \ \ \ Named Entity Classification & 0.8812 & 0.9026 \\ 
    Coreference & 0.9302 & 0.9324 \\ 
    Entity Linking & 0.9280 & 0.9320 \\ 
    Relation & 0.6594 & 0.8729 \\
    \ \ \ Relation Detection & 0.7686 & 0.8727 \\ 
    \ \ \ Relation Classification & 0.8118 & 0.9666 \\ 
   \bottomrule
    \end{tabular}
\end{minipage}
    \label{ch_dwie:tab:kappas_pass}
\end{table}}
\renewcommand{\thefootnote}{\arabic{footnote}}
In order to assess and further improve the quality of our dataset we re-annotate a 100 randomly selected news articles (12.5\% of the articles used in the previous annotation rounds) 
from scratch. This work is done by a second independent expert annotator. The annotations in this pass are performed by following the already defined annotation schema based on the annotation process in the \textit{exploratory} and \textit{schema-driven} passes. We use this second annotated subset to measure the inter-annotator agreement and subsequently determine the parts of the dataset that still need to be improved. \tabref{ch_dwie:tab:kappas_pass} compares the kappa scores before and after this refinement pass for each of the tasks (see \appref{ch_dwie:app:kappa_agreement_details} for details on how the kappa score is calculated). We observe that, after the refinement, all of the kappa scores are above 0.85, which is considered a `strong' \cite{mchugh2012interrater} to `almost perfect' \cite{landis1977measurement} agreement.

Note that the revisions were seeded by and evaluated on the subset of 100 re-annotated articles. However, we argue that the 
inter-annotator refinement
improved the annotation consistency of the entire dataset, given that the reviewed entity and relation types are used in more than 99.4\% of all annotations in \datasetname. 
  
\section{Model architecture}
\label{ch_dwie:sec:models}

\begin{figure}[!t]
\centering
\includegraphics[width=1.0\columnwidth, trim={0cm 2.8cm 8.5cm 0.2cm},clip]{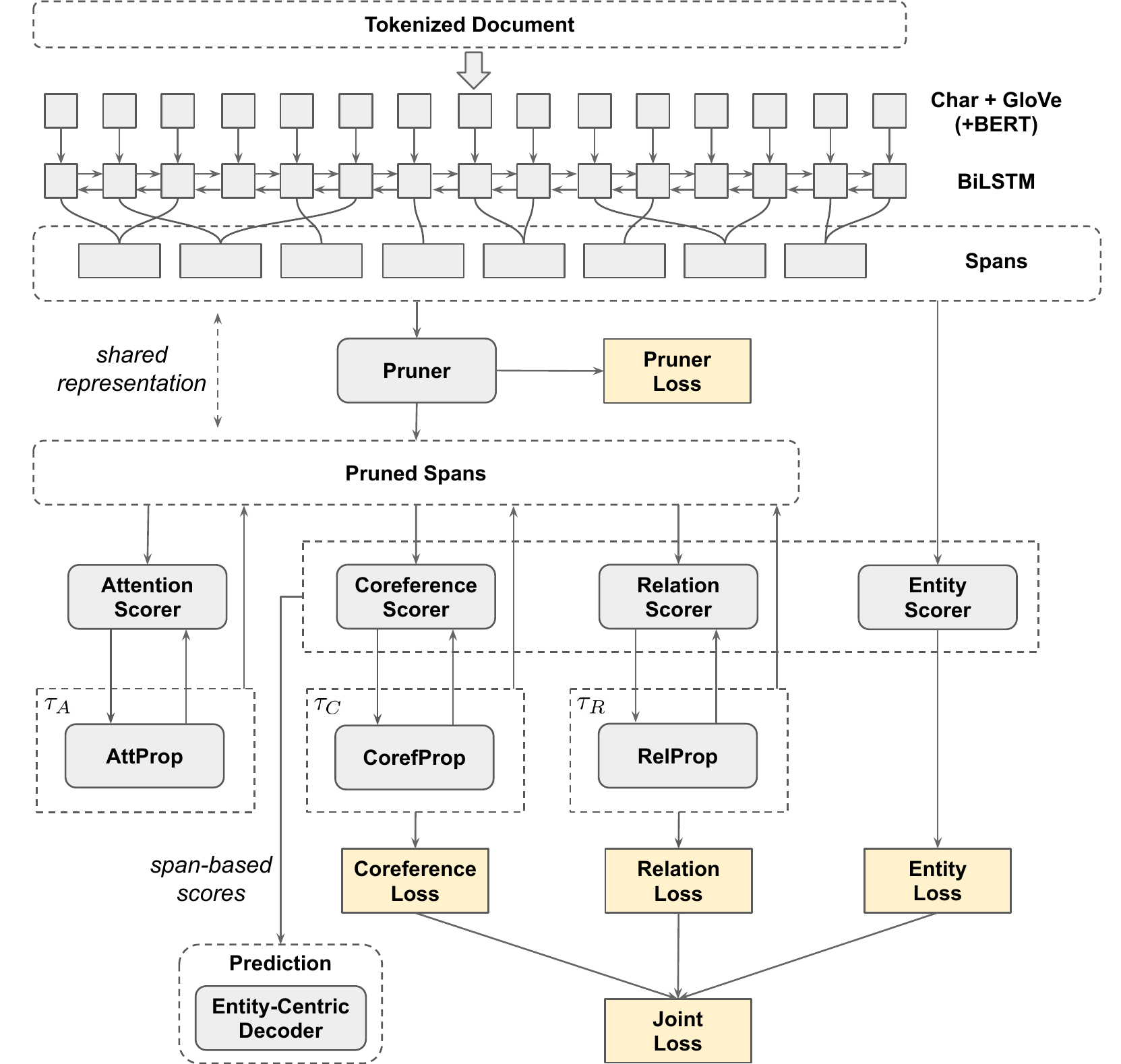}
\captionsetup{singlelinecheck=off}
\caption[Architecture of~\datasetname~model]{Architecture of our model; the span-oriented approach makes it possible to execute \textit{coreference} (\secref{ch_dwie:subsubsec:coref}) and \textit{relation} (\secref{ch_dwie:subsubsec:relation}) scorers independently from \textit{entity scorer}  (\secref{ch_dwie:subsubsec:entity_module}). However, a pruning step (described in \secref{ch_dwie:sec:shared_model}) is needed in order to 
limit
the memory required to perform matrix operations on span representations involved in graph propagation (\propformat{AttProp}, \propformat{CorefProp}, \propformat{RelProp})(\secref{ch_dwie:sec:graph_prop_model}) as well as in the attention, coreference and relation scorer modules. The \textit{pruned spans} share the same representation with the rest of the \textit{spans} (\textit{shared representation}). This way, the update in span representations caused by the graph propagation modules also affects the \textit{entity scorer}. Our \propformat{AttProp} graph propagation method runs independently from coreference, relation, and entity scorers, enabling its use in combination with any task. Finally, the \textit{entity-centric decoder} (\secref{ch_dwie:sec:decoding}) uses the entity clusters predicted by the \textit{coreference scorer} to convert the span-based predictions from the \textit{relation} and \textit{entity} scorers to entity-centric ones.}
\label{ch_dwie:fig:model_architecture}
\end{figure}

In this section we introduce the end-to-end architecture used to compare the performance of models trained on the separate tasks  with the models that are trained jointly for multiple tasks  
on the \datasetname~dataset. 
The main component of our approach is the use of Graph Neural Networks~\cite{scarselli2008graph,li2015gated,xu2018powerful,wu2020comprehensive}, relying on propagation techniques in both single-task and joint setups. More specifically, we implement span-based 
graph message passing
on coreference (\propformat{CorefProp})~\cite{lee2018higher,luan2019general} and relation levels (\propformat{RelProp})~\cite{luan2019general}. Additionally, 
we introduce a latent attentive propagation method (\propformat{AttProp}) which 
is not driven by annotations of any task in particular
and, 
as a result, 
can be freely applied to any task or joint combination of tasks. The interconnection between the different components of our model architecture is depicted in \figref{ch_dwie:fig:model_architecture}. 
It is based on the 
\textit{span-based architecture} introduced in \cite{lee2017end}, 
which supports training on the space of all entity spans simultaneously, dynamically updating span representations by using the graph propagation approach (further detailed in \secref{ch_dwie:sec:graph_prop_model}). Recent works have shown that this idea has the potential for improved effectiveness (albeit at a higher computational cost)~\cite{lee2018higher,luan2019general,dixit2019span,fei2020boundaries}, compared to 
more traditional sequence-labeling approaches 
\cite{lample2016neural,ma2016end,luan2017scientific,katiyar2018nested}.
More concretely, the use of a span-based approach where all the spans are shared between the individual task modules avoids the cascading of errors from the entity mention identification module (\textit{entity scorer} in \figref{ch_dwie:fig:model_architecture}) to the rest of the tasks. 

The most similar architecture to ours in using joint span-based neural graph IE is DyGIE~\cite{luan2019general} and its successor DyGIE$++$~\cite{wadden2019entity}. Our model is described in detail below, but here we already list the aspects in which it differs from these models:
\begin{enumerate}
    \item We introduce the graph propagation technique \propformat{AttProp} (see \linebreak  \secref{ch_dwie:sec:graph_prop_model}), which is not directly conditioned on a particular task and can be used in  single-task (for each of the tasks) as well as joint settings. 
    \item We define a coreference architecture that, unlike previous work in span-based coreference resolution~\cite{lee2017end,lee2018higher}, allows to also account for singleton entities in the \datasetname~dataset (see Sections~\ref{ch_dwie:subsubsec:coref} and \ref{ch_dwie:subsubsec:single_coref}) by using an additional \textit{pruner loss}, which turns out essential for the single model focusing on end-to-end coreference resolution. 
    \item Due to the document-level nature of \datasetname, we run graph propagations on the whole document. This contrasts with a sentence-based approach adopted initially in the DyGIE/DyGIE$++$ architectures. It also drives some changes such as the use of a single \textit{pruner} (see \secref{ch_dwie:sec:shared_model}) to extract spans used in coreference and 
    RE
    modules. Similarly, instead of applying the shared BiLSTM sentence by sentence as in \cite{luan2019general} and \cite{wadden2019entity}, we do it on the entire document, in order to allow capturing cross-sentence dependencies for document-level relations and entity clusters in \datasetname.
    \item We add an additional decoding step (see \secref{ch_dwie:sec:decoding}) needed to transform mention-based predictions for 
    RE
    and NER tasks into entity-based ones, as required by the entity-centric nature of \datasetname, and propose corresponding evaluation metrics (see \secref{ch_dwie:sec:metrics}).
    \item Finally, we make changes in the loss and prediction components to support multi-label classification (in NER and RE) as required in \datasetname. 
\end{enumerate}

\subsection{Span-based representation}
\label{ch_dwie:sec:shared_model}
The input to our model consists of document-level annotation instances. Each document $D$ from the considered document collection $\mathcal{D}$ is represented by its sequence $T$ of tokens. 
These tokens are represented internally as a concatenation of GloVe~\cite{pennington2014glove} and character embeddings~\cite{ma2016end}. We also experiment with additionally concatenating BERT~\cite{devlin2019bert} contextualized embeddings. Since BERT is run on a sub-token level, to the representation of each token we only concatenate the BERT-based representation of the first sub-token, as originally proposed by \cite{devlin2019bert}. This input is fed into a BiLSTM layer in order to obtain the output token representations by concatenating the forward and backward LSTM hidden states. The BiLSTM outputs for the considered document $D$ are written on the token level as $\textbf{e}_i\in\mathbb{R}^m\;(i=1,\ldots,\vert T\vert)$.
These are converted into \textit{span representations}. The set of all possible spans for $D$, up to maximum span width $w_{\mathrm{max}}$ (which is a hyperparameter of the model), is written as
$S= \{s_1, \dotsc, s_{|S|}\}$.
The number of spans 
can be calculated as follows, 
\begin{equation}
|S| = \sum\limits_{k=1}^{w_{\mathrm{max}}} |T| - k + 1 \;={w_{\mathrm{max}}}\left(|T| - \frac{w_{\mathrm{max}}-1}{2}\right)\label{ch_dwie:eq:spans}\end{equation}
We obtain the representation $\textbf{g}^0_i$ for span $s_i$, ranging from token $l$ to token $r$, by concatenating their respective BiLSTM states $\textbf{e}_{l}$ and $\textbf{e}_{r}$ with an embedding $\boldsymbol{\psi}_{r-l}$ for the span width $w_i = r-l$
\begin{equation}
\textbf{g}^0_i = [\textbf{e}_{l}; \textbf{e}_{r}; \boldsymbol{\psi}_{r-l}] \label{ch_dwie:eq:spanrepr}
\end{equation}

As seen from \equref{ch_dwie:eq:spans}, the number of possible spans scales approximately linearly with the maximum span width $w_{\mathrm{max}}$, as well as the document length $\vert T\vert$ (assuming $w_{\mathrm{max}}\ll\vert T\vert$). This leads to a strongly increased set of spans, as compared to previous works where $\vert S\vert$ scales with the length of individual sentences rather than entire documents~\cite{luan2019general,wadden2019entity}.
In order to mitigate the required memory of our model, 
we use a shared \textit{pruner} to reduce $S$ to a smaller set $P$ of candidate spans
to be used by the coreference and 
RE
scorers and in the graph propagation modules (see further). The choice of using a single pruner contrasts with similar work in \cite{luan2019general} and \cite{wadden2019entity} where two separate pruners are used, one for the relation task, and another for coreference. Our design choice is based on the fact that both of these tasks 
use the same document-level entity mentions. This contrasts with datasets used in \cite{luan2019general} and \cite{wadden2019entity} where, while the coreference is defined on the document-level, the relations are sentence-based. 

Finally, we use graph propagation to iteratively refine the pruned spans representations. Three graph propagation mechanisms are compared in the experiments. Our own contribution is the attention-based graph propagation method \propformat{AttProp}, where the span representations are updated in $\tau_A$ iterations. Alternatively, $\tau_C$ iterations of \propformat{CorefProp}~\cite{lee2018higher,luan2019general} can be performed, or $\tau_R$ iterations of \propformat{RelProp}~\cite{luan2019general}.  

The span representation of a particular span $s_i$ after iteration $t$ is denoted as
$\textbf{g}_i^t$ in our notation. The details of graph propagation are explained in \secref{ch_dwie:sec:graph_prop_model}. Note that in theory several of these graph propagation techniques could be accumulated, but in our setting the benefits thereof in terms of model effectiveness were minor, at a significantly higher computational cost. Therefore, in our experiments, we only compare models without graph propagation with models applying a single form of graph propagation. 
To keep the sections introducing the models clear, we will write $\tau$ to denote the number of propagations in general (which could be 0, or any of ${\tau_A, \tau_C \mathrm{\ or\ } \tau_R}$, depending on the chosen experiments and considered model components).

\subsection{Joint model for entity recognition, coreference resolution, and relation extraction}
\label{ch_dwie:sec:joint_training}

In this section, we present the joint model including recognition of entity mentions as belonging to $L_T$ types (introduced as NER), the clustering of the entity mentions into entities (coreference resolution), and identifying relations between entities, all on the document level. The building blocks responsible for the three subtasks are discussed next, as well as the total loss of the joint model. 
The details of the graph propagation mechanisms are then provided further on (\secref{ch_dwie:sec:graph_prop_model}). 

\subsubsection{Entity mention module} \label{ch_dwie:subsubsec:entity_module}
All spans $s_i$ (up to width $w_\mathrm{max}$) of the considered document\footnote{For convenience, the subscript $D$ indicating the current document is left out in the equations of this section.} are scored by feeding their representation (starting from \equref{ch_dwie:eq:spanrepr} and potentially updated after $\tau$ graph propagation iterations) into the feed-forward neural network (FFNN) written
as
$\mathbfcal{F}_{\text{mention}}$, with as many outputs as there are entity types:
\begin{align}
    \boldsymbol{\Phi}^{\tau}_{\mathrm{mention}}(s_i) = \mathbfcal{F}_{\mathrm{mention}}(\textbf{g}^{\tau}_i). \label{ch_dwie:eq:ner1}
\end{align}
Throughout this section, we will maintain the same notation of $\mathbfcal{F}(\mathbf{x})$ to denote a FFNN that takes as input a vector $\mathbf{x}$ and produces a vector of scores, and $\mathcal{F}(\mathbf{x})$ 
 to refer to a FFNN with a scalar output.

The probability of each label being valid for the considered span is modeled by component-wise application of a sigmoid ($\sigma(x)=1/(1+e^{-x})$) to these scores $\boldsymbol{\Phi}^\tau_{\mathrm{mention}}(s_i)\in \mathbb{R}^{L_T}$ (with $L_T$  the number of entity 
tags). 
The log probability of the ground truth mention labels for all spans of document $D$ is given by
\begin{equation}
    \begin{split}
    \log P_{\mathrm{mention}}\left(E^*\vert G^\tau\right) = \sum_{i=1}^{\vert S\vert}\sum_{l=1}^{L_T} I_{i,l} \log\sigma(\boldsymbol{\Phi}^{\tau}_{\mathrm{mention}}(s_i)_l)
    + (1-I_{i,l}) \\ \log\big( 1 - \sigma\left(\boldsymbol{\Phi}^{\tau}_{\mathrm{mention}}(s_i)_l\right)\big),
    \label{ch_dwie:eq:ner2}
    \end{split}
\end{equation}
in which $E^*$ represents the set of ground truth mention labels for all spans in the document, and $I_{i,l}\in\{0,1\}$ is the ground truth indicator label for mention 
tag
$l$ of span $s_i$. $G^\tau$ denotes the set of all considered span representations for the current document. The superscript $\tau$ reflects the fact that, in case graph propagation is applied, the subset of $\vert P\vert$ representations (for the spans retained after pruning) have been updated over $\tau$ iterations.  By summing over all entity types ($l=1,\ldots,L_T$), we account for the fact that a particular span can have multiple associated entity tags (i.e., the considered NER task is multi-label). 
At inference time, spans get assigned those entity types for which the corresponding score $\boldsymbol{\Phi}^{\tau}_{\mathrm{mention}}(s_i)>0$. Note that not all valid entity mentions necessarily get an entity type assigned: if the relation extractor determines that a span is part of a relation, it effectively becomes an entity mention, even if none of the pre-defined types is considered applicable by the entity scorer.

\subsubsection{Coreference module}\label{ch_dwie:subsubsec:coref}
While the entity scoring is performed on all span representations $S$, this is not possible for the coreference and relation scorers, due to memory limitations. The latter scorers predict on \textit{pruned spans}, as shown in \figref{ch_dwie:fig:model_architecture}. How the pruner is trained jointly with the model, is described in \secref{ch_dwie:subsubsec:pruner}. 
In order to avoid confusion by introducing additional notations, we list the spans in the pruned set $P$ as $s_1,\ldots,s_{\vert P\vert}$, according to their original order in the text. 

The module for coreference resolution is based on pairwise scoring of the pruned spans from~$P$. Following ideas from \cite{lee2017end,lee2018higher,luan2018multi,luan2019general}, 
for any span $s_j$, scores with respect to each of the preceding (also referred to as `antecedent') spans $s_i\;(i\leq j)$ in the document are calculated with a neural network $\mathcal{F}_{\mathrm{coref}}$:
\begin{equation}
\Phi^\tau_{\mathrm{coref}}(s_i,s_j) = \mathcal{F}_{\mathrm{coref}} \left([\textbf{g}^\tau_i; \textbf{g}^\tau_j; \textbf{g}^\tau_i \odot \textbf{g}^\tau_j; \boldsymbol{\varphi}_{i,j}]\right).
\label{ch_dwie:eq:coref1} 
\end{equation}
This expression scores the compatibility between spans $s_i$ and $s_j$, taking as input the concatenation of their respective span representations (after $\tau$ propagation iterations), their component-wise product, and an embedding $\boldsymbol{\varphi}_{i,j}$ representing their distance in terms of the number of ordered candidate spans from $s_i$ to $s_j$. 

In order to deal with non-coreferent or incorrect spans, previous work in span-based coreference~\cite{lee2017end,lee2018higher} defines a dummy antecedent $\epsilon$ to which all non-coreferent or invalid spans point.  
While this approach is effective in datasets that do not contain singleton entity clusters, such as OntoNotes-based CoNLL-2012~\cite{pradhan2012conll}, it does not allow to distinguish between valid singleton entity mentions and invalid mention spans. 
This makes it unsuitable to use on \datasetname, since it contains singleton entity clusters, consisting of a single mention. In fact, 66.4\% of the entity clusters in {\datasetname} are singletons. Furthermore, the current official CoNLL-2012 evaluation script\footnote{\url{https://github.com/conll/reference-coreference-scorers}} based on \cite{pradhan2014scoring} accounts for scenarios where either the dataset or the predicted mentions are singletons, which 
has a direct impact on the established
B-CUBED~\cite{bagga1998algorithms} and CEAF$_{\text{e}}$~\cite{luo2005coreference} coreference scores. 
In order to tackle the singleton entity cluster detection in our coreference model, we 
propose to start from $\Phi^\tau_{\mathrm{coref}}(s_j,s_j)$\footnote{This would be replaced with $\Phi^\tau_{\mathrm{coref}}(\epsilon, s_j)$ in the \textit{dummy-based} formulation defined in \cite{lee2017end}.} as a self-coreference span score. By applying the correct target in the coreference loss, it allows indicating that either the span $s_j$ is not a valid mention, or that it is a valid mention that is not co-referenced with any antecedent span. 

The log probability of the ground truth coreference labels of document $D$ is given by
\begin{equation}
\log P_{\mathrm{coref}}\left(C^*\vert G^\tau\right) = \sum_{j=1}^{\vert P\vert} \log \frac{ \sum\limits_{s^*\in S_j^*}\exp\left(\Phi^\tau_{\mathrm{coref}}(s^*,s_j)\right)  }{ \sum\limits_{i=1}^j\exp\left(\Phi^\tau_{\mathrm{coref}}(s_i,s_j)\right)}.
\label{ch_dwie:eq:coref2}
\end{equation}
The set of ground truth coreference labels is indicated as $C^*$. The summation over $j$ represents the contribution to the log likelihood of the correct antecedent labels for each span $s_j$ in the pruned set $P$. The individual terms in the right-hand side correspond to the log probability of the correct antecedent labels for a particular span $s_j$.  In the denominator, the summation ranges from the first span, up to span $s_j$ itself (i.e., for the self-coreference score), but not beyond it (given that only \emph{antecedents} in the sorted sequence of pruned spans are considered). The numerator contains the contributions from the potentially multiple ground truth antecedents for span $s_j$. This stems from the fact that multiple antecedent mentions may belong to the same cluster as $s_j$, which all contribute to the probability of the correct antecedent labels. The set of ground truth antecedents corresponding to span $s_j$ is written $S_j^*$.

At inference time, the highest scoring antecedent for span $s_j$ (including $s_j$ itself) is picked. Due to the idea of only predicting \emph{antecedents}, picking any of the ground truth antecedents leads to the correct mention clusters
\cite{durrett2013easy, wiseman2015learning,lee2017end,lee2018higher}.

\subsubsection{Relation module}
\label{ch_dwie:subsubsec:relation}
Similar to the coreference module (\equref{ch_dwie:eq:coref1}), we score span pairs using an FFNN 
\begin{align}
\boldsymbol{\Phi}^\tau_{\mathrm{relation}}(s_i,s_j) = \mathbfcal{F}_{\mathrm{relation}} \left([\textbf{g}^\tau_i; \textbf{g}^\tau_j; \textbf{g}^\tau_i \odot \textbf{g}^\tau_j; \boldsymbol{\varphi}_{i,j}]\right),
\label{ch_dwie:eq:rel1} 
\end{align}
where
$\boldsymbol{\varphi}_{i,j}$ is again the distance embedding as introduced in \secref{ch_dwie:subsubsec:coref}. 
$\boldsymbol{\Phi}^\tau_{\mathrm{relation}}(i,j) \in \mathbb{R}^{L_R}$ is a vector representing relation span pair scores
for each of the $L_R$ possible relation types between spans $s_i$ and $s_j$.

The log probability of the ground truth relation labels of document $D$ is given by
\begin{equation}
\begin{split}
\log P_{\mathrm{relation}} \left(R^* \vert G^\tau \right) = \sum\limits_{{i,j=1}}^{|P|} \sum\limits_{l=1}^{L_R}  I_{i,j,l} \log \sigma(\boldsymbol{\Phi}^\tau_{\mathrm{relation}}(s_i,s_j)_l) + \\ (1-I_{i,j,l}) \log\left(1 - \sigma(\boldsymbol{\Phi}^\tau_{\mathrm{relation}}(s_i,s_j)_l)\right), \label{ch_dwie:eq:rel2} 
\end{split} 
\end{equation}
in which $R^*$ represents the set of ground truth relation labels for all combination of pruned span pairs in the document, and $I_{i,j,l} \in \{0,1\}$ is the ground truth indicator label for relation type $l$ of the span pair $(s_i, s_j)$. Note that all $\vert P\vert^2$ pruned span pairs are considered, since the order of the spans in the relation matters (unlike the coreference case).
By summing over all possible relation types $L^R$, we account for the fact that a particular relation between two spans can be multi-label (which is the case for more than 30\% of relations, as shown in \Tabref{ch_dwie:tab:relations_multilabel_stats}).

Since this model is run in parallel with the coreference module, it is used to predict relations only between entity mentions and not entity clusters. During inference, candidate relations are accepted when \linebreak $\boldsymbol{\Phi}^\tau_{\mathrm{relation}}(s_i, s_j)_l > 0$.

\subsubsection{Span pruner}\label{ch_dwie:subsubsec:pruner}
The span pruner is an FFNN, denoted $\mathcal{F}_{\text{pruner}}$, that scores all spans $s_i$ based on their initial representation $\mathbf{g}_i^0$, after which only the highest scoring spans are retained in the pruned span set $P$.  In our experiments $P$ contains the top $0.2\,\vert T\vert$ highest scoring spans, which covers more than $98\%$ of all the ground truth mention spans in the \datasetname~dataset. We represent the pruner score for span $s_i$ as 
\begin{align}
     \Phi_{\mathrm{pruner}}\left(s_i\right) = \mathcal{F}_{\mathrm{pruner}} \left(\textbf{g}^0_i\right).
     \label{ch_dwie:eq:pruner1} 
\end{align}
Several strategies can be used to train the pruner. One option is to directly optimize the probability of the pruner to detect the spans of correct entity mentions. With $S^*$ the set of spans with at least one ground truth entity type, and $I_i\in\{0,1\}$ an indicator for whether $s_i\in S^*$, the corresponding log likelihood can be written as 
\begin{equation}
\begin{split}
\log P_{\mathrm{pruner}} \left(S^*\vert G^0 \right) = \sum_{i=1}^{\vert S\vert} I_i \log\sigma\left({\Phi}_{\mathrm{pruner}}(s_i)\right)
    + \\ (1-I_i) \log\left( 1 - \sigma\left({\Phi}_{\mathrm{pruner}}(s_i)\right)\right),
    \label{ch_dwie:eq:pruner_loss}  
\end{split}
\end{equation}
leading to a separate pruner loss term. 
Alternatively, the pruner can be trained indirectly by adapting the mention score from \equref{ch_dwie:eq:ner1}, the coreference score from \equref{ch_dwie:eq:coref1} or the relation score from \equref{ch_dwie:eq:rel1} as follows:
\begin{align}
     \tilde\Phi_{\mathrm{mention}}^\tau\left(s_i\right) &= \Phi_{\mathrm{mention}}^\tau\left(s_i\right) + \Phi_{\mathrm{pruner}}\left(s_i\right)
     \label{ch_dwie:eq:pruner2}
     \\
     \tilde\Phi_{\mathrm{coref}}^\tau\left(s_i, s_j\right) &= \Phi^\tau_{\mathrm{coref}}\left(s_i, s_j\right) + \Phi_{\mathrm{pruner}}\left(s_i\right)
     \label{ch_dwie:eq:pruner3}
     \\
    \tilde\Phi_{\mathrm{relation}}^\tau\left(s_i, s_j\right) &= \Phi^\tau_{\mathrm{relation}}\left(s_i, s_j\right) + \Phi_{\mathrm{pruner}}\left(s_i\right)
    \label{ch_dwie:eq:pruner4}      
\end{align}
for use in the expressions \equref{ch_dwie:eq:ner2}, \equref{ch_dwie:eq:coref2} and \equref{ch_dwie:eq:rel2}, respectively. As such, higher pruner scores would directly correspond to higher mention or coreference scores, and lead to a meaningful ranking of spans according to pruner scores. 
All three strategies seem to work on a similar level, but for the presented joint model experiments, we use the indirect training through the coreference module, as in \equref{ch_dwie:eq:pruner3}. Note that we did not experiment with training the pruner through the relation module, because it would be trained only on those spans involved in relations, which is a mere subset of all valid mentions.

\subsubsection{Joint model}\label{ch_dwie:subsubsec:joint_model}
We perform joint training in order to explore the degree to which the graph propagation techniques (see  \secref{ch_dwie:sec:graph_prop_model}) affect related tasks in \datasetname. For instance, we expect that performing a coreference propagation can have a positive impact on the NER task. We hypothesize that enriching the entity spans with broader contextual information coming from other mention spans in the cluster, can improve the effectiveness of the entity module.
Furthermore, given the entity-centric nature of \datasetname, the mention-based predictions for NER and RE
have to be grouped in coreference clusters (see section \ref{ch_dwie:sec:decoding} for details), which makes it necessary to execute these tasks jointly with the coreference task. 

The joint loss for each document $D$ is a weighted sum of the individual loss functions of the subtasks: 
\begin{equation}
    \begin{split}
    \mathcal{L}^{\mathrm{joint}}_D = \sum_{(E^*, C^*, R^*)} 
    \lambda_E \log P_{\mathrm{mention}}\left(E^*\vert G^\tau\right) 
    +\lambda_C \log P_{\mathrm{coref}}\left(C^*\vert G^\tau\right)
    + \\ \lambda_R \log P_{\mathrm{relation}} \left(R^* \vert G^\tau \right),
    \label{ch_dwie:eq:jointloss}
    \end{split}
\end{equation}
in which $\lambda_E$, $\lambda_C$, and $\lambda_R$ are hyperparameters of the joint model.

\subsection{Decoding and prediction}
\label{ch_dwie:sec:decoding}
Unlike previous datasets used in span-based predictions~\cite{luan2018multi,kulkarni2018annotated,walker2006ace,doddington2004automatic} where the relation and entity extraction 
are done on the mention-level, \datasetname~is an entity-centric dataset. During inference, this requires an additional decoding step to cluster the mention-based span-dependent predictions into entity-centric ones. The component responsible for this decoding in the proposed architecture is the \textit{entity-centric decoder} (see \figref{ch_dwie:fig:model_architecture}). The pseudo-code in \Algref{ch_dwie:alg:entity_centric} summarizes the steps performed by this component. First, the decoder receives as \textit{input} 
the predicted span clusters ($p\_cl$), entity mentions ($p\_men$) and relations between spans ($p\_rel$) obtained from the scores calculated in 
\equref{ch_dwie:eq:pruner3},
\equref{ch_dwie:eq:ner1} and \equref{ch_dwie:eq:rel1}, respectively. Next, the predicted entity mentions are 
connected with
the respective clusters by using the dictionary $C$ that maps mention spans to cluster ids (lines 3--12 in \Algref{ch_dwie:alg:entity_centric}). 
Specifically, 
each of the entity clusters is assigned the union of the entity types predicted for any of the 
mention spans 
inside the cluster (line 11 in \Algref{ch_dwie:alg:entity_centric}). If the predicted entity mention can not be located inside the predicted clusters, a new singleton cluster is added (lines 5--6 in \Algref{ch_dwie:alg:entity_centric}). Finally, all the pairwise predicted relations on the mention level ($p\_rel$) between members of two different clusters are assigned as predicted relations between the (cluster-level) entities (lines 13--20 in \Algref{ch_dwie:alg:entity_centric}). Similarly as with entity mentions, the dictionary $C$ is used to map the 
mention spans ($span\_h$ and $span\_t$) of a particular relation type $rel\_type$ to the corresponding cluster ids. 
Furthermore, the relations added between two clusters are the union of all the relations predicted between 
any pair of
mentions 
inside these clusters (line~18 in \Algref{ch_dwie:alg:entity_centric}). 

\renewcommand\baselinestretch{1}\small
\begin{algorithm}
\begin{flushleft}
\textbf{Input:} predicted clusters ($p\_cl$), entity mentions ($p\_men)$ and relations between mentions ($p\_rel$): \vspace*{-2mm}
\end{flushleft}
\begin{enumerate}
    \item $p\_cl$ is a dictionary (map) that maps cluster ids to mention spans \vspace{-3mm}
    \item $p\_men$ is list of tuples $\langle$predicted span, predicted tag$\rangle$ \vspace{-3mm}
    \item $p\_rel$ is list of tuples $\langle$predicted head span, predicted relation, predicted tail span$\rangle$ 
\end{enumerate}\vspace*{-1mm}
\begin{flushleft}
\textbf{Output:} clusters ($p\_cl$), decoded entities ($d\_ent$) and relations between entities ($d\_rel$)
\end{flushleft}
  \begin{algorithmic}[1]
\State Initialize $d\_ent, d\_rel\gets\mathrm{empty\ dictionary\ (map)}$ 
\State $C \gets$ transformed $p\_cl$ that maps spans to cluster ids \vspace*{1mm}
\LeftComment{Decode entity mentions ($p\_men$) to entities ($d\_ent$) (lines 3--12)}
\For{$span,tag$ \textbf{in} $p\_men$}
    \If{$span$ \textbf{not in} $C.keys()$}
        \State $C[span] \gets$ new concept id
        \State $p\_cl[C[span]] \gets \mathrm{list}([span])$
    \EndIf
    \If{$C[span]$ \textbf{not in} $d\_ent.keys()$}
        \State $d\_ent[C[span]] \gets$ empty set
    \EndIf
    \State $d\_ent[C[span]].\mathrm{add}(tag)$
\EndFor
\vspace*{1mm}
\LeftComment{Decode relations between mentions ($p\_rel$) to relations between entities ($d\_rel$) (lines 13--20)}
\For {$span\_h, rel\_type, span\_t$ \textbf{in} $p\_rel$}
    \If{$(span\_h$ \textbf{in} $C.keys())$ \textbf{and} $(span\_t$ \textbf{in} $C.keys())$}
        \If{$\langle C[span\_h],C[span\_t] \rangle$ \textbf{not in} $d\_rel.keys()$}
            \State $d\_rel[\langle C[span\_h],C[span\_t] \rangle] \gets$ empty set
        \EndIf
        \State $d\_rel[\langle C[span\_h],C[span\_t] \rangle].\mathrm{add}(rel\_type)$
    \EndIf
\EndFor
 \caption{Entity-centric decoder for the \textit{Joint} model.}\label{ch_dwie:alg:entity_centric}
\end{algorithmic}
\end{algorithm}
\renewcommand\baselinestretch{1.0}\normalsize

\subsection{Graph propagation mechanisms}\label{ch_dwie:sec:graph_prop_model}
In order to evaluate the impact of graph-based propagation of contextual information between the spans, 
we propose \propformat{AttProp}, and reimplement the \propformat{CorefProp} and \propformat{RelProp} graph propagation algorithms. 
\cite{lee2018higher} proposed the gated graph propagation update function for use on coreference resolution, which was then successfully applied in a joint multi-task setting by \cite{luan2019general,wadden2019entity}.
The graph propagation equations are written as:
\begin{align}
    \textbf{f}^t_x(s_i) &= \sigma\left(\mathbfcal{F}_{x} ([\textbf{g}^t_i; \textbf{u}^t_x(s_i)])\right), \label{ch_dwie:eq:update_spans1}
    \\ 
    \textbf{g}_i^{t+1} &= \textbf{f}^t_x(s_i) \odot \textbf{g}^t_i + \left(1 - \textbf{f}^t_x(s_i)\right) \odot \textbf{u}^t_x(s_i),
    \label{ch_dwie:eq:update_spans2}
\end{align}
where in our case $x \in \{A, C, R\}$ denotes \propformat{AttProp}, \propformat{CorefProp}, and \propformat{RelProp}, respectively.
The $n$-dimentional vector $\textbf{f}^t_x(s_i)$, produced by the single-layer FFNN $\mathbfcal{F}_{x}$ can be interpreted as 
a gating vector that acts as a switch between the current span representations $\textbf{g}^t_i \in \mathbb{R}^n$, and the update span vector $\textbf{u}^t_x(s_i) \in \mathbb{R}^n$. The various graph propagation methods differ in how $\textbf{u}^t_x(s_i)$ is calculated.

\boldpartitle{\propformat{CorefProp}} The coreference confidence score between span $s_i$ and $s_j$ for propagation iteration \textit{t} is denoted as $P^t_C(s_i,s_j)$ and calculated as follows,
\begin{align}
    P^t_C(s_i,s_j) = \dfrac{\text{exp}\left(\tilde\Phi^t_{\mathrm{coref}}(s_i,s_j)\right)}{\sum\limits_{i'=1}^{j} \text{exp}\left(\tilde\Phi^t_{\mathrm{coref}}(s_{i'},s_{j})\right)},
    \label{ch_dwie:eq:coref_propagation} 
\end{align}
in which $i' \in \{1, \dotsc, j\}$ refers to all antecedent spans $s_{i'}$ to span $s_j$ in the pruned span set. 
Note that the coreference scores according to \equref{ch_dwie:eq:pruner3} are used. This means the confidence scores not only reflect whether the considered spans are compatible, but also whether the individual spans are likely to be retained by the pruner as potential entity mentions.
In order to perform a \propformat{CorefProp} graph iteration, the span update vector $\textbf{u}^t_C(i) \in \mathbb{R}^n$ is first calculated as a weighted average of the current representation of span $s_j$ and all of its antecedents 
\begin{align}
    \textbf{u}^t_C(s_j) = \sum\limits_{i=1}^{j}P^t_C(s_i,s_j) \> \textbf{g}^t_i,
    \label{ch_dwie:eq:coref_update_vector} 
\end{align}
in which the weighting coefficients quantify the coreference compatibility of the corresponding span with $s_j$.
After that, the update equations \equref{ch_dwie:eq:update_spans1} and \equref{ch_dwie:eq:update_spans2} are applied.

\boldpartitle{\propformat{RelProp}} Similarly as with \propformat{CorefProp},
a relation span update vector is calculated as formalized next,
\begin{align}
    \textbf{u}^t_R(s_j) = \sum\limits_{i=1}^{|P|} \left(\textbf{A}_R\; f\left(\boldsymbol{\Phi}^t_{\mathrm{relation}}(s_i,s_j)\right) \right)   \>  \odot \textbf{g}^t_i, \label{ch_dwie:eq:relation_update_vector}    
\end{align}
where $\textbf{A}_R \in \mathbb{R}^{n\times L_R}$ is a trainable projection tensor, and $f$ is a non-linear activation function (ReLU). Similarly as in \equref{ch_dwie:eq:coref_update_vector}, the update vector can be interpreted as a weighted sum of all span representations, with the additional expressiveness stemming from the projection matrix $\textbf{A}_R$ in accounting for the relation scores.

\boldpartitle{\propformat{AttProp}} In order to measure the impact of the `supervised' \propformat{CorefProp} and \propformat{RelProp} propagation techniques described by equations (\ref{ch_dwie:eq:coref_propagation})-(\ref{ch_dwie:eq:relation_update_vector}) above, we introduce a latent attentive propagation. Unlike \propformat{CorefProp} and \propformat{RelProp} that are driven by the task-specific confidence propagation scores $\Phi_{\mathrm{coref}}^t(s_i,s_j)$ and $\boldsymbol{\Phi}_{\mathrm{relation}}^t(s_i,s_j)$, \propformat{AttProp} is influenced only by latent attention weights between all the pruned spans $P$ calculated as follows,  
\begin{align}
\Phi^t_{\mathrm{att}}(s_i,s_j) = \mathcal{F}_{\mathrm{att}} \left([\textbf{g}^t_i; \textbf{g}^t_j; \textbf{g}^t_i \odot \textbf{g}^t_j; \boldsymbol{\varphi}_{i,j}]\right), \label{ch_dwie:eq:att_scores} 
\end{align}
where $\boldsymbol{\varphi}_{i,j}$ is the distance feature embedding function between spans $s_i$ and $s_j$, and $\Phi^t_{\mathrm{att}}(s_i,s_j)$ is the attention score between these spans.
This score is normalized with a softmax to get the $P^t_{A}(s_i,s_j)$ confidence score
\begin{align}
    P^t_{A}(s_i,s_j) = \dfrac{\text{exp}\left(\Phi^t_{\mathrm{att}}(s_i,s_j)\right)}{\sum\limits_{j'=1}^{|P|} \text{exp}\left(\Phi^t_{\mathrm{att}}(s_i,s_{j'})\right)}.
    \label{ch_dwie:eq:att_propagation} 
\end{align}
The span update vector $\textbf{u}^t_A(s_i) \in \mathbb{R}^n$ 
is calculated as a weighted sum of all the $P$ span representations as opposed to only antecedents in \propformat{CorefProp}
\begin{align}
    \textbf{u}^t_A(s_i) = \sum\limits_{j=1}^{|P|}P^t_A(s_i,s_j) \> \textbf{g}^t_j.
    \label{ch_dwie:eq:att_update_vector} 
\end{align}

\subsection{Single task models}
\label{ch_dwie:sec:single_task_models}

In this section we shortly describe independent baseline models for the three individual core tasks under study in this paper, as training these models not entirely corresponds to merely minimizing the corresponding loss term from the total loss \equref{ch_dwie:eq:jointloss}.

\subsubsection{Single entity recognition model}
\label{ch_dwie:sec:single_ner_model}
The single-task NER model is designed for detecting and correctly labeling the individual entity spans, and is based on \equref{ch_dwie:eq:ner2}. However, even for the single models, the graph propagation mechanism \propformat{AttProp} may be useful, but for that the pruner needs to be jointly trained with the model. This is obtained by augmenting the mention loss $-\log P_{\mathrm{mention}}(E^*\vert G^\tau)$ with the pruner loss $-\log P_{\mathrm{pruner}} \left(S^*\vert G^0 \right)$ according to \equref{ch_dwie:eq:pruner_loss}. 

\subsubsection{Single coreference resolution model}
\label{ch_dwie:subsubsec:single_coref}
The single-task end-to-end coreference model needs to detect mentions and correctly cluster them. Here again, the standard coreference loss \linebreak $-\log P_{\mathrm{coref}}\left(C^*\vert G^\tau\right)$ according to 
\equref{ch_dwie:eq:pruner3} and
\equref{ch_dwie:eq:coref2} is extended with the  pruner loss $-\log P_{\mathrm{pruner}} (S^*\vert G^0 )$.
This turned out essential for correctly predicting the singleton clusters. 

\subsubsection{Single relation extraction model}
The single relation extraction model is trained to detect mentions as well as the correct pairwise relations between mentions (i.e., without the coreference step). 
In order to train the pruner as well, the standard relation score is extended as described in \equref{ch_dwie:eq:pruner4}
before calculating the loss \linebreak $- \log P_{\mathrm{relation}} \left(R^* \vert G^\tau \right)$ based on \equref{ch_dwie:eq:rel2}. 

\section{Entity-centric metrics}
\label{ch_dwie:sec:metrics}

\begin{figure}
\centering
\includegraphics[width=0.9\columnwidth]{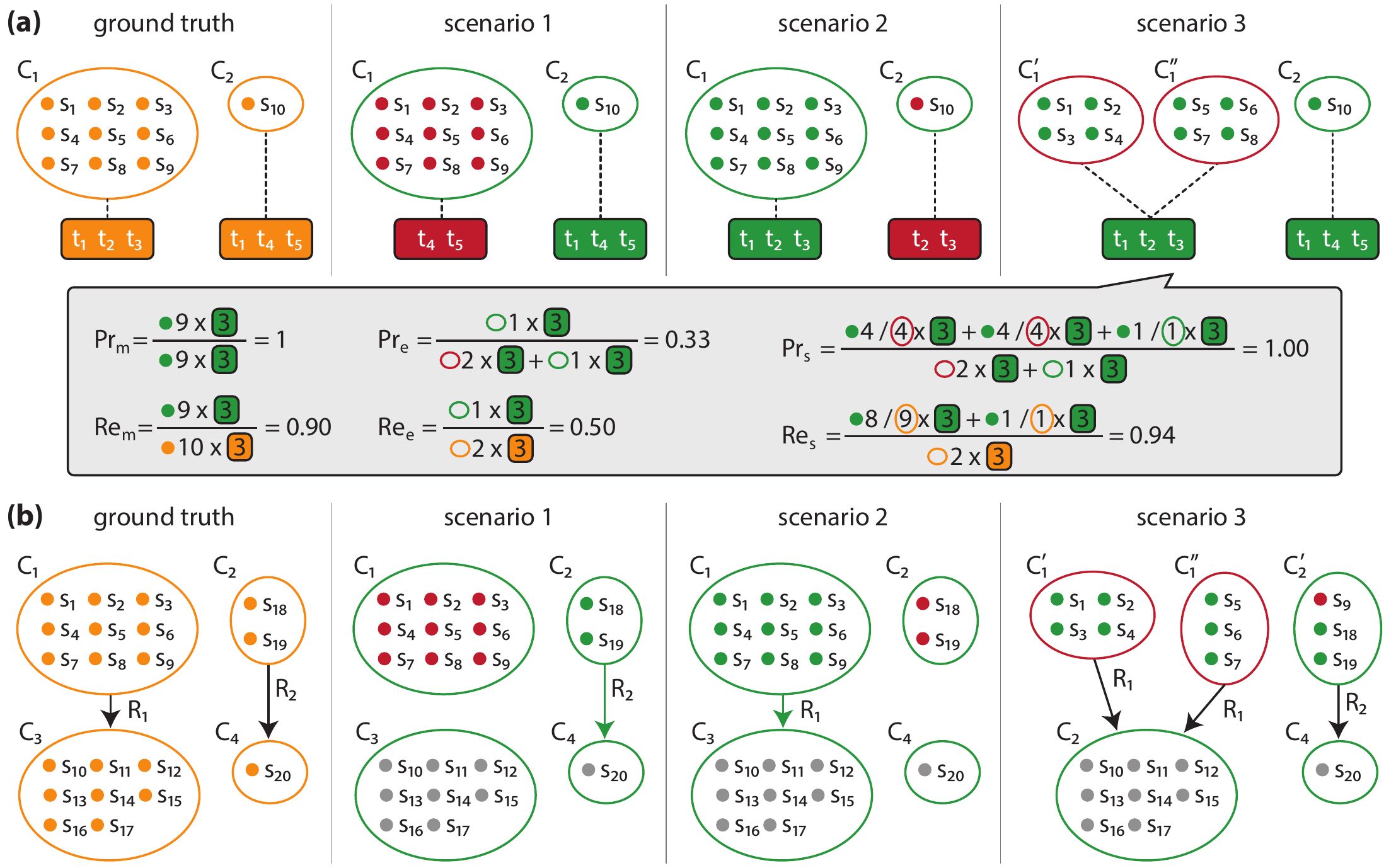}
\caption[Illustration of entity prediction scenarios for \textit{NER} and \textit{relation extraction} tasks]{Illustration of entity prediction scenarios for \textbf{(a)}~NER and \textbf{(b)}~relation extraction, with large clusters ($C_1, C_3$) and smaller ones ($C_2, C_4$). \emph{Scenario 1} erroneously labels the large one, \emph{scenario 2} incorrectly labels the small one, \emph{scenario 3} incorrectly splits up the large one and makes a mistake for one of its mentions, $s_9$. The formulas in the grey box illustrate the calculation of mention-level 
($\mathrm{Pr_m}$, $\mathrm{{Re}_m}$),
hard entity-level ($\mathrm{{Pr}_e}$, $\mathrm{{Re}_e}$) and soft entity-level ($\mathrm{{Pr}_s}$, $\mathrm{{Re}_s}$) precision and recall for NER in \emph{scenario 3}. Note that in~(b), the mention dots are colored for correct (green) and incorrect (red) relation heads only.}
\label{ch_dwie:fig:metric_motivation}
\end{figure}

Unlike the currently widespread datasets that use a mention-driven approach to annotate named entities \cite{sang2003introduction,derczynski2017results,weischedel2011ontonotes,bekoulis2017reconstructing}, relations \cite{augenstein2017semeval,song2015light,doddington2004automatic,ji2017overview,kim2003genia,luan2018multi, bekoulis2017reconstructing} and entity linking \cite{bentivogli2010extending,riedel2010modeling,hoffart2011robust}, \datasetname~is entirely entity-centric.
As explained before, we group entity mentions $s_i$ referring to the same entity into clusters $C_k$. While we can, and will, adopt the traditional coreference measures as defined by \cite{pradhan2014scoring} to judge this cluster formation, the NER and relation extraction (RE) evaluation (using precision, recall and $\mathrm{F_1}$) can be done either on
\begin{enumerate*}[(i)]
    \item mention level, or
    \item entity (cluster) level.
\end{enumerate*}
The first option however would have the metrics being dominated by the more frequently occurring entities, while the second would penalize mistakes in the clustering (since partially correctly identified clusters would be seen as completely incorrect). This is illustrated in \figref{ch_dwie:fig:metric_motivation} and the corresponding performance metrics in \tabref{ch_dwie:tab:metric_motivation}, where scenarios~1 and 2 highlight the effect of making labeling mistakes on the cluster level for different sizes, and scenario~3 highlights the pessimistic view of hard entity-level metrics in case of clustering mistakes. Note that we indicate the mention-level metrics with subscript $m$, while the (hard) entity-level metrics will have subscript with $e$.

Because the (hard) entity-level metrics in our opinion overly penalize clustering mistakes 
(cf.\ scenario~3), 
we propose a variant of entity-level evaluation which we term \emph{soft} entity-level metrics (denoted by subscript~$s$). 
Basically, instead of adopting a binary count of 1 (all mentions correct) or 0 (as soon as a single mention is missed) on an entity cluster level, we rather count the fraction of its mentions that are correctly labeled. This is illustrated in the formula part of \figref{ch_dwie:fig:metric_motivation}(a) for NER, and below we present the adopted formulas in detail. Note that in case clusters are completely predicted correctly, the soft entity-level metrics are the same as hard entity-level metrics (and thus avoid the metric being dominated by frequent mentions, as in the mention-level case).

The formal definition of the metrics depends on counting true positives $\textit{tp}_p(l)$ and $\textit{tp}_g(l)$, false positives $\textit{fp}(l)$, and false negatives $\textit{fn}(l)$ for a particular NER tag/relation type $l$, which are specified in \eqsref{ch_dwie:eq:soft_m_tp_generic}{ch_dwie:eq:soft_m_negatives_generic}. These and other notation definitions are summarized in \tabref{ch_dwie:tab:metric_symbols}. 
Further, note that we define two true positives for a particular label $l$, because of the potential difference between predicted and ground truth clusters: $\textit{tp}_p(l)$ 
sums
fractions of \emph{predicted} clusters and is used to calculate the precision $\mathrm{Pr}_\mathrm{s}$ in \equref{ch_dwie:eq:soft_metrics}, while $\textit{tp}_g(l)$ considers \emph{ground truth} clusters and is used for the recall $\mathrm{Re}_\mathrm{s}$ in \equref{ch_dwie:eq:soft_metrics}.
This allows us to preserve the \textit{cluster-based} relationships between true positives, false positives and false negatives as described for expressions $\textit{tp}_p(l) + \textit{fp}(l)$ and $\textit{tp}_g(l) + \textit{fn}(l)$ in \Tabref{ch_dwie:tab:metric_cluster_sum}. Thus our soft entity-level metrics are still cluster-based, while accounting for the mention-level predictions.

\renewcommand\baselinestretch{1}
{\begin{table}[]
\centering
\caption[Comparison of used metrics for NER and relation extraction tasks]{Comparison of different metrics for the example scenarios depicted in \figref{ch_dwie:fig:metric_motivation}, for \textbf{(a)}~NER and \textbf{(b)}~relation extraction.}
\label{ch_dwie:tab:metric_motivation}
\resizebox{1.0\textwidth}{!}{\begin{tabular}{l l ccc ccc ccc}
  \toprule 
& & \multicolumn{3}{c}{\textbf{Mention-Level}} &  \multicolumn{3}{c}{\textbf{Hard Ent-Level}} & \multicolumn{3}{c}{\textbf{Soft Ent-Level}} \\
\cmidrule(lr){3-5} \cmidrule(lr){6-8} \cmidrule(lr){9-11} 
 & & $\mathbf{Pr_m}$ & $\mathbf{Re_m}$ & $\mathbf{F_{1,m}}$
 & $\mathbf{Pr_s}$ & $\mathbf{Re_s}$ & $\mathbf{F_{1,s}}$
 & $\mathbf{Pr_e}$ & $\mathbf{Re_e}$ & $\mathbf{F_{1,e}}$
\\
\midrule
\multirow{4}{*}{\textbf{(a)~NER}} & Gr. Truth & 1.000 & 1.000 & 1.000 & 1.000 & 1.000 & 1.000 & 1.000 & 1.000 & 1.000 \\
& Scenario 1 & 0.143 & 0.100 & 0.118 & 0.600 & 0.500 & 0.545 &  0.600 & 0.500 & 0.545 \\
& Scenario 2 & 0.931 & 0.900 & 0.915 & 0.600 & 0.500 & 0.545 & 0.600 & 0.500 & 0.545 \\
& Scenario 3 & 1.000 & 0.900 & 0.947 & 0.333 & 0.500 & 0.400 & 1.000 & 0.944 & 0.971 \\
\midrule
\multirow{4}{*}{\textbf{(b)~RE}} & Gr. Truth & 1.000 & 1.000 & 1.000 & 1.000 & 1.000 & 1.000 & 1.000 & 1.000 & 1.000 \\
&Scenario 1 & 1.000 & 0.027 & 0.053 & 1.000 & 0.500 & 0.667 & 1.000 & 0.500 & 0.667 \\
&Scenario 2 & 1.000 & 0.973 & 0.986 & 1.000 & 0.500 & 0.667 & 1.000 & 0.500 & 0.667 \\
&Scenario 3 & 0.983 & 0.783 & 0.872 & 0.000 & 0.000 & 0.000 & 0.889 & 0.889 & 0.889 \\
\bottomrule
\end{tabular}}
\end{table}}


\begin{align}
    \textit{tp}_p(l) &= \sum\limits_{C_p \in P_C(l)}{\frac{|C_p \cap G_M(l) |}{|C_p|}}, & 
    \textit{tp}_g(l) &= \sum\limits_{C_g \in G_C(l)}{\frac{|C_g \cap P_M(l) |}{|C_g|}} \label{ch_dwie:eq:soft_m_tp_generic} \\ 
    \textit{fp}(l) &= |P_C(l)| - \textit{tp}_p(l), & 
    \textit{fn}(l) &= |G_C(l)| - \textit{tp}_g(l) \label{ch_dwie:eq:soft_m_negatives_generic}
\end{align}

Our soft entity-level precision, recall and F$_1$ metrics are 
formally defined as follows,
where $L$ refers to either the number of all possible tags
for NER or the number of all possible relation types
for RE:

\begin{equation}
    \mathrm{Pr}_\mathrm{s} = \frac{\sum\limits_{l=1}^L\textit{tp}_p(l)}{\sum\limits_{l=1}^L\textit{tp}_p(l) + \textit{fp}(l)},\qquad 
    \mathrm{Re}_\mathrm{s} = \frac{\sum\limits_{l=1}^L\textit{tp}_g(l)}{\sum\limits_{l=1}^L\textit{tp}_g(l) + \textit{fn}(l)}, 
    \qquad
    \mathrm{F}_\mathrm{1,s} = \frac{2 \cdot  \mathrm{Pr}_\mathrm{s} \cdot \mathrm{Re}_\mathrm{s}}{\mathrm{Pr}_\mathrm{s} + \mathrm{Re}_\mathrm{s}} \label{ch_dwie:eq:soft_metrics} 
\end{equation}

\begin{table}
    \centering
    \caption[The interpretation of metric components for NER and relation extraction tasks]{The relations between the weighted true positives by the size of predicted ($\textit{tp}_p(l)$) and ground truth ($\textit{tp}_g(l)$) entity clusters allows us to achieve the constraints needed for the denominators of precision ($\textit{tp}_p(l) + \textit{fp}(l)$) and recall ($\textit{tp}_p(l) + \textit{fn}(l)$) functions (\equref{ch_dwie:eq:soft_metrics}) in terms of the number of entity clusters.}
    \label{ch_dwie:tab:metric_cluster_sum}

    \small
    \begin{tabular}{lm{4.2cm}m{4.2cm}}
    \toprule
    \textbf{Expression} &
    \multicolumn{1}{c}{\textbf{(a)~Meaning for NER}} &
    \multicolumn{1}{c}{\textbf{(b)~Meaning for RE
    }}
    \\
    \midrule
    $\textit{tp}_p(l) + \textit{fp}(l)$ & 
    Number of \emph{predicted} entity clusters with tag $l$. &
    Number of \emph{predicted} relations of type $l$ between entity clusters.
    \\
    $\textit{tp}_g(l) + \textit{fn}(l)$ &
    Number of \emph{ground truth} entity clusters with tag $l$. & 
    Number of \emph{ground truth} relations of type $l$ between entity clusters.
    \\    
    \bottomrule
    \end{tabular}
\end{table}

\begin{table}
    \centering
    
    \caption[Definition of symbols involved in NER and relation extraction metric formulation]{Short definition of the symbols and expressions involved in our \textit{soft-entity level} metric formulation in  \eqsref{ch_dwie:eq:soft_m_tp_generic}{ch_dwie:eq:soft_metrics} for both NER and RE tasks.}
    \label{ch_dwie:tab:metric_symbols}
    
    \small
    \begin{tabular}{lm{4.3cm}m{4.6cm}}
    \toprule
    \multicolumn{1}{c}{\textbf{Symbol}} &    
    \multicolumn{1}{c}{\textbf{(a)~Meaning for NER}} &
    \multicolumn{1}{c}{\textbf{(b)~Meaning for RE}} \\
    \midrule
    $P_C(l)$ & Set of predicted entity clusters with tag $l$. & Set of predicted relations of type $l$ between the predicted entity clusters. \\
    $C_p \in P_C(l)$ & Set of predicted entity mentions for a particular entity cluster in $P_C(l)$. & Set of relations between the predicted entity mentions for a particular pair of related entity clusters in $P_C(l)$. \\
    $G_C(l)$ & Set of ground truth entity clusters annotated with tag $l$. & Set of ground truth relations of type $l$ between the ground truth entity clusters. \\
    $C_g \in G_C(l)$ & Set of ground truth entity mentions for a particular entity cluster in $G_C(l)$. & Set of relations between the ground truth entity mentions for a particular pair of related entity clusters in $G_C(l)$ \\
    $P_M(l)$ & Set of predicted entity mentions with tag $l$. & Set of predicted relations of type $l$ between the predicted entity mentions. \\ 
    $G_M(l)$ & Set of ground truth entity mentions annotated with tag $l$. & Set of ground truth relations of type $l$ between the ground truth entity mentions. \\    
    \midrule 
    $\textit{tp}_p(l)$ & Number of true positive predictions of tag $l$ on mentions re-weighted by predicted cluster sizes. & Number of true positive predictions of relation type $l$ between mentions re-weighted by the number of mention level relations between the connected pairs of predicted clusters. \\
    $\textit{tp}_g(l)$ & Number of true positive mention level predictions of tag $l$ re-weighted by ground truth cluster sizes. & Number of true positive predictions of relation type $l$ between mentions re-weighted by the number of mention level relations between the connected pairs of ground truth clusters. \\
    $\textit{fp}(l)$ & Number of false positive mention level predictions of tag $l$ re-weighted by predicted cluster sizes. & Number of false positive predictions of relation type $l$ between mentions re-weighted by the number of mention level relations between the connected pairs of predicted clusters. \\
    $\textit{fn}(l)$ & Number of false negative mentions with ground truth tag $l$ re-weighted by ground truth cluster sizes. & Number of false negative relations of type $l$ between mentions re-weighted by the number of mention level relations between the connected pairs of ground truth clusters. \\
    \bottomrule
    \end{tabular}
\end{table}

\section{Experimental results}
\label{ch_dwie:sec:results}
\subsection{Experimental setup}
\label{ch_dwie:sec:exp_setup}

We 
train and evaluate
our model 
as described in \secref{ch_dwie:sec:models}
on three tasks: NER, coreference, and 
 relation extraction (RE)
independently and jointly. We experiment with three main model 
variations:
\begin{enumerate}
    \item \textbf{Single}: Experiments 
    on individual tasks by training with the respective loss functions as described in \secref{ch_dwie:sec:single_task_models}. 
    \item \textbf{Joint}: Experiments 
    jointly on all three tasks using pre-trained \emph{GloVe representations}\footnote{\url{http://nlp.stanford.edu/data/glove.840B.300d.zip}} concatenated to character embeddings in the shared input layer (see \figref{ch_dwie:fig:model_architecture}). For training we use the joint loss defined in  \secref{ch_dwie:sec:joint_training}. 
    \item \textbf{Joint+BERT}: 
    as in the \emph{Joint} setting, experiments jointly on all three tasks, but using pre-trained BERT$_{\mathrm{BASE}}$ embeddings\footnote{\url{https://storage.googleapis.com/bert_models/2018_10_18/cased_L-12_H-768_A-12.zip}} 
    concatenated to the GloVe and character embeddings. 
    We use an input window size of 250 tokens and concatenate the last 2 hidden layers of BERT to get token representations. 
\end{enumerate}
Additionally, for each of the three model setups we experiment with the graph propagation techniques defined in \secref{ch_dwie:sec:graph_prop_model}. 
To maximize result consistency, we train each model 5 times and report the average of these 5 results for each of the experiments. 

We use a single-layer BiLSTM with forward and backward hidden states of 200 dimensions each.
All our FFNNs used to obtain confidence scores ($\mathcal{F}_{\mathrm{pruner}}$, $\mathcal{F}_{\mathrm{coref}}$, $\mathbfcal{F}_{\mathrm{mention}}$, $\mathbfcal{F}_{\mathrm{relation}}$, and $\mathcal{F}_{\mathrm{att}}$) have two 150-dimensional hidden layers trained with a dropout of 0.4. We set the maximum span width ${w_{\mathrm{max}}}$ to 5 and the pruner ratio to 0.2 of the total number of tokens in a document. For training, we use Adam with a learning rate of 
\num{1e-3} for 100 epochs with a linear decay of 0.1 starting at epoch 15. 
{\renewcommand\baselinestretch{1}\begin{table}
	\centering
\caption[Main results of~\datasetname~model]{Main results of the experiments grouped in three model setups: \begin{enumerate*}[(i)]
	\item \textit{Single} models trained individually,
	\item \textit{Joint} model trained using as input GloVe and character embeddings, and 
	\item \textit{Joint+BERT} model trained on BERT$_{\mathrm{BASE}}$ embeddings.
	\end{enumerate*} 
	 To report the results, we use MUC, CEAF$_\text{e}$, B$^\text{3}$ as well as the average (Avg.) of these three metrics for \textit{coreference resolution}. For NER and RE we use 
	 mention-level (F$_\text{1,m}$), hard entity-level (F$_\text{1,e}$), and soft entity-level (F$_\text{1,s}$) metrics described in \secref{ch_dwie:sec:metrics}. In bold we mark the best results for each model setup, the best overall results are underlined. Note that the metrics are expressed in percentage points.}
	\label{ch_dwie:tab:main_results}
	\setlength{\tabcolsep}{4pt}
	\renewcommand{\arraystretch}{1.0}
	\resizebox{1.0\textwidth}{!}{\begin{tabular}{l cccc c ccc c ccc}
	\toprule
	&	\multicolumn{4}{c}{\textbf{Coreference} $\mathbf{F_1}$} && \multicolumn{3}{c}{\textbf{NER} $\mathbf{F_1}$} && \multicolumn{3}{c}{\textbf{RE} $\mathbf{F_1}$} \\ 
\cmidrule(lr){2-5} \cmidrule(lr){7-9} \cmidrule(lr){11-13}
	 \textbf{Model Setup} & 
	 \textbf{MUC} & 
	 \textbf{CEAF}$\mathbf{_{e}}$ & 
	 $\mathbf{B^3}$ & 
	 \textbf{Avg.} & 
	 \hspace{.5em} & 
	 $\mathbf{F_{1,m}}$ & 
	 $\mathbf{F_{1,e}}$ & 
	 $\mathbf{F_{1,s}}$ & 
	 \hspace{.5em} & 
	 $\mathbf{F_{1,m}}$ & 
	 $\mathbf{F_{1,e}}$ & 
	 $\mathbf{F_{1,s}}$ \\ 
	 \midrule 
	Single & 92.8 & 90.9 & 88.2 & 90.6 && 85.7 & - & - && 68.2 & - & - \\ 
 	\ \ \propformat{+AttProp} & \textbf{93.2} & \textbf{91.5} & \textbf{88.7} & \textbf{91.1} && \textbf{87.1} & - & - && \textbf{71.3} & - & - \\ 
  	\ \ \propformat{+CorefProp} & 92.8 & 90.9 & 88.3 & 90.7 && - & - & - && - & - & - \\ 
 	\ \ \propformat{+RelProp} & - & - & - & - && - & - & - && 68.2 & - & - \\ 
 	 \midrule
	Joint & 92.5 & \textbf{90.5} & \textbf{87.3} & \textbf{90.1} && 85.4 & 71.7 & 84.4 && 68.1 & 46.8 & 66.5 \\
 	\ \ \propformat{+AttProp} & 92.3 & 90.4 & 87.3 & 90.0 && 87.1 & 72.9 & \textbf{86.1} && \textbf{72.1} & \textbf{50.4} & \textbf{72.1} \\ 
 	\ \ \propformat{+CorefProp} & 92.3 & 90.3 & 87.2 & 89.9 && \textbf{87.2} & \textbf{73.2} & 86.0 && 71.6 & 50.2 & 71.0 \\ 
 	\ \ \propformat{+RelProp} & \textbf{92.6} & 90.2 & 86.8 & 89.9 && 86.7 & 72.4 & 85.2 && 69.5 & 48.2 & 68.8 \\ 
 	 \midrule
	Joint+BERT & \underline{\textbf{93.8}} & \underline{\textbf{92.1}} & \underline{\textbf{89.0}} & \underline{\textbf{91.6}} && 87.6 & 74.2 & 86.4 && 70.6 & 48.7 & 68.9 \\ 
 	\ \ \propformat{+AttProp} & 93.2 & 91.4 & 88.6 & 91.1 && \underline{\textbf{88.8}} & 74.2 & \underline{\textbf{87.7}} && 72.3 & \underline{\textbf{50.4}} & \underline{\textbf{73.0}} \\ 
 	\ \ \propformat{+CorefProp} & 93.5 & 91.8 & 88.7 & 91.3 && 88.7 & 74.4 & 87.4 && \underline{\textbf{72.7}} & 50.0 & 71.9 \\ 
 	\ \ \propformat{+RelProp} & 93.7 & 91.8 & 88.7 & 91.4 && 88.4 & \underline{\textbf{74.8}} & 87.0 && 72.0 & 49.9 & 71.4 \\ 
 	\bottomrule 
	\end{tabular}}
\end{table}}

\subsection{Results and analyses}
\label{ch_dwie:sec:results_and_analysis}
\begin{figure}
    \centering
    \includegraphics[width=1.0\columnwidth]{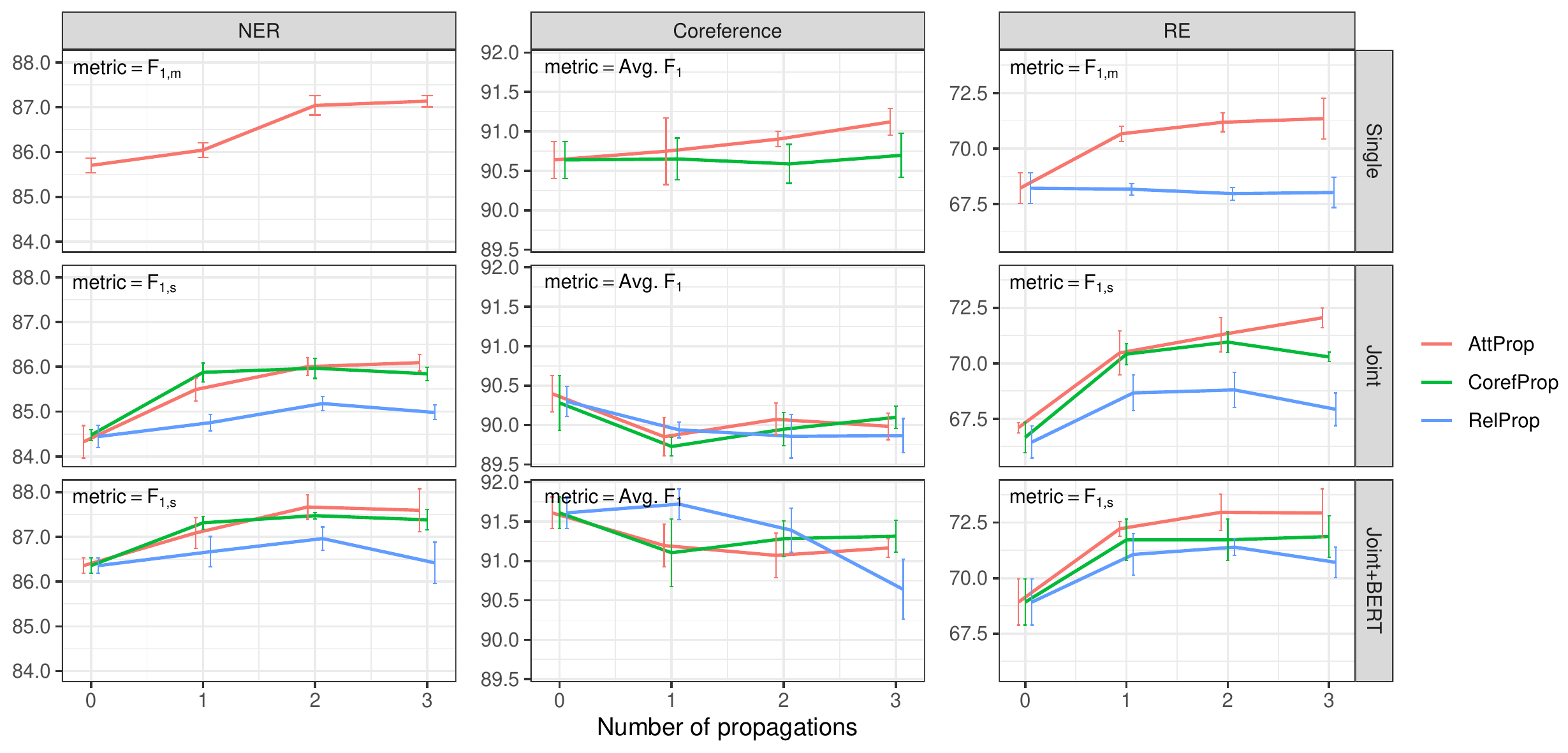}
    \caption[Impact of graph propagations on performance metrics]{Impact of \propformat{AttProp}, \propformat{CorefProp} and \propformat{RelProp} graph propagations on performance metrics for each of the \emph{Single}, \emph{Joint} and \emph{Joint+BERT} model setups. Note the different Y-axis scales.}
    \label{ch_dwie:fig:res_prop}
\end{figure}

 \Tabref{ch_dwie:tab:main_results} gives an overview of the results achieved in \textit{Single} as well as \textit{Joint} and \textit{Joint + BERT} setups. Additionally, 
 \figref{ch_dwie:fig:res_prop}
 illustrates the impact of the number of graph propagation iterations for each of the span graph propagation methods on the final results.
 
 First, we observe a general improvement in all our \emph{Single} tasks when using graph propagation techniques. More specifically, 
 our proposed latent 
 \propformat{AttProp}
 achieves superior results compared to the relation (\propformat{RelProp}) and coreference (\propformat{CorefProp}) propagations when added to the \textit{Single} setup. The biggest 
 improvement across iterations (see \figref{ch_dwie:fig:res_prop}) is for the single 
 RE
 task 
 mention-level $\mathrm{F_{1,m}}$ score with a boost of ${\sim}3$ percentage points when incorporating \propformat{AttProp}. We also observe an improvement of ${\sim}1.5$ percentage points in $\mathrm{F_{1,m}}$ for the NER task and a consistent but smaller improvement of $0.5$ $\mathrm{F_1}$ percentage points for the coreference task. These results illustrate the effectiveness of \propformat{AttProp} when applied to single task models. 

A further improvement in results is achieved by training our model \emph{jointly} (see the {Joint} setup in \tabref{ch_dwie:tab:main_results} and graphs in \figref{ch_dwie:fig:res_prop}) for NER and 
RE
tasks. This illustrates that, besides the positive effect of neural graph propagation on single task models, training our model jointly has an additional benefit by exploiting the interaction between tasks. In particular, this effect can be seen for RE, where our \textit{Joint} model achieves a boost in performance of $0.8$ percentage points for the mention-level $\mathrm{F_{1,m}}$ metric compared to the best result for the \textit{Single} setup. Furthermore, our \propformat{AttProp} graph propagation method achieves the best performance on all the metrics for the RE task in the \textit{Joint} setting with up to $\sim{5.5}$ percentage points improvement in our newly proposed $\mathrm{F_{1,s}}$ metric. 
Additionally, we observe a beneficial effect of graph propagation for the
NER
task in the \textit{Joint} setup with slightly better results for the $\mathrm{F_{1,m}}$ metric compared to the \textit{Single} setting. Our \propformat{AttProp} technique performs on par with \propformat{CorefProp}, outperforming the latter by a small margin in terms of $\mathrm{F_{1,s}}$ metric. 

Similarly to the \textit{Joint} model variation, we observe benefits when using graph propagation techniques in the \textit{Joint+BERT} models.
\tabref{ch_dwie:tab:analysis_deltas_props} illustrates the deltas in performance for the NER and 
relation extraction 
tasks. 
This way, we can see more clearly the difference in impact of our neural message passing methods grouped by the model setup and metric type.  
First, we observe that the general performance boost from using graph propagation techniques is lower in \textit{Joint+BERT} than in the \textit{Joint} setup. We hypothesize that this effect is due to the fact that BERT itself has a better long-range context extraction due to the attention-based mechanism, which spans the input window as opposed to purely local (non-contextualized) GloVe embeddings used in the \textit{Joint} setting. This is in line with the findings in \cite{han2020novel}, \cite{wadden2019entity}, and \cite{wu2019enriching} that show the advantage of using 
large BERT input window sizes 
to produce better IE results.
Second, we observe that our \propformat{AttProp} method achieves consistently superior performance on our proposed soft entity-level metric $\mathrm{F_{1,s}}$, capturing thus better the mention-based predictions as weighted by their cluster sizes. Finally, from \tabref{ch_dwie:tab:analysis_deltas_props}(b) we notice that adding BERT to our joint model does not affect the boost in performance caused by the \propformat{RelProp} method for 
relation extraction. 
We hypothesize that this is due to the fact that \propformat{RelProp} propagation can capture relational semantics that 
goes beyond
BERT's contextual span representation similarity (which mainly drives the positive impact of \textit{Joint+BERT}). 
{\renewcommand\baselinestretch{1}\begin{table}
\centering
\caption[Deltas of improvement in performance for each graph propagation method]{Deltas of improvement in performance for each of the graph propagation methods (\propformat{AttProp}, \propformat{CorefProp}, \propformat{RelProp}) in $\mathrm{F_1}$ scores for \textbf{(a)}~NER and \textbf{(b)}~relation extraction tasks.}
\label{ch_dwie:tab:analysis_deltas_props}
\begin{tabular}{c l ccc c ccc}
\toprule
& & 
\multicolumn{3}{c}{\textbf{Joint}} & & 
\multicolumn{3}{c}{\textbf{Joint+BERT}} \\ 
\cmidrule(lr){3-5} \cmidrule(lr){7-9} 
& & 
$\mathbf{F_{1,m}}$ & 
$\mathbf{F_{1,e}}$ & 
$\mathbf{F_{1,s}}$ & 
&
$\mathbf{F_{1,m}}$ & 
$\mathbf{F_{1,e}}$ & 
$\mathbf{F_{1,s}}$ \\
\midrule 
\multirow{3}{*}{\textbf{(a)~NER}} &
$\Delta\mathrm{\ }$\propformat{AttProp} & 1.69 & 1.18 & \textbf{1.67} &  & \textbf{1.16} & $\mathrm{-0.02}$ & \textbf{1.31} \\ 
& $\Delta\mathrm{\ }$\propformat{CorefProp} & \textbf{1.78} & \textbf{1.50} & 1.54 &  & 1.05 & 0.20 & 1.02 \\ 
& $\Delta\mathrm{\ }$\propformat{RelProp} & 1.33 & 0.70 & 0.75 &  & 0.78 & \textbf{0.56} & 0.60 \\ 
\midrule 
\multirow{3}{*}{\textbf{(b)~RE}} &
$\Delta\mathrm{\ }$\propformat{AttProp} & \textbf{3.97} & \textbf{3.62} & \textbf{5.56} &  & 1.66 & \textbf{1.69} & \textbf{4.05} \\ 
& $\Delta\mathrm{\ }$\propformat{CorefProp} & 3.48 & 3.45 & 4.47 &  & \textbf{2.02} & 1.29 & 2.95 \\ 
& $\Delta\mathrm{\ }$\propformat{RelProp} & 1.35 & 1.47 & 2.32 &  & 1.37 & 1.20 & 2.48 \\ 
\bottomrule
\end{tabular}
\end{table}}

Unlike for the NER and 
RE
tasks, where we observe a consistent positive impact of span graph propagation and joint modeling across all our experiments, the impact on the \emph{coreference} task is not clear. Our experiments on \textit{Single} setup show small, but constant improvement of the Avg.-$\mathrm{F_1}$ score with the number of \propformat{AttProp} propagation iterations (see \figref{ch_dwie:fig:res_prop}).
However, in our \textit{Joint} and \textit{Joint+BERT} setups the graph propagation appears to not have any positive impact on Avg.-$\mathrm{F_1}$ coreference scores.
We hypothesize that the main reason for this phenomenon lies in the coreference annotations in \datasetname: since we only annotate clusters of proper nouns, leaving out the nominal (\eg ``the prime minister'') and anaphoric expressions (\eg ``he'', ``she'', ``they'', etc), there might be little to no additional benefit in propagating 
information between co-referenced entity mentions, 
since the representation of proper nouns likely is not much influenced by textual context
(\eg the span ``Merkel'' can have very similar span representation to ``Angela Merkel'', gaining nothing in adding contextual graph propagation). 

Additionally, we explore in more detail the effect of the number of \propformat{AttProp}, \propformat{CorefProp}, \propformat{RelProp} graph propagation 
iterations on the final $\mathrm{F_1}$ score of all the tasks in 
\figref{ch_dwie:fig:res_prop}.
We observe that the number of iterations have a decreasing effect on the improvement of performance for the NER and 
RE
tasks.
Furthermore, the positive effect of \propformat{CorefProp} and \propformat{RelProp} tends to saturate or even become negative after 1 or 2 iterations. This is in line with findings of \cite{luan2019general} on other datasets, where the performance peak is usually achieved at 2 graph propagation iterations.
For our \propformat{AttProp} however, we observe that the positive effect of additional iterations tends to persist longer, particularly on the \textit{Joint} setup where the positive effect of \propformat{AttProp} seems to be still growing after the last iteration~(3) in our experiments.

\section{Conclusion and future work}
\label{ch_dwie:sec:conclusion}

In this work we introduced \datasetname, a manually annotated multi-task dataset that comprises Named Entity Recognition, Coreference, Relation Extraction and Entity Linking as main tasks. We highlight how \datasetname~is different from the mainstream datasets by focusing on document-level and entity-centric annotations. This also makes the predictions on this dataset more challenging by having not only to consider explicit, but also implicit document-level interactions between entities. 
Furthermore, we showed how Graph Neural Networks can help to tackle this issue by propagating local contextual mention span information on a document level for a single task as well as across the tasks on the \datasetname~dataset. 
We experiment with known graph propagation techniques driven by the scores of the coreference resolution (\propformat{CorefProp}) and relation extraction (\propformat{RelProp}) components, as well as introduced a new latent task-independent attention-based graph propagation method (\propformat{AttProp}). We demonstrated that, without relying on the task-specific scorers, \propformat{AttProp} can boost the performance of single-task as well as joint models, performing on par and even outperforming significantly in some scenarios the \propformat{RelProp} and \propformat{CorefProp} graph propagations.  

In future work we will aim to integrate an entity linking component into our joint architecture. As a consequence, we expect to obtain a further boost in performance of different tasks included in \datasetname~by taking advantage of the information coming from Wikipedia 2018, the reference knowledge base for the entity linking annotations. Conversely, we conjecture that the results of the entity linking component can be improved when training it jointly with other tasks, such as NER and coreference resolution. Finally, we plan extending the coreference annotations to include nominal and anaphoric expressions. We expect that including these diverse mention types, whose initial span embedding representation can be different from coreferenced named entities, will make our coreference resolution task more challenging, allowing to investigate further the potential benefits of using graph-based neural networks. 

\section*{Acknowledgements}
\noindent Part of the research leading to these results has received funding from
\begin{enumerate*}[(i)]
\item the European Union's Horizon 2020 research and innovation programme under grant agreement no.\ 761488 for the CPN project,\footnote{\url{https://www.projectcpn.eu/}} and
\item the Flemish Government under the ``Onderzoeksprogramma Artifici\"{e}le Intelligentie (AI) Vlaanderen'' programme.
\end{enumerate*}
\\ 
\\ 
\\
\\
\\ 
\\ 
\\
\\
\\ 
\\ 
\\
\\
\\ 
\\ 
\\
\\
\\ 
\\ 
\\
\\ 
\\

\section*{Appendix}
\label{ch_dwie:sec:appendix}

\renewcommand\baselinestretch{1}\small
\subsection*{Dataset insights}
\setlength{\intextsep}{1pt plus 1pt minus 1pt}
\label{ch_dwie:app:dataset_insights}
\begin{table}[H]
\centering
\caption[Statistics depicting the hierarchical structure of entity types]{Statistics depicting the hierarchical structure of entity types described in \secref{ch_dwie:sec:schema_driven_pass}. Only the most frequent entity types/subtypes are shown (\% Mentions $>$ 0.5\%)}
\resizebox{0.9\textwidth}{!}{
\begin{tabular}{lllll} 
\toprule
\textbf{Entity Type} & \textbf{\# Entities} & \textbf{\% Entities} & \textbf{\# Mentions} & \textbf{\% Mentions} \\ 
\midrule
	\textit{\hspace{0.00cm}ENTITY} & \textit{13,151} & \textit{56.9}\% & \textit{30,719} & \textit{70.8}\% \\ 
	\hspace{0.30cm}location & 4,957 & 21.4\% & 11,548 & 26.6\% \\ 
	\hspace{0.60cm}gpe & 3,965 & 17.1\% & 9,830 & 22.7\% \\ 
	\hspace{0.90cm}gpe0 & 2,225 & 9.6\% & 6,559 & 15.1\% \\ 
	\hspace{0.90cm}gpe2 & 1,497 & 6.5\% & 2,873 & 6.6\% \\ 
	\hspace{0.90cm}gpe1 & 244 & 1.1\% & 406 & 0.9\% \\ 
	\hspace{0.60cm}regio & 479 & 2.1\% & 916 & 2.1\% \\ 
	\hspace{0.60cm}facility & 259 & 1.1\% & 385 & 0.9\% \\ 
	\hspace{0.30cm}organization & 3,434 & 14.8\% & 8,165 & 18.8\% \\ 
	\hspace{0.60cm}media & 659 & 2.8\% & 984 & 2.3\% \\ 
	\hspace{0.60cm}igo & 547 & 2.4\% & 1,992 & 4.6\% \\ 
	\hspace{0.90cm}so & 171 & 0.7\% & 912 & 2.1\% \\ 
	\hspace{0.60cm}party & 381 & 1.6\% & 949 & 2.2\% \\ 
	\hspace{0.60cm}company & 368 & 1.6\% & 932 & 2.1\% \\ 
	\hspace{0.60cm}sport\_team & 367 & 1.6\% & 1,106 & 2.5\% \\ 
	\hspace{0.60cm}governmental\_organization & 342 & 1.5\% & 636 & 1.5\% \\ 
	\hspace{0.90cm}agency & 228 & 1.0\% & 444 & 1.0\% \\ 
	\hspace{0.60cm}armed\_movement & 108 & 0.5\% & 374 & 0.9\% \\ 
	\hspace{0.30cm}person & 3,390 & 14.7\% & 8,259 & 19.0\% \\ 
	\hspace{0.60cm}politician & 1,184 & 5.1\% & 3,326 & 7.7\% \\ 
	\hspace{0.90cm}head\_of\_state & 380 & 1.6\% & 1,271 & 2.9\% \\ 
	\hspace{0.90cm}head\_of\_gov & 247 & 1.1\% & 673 & 1.6\% \\ 
	\hspace{0.90cm}minister & 217 & 0.9\% & 458 & 1.1\% \\ 
	\hspace{0.60cm}sport\_player & 405 & 1.8\% & 844 & 1.9\% \\ 
	\hspace{0.60cm}artist & 260 & 1.1\% & 586 & 1.4\% \\ 
	\hspace{0.60cm}politics\_per & 209 & 0.9\% & 457 & 1.1\% \\ 
	\hspace{0.60cm}manager & 104 & 0.4\% & 297 & 0.7\% \\ 
	\hspace{0.60cm}offender & 75 & 0.3\% & 347 & 0.8\% \\ 
	\hspace{0.30cm}misc & 823 & 3.6\% & 1,646 & 3.8\% \\ 
	\hspace{0.60cm}work\_of\_art & 174 & 0.8\% & 247 & 0.6\% \\ 
	\hspace{0.30cm}event & 354 & 1.5\% & 701 & 1.6\% \\ 
	\hspace{0.60cm}sport\_competition & 183 & 0.8\% & 410 & 0.9\% \\ 
	\hspace{0.30cm}ethnicity & 84 & 0.4\% & 242 & 0.6\% \\ 
	\textit{\hspace{0.00cm}VALUE} & \textit{5,903} & \textit{25.5}\% & \textit{7,104} & \textit{16.4}\% \\ 
	\hspace{0.30cm}time & 2,907 & 12.6\% & 3,608 & 8.3\% \\ 
	\hspace{0.30cm}role & 2,390 & 10.3\% & 2,865 & 6.6\% \\ 
	\hspace{0.30cm}money & 606 & 2.6\% & 631 & 1.5\% \\ 
	\textit{\hspace{0.00cm}OTHER} & \textit{2,724} & \textit{11.8}\% & \textit{5,482} & \textit{12.6}\% \\ 
	\hspace{0.30cm}gpe0-x & 1,596 & 6.9\% & 3,827 & 8.8\% \\ 
	\hspace{0.30cm}footer & 413 & 1.8\% & 413 & 1.0\% \\ 
	\hspace{0.30cm}loc-x & 353 & 1.5\% & 585 & 1.3\% \\ 
	\hspace{0.30cm}religion-x & 235 & 1.0\% & 486 & 1.1\% \\ 
\midrule
\textbf{TOTAL} & 23,130 & 100.0\% & 43,373 & 100.0\% \\ 
\bottomrule
\end{tabular}
}
\label{ch_dwie:tab:entities_stats}
\end{table}

\newgeometry{margin=1.5cm}
\begin{landscape}
\begin{table}[h]
    \centering
\caption[Illustration of NER entity types in \datasetname]{Illustration of NER entity types in \datasetname. Each cells contains possible entity subtypes (of different hierarchy levels) corresponding to the respective parent entity type (column) and topic (row).}
\label{ch_dwie:tab:ner_labels_matrix}
\footnotesize
    \resizebox{\columnwidth}{!}{\begin{tabular}{l>{\raggedright\arraybackslash}p{5.4cm}>{\raggedright\arraybackslash}p{5.4cm}>{\raggedright\arraybackslash}p{2.4cm}>{\raggedright\arraybackslash}p{2.4cm}>{\raggedright\arraybackslash}p{7.4cm}}
\toprule
         & \multicolumn{1}{c}{\entityname{person}}  & \multicolumn{1}{c}{\entityname{organization}}  & \multicolumn{1}{c}{\entityname{event}}  & \multicolumn{1}{c}{\entityname{location}}  & \multicolumn{1}{c}{\entityname{misc}} \\
         \midrule
         \topicname{politics} & head\_of\_gov, head\_of\_state, minister, politician\_regional, politician\_local, politician\_national, candidate, politician, politics\_per, activist, gov\_per & politics\_institution, politics\_org, party, ngo, igo, so, policy\_institute, movement, agency, ministry, military\_alliance & summit\_meeting, scandal, politics\_event & politics\_facility & politics\_misc, project, treaty, report \\ 
         \midrule
         \topicname{culture} & character, culture\_per, artist, writer, actor, filmmaker, musician, photographer & music\_band, culture\_org, theatre\_org, dance\_org & festival, film\_festival & culture\_facility & art\_title, culture\_title, exhibition\_title, culture\_misc, work\_of\_art, book\_title, film\_title, tv\_title, music\_title, theatre\_title, musical\_title, film\_award, book\_award, music\_award, tv\_award, column\_title, game, comic, radio\_title, dance\_title, opera \\ 
         \midrule
         \topicname{education} & teacher, education\_per, education\_student & education\_org &  & education\_facility & education\_study \\ 
         \midrule
         \topicname{religion} & deity, clergy & religion\_org & religious\_event & religion\_facility & religion, religion\_misc \\ 
         \midrule
         \topicname{human} & royalty &  &  &  & film\_award, book\_award, award, music\_award, tv\_award, sport\_award \\ 
         \hline 
\topicname{conflict} & military\_personnel, military\_rebel & army, military\_alliance, armed\_movement & war, protest & military\_facility & military\_equipment, military\_mission \\ 
         \midrule 
\topicname{media} & journalist & media &  &  &  \\ 
         \midrule 
\topicname{science} & researcher, science\_per & research\_center &  &  & species, research\_journal, technology \\ 
         \midrule 
\topicname{sport} & sport\_player, sport\_coach, sport\_head, sport\_referee, sport\_person & sport\_team, sport\_org & sport\_competition & sport\_facility & sport\_award \\ 
         \midrule 
\topicname{labor} & union\_head, union\_member, union\_rep, union\_per & union &  &  &  \\ 
         \midrule 
\topicname{business} & manager, employee, business\_per & company, business\_org, brand, trade\_fair, market\_exchange, advocacy &  & business\_facility & product, market\_index, business\_misc \\ 
         \midrule 
\topicname{health} & health\_per & health\_org &  & health\_facility & health\_disease, health\_drug \\ 
         \midrule 
\topicname{justice} & offender, advisor, victim, judge, police\_per, justice\_per & court, criminal\_org, police\_org, justice\_org &  & prison & justice\_misc, case \\ 
         \midrule 
\topicname{weather} &  &  & storm &  & \\ 
         \bottomrule
       \end{tabular}}
\end{table}
\end{landscape}
\restoregeometry
\clearpage

\renewcommand\baselinestretch{1}\small \Tabref{ch_dwie:tab:entity_linking_stats} describes the statistics of linked entities with respect to the total number of entities in each of the \textit{Entity} subtypes. The columns \textit{\% Linked Entities} and \textit{\% Linked Mentions} indicate the percentage of annotated linked entities and mentions with respect to the total number of annotated entities/mentions in a particular \textit{Entity} type category. Furthermore, we calculate two accuracies on test split when linking the entity mention with the most frequent entity link used either in \datasetname: \begin{enumerate*}[(i)]
    \item training set of \datasetname~dataset (``Acc. Prior Train''), or
    \item Wikipedia corpus (``Acc. Prior Wiki'')
\end{enumerate*}. 
Overall, using prior linking annotations from Wikipedia gives 9 percentage points better performance (79.0\%) than when using train set (70.0\%). This difference is explained by the fact that Wikipedia has much larger corpus to calculate the prior linking information from. Nevertheless, we still observe that for some entity types such as \textit{sport\_team} and  \textit{media} the accuracy based on \datasetname~training set prior is higher. This suggests the use of domain-specific language to refer to some entities in \datasetname~not used in a more general Wikipedia domain.
\setlength{\intextsep}{12.0pt plus 2.0pt minus 2.0pt}
\begin{table}[!htb]
\centering
\caption[Entity linking statistics in \datasetname]{Entity linking statistics, only the top 5 types and subtypes with largest number of linked entities are showed. The \textit{total} is calculated on all the entity types. The accuracy (both for most likely prior links on train and Wiki corpora) is computed on test set. }
\resizebox{0.9\textwidth}{!}{\begin{tabular}{lllllll} 
\toprule
\begin{tabular}{@{}c@{}}\textbf{\ } \\ \textbf{Entity Type} \end{tabular}& \begin{tabular}{@{}l@{}} \textbf{\# Linked} \\ \textbf{Entities} \end{tabular} & \begin{tabular}{@{}l@{}} \textbf{\% Linked} \\ \textbf{Entities} \end{tabular} & \begin{tabular}{@{}l@{}} \textbf{\# Linked} \\  \textbf{Mentions} \end{tabular} & \begin{tabular}{@{}l@{}} \textbf{\% Linked} \\  \textbf{Mentions} \end{tabular} & \begin{tabular}{@{}l@{}} \textbf{Acc. Prior} \\ \textbf{Train} \end{tabular} & \begin{tabular}{@{}l@{}} \textbf{Acc. Prior} \\ \textbf{Wiki} \end{tabular} \\ 
\midrule
	\textit{\hspace{0.00cm}LOCATION} & \textit{4,863} & \textit{98.1}\% & \textit{11,496} & \textit{99.5}\% & \textit{85.7\%} & \textit{92.9\%} \\ 
	\hspace{0.30cm}gpe & 3,938 & 99.3\% & 9,810 & 99.8\% & 89.8\% & 95.6\% \\ 
	\hspace{0.30cm}regio & 456 & 95.2\% & 889 & 97.1\% & 83.3\% & 76.3\% \\ 
	\hspace{0.30cm}facility & 229 & 88.4\% & 381 & 99.0\% & 19.7\% & 73.8\% \\ 
	\hspace{0.30cm}waterbody & 90 & 98.9\% & 145 & 100.0\% & 83.3\% & 91.7\% \\ 
	\hspace{0.30cm}district & 37 & 94.9\% & 45 & 100.0\% & 33.3\% & 33.3\% \\ 
	\textit{\hspace{0.00cm}ORGANIZATION} & \textit{3,145} & \textit{91.6}\% & \textit{8,029} & \textit{98.3}\% & \textit{69.8\%} & \textit{70.8\%} \\ 
	\hspace{0.30cm}media & 622 & 94.4\% & 979 & 99.5\% & 81.8\% & 59.5\% \\ 
	\hspace{0.30cm}igo & 525 & 96.0\% & 1,952 & 98.0\% & 76.4\% & 78.8\% \\ 
	\hspace{0.30cm}party & 358 & 94.0\% & 897 & 94.5\% & 77.5\% & 66.7\% \\ 
	\hspace{0.30cm}company & 320 & 87.0\% & 923 & 99.0\% & 67.6\% & 89.7\% \\ 
	\hspace{0.30cm}sport\_team & 366 & 99.7\% & 1,105 & 99.9\% & 71.0\% & 47.5\% \\ 
	\textit{\hspace{0.00cm}PERSON} & \textit{2,627} & \textit{77.5}\% & \textit{8,217} & \textit{99.5}\% & \textit{45.7\%} & \textit{69.4\%} \\ 
	\hspace{0.30cm}politician & 1,162 & 98.1\% & 3,324 & 99.9\% & 66.0\% & 78.1\% \\ 
	\hspace{0.30cm}sport\_player & 404 & 99.8\% & 843 & 99.9\% & 34.4\% & 71.3\% \\ 
	\hspace{0.30cm}artist & 246 & 94.6\% & 567 & 96.8\% & 0.0\% & 29.4\% \\ 
	\hspace{0.30cm}politics\_per & 126 & 60.3\% & 456 & 99.8\% & 23.7\% & 42.1\% \\ 
	\hspace{0.30cm}manager & 58 & 55.8\% & 296 & 99.7\% & 22.2\% & 33.3\% \\ 

	\textit{\hspace{0.00cm}MISC} & \textit{607} & \textit{73.8}\% & \textit{1,532} & \textit{93.1}\% & \textit{58.4\%} & \textit{73.4\%} \\ 
	\hspace{0.30cm}work\_of\_art & 142 & 81.6\% & 246 & 99.6\% & 0.0\% & 100.0\% \\ 
	\hspace{0.30cm}award & 72 & 80.0\% & 186 & 94.9\% & 63.6\% & 81.8\% \\ 
	\hspace{0.30cm}treaty & 60 & 74.1\% & 149 & 99.3\% & 66.7\% & 50.0\% \\ 
	\hspace{0.30cm}product & 50 & 76.9\% & 146 & 98.6\% & 52.0\% & 92.0\% \\ 
	\hspace{0.30cm}species & 10 & 25.0\% & 14 & 18.4\% & 0.0\% & 100.0\% \\ 
	\textit{\hspace{0.00cm}EVENT} & \textit{320} & \textit{90.4}\% & \textit{683} & \textit{97.4}\% & \textit{49.4\%} & \textit{67.1\%} \\ 
	\hspace{0.30cm}sport\_competition & 163 & 89.1\% & 397 & 96.8\% & 64.6\% & 87.5\% \\ 
	\hspace{0.30cm}summit\_meeting & 15 & 68.2\% & 37 & 92.5\% & 100.0\% & 100.0\% \\ 
	\hspace{0.30cm}holiday & 21 & 95.5\% & 39 & 97.5\% & 100.0\% & 100.0\% \\ 
	\hspace{0.30cm}history & 17 & 89.5\% & 30 & 100.0\% & 100.0\% & 100.0\% \\ 
	\hspace{0.30cm}protest & 14 & 100.0\% & 22 & 100.0\% & 80.0\% & 100.0\% \\ 
\midrule
\textbf{TOTAL} & \textbf{13,086} & \textbf{56.6\%} & \textbf{28,482} & \textbf{65.7\%} & \textbf{70.0\%} & \textbf{79.0\%} \\ 
\bottomrule
\end{tabular}}
\label{ch_dwie:tab:entity_linking_stats}
\end{table}

\clearpage

\begin{table}[t]
\centering
\caption[Main named entity tag categories in \datasetname]{Main named entity tag categories with statistics of the number and \% of covered entities and mentions as well as the number of classes in each and average number of labels per entity cluster.}
\resizebox{1.0\textwidth}{!}{
\begin{tabular}{llllllll}
\toprule
\begin{tabular}{@{}l@{}}\textbf{Entity Tag} \\ \textbf{Category}\end{tabular} & \textbf{\# Entities} & \textbf{\% Entities} & \textbf{\# Mentions} & \textbf{\% Mentions} & \textbf{\# Classes} & \begin{tabular}{@{}l@{}}\textbf{Labels per} \\ \textbf{Entity}\end{tabular} \\\midrule
	type & 21,745 & 94.0\% & 43,122 & 99.4\% & 174 & 2.9 \\ 
	topic & 7,843 & 33.9\% & 18,359 & 42.3\% & 14 & 1.0 \\ 
	iptc & 7,059 & 30.5\% & 17,195 & 39.6\% & 114 & 1.3 \\ 
	gender & 3,352 & 14.5\% & 8,200 & 18.9\% & 2 & 1.0 \\ 
	slot & 3,232 & 14.0\% & 14,983 & 34.5\% & 7 & 1.2 \\ 
\midrule
	\textbf{TOTAL} & 23,130 & 100.0\% & 43,373 & 100.0\% & 311 & 4.0 \\ 
\bottomrule
\end{tabular}
}
\label{ch_dwie:tab:main_entity_types}
\end{table}

\setlength{\tabcolsep}{5pt}
\Tabref{ch_dwie:tab:main_entity_types} illustrates the number of annotated entities and mentions per each tag category (type, topic, iptc, gender and slot). It also showcases the multi-label nature of entity classification task in \datasetname, with the average number of labels per entity of 4.0. 

\Tabref{ch_dwie:tab:relations_multilabel_stats} illustrates the number and percentage of related entities and mentions of our dataset grouped by the number of relation labels. 
It also compares with other entity-centric 
RE
datasets, namely BC5CDR~\cite{li2016biocreative,wei2015overview} and DocRED~\cite{yao2019docred} datasets. 
\setlength{\tabcolsep}{5pt}
\renewcommand{\arraystretch}{1.0}
\begin{table}[h]
\centering
\caption[Multi-label relation types statistics in \datasetname]{This table groups the number of related pairs in \datasetname~by the number of assigned relation labels to each of these pairs. We compare with other two entity-centric datasets: BC5CDR and DocRED.}
\resizebox{1.0\textwidth}{!}{
\begin{tabular}{lllllll} 
\toprule
 & \multicolumn{4}{c}{\textbf{\datasetname}} & \multicolumn{1}{c}{\textbf{BC5CDR}} & \multicolumn{1}{c}{\textbf{DocRED}} \\
\cmidrule(lr){2-5} \cmidrule(lr){6-6} \cmidrule(lr){7-7}

\textbf{\# Relation} & \textbf{\# Related} & \textbf{\% Related} & \textbf{\# Related} & \textbf{\% Related} & \textbf{\% Related} & \textbf{\% Related} \\
\textbf{labels} & \textbf{ent.\ pairs} & \textbf{ent.\ pairs} & \textbf{mention pairs} & \textbf{mention pairs} & \textbf{ent.\ pairs} & \textbf{ent.\ pairs} \\
\midrule
	1 & 12,856 & 76.32\% & 112,708 & 69.40\% & 100\% & 92.89\%\\ 
	2 & 3,101 & 18.41\% & 34,948 & 21.52\% & 0\% & 6.82\% \\ 
	3 & 884 & 5.25\% & 14,650 & 9.02\% & 0\% & 0.26\% \\ 
	4 & 3 & 0.02\% & 100 & 0.06\% & 0\% & 0.03\% \\ 
\midrule
	\textbf{TOTAL} & 16,844 & 100.0\% & 162,406 & 100.0\% & 100.0\% & 100.0\% \\ 
\bottomrule
\end{tabular}
}
\label{ch_dwie:tab:relations_multilabel_stats}
\end{table}

\setlength{\tabcolsep}{5pt}
\renewcommand{\arraystretch}{1.0}
\begin{table}[!htb]
\centering
\caption[Relation type statistics in \datasetname]{Relation type statistics. We compare the number of related entity and mention pairs per relation type. Only the most frequent relation types are shown (\% Related Mention Pairs $>$ \num{0.1}\%)}
\footnotesize
\resizebox{0.7\textwidth}{!}{
\begin{tabular}{lllll} 
\toprule
\begin{tabular}{@{}l@{}}\textbf{Relation} \\ \textbf{Type}\end{tabular} & \begin{tabular}{@{}l@{}}\textbf{\# Related} \\ \textbf{Ent. Pairs}\end{tabular}& \begin{tabular}{@{}l@{}}\textbf{\% Related} \\ \textbf{Ent. Pairs}\end{tabular}& \begin{tabular}{@{}l@{}}\textbf{\# Related} \\ \textbf{Men. Pairs}\end{tabular}& \begin{tabular}{@{}l@{}}\textbf{\% Related} \\ \textbf{Men. Pairs}\end{tabular} \\\midrule
	based\_in0 & 2,361 & 14.0\% & 18,771 & 11.6\% \\ 
	in0 & 2,120 & 12.6\% & 15,810 & 9.7\% \\ 
	citizen\_of & 1,969 & 11.7\% & 25,752 & 15.9\% \\ 
	based\_in0-x & 1,882 & 11.2\% & 12,211 & 7.5\% \\ 
	citizen\_of-x & 1,844 & 10.9\% & 17,049 & 10.5\% \\ 
	member\_of & 1,616 & 9.6\% & 19,953 & 12.3\% \\ 
	gpe0 & 1,569 & 9.3\% & 18,110 & 11.2\% \\ 
	in0-x & 1,474 & 8.8\% & 8,784 & 5.4\% \\ 
	agent\_of & 954 & 5.7\% & 15,776 & 9.7\% \\ 
	head\_of & 564 & 3.3\% & 7,710 & 4.7\% \\ 
	agency\_of & 435 & 2.6\% & 4,775 & 2.9\% \\ 
	player\_of & 401 & 2.4\% & 5,692 & 3.5\% \\ 
	agency\_of-x & 382 & 2.3\% & 2,108 & 1.3\% \\ 
	head\_of\_state & 380 & 2.3\% & 7,986 & 4.9\% \\ 
	head\_of\_state-x & 343 & 2.0\% & 3,853 & 2.4\% \\ 
	appears\_in & 294 & 1.7\% & 4,555 & 2.8\% \\ 
	vs & 281 & 1.7\% & 7,187 & 4.4\% \\ 
	head\_of\_gov & 273 & 1.6\% & 4,015 & 2.5\% \\ 
	head\_of\_gov-x & 247 & 1.5\% & 2,383 & 1.5\% \\ 
	minister\_of & 234 & 1.4\% & 2,280 & 1.4\% \\ 
	minister\_of-x & 213 & 1.3\% & 1,629 & 1.0\% \\ 
	based\_in2 & 185 & 1.1\% & 971 & 0.6\% \\ 
	event\_in0 & 181 & 1.1\% & 843 & 0.5\% \\ 
	part\_of & 164 & 1.0\% & 2,858 & 1.8\% \\ 
	in2 & 157 & 0.9\% & 1,055 & 0.6\% \\ 
	created\_by & 134 & 0.8\% & 945 & 0.6\% \\ 
	agent\_of-x & 125 & 0.7\% & 897 & 0.6\% \\ 
	award\_received & 111 & 0.7\% & 969 & 0.6\% \\ 
	institution\_of & 105 & 0.6\% & 2,113 & 1.3\% \\ 
	ministry\_of & 81 & 0.5\% & 666 & 0.4\% \\ 
	coach\_of & 65 & 0.4\% & 1,211 & 0.7\% \\ 
	won\_vs & 61 & 0.4\% & 1,531 & 0.9\% \\ 
	spouse\_of & 55 & 0.3\% & 599 & 0.4\% \\ 
	directed\_by & 44 & 0.3\% & 318 & 0.2\% \\ 
	is\_meeting & 41 & 0.2\% & 968 & 0.6\% \\ 
	event\_in2 & 40 & 0.2\% & 259 & 0.2\% \\ 
	spokesperson\_of & 39 & 0.2\% & 177 & 0.1\% \\ 
	plays\_in & 38 & 0.2\% & 330 & 0.2\% \\ 
	gpe1 & 35 & 0.2\% & 135 & 0.1\% \\ 
	product\_of & 31 & 0.2\% & 334 & 0.2\% \\ 
	parent\_of & 22 & 0.1\% & 281 & 0.2\% \\ 
	child\_of & 22 & 0.1\% & 281 & 0.2\% \\ 
	based\_in1 & 22 & 0.1\% & 376 & 0.2\% \\ 
	signed\_by & 20 & 0.1\% & 521 & 0.3\% \\ 
	law\_of & 16 & 0.1\% & 286 & 0.2\% \\
\midrule
	\textbf{TOTAL} & 16,844 & 100.0\% & 162,406 & 100.0\% \\ 
\bottomrule
\end{tabular}
}
\label{ch_dwie:tab:relations_stats}
\end{table}
\clearpage

\subsection*{Inter-annotator agreement calculations}
\label{ch_dwie:app:kappa_agreement_details}
 In order to measure the agreement we use Cohen's kappa coefficient \cite{cohen1960coefficient}, defined as
\begin{align}
\kappa = \dfrac{p_o - p_e}{1 - p_e} \label{ch_dwie:eq:kappa_generic} 
\end{align}
where $p_o$ represents the observed agreement between the two annotators and $p_e$ is the expected agreement between the annotators (\ie agreement by chance). 
More specifically, in our case we calculate the observed probability $p_o$ as in \equref{ch_dwie:eq:kappa_observed} where $N$ is the number of annotated items, $A_{i,j}$ is the annotation made by annotator $i$ for item $j$, and 
$\mathbbm{1}\{ A_{1,j} = A_{2,j} \}$
returns $1$ if $A_{1,j}$ is equal to $A_{2,j}$ and $0$ otherwise. Thus, $p_o$ can be interpreted as the fraction of the labels two annotators agree, also called \textit{percent agreement} \cite{mchugh2012interrater,scott1955reliability}.
\begin{align}
p_o = \dfrac{\sum\limits_{j=1}^N{\mathbbm{1}\{ A_{1,j} = A_{2,j} \}}}{N} \label{ch_dwie:eq:kappa_observed} 
\end{align}
To calculate the expected agreement probability we use the formulation in \equref{ch_dwie:eq:kappa_expected}. It can be interpreted as
the probability that both annotators, when randomly distributing all of their label annotations among the items to be annotated, assign the same label to a given item.
\begin{align}
p_e = \sum\limits_{l=1}^L{\frac{n_{1,l}}{N} \> \frac{n_{2,l}}{N}} 
\label{ch_dwie:eq:kappa_expected} 
\end{align}
In this context, $n_{i,l}$ is the number of items the annotator $i$ annotated with label $l$ and $L$ is the total number of labels. For multi-label annotations where it is possible to assign multiple classes for a particular annotation item (\ie named entity and relation types), we report a weighted kappa score. 
\clearpage
\subsection*{Relation consistency rules}
\label{ch_dwie:app:rel_rules_list}
\setcounter{table}{0}
\setcounter{figure}{0}
This appendix enumerates the logical predicates used as a consistency check in our dataset. 
\allowdisplaybreaks
\setcounter{equation}{0}
\begin{align}
	\relationalign{spouse\_of}{Y}{X} \implies \relationalign{spouse\_of}{X}{Y} \label{ch_dwie:eq:rel_cons1} \\ 
	\relationalign{vs}{Y}{X} \implies \relationalign{vs}{X}{Y} \label{ch_dwie:eq:rel_cons2} \\ 
	\relationalign{won\_vs}{X}{Y} \implies \relationalign{vs}{X}{Y} \label{ch_dwie:eq:rel_cons3} \\ 
 	\relationalign{won\_vs}{X}{Y} \implies \relationalign{vs}{Y}{X} \label{ch_dwie:eq:rel_cons4} \\ 
 	\relationalign{child\_of}{Y}{X} \implies \relationalign{parent\_of}{X}{Y} \label{ch_dwie:eq:rel_cons5} \\ 
 	\relationalign{parent\_of}{Y}{X} \implies \relationalign{child\_of}{X}{Y} \label{ch_dwie:eq:rel_cons6} \\ 
 	\relationalign{ministry\_of}{X}{Y} \implies \relationalign{agency\_of}{X}{Y} \label{ch_dwie:eq:rel_cons7} \\ 
 	\relationalign{agency\_of\text{-}x}{X}{Z} \land \relationalign{gpe0}{Z}{Y} \implies \relationalign{agency\_of}{X}{Y} \label{ch_dwie:eq:rel_cons8} \\ 
 	\relationalign{agency\_of}{X}{Y} \land \relationalign{gpe0}{Z}{Y} \implies \relationalign{agency\_of\text{-}x}{X}{Z} \label{ch_dwie:eq:rel_cons9} \\ 
 	\relationalign{agent\_of\text{-}x}{X}{Z} \land \relationalign{gpe0}{Z}{Y} \implies \relationalign{agent\_of}{X}{Y} \label{ch_dwie:eq:rel_cons10} \\ 
 	\relationalign{agent\_of}{X}{Y} \land \relationalign{gpe0}{Z}{Y} \implies \relationalign{agent\_of\text{-}x}{X}{Z} \label{ch_dwie:eq:rel_cons11} \\ 
 	\relationalign{minister\_of}{X}{Y} \implies \relationalign{agent\_of}{X}{Y} \label{ch_dwie:eq:rel_cons12} \\ 
 	\relationalign{head\_of\_gov}{X}{Y} \implies \relationalign{agent\_of}{X}{Y} \label{ch_dwie:eq:rel_cons13} \\ 
 	\relationalign{head\_of\_state}{X}{Y} \implies \relationalign{agent\_of}{X}{Y} \label{ch_dwie:eq:rel_cons14} \\ 
 	\relationalign{citizen\_of\text{-}x}{X}{Z} \land \relationalign{gpe0}{Z}{Y} \implies \relationalign{citizen\_of}{X}{Y} \label{ch_dwie:eq:rel_cons15} \\ 
 	\relationalign{citizen\_of}{X}{Y} \land \relationalign{gpe0}{Z}{Y} \implies \relationalign{citizen\_of\text{-}x}{X}{Z} \label{ch_dwie:eq:rel_cons16} \\ 
 	\relationalign{minister\_of\text{-}x}{X}{Z} \land \relationalign{gpe0}{Z}{Y} \implies \relationalign{minister\_of}{X}{Y} \label{ch_dwie:eq:rel_cons17} \\ 
 	\relationalign{minister\_of}{X}{Y} \land \relationalign{gpe0}{Z}{Y} \implies \relationalign{minister\_of\text{-}x}{X}{Z} \label{ch_dwie:eq:rel_cons18} \\ 
 	\relationalign{head\_of\_state\text{-}x}{X}{Z} \land \relationalign{gpe0}{Z}{Y} \implies \relationalign{head\_of\_state}{X}{Y} \label{ch_dwie:eq:rel_cons19} \\ 
 	\relationalign{head\_of\_state}{X}{Y} \land \relationalign{gpe0}{Z}{Y} \implies \relationalign{head\_of\_state\text{-}x}{X}{Z} \label{ch_dwie:eq:rel_cons20} \\ 
 	\relationalign{head\_of\_gov\text{-}x}{X}{Z} \land \relationalign{gpe0}{Z}{Y} \implies \relationalign{head\_of\_gov}{X}{Y} \label{ch_dwie:eq:rel_cons21} \\ 
 	\relationalign{head\_of\_gov}{X}{Y} \land \relationalign{gpe0}{Z}{Y} \implies \relationalign{head\_of\_gov\text{-}x}{X}{Z} \label{ch_dwie:eq:rel_cons22} \\ 
 	\relationalign{in0\text{-}x}{X}{Z} \land \relationalign{gpe0}{Z}{Y} \implies \relationalign{in0}{X}{Y} \label{ch_dwie:eq:rel_cons23} \\ 
 	\relationalign{in0}{X}{Y} \land \relationalign{gpe0}{Z}{Y} \implies \relationalign{in0\text{-}x}{X}{Z} \label{ch_dwie:eq:rel_cons24} \\ 
 	\relationalign{in2}{X}{Z} \land \relationalign{in0}{Z}{Y} \implies \relationalign{in0}{X}{Y} \label{ch_dwie:eq:rel_cons25} \\ 
 	\relationalign{in1}{X}{Z} \land \relationalign{in0}{Z}{Y} \implies \relationalign{in0}{X}{Y} \label{ch_dwie:eq:rel_cons26} \\ 
 	\relationalign{based\_in2}{X}{Z} \land \relationalign{in0}{Z}{Y} \implies \relationalign{based\_in0}{X}{Y} \label{ch_dwie:eq:rel_cons27} \\ 
 	\relationalign{based\_in1}{X}{Z} \land \relationalign{in0}{Z}{Y} \implies \relationalign{based\_in0}{X}{Y} \label{ch_dwie:eq:rel_cons28} \\ 
  	\relationalign{agency\_of}{X}{Y} \land \relationalignsingle{gpe0}{Y} \implies \relationalign{based\_in0}{X}{Y} \label{ch_dwie:eq:rel_cons29} \\ 
 	\relationalign{event\_in2}{X}{Z} \land \relationalign{in0}{Z}{Y} \implies \relationalign{event\_in0}{X}{Y} \label{ch_dwie:eq:rel_cons30} \\ 
 	\relationalign{event\_in1}{X}{Z} \land \relationalign{in0}{Z}{Y} \implies \relationalign{event\_in0}{X}{Y} \label{ch_dwie:eq:rel_cons31} \\ 
 	\relationalign{head\_of}{X}{Y} \implies \relationalign{member\_of}{X}{Y} \label{ch_dwie:eq:rel_cons32} \\ 
 	\relationalign{coach\_of}{X}{Y} \implies \relationalign{member\_of}{X}{Y} \label{ch_dwie:eq:rel_cons33} \\ 
 	\relationalign{spokesperson\_of}{X}{Y} \implies \relationalign{member\_of}{X}{Y} \label{ch_dwie:eq:rel_cons34} \\ 
 	\relationalign{member\_of}{X}{Y} \land \relationalignsingle{sport\_player}{X} \implies \relationalign{player\_of}{X}{Y} \label{ch_dwie:eq:rel_cons35} \\ 
 	\relationalign{mayor\_of}{X}{Y} \implies \relationalign{head\_of\_gov}{X}{Y} \label{ch_dwie:eq:rel_cons36} \\ 
 	\relationalign{directed\_by}{X}{Y} \implies \relationalign{created\_by}{X}{Y} \label{ch_dwie:eq:rel_cons37} \\ 
 	\relationalign{character\_in}{X}{Y} \land \relationalign{played\_by}{X}{Z} \implies \relationalign{plays\_in}{Z}{Y} \label{ch_dwie:eq:rel_cons38} \\ 
 	\relationalign{institution\_of}{X}{Y} \implies \relationalign{part\_of}{X}{Y} \label{ch_dwie:eq:rel_cons39} \\ 
 	\relationalign{based\_in0\text{-}x}{X}{Z} \land \relationalign{gpe0}{Z}{Y} \implies \relationalign{based\_in0}{X}{Y} \label{ch_dwie:eq:rel_cons40} \\ 
 	\relationalign{based\_in0}{X}{Y} \land \relationalign{gpe0}{Z}{Y} \implies \relationalign{based\_in0\text{-}x}{X}{Z} \label{ch_dwie:eq:rel_cons41} 
\end{align}


\renewcommand*{\thesection}{\thechapter.\arabic{section}}       


\newcommand{\originalaida}{AIDA}

\newcommand{\ELonly}{\textsf{Linker}}
\newcommand{\Baseline}{\textsf{Standalone}}

\newcommand{\dwiedataset}{DWIE}
\newcommand{\CEAFe}{CEAF$_\textrm{e}$}
\newcommand{\ouraidaseta}{AIDA$^{+}_\textrm{a}$}
\newcommand{\ouraidasetb}{AIDA$^{+}_\textrm{b}$}
\newcommand{\clinker}{{\textsf{Local}}}
\newcommand{\mtt}{{\textsf{Global}}}
\newcommand{\corefonly}{\textsf{Coref}}

\definecolor{myRed}{RGB}{204,0,0}
\definecolor{myGreen}{RGB}{0,84,0}

\graphicspath{{klim_ch_coreflinker/figures/}}

\widowpenalty100000
\clubpenalty100000

\hyphenation{}

\chapter[Towards Consistent Document-Level Entity Linking: Joint Models for Entity Linking and Coreference Resolution]{Towards Consistent Document-level Entity Linking: Joint Models for Entity Linking and Coreference Resolution}
\label{chap:coreflinker}

\renewcommand\evenpagerightmark{{\scshape\small Chapter \arabic{chapter}}}
\renewcommand\oddpageleftmark{{\scshape\small Towards Consistent Document-Level Entity Linking}}

\renewcommand{\bibname}{References}


\begin{flushright}
\end{flushright}

\noindent%
\emph{%
In this chapter, we present a more detailed model on how the entity-centric approach can be used for \textit{entity linking} task. The \textit{entity linking} consists in mapping the anchor \textit{mentions} in text to target \textit{entities} that describe
them in a Knowledge Base (KB) (\eg Wikipedia). In our work, we showcase that this task can be improved by considering performing entity linking on the coreference cluster level instead of on each of the mentions individually. By adopting this approach, our joint model is able to use the information of all the coreferent mentions at once when choosing the candidate entity. As a result, this leads to more consistent predictions among mentions referring to the same concept, especially boosting the performance on corner cases consisting of unpopular mentions.
}

\begin{center}
\par{$\star\star\star$}
\end{center}
\vspace{0.15in}

\par{\noindent\large{\textbf{K.~Zaporojets, J.~Deleu, Y.~Yiang, T.~Demeester and C.~Develder}}}
\vspace{0.1in}
\par{\noindent\textbf{In Proceedings of the ACL 2022}}
\vspace{0.15in}

\par{\noindent\bf{Abstract}}
We consider the task of 
document-level 
entity linking (EL), where it is important to make consistent decisions for entity mentions over the full document jointly.
We aim to leverage explicit ``connections'' among mentions within the document itself:~we propose to join EL and coreference resolution (coref) in a \textit{single} structured prediction task over directed trees and use a globally normalized model to solve it. This contrasts with related works where two separate models are trained for each of the tasks and additional logic is required to merge the outputs.~Experimental results 
on two datasets show a boost of up to +5\% F1-score on both coref and EL tasks, compared to their standalone counterparts.
For a subset of hard cases, with individual mentions lacking the correct EL in their candidate entity list, we obtain a 
+50\%
increase in 
accuracy.\footnote{Our code, models and \ouraida~dataset will be released on \url{https://github.com/klimzaporojets/consistent-EL} }

\section{Introduction}
\label{ch_coreflinker:sec:intro}
In this paper we explore a principled approach to solve entity linking (EL) jointly with coreference resolution (coref). Concretely, we formulate coref+EL
as a \textit{single} structured task over directed trees that conceives EL and coref as two complementary components: a coreferenced cluster can only be linked to a single entity or NIL (\ie a non-linkable entity), and all mentions linking to the same entity are coreferent. This contrasts with previous attempts to join coref+EL \cite{hajishirzi2013joint, dutta2015c3el, angell2021clustering} where coref and EL models are trained separately and additional logic is required to merge the predictions of both tasks. 
\begin{figure}[t]
\centering
\includegraphics[width=.6\columnwidth]{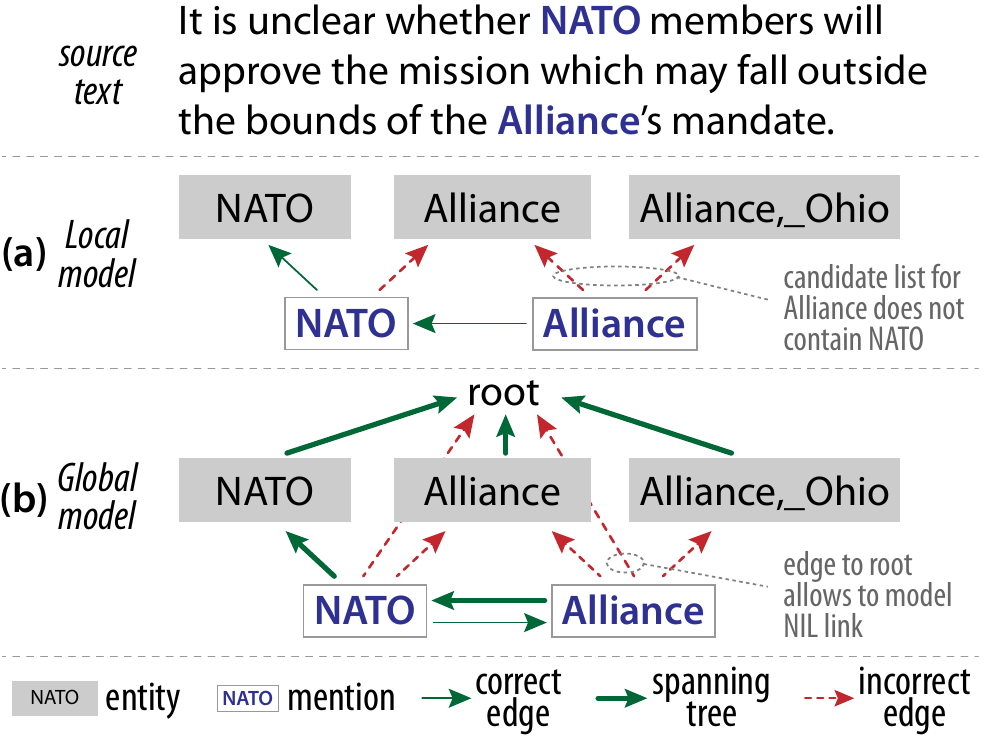}
\captionsetup{singlelinecheck=off}
\caption[Illustration of the explored graph models]{Illustration of our 2 explored graph models:
\begin{enumerate*} [(a)]
    \item {\clinker} where edges are only allowed from spans to antecedents or candidate entities, and
    \item {\mtt} where the prediction involves a spanning tree over all nodes.  
\end{enumerate*}}
\label{ch_coreflinker:fig:architectures}
\end{figure}

Our first approach ({\clinker} in \figref{ch_coreflinker:fig:architectures}(a)) is motivated by current state-of-the-art 
coreference resolution models 
\cite{joshi2019bert,wu2020corefqa}
that predict a single antecedent for each span to resolve.
We extend this architecture by also considering entity links as potential antecendents:
in the example of \figref{ch_coreflinker:fig:architectures}, the mention ``Alliance'' can be either connected to its antecedent mention ``NATO'' or to any of its candidate links (\emph{Alliance} or \emph{Alliance,\_Ohio}). 
While straightforward,
this approach cannot solve 
cases where the first coreferenced mention does not include the correct entity in its candidate list
(\eg if the order of ``NATO'' and ``Alliance'' mentions in \figref{ch_coreflinker:fig:architectures} would be reversed). 
We therefor propose a second approach, {\mtt}, which by construction overcomes this inherent limitation by using bidirectional connections between mentions. 
Because that implies cycles could be formed, we resort to solving a maximum spanning tree problem. 
Mentions that refer to the same entity form a cluster, represented as a subtree rooted by the single entity they link to. 
To encode the overall document's clusters in a single spanning tree, we introduce a virtual \textit{root} node
(see \figref{ch_coreflinker:fig:architectures}(b)).\footnote{Coreference clusters without a linked entity, \ie a NIL cluster, have a link of a mention directly to the root.}

This paper contributes: \begin{enumerate*}[(i)]
    \item 2 architectures ({\clinker} and {\mtt}) for joint entity linking (EL) and corefence resolution, 
    \item an extended AIDA dataset \cite{hoffart2011robust}, adding new annotations of linked and NIL coreference clusters,
    \item experimental analysis on 2 datasets where our joint coref+EL models achieve up to +5\% F1-score on both tasks compared to standalone models. We also show up to 
    +50\% in accuracy
    for hard cases of EL 
    where
    entity mentions 
    lack
    the correct entity in their candidate list. 
\end{enumerate*}

\section{Architecture}
\label{ch_coreflinker:sec:architecture}

Our model takes as input
\begin{enumerate*}[(i)]
    \item the full document text, and
    \item an \emph{alias table} with entity candidates for each of the possible spans. 
\end{enumerate*} 
Our end-to-end approach allows to jointly predict the mentions, entity links and coreference relations between them. 

\subsection{Span and entity representations}

We use SpanBERT (base) from \cite{joshi2020spanbert} to obtain \emph{span} representations $\textbf{g}_i$ for a particular span $s_i$. 
Similarly to \cite{luan2019general,xu2020revealing}, we apply an additional pruning step to keep only the top-$N$ spans based on the pruning score $\Phi_{\mathrm{p}}$ from a feed-forward neural net (FFNN): 
\begin{equation}
\Phi_{\mathrm{p}}(s_i) = \mathrm{FFNN}_{P}( \textbf{g}_i).
\end{equation}

For a candidate entity $e_{j}$ of span $s_i$ we 
will obtain
representation as $\textbf{e}_{j}$ 
(which is further detailed in \secref{ch_coreflinker:sec:experimental_setup}). 

\subsection{Joint approaches}
\label{ch_coreflinker:sec:joint_approaches}
We propose two 
methods for joint coreference and EL.
The first, {\clinker}, is motivated by end-to-end span-based coreference resolution models 
\cite{lee2017end,lee2018higher} 
that optimize the marginalized probability of the correct antecedents for each given span.
We extend this local marginalization to include the span's candidate entity links. Formally, the 
modeled 
probability of 
$y$ 
(text span or candidate entity) 
being the antecedent of span $s_i$ is: 
\begin{equation}
P_{\mathrm{cl}}(y | s_i) = \dfrac{\exp\big(\Phi_{\mathrm{cl}}(s_i,y)\big)}{\sum_{y'\in \mathcal{Y}(s_i)} \exp \big( \Phi_{\mathrm{cl}}(s_i,y') \big) }, \label{ch_coreflinker:eq:softmax}
\end{equation}
where $\mathcal{Y}(s_i)$ is the set of antecedent spans unified with the candidate entities for $s_i$. For antecedent \emph{spans} $\{s_j: j < i\}$ the score $\Phi_{\mathrm{cl}}$ is defined as:
\begin{align}
\medmath{\Phi_{\mathrm{cl}}(s_i, s_j) = \Phi_{\mathrm{p}}(s_i) + \Phi_{\mathrm{p}}(s_j) + \Phi_{\mathrm{c}}(s_i, s_j)}, \label{ch_coreflinker:eq:span_cl1} \\ 
\medmath{\Phi_{\mathrm{c}}(s_i, s_j) = \mathrm{FFNN}_{C}([ \textbf{g}_i; \textbf{g}_{j}; \textbf{g}_{i} \odot \textbf{g}_{j}; \boldsymbol{\varphi}_{i,j}])}, \label{ch_coreflinker:eq:span_cl2}
\end{align}
where $\boldsymbol{\varphi}_{i,j}$ is an embedding encoding the distance\footnote{Measured in number of spans, after pruning.} between spans $s_i$ and $s_j$.
Similarly, for a particular candidate \emph{entity} $e_{j}$, the score $\Phi_{\mathrm{cl}}$ is:
\begin{align}
    \Phi_{\mathrm{cl}}(s_i, e_{j}) = \Phi_{\mathrm{p}}(s_i) + \Phi_{{\ell}}(s_i, e_{j}), \label{ch_coreflinker:eq:entity_cl1} \\
    \Phi_{\ell}(s_i, e_{j}) = \mathrm{FFNN}_{L}([\textbf{g}_i; \textbf{e}_{j}]). \label{ch_coreflinker:eq:entity_cl2}
\end{align}
An example graph of mentions and entities with edges for which aforementioned scores $\Phi_{\mathrm{cl}}$ would be calculated is sketched in \figref{ch_coreflinker:fig:architectures}(a). 
While simple, this approach fails
to correctly solve EL when the correct entity is only present in the candidate lists of mention spans occurring later in the text (since 
earlier
mentions have no access to it).

To solve EL in the general case, even
when the first mention does not have the correct entity, we propose bidirectional connections between mentions, thus leading to a maximum spanning tree problem in our {\mtt} approach.
Here we define a score for a (sub)tree $t$, noted as $\Phi_\mathrm{tr}(t)$:
\begin{equation}
   \Phi_\mathrm{tr}(t) = \sum_{(i,j) \in t} \Phi_{\mathrm{cl}}(u_i, u_j), \label{ch_coreflinker:eq:mtt1}   
\end{equation}
where $u_i$ and $u_j$ are two connected nodes (\ie \emph{root}, candidate entities or spans) in $t$.
For a ground truth cluster $c \in C$ (with $C$ being the set of all such clusters), with its set\footnote{For a single cluster annotation, indeed it is possible that multiple correct trees can be drawn.} of correct subtree representations $\mathcal{T}_c$, we model the cluster's likelihood with its subtree scores. We minimize the negative log-likelihood $\mathcal{L}$ of all clusters:
\begin{align}
\mathcal{L} &= - \log \frac{\prod_{c \in C} \sum_{t \in \mathcal{T}_c} \exp \big(\Phi_\mathrm{tr}(t)\big)}{\sum_{t \in \mathcal{T}_\textit{all}} \exp \big( \Phi_\mathrm{tr}(t) \big)}. \label{ch_coreflinker:eq:loss_naive}
\end{align}%
Naively enumerating all possible spanning trees ($\mathcal{T}_\textit{all}$ or $\mathcal{T}_c$) implied by this equation is infeasible, since their number is exponentially large.
We use the adapted Kirchhoff's Matrix Tree Theorem 
(MTT; \cite{koo2007structured,william1984tutte})
to solve this:
the sum of the weights of the spanning trees in a directed graph rooted in \emph{r} is equal to the determinant of the Laplacian matrix of the graph with the row and column corresponding to \emph{r} removed (\ie the \emph{minor} of the Laplacian with respect to \emph{r}). This way, \equref{ch_coreflinker:eq:loss_naive} can be rewritten as
\begin{align}
    \mathcal{L} &= - \log \frac{\prod_{c \in C} {\det \Big(\mathbf{\hat{L}}_{c} \big(\mathbf{\Phi_\mathrm{cl}}\big) \Big)}}{\det \Big(\mathbf{L}_{r}\big(\mathbf{\Phi_\mathrm{cl}}\big)\Big)}, \label{ch_coreflinker:eq:loss_mtt}
\end{align}
where $\mathbf{\Phi_\mathrm{cl}}$ is the weighted adjacency matrix of the graph, and $\mathbf{L}_{r}$ is the minor of the Laplacian with respect to the root node $r$. An entry in the Laplacian matrix is 
calculated
as
\begin{align}
    \medmath{L_{i,j} = 
    \begin{cases}
      \sum\limits_{k} \exp(\Phi_{\mathrm{cl}}(u_k, u_j)) & \text{if $i = j$}\\
      - \exp(\Phi_{\mathrm{cl}}(u_i, u_j)) & \text{otherwise}
    \end{cases}}, \label{ch_coreflinker:eq:laplacian}
\end{align}
Similarly, $\mathbf{\hat{L}}_{c}$ is a \textit{modified Laplacian} matrix where the first row is replaced with the root $r$ selection scores $\Phi_{\mathrm{cl}}(r, u_j)$. 
For clarity, Appendix~\ref{ch_coreflinker:sec:appendix} presents a toy example with detailed steps to calculate the loss in \equref{ch_coreflinker:eq:loss_mtt}.

To calculate the scores of each of the entries $\Phi_\textrm{cl} (u_i, u_j)$ to $\mathbf{\Phi_\mathrm{cl}}$ matrix in 
eqs.~(\ref{ch_coreflinker:eq:mtt1}) and~(\ref{ch_coreflinker:eq:loss_mtt})
for {\mtt}, we use the same approach as in {\clinker} for edges between two mention spans, or between a mention and entity. 
For the directed edges between the root $r$ and a candidate entity $e_j$ we choose $\Phi_{\mathrm{cl}}(r, e_j)=0$.
Since we represent NIL clusters by edges from the 
mention spans directly to the root,
we also need scores for them: we use \equref{ch_coreflinker:eq:span_cl1} with $\Phi_\mathrm{p}(r)=0$.
We use 
Edmonds'
algorithm \cite{edmonds1967optimum} for decoding the maximum spanning tree.  
\section{Experimental setup}
\label{ch_coreflinker:sec:experimental_setup}

We considered two datasets to evaluate our proposed models: {\dwiedataset} \cite{zaporojets2021dwie} and {\originalaida} \cite{hoffart2011robust}.
Since {\originalaida} essentially does not contain coreference information, we had to extend it by
\begin{enumerate*}[(i)]
    \item adding missing mention links in order to make annotations consistent on the coreference cluster level, and 
    \item annotating NIL coreference clusters.
\end{enumerate*}
We note this extended dataset as {\ouraida}. See \tabref{ch_coreflinker:tab:dataset} for the details. 

As input to our models, for {\dwiedataset} we generate spans of up to 5 tokens.
For each mention span $s_i$, we find candidates from a dictionary of entity surface forms used for hyperlinks in Wikipedia.
We then keep the top-16 candidates based on the prior for that surface form, as per \cite[\S3]{yamada2016joint}.
Each of those candidates $e_{j}$ is represented using a Wikipedia2Vec embedding $\textbf{e}_{j}$ \cite{yamada2016joint}.\footnote{We use Wikipedia version 20200701.} 
For {\ouraida}, we use 
the spans, entity candidates, and entity representations
from \cite{kolitsas2018end}.\footnote{\url{https://github.com/dalab/end2end_neural_el}}
\begin{table}[t]
    \centering
    \resizebox{0.7\columnwidth}{!}
    {\begin{tabular}{lcccc}
        \toprule
         \multirow{2}{*}{Dataset} & \# Linked & \# NIL & Linked & \# NIL \\
          & clusters & clusters & mentions & mentions \\         
         \midrule
	 \dwiedataset & 11,967 & 9,935 & 28,482 & 14,891 \\ 
	 \originalaida & 16,673 & - & 27,817 & 7,112 \\ 
	 \ouraida & 16,775 & 4,284 & 28,813 & 6,116 \\     \bottomrule
    \end{tabular}}
    \caption[Datasets statistics]{Datasets statistics.} 
    \label{ch_coreflinker:tab:dataset}
\end{table}

To assess the performance of our joint coref+EL models {\clinker} and {\mtt}, we also provide 
\Baseline~implementations for coref and EL tasks. The \Baseline~coref model is trained using only the coreference component of our joint architecture (\eqsref{ch_coreflinker:eq:softmax}{ch_coreflinker:eq:span_cl2}), while the EL model is based only on the linking component (\equref{ch_coreflinker:eq:entity_cl2}).

As performance metrics, for coreference resolution we calculate the average-F1 score of commonly used MUC~\cite{vilain1995model}, B$^3$~\cite{bagga1998algorithms} and {\CEAFe} \cite{luo2005coreference} metrics as implemented by \cite{pradhan2014scoring}.
For EL, we use
\begin{enumerate*}[(i)]
    \item \emph{mention}-level F1 score (EL$_\mathrm{m}$), and
    \item \emph{cluster}-level \textit{hard} F1 score (EL$_\mathrm{h}$) that counts a true positive only if both the coreference cluster (in terms of all its mention spans) and the entity link are correctly predicted. 
\end{enumerate*}
These EL metrics are executed in a \emph{strong matching} setting that requires predicted spans to exactly match the boundaries of gold mentions.
Furthermore, for EL we only report the performance on non-NIL 
mentions, leaving the study of NIL links for future work. 

Our experiments will answer the following research questions:~\begin{enumerate*}[label=\textbf{(Q\arabic*)}]
   \item \label{ch_coreflinker:it:q-performance} How does performance of our joint coref+EL models compare to \Baseline~
   models?
    \item \label{ch_coreflinker:it:q-coherence} Does jointly solving coreference resolution and EL enable more coherent EL predictions?
    \item \label{ch_coreflinker:it:q-corner} How do our joint models perform on hard cases where some individual entity mentions do not have the correct candidate?
\end{enumerate*}

\section{Results}
\label{ch_coreflinker:sec:results}
\begin{table*}[t]
\centering
\resizebox{1.0\textwidth}{!}{
\begin{tabular}{c ccc c ccc c ccc}
\toprule
& \multicolumn{3}{c}{\dwiedataset} &&  \multicolumn{3}{c}{\ouraidaseta} && \multicolumn{3}{c}{\ouraidasetb} \\ 
\cmidrule(lr){2-4}\cmidrule(lr){6-8}\cmidrule(lr){10-12} 
Setup & EL$_\mathrm{m}$ & EL$_\mathrm{h}$ & Coref && EL$_\mathrm{m}$ & EL$_\mathrm{h}$ & Coref && EL$_\mathrm{m}$ & EL$_\mathrm{h}$ & Coref \\ 
         \midrule
         \Baseline & 88.7${\scriptstyle \pm\text{0.1}}$ & 78.4${\scriptstyle \pm\text{0.2}}$ & 94.5${\scriptstyle \pm\text{0.1}}$ && 86.2${\scriptstyle \pm\text{0.4}}$ & 80.7${\scriptstyle \pm\text{0.5}}$ & 93.8${\scriptstyle \pm\text{0.1}}$ && 79.1${\scriptstyle \pm\text{0.3}}$ & 74.0${\scriptstyle \pm\text{0.3}}$ & 91.5${\scriptstyle \pm\text{0.3}}$ \\
         \clinker & 90.5${\scriptstyle \pm\text{0.4}}$ & 83.4${\scriptstyle \pm\text{0.4}}$ & 94.4${\scriptstyle \pm\text{0.2}}$ && 87.5${\scriptstyle \pm\text{0.2}}$ & 83.1${\scriptstyle \pm\text{0.2}}$ & 94.7${\scriptstyle \pm\text{0.1}}$ && \textbf{79.9}${\scriptstyle \pm\textbf{0.4}}$ & 75.8${\scriptstyle \pm\text{0.3}}$ & \textbf{92.3${\scriptstyle \pm\text{0.1}}$} \\
         \mtt & \textbf{90.7}${\scriptstyle \pm\textbf{0.3}}$ & \textbf{83.9}${\scriptstyle \pm\text{0.5}}$ & \textbf{94.7}${\scriptstyle \pm\text{0.2}}$ && \textbf{87.6}${\scriptstyle \pm\textbf{0.2}}$ & \textbf{83.7}${\scriptstyle \pm\text{0.3}}$ & \textbf{95.1}${\scriptstyle \pm\text{0.1}}$ && 79.6${\scriptstyle \pm\text{0.4}}$ & \textbf{76.0}${\scriptstyle \pm\text{0.4}}$ & 92.2${\scriptstyle \pm\text{0.2}}$ \\ 
\bottomrule 
	\end{tabular}
}
	\caption[General experimental results]{Experimental results 
	(F1 scores defined in \secref{ch_coreflinker:sec:experimental_setup}) 
	using the 
	\Baseline~coreference and EL
	models compared to our joint architectures (\clinker~and \mtt), on \dwiedataset~and \ouraida~datasets.}
	\label{ch_coreflinker:tab:overview_results}
\end{table*}
\Tabref{ch_coreflinker:tab:overview_results} shows the results of our compared models for EL and 
coreference resolution 
tasks.
Answering \ref{ch_coreflinker:it:q-performance}, we observe a general improvement in performance of our coref+EL joint models ({\clinker} and {\mtt}) compared to \Baseline~
on the EL task.
Furthermore, this difference is bigger when using our cluster-level \textit{hard} metrics.
This also answers \ref{ch_coreflinker:it:q-coherence} by indicating that the joint models tend to produce more coherent cluster-based predictions.
To make this more explicit, \Tabref{ch_coreflinker:tab:cluster_coherence} compares the accuracy for singleton clusters (\ie clusters composed by a single entity mention), denoted as $S$, to that of clusters composed by multiple mentions, denoted as $M$.
We observe that the difference in performance between our joint models and
\Baseline~is bigger on $M$ clusters (with a consistent 
superiority 
of \mtt), indicating that our approach indeed produces more coherent predictions for mentions that refer to the same concept. 
\begin{table}[t]
    \centering
    \resizebox{0.6\columnwidth}{!}{
    \begin{tabular}{l cc c cc c cc}
        \toprule
         & \multicolumn{2}{c}{\dwiedataset} && \multicolumn{2}{c}{\ouraidaseta} && \multicolumn{2}{c}{\ouraidasetb}  \\
         \cmidrule(lr){2-3}\cmidrule(lr){5-6}\cmidrule(lr){8-9} 
         Setup & $S$ & $M$ && $S$ & $M$ && $S$ & $M$ \\
         \midrule
         \Baseline & 80.4 & 69.5 && 82.9 & 70.7 && 77.0 & 57.0 \\
         \clinker & \textbf{82.6} & 78.6 && 84.9 & 74.8 && \textbf{79.8} & 61.4 \\
         \mtt & \textbf{82.6} & \textbf{80.0} && \textbf{85.1} & \textbf{76.8} && 79.3 & \textbf{63.0} \\                  
\bottomrule
    \end{tabular}
    } 
    \caption[Accuracy for singletons and multiple mentions clusters]{Cluster-based accuracy of link prediction on singletons ($S$) and clusters of multiple mentions ($M$).} 
    \label{ch_coreflinker:tab:cluster_coherence}
\end{table}
Further analysis reveals that this difference in performance is even higher for a more complex scenario where the clusters contain mentions with different surface forms (not 
shown 
in the table). 
\begin{table}[t]
    \centering
    \small
    \begin{tabular}{lccc}
        \toprule
         {Setup} & \dwiedataset & \ouraidaseta & \ouraidasetb  \\
         \midrule
	 \Baseline & 0.0 & 0.0 & 0.0  \\ 
	 \clinker & 41.7 & 27.4 & 26.9  \\ 
	 \mtt & \textbf{57.6} & \textbf{50.2} & \textbf{29.7} \\     \bottomrule
    \end{tabular}
    \caption[Accuracy for mentions without correct entity in candidate list]{EL accuracy for corner case mentions where the correct entity is not in the mention's candidate list.} 
    \label{ch_coreflinker:tab:results_corner}
\end{table}

In order to tackle research question \ref{ch_coreflinker:it:q-corner}, we study the accuracy of our models on the important corner case that involves mentions without correct entity in their candidate lists.
This is illustrated in \Tabref{ch_coreflinker:tab:results_corner}, which focuses on such mentions 
in clusters where at least one mention contains the correct entity in its candidate list.
As expected, the 
\Baseline~model cannot link such mentions, as it is limited to the local 
candidate list.
In contrast, both our joint approaches can solve some of these cases by using the correct candidates from other mentions in the cluster, with a superior performance of our {\mtt} model compared to the {\clinker} one.

\section{Related work}
\label{ch_coreflinker:sec:related}
\textbf{Entity Linking:} Related work in entity linking (EL) tackles the document-level linking coherence by exploring relations between entities \cite{kolitsas2018end,yang2019learning,le2019boosting}, or entities and mentions \cite{le2018improving}.
More recently, contextual BERT-driven \cite{devlin2019bert} language models have been used for the EL task  \cite{broscheit2019investigating,de2020autoregressive,de2021highly,yamada2020global} by jointly embedding mentions and entities. 
In contrast, we explore a cluster-based EL approach where the coherence is achieved on \textit{coreferent} entity mentions level.

\noindent\textbf{Coreference Resolution:} Span-based antecedent-ranking coreference resolution \cite{lee2017end,lee2018higher} has 
seen
a recent boost by using SpanBERT representations \cite{xu2020revealing,joshi2020spanbert,wu2020corefqa}.
We extend this approach in our {\clinker} joint coref+EL architecture.
Furthermore, 
we rely on Kirchhoff's Matrix Tree Theorem \cite{koo2007structured,william1984tutte} to efficiently train a more expressive spanning tree-based {\mtt} method. 

\noindent\textbf{Joint EL+Coref:} \cite{fahrni2012jointly} 
introduce a more expensive rule-based Integer Linear Programming component to jointly predict coref and EL. 
\cite{durrett2014joint} jointly train coreference and entity linking without enforcing single-entity per cluster consistency. 
More recently, \cite{angell2021clustering,agarwal2021entity} use additional logic to achieve consistent cluster-level entity linking.
In contrast, our 
proposed
approach constrains the space of the predicted spanning trees on a structural level (see \figref{ch_coreflinker:fig:architectures}). 
\section{Conclusion}
\label{ch_coreflinker:sec:conclusion}
We propose two end-to-end models to solve entity linking and coreference resolution tasks in a joint setting. 
Our joint architectures achieve superior performance compared to the standalone counterparts. 
Further analysis reveals that this boost in performance is driven by more coherent predictions on the level of mention clusters (linking to the same entity) and extended candidate entity coverage. 
\pagebreak
\section*{Appendix}
\label{ch_coreflinker:sec:appendix}
\begin{figure}[t]
\centering
\includegraphics[width=.60\columnwidth,trim={0.4cm 22.5cm 15.5cm 1cm},clip]{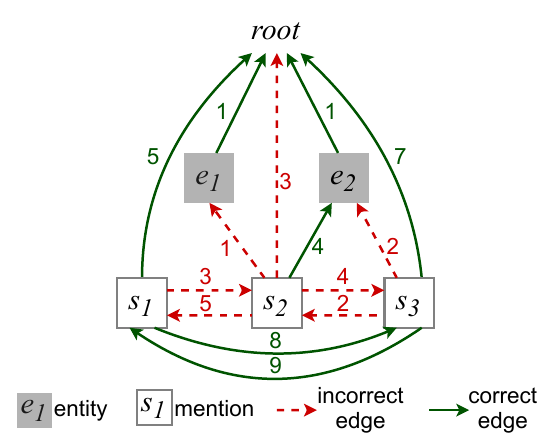}
\captionsetup{singlelinecheck=off}
\caption[Illustrative graph example of {\mtt} model]{Illustrative graph example of {\mtt} model.~The weights of the edges correspond to $\exp(\mathbf{\Phi_{\mathrm{cl}}})$ (see \equref{ch_coreflinker:eq:ex_exp_weight_denominator}). }
\label{ch_coreflinker:fig:illustrative_example}
\end{figure}

In this appendix we will provide a clarifying artificial example 
in order to walk the reader step by step through MTT (\eqsref{ch_coreflinker:eq:loss_mtt}{ch_coreflinker:eq:laplacian}) applied in our \mtt~approach.
The graph of the example is illustrated in \figref{ch_coreflinker:fig:illustrative_example} and is composed by nodes representing $root$ ($r$), entities $e_1$ and $e_2$, and spans $s_1$, $s_2$ and $s_3$. 
The span $s_2$ is associated with candidate entity set $\{e_1, e_2\}$ (\ie represented by edges from $s_2$ to $e_1$ and $e_2$), and $s_3$ with $\{e_2\}$ (\ie represented by the edge from $s_3$ to $e_2$). The candidate entity set of $s_1$ is empty.  
The nodes are grouped in two ground truth clusters: NIL cluster $c_1 = \{s_1, s_2\}$, and linked cluster $c_2 = \{e_2, s_2\}$. 

\renewcommand{\kbldelim}{[}
\renewcommand{\kbrdelim}{]}
The exponential of weighted adjacency matrix\footnote{For simplicity, the weights are small integers.} $\mathbf{\Phi_\mathrm{cl}}$ of the presented example is:
\begin{equation}
\small 
  \exp(\mathbf{\Phi_{\mathrm{cl}}}) = \kbordermatrix{
    & r & e_1 & e_2 & s_1 & s_2 & s_3 \\
    r & 0 & \textcolor{myGreen}{1} & \textcolor{myGreen}{1} & \textcolor{myGreen}{5} & \textcolor{myRed}{3} & \textcolor{myGreen}{7} \\
    e_1 & 0 & 0 & 0 & 0 & \textcolor{myRed}{1} & 0 \\
    e_2 & 0 & 0 & 0 & 0 & \textcolor{myGreen}{4} & \textcolor{myRed}{2} \\
    s_1 & 0 & 0 & 0 & 0 & \textcolor{myRed}{5} & \textcolor{myGreen}{9} \\
    s_2 & 0 & 0 & 0 & \textcolor{myRed}{3} & 0 & \textcolor{myRed}{2} \\
    s_3 & 0 & 0 & 0 & \textcolor{myGreen}{8} & \textcolor{myRed}{4} & 0 
  },\label{ch_coreflinker:eq:ex_exp_weight_denominator}
\end{equation}
where the weights of incorrect edges are represented in \textcolor{myRed}{red} (\ie \textcolor{myRed}{red} dashed edges in \figref{ch_coreflinker:fig:illustrative_example}),
the weights of the correct edges in \textcolor{myGreen}{green} (\ie \textcolor{myGreen}{green} edges in \figref{ch_coreflinker:fig:illustrative_example}),
and the weights between disconnected nodes are set to 0. 

In order to compute the \textit{denominator} of the loss function in \equref{ch_coreflinker:eq:loss_mtt},
the Laplacian of the matrix in \equref{ch_coreflinker:eq:ex_exp_weight_denominator} is calculated as described in \equref{ch_coreflinker:eq:laplacian}, and the row and column corresponding to root $r$ removed (\ie the \textit{minor} $\mathbf{L}_{r}$ with respect to the root):
\begin{equation}
    \mathbf{L}_{r} = \kbordermatrix{
     & e_1 & e_2 & s_1 & s_2 & s_3 \\
    e_1  & 1 & 0 & 0 & -1 & 0 \\
    e_2 & 0 & 1 & 0 & -4 & -2 \\
    s_1 & 0 & 0 & 16 & -5 & -9 \\
    s_2 & 0 & 0 & -3 & 17 & -2 \\
    s_3 & 0 & 0 & -8 & -4 & 20 
  }.
\end{equation}
Following Kirchhoff's Matrix Tree Theorem \cite{koo2007structured,william1984tutte}, the determinant of $\mathbf{L}_{r}$ equals to the sum of the weights of all possible spanning trees of the graph represented in \figref{ch_coreflinker:fig:illustrative_example}: 
\begin{equation}
    \det(\mathbf{L}_{r}) = 3600 = \sum_{t \in \mathcal{T}_\textit{all}} \exp \big( \Phi_\mathrm{tr}(t) \big). \label{ch_coreflinker:eq:denom}
\end{equation}

In order to compute the \textit{numerator} of the loss function in \equref{ch_coreflinker:eq:loss_mtt} (\ie the sum of the weights of the spanning trees of ground truth clusters), we first mask out (set to zero) all the weights assigned to incorrect edges:
\begin{equation}
\small
  \exp(\mathbf{\Phi_{\mathrm{cl}}})' = \kbordermatrix{
    & r & e_1 & e_2 & s_1 & s_2 & s_3 \\
    r & 0 & \textcolor{myGreen}{1} & \textcolor{myGreen}{1} & \textcolor{myGreen}{5} & \textcolor{myRed}{0} & \textcolor{myGreen}{7} \\
    e_1 & 0 & 0 & 0 & 0 & \textcolor{myRed}{0} & 0 \\
    e_2 & 0 & 0 & 0 & 0 & \textcolor{myGreen}{4} & \textcolor{myRed}{0} \\
    s_1 & 0 & 0 & 0 & 0 & \textcolor{myRed}{0} & \textcolor{myGreen}{9} \\
    s_2 & 0 & 0 & 0 & \textcolor{myRed}{0} & 0 & \textcolor{myRed}{0} \\
    s_3 & 0 & 0 & 0 & \textcolor{myGreen}{8} & \textcolor{myRed}{0} & 0 
  }
\end{equation}
Next, the \textit{modified Laplacian} (\ie Laplacian with the first row replaced by root $r$ selection weights) $\mathbf{\hat{L}}$ is calculated for both clusters $c_1$ and $c_2$:
\begin{align}
    \mathbf{\hat{L}}_{c_1} = \kbordermatrix{
     & s_1 & s_3 \\
    r & 5 & 7  \\
    s_3 & -8 & 9 
  } \\
    \mathbf{\hat{L}}_{c_2} = \kbordermatrix{
     & e_2 & s_2 \\
    r & 1 & 0  \\
    s_2 & 0 & 4 
  }
\end{align}
The determinants of $\mathbf{\hat{L}}_{c_1}$ and $\mathbf{\hat{L}}_{c_2}$ equal to the sum of the weights of all spanning trees connecting the nodes in clusters $c_1$ and $c_2$ respectively:
\begin{equation}
    \det(\mathbf{\hat{L}}_{c_1}) = 101 = \sum_{t \in \mathcal{T}_{c_1}} \exp \big(\Phi_\mathrm{tr}(t)\big) \label{ch_coreflinker:eq:numer_1}
\end{equation}
\begin{equation}
    \det(\mathbf{\hat{L}}_{c_2}) = 4 = \sum_{t \in \mathcal{T}_{c_2}} \exp \big(\Phi_\mathrm{tr}(t)\big) \label{ch_coreflinker:eq:numer_2}
\end{equation}
Finally, in order to calculate the final loss, we replace the obtained results in eqs.~(\ref{ch_coreflinker:eq:denom}), (\ref{ch_coreflinker:eq:numer_1}), and (\ref{ch_coreflinker:eq:numer_2})
in the loss function of \equref{ch_coreflinker:eq:loss_mtt}: 
\begin{equation}
    \mathcal{L} = - \log \frac{101*4}{3600}. \label{ch_coreflinker:eq:ex_loss}
\end{equation}
\textit{Note}: strictly speaking, there are \textit{three} clusters rooted in \textit{root} in the graph of \figref{ch_coreflinker:fig:illustrative_example}, the third one being $c_3=\{e_1\}$, whose exponential weight is 1 by definition of $\Phi_{\mathrm{cl}}(r, e_j) = 0$ (see \secref{ch_coreflinker:sec:joint_approaches}), and has no impact in calculation of the loss function in \equref{ch_coreflinker:eq:ex_loss}. 

\renewcommand*{\thesection}{\thechapter.\arabic{section}}       


\graphicspath{{klim_ch_injecting/figures/}}

\newcommand{\docreddataset}{DocRED}

\newcommand{\highlights}[1]{\textcolor[RGB]{0,104,180}{#1}}
\newcommand{\klimcomment}[1]{\textcolor[rgb]{0,0.5,0.5}{\underline{KZ}: #1}}
\newcommand{\oldignore}[1]{}
\newcommand{\red}[1]{\textcolor{red}{#1}}
\definecolor{RoyalBlue}{cmyk}{1, 0.50, 0, 0}
%

\chapter{Injecting Knowledge Base Information into End-to-End Joint Entity and Relation Extraction and Coreference Resolution}
\label{chap:injecting_knowledge}

\renewcommand\evenpagerightmark{{\scshape\small Chapter \arabic{chapter}}}
\renewcommand\oddpageleftmark{{\scshape\small Injecting Knowledge Base Information}}

\renewcommand{\bibname}{References}


\begin{flushright}
\end{flushright}


\noindent 
\emph{
In this chapter we adopt a slightly different approach involving entities: instead of using purely textual information to solve information extraction tasks such as relation extraction, we also study how the information of entities from Knowledge Base can be integrated. We achieve significant improvement on all the evaluated tasks by injecting information both from Wikipedia, as well as from Wikidata KBs. Furthermore, while the tasks we are tackling are annotated and defined on \textit{named entity} level, the information we inject in our text comes from all the existing entities defined in the experimented KBs. We find that this unsupervised technique is still able to detect the entities that are more relevant for a particular text.  
}

\begin{center}
\par{$\star\star\star$}
\end{center}
\vspace{0.15in}

\par{\noindent\large{\textbf{S.~Verlinden\textsuperscript{$*$}, K.~Zaporojets\textsuperscript{$*$}, J.~Deleu, T.~Demeester and C.~Develder}}}
\vspace{0.1in}
\par{\noindent\textbf{Findings of the Association for Computational Linguistics: ACL-IJCNLP 2021}}
\vspace{0.15in}


\begingroup\renewcommand\thefootnote{$*$}
\footnotetext{Equal contribution}
\endgroup
\par{\noindent\bf{Abstract}}
We consider a joint information extraction (IE) model, solving named entity recognition, coreference resolution and relation extraction 
jointly over the whole document.
In particular, we study how to inject information from a knowledge base (KB) in such IE model, based on unsupervised entity linking. 
The used KB entity representations are learned from either
(i)~hyperlinked text documents (Wikipedia), or
(ii)~a knowledge graph (Wikidata), and
appear complementary in raising IE performance.
Representations of corresponding entity linking (EL) candidates are added 
to text span representations of the input document, and we experiment with  
(i)~taking a weighted average of the EL candidate representations based on their prior (in Wikipedia), and
(ii)~using an attention scheme over the EL candidate list.
Results demonstrate an increase of up to 5\% F1-score for the evaluated IE tasks on two datasets.
Despite a strong performance of the prior-based model, our quantitative and qualitative analysis reveals the advantage of using the attention-based approach.

\section{Introduction}
\label{ch_injecting:sec:introduction}
Information extraction (IE) comprises several subtasks, \eg named entity recognition (NER), coreference resolution (coref), relation extraction (RE). State-of-the-art results mainly report performance on single tasks, usually solving them on a sentence level (especially NER, RE).
However, in practice, IE system decisions should be consistent on the document level, \eg when processing news articles to automatically link entities (aside from potentially learning, \eg new relations).
Yet, the challenge of solving the tasks jointly on a document level has not received as much attention and remains hard \cite{durrett2014joint, yao2019docred, zaporojets2021dwie}.

On the other hand, it is well established that IE models benefit from incorporating background information of knowledge bases (KBs).
Still, so far this has been shown from the perspective of solving individual tasks such as relation classification or entity typing (\eg \cite{peters2019knowledge, liu2020k}).
Integrating KBs in joint models, realizing and analyzing the more complex end-to-end setting, has been left unexplored.

In terms of the nature of KBs adopted in IE, current approaches use either
\begin{enumerate*}[(i)]
\item structured knowledge \emph{graphs} comprising \texttt{(subj,rel,obj)} triples, \eg Wikidata \cite{yang2017leveraging,han2018neural,zhang2019long}, or
\item \emph{textual} descriptions, usually in hyperlinked  documents, \eg Wikipedia \cite{
martins2019joint,
yamada2020luke}.
\end{enumerate*}
It has not been established to what extent KB-text and KB-graph 
entity
representations complement each other in boosting IE performance.

We address both research gaps of 
\begin{enumerate*}[(a)]
\item integrating KB information into a joint end-to-end IE model for solving named entity recognition, coreference resolution and relation extraction, and
\item analyzing what KB representation is more beneficial for IE, either 
  \emph{KB-graph} trained on Wikidata, or 
  \emph{KB-text} trained directly on Wikipedia.
\end{enumerate*}
We particularly contribute:
\begin{enumerate*}[(i)]
\item a first span-based end-to-end architecture incorporating KB knowledge in a joint entity-centric setting, exploiting unsupervised entity linking (EL) to select KB entity candidates,
\item exploration of prior- and attention-based mechanisms to combine the EL candidate representations into the model,
\item assessment of the complementarity of KB-graph and KB-text representations, and
\item consistent gains of up to 5\% F1-score when incorporating KB knowledge in 3 document-level IE tasks evaluated on 2 different datasets. 
\end{enumerate*}

\oldignore{
\textbf{------------ OPTION 2: start from KB-injection to build story --------}
It is well established that models to solve information extraction (IE) tasks can benefit from incorporating background information contained in knowledge bases (KBs).
Current approaches make use of KBs comprising either
\begin{enumerate*}[(i)]
\item \emph{textual} descriptions, usually in hyperlinked  documents, \eg Wikipedia \cite{zhang2019ernie,yamada2020luke}, or
\item structured knowledge \emph{graphs} (KGs), \eg Wikidata \cite{yang2017leveraging,han2018neural,zhang2019long}.
\end{enumerate*}
It has not been established to what extent such KB text-based and graph-based representations complement each other in boosting IE performance.

Further, efforts in leveraging KB background knowledge in IE have mainly focused on solving individual tasks (\eg \cite{peters2019knowledge, liu2020k}), such as relation classification or entity typing. Integrating KBs into joint models, thus realizing and analyzing the more complex end-to-end setting, has been left unexplored. 

The current paper addresses both research gaps through exploring the impact of integrating two types of KBs into a joint end-to-end IE setting, simultaneously solving named entity recognition, coreference resolution and relation extraction on a document level. 
As KBs we consider 
\begin{enumerate*}[(i)]
\item a Wikipedia corpus of hyperlinked entity pages, and
\item a Wikidata knowledge graph comprising \texttt{(subj,rel,obj)} triples for the same.
\end{enumerate*}
Specifically, we derive text span representations for the input document, and use heuristic string matching as an unsupervised entity linking (EL) system to find candidate entities for it from the KB. For the selected entities, we add KB entity representations to the spans before feeding them as input to the specific IE task modules. Entity representations are either:
\begin{enumerate*}[(i)]
\item \emph{KB-text}: trained on Wikipedia \cite{yamada2016joint} , or
\item \emph{KB-graph}: trained on Wikidata \cite{joulin2017fast}.
\end{enumerate*}
To combine the multiple EL candidate representations, we experiment with different prior- and attention-based mechanisms.
Our results show the complementary nature of text- and graph-based KB entity representations, achieving best performance by combining them.
}

\oldignore{
The background information contained in the knowledge bases (KBs) can be used to boost the performance of Information Extraction (IE) tasks. 
However, most of the 
related work has either
focused on exploring the impact of background knowledge from KB entity description corpora (\eg Wikipedia) \cite{zhang2019ernie,yamada2020luke} or from purely structured knowledge graphs (KGs; \eg Wikidata) \cite{han2018neural,yang2017leveraging,zhang2019long} independently without exploring how these two sources can complement each other. 
Additionally, 
current efforts in leveraging background knowledge 
have focused on solving individual
IE tasks \cite{liu2020k,peters2019knowledge} such as relation classification or entity typing, leaving unexplored its impact in a more complex end-to-end setting.

In this work, we tackle these research gaps by 
exploring the impact of integrating two background knowledge sources in a joint entity-centric end-to-end IE setting: \begin{enumerate*}[(i)]
    \item Wikidata knowledge graph composed by \texttt{(subj,rel,obj)} tuples, and
    \item Wikipedia corpus composed by hyperlinked web pages
\end{enumerate*}. 
In order to achieve this, we associate each of the textual spans in the input document with top entity linking candidate entities based on the prior probability. 
The candidate entities are represented with KB embeddings trained either on Wikidata KG \cite{joulin2017fast} or on Wikipedia hyperlinked web pages \cite{yamada2016joint}. 
We experiment with different prior and attention-based mechanisms to combine the information from the entity representations in the candidate list. Our experimental results show the complementary nature of the information from the Wikidata and Wikipedia, achieving the biggest performance boost when using both of the background knowledge sources jointly. 
}

\oldignore{
Our contribution is four-fold:
\begin{enumerate*}[(i)]
\item propose a 
first
span-based end-to-end architecture
to incorporate background knowledge in a joint entity-centric setting,
\item explore attention and prior-based mechanisms to combine the EL candidate embeddings,
\item Explore the impact of KG vs KB-derived embeddings, 
\item obtain consistent gains of up to 5 \% F1 score when incorporating background knowledge in three document-level IE tasks evaluated on two different
datasets. 
\end{enumerate*}
}

\section{Model}
\label{ch_injecting:sec:model}
\begin{figure*}[ht]
\centering
\includegraphics[width=1.0\textwidth]{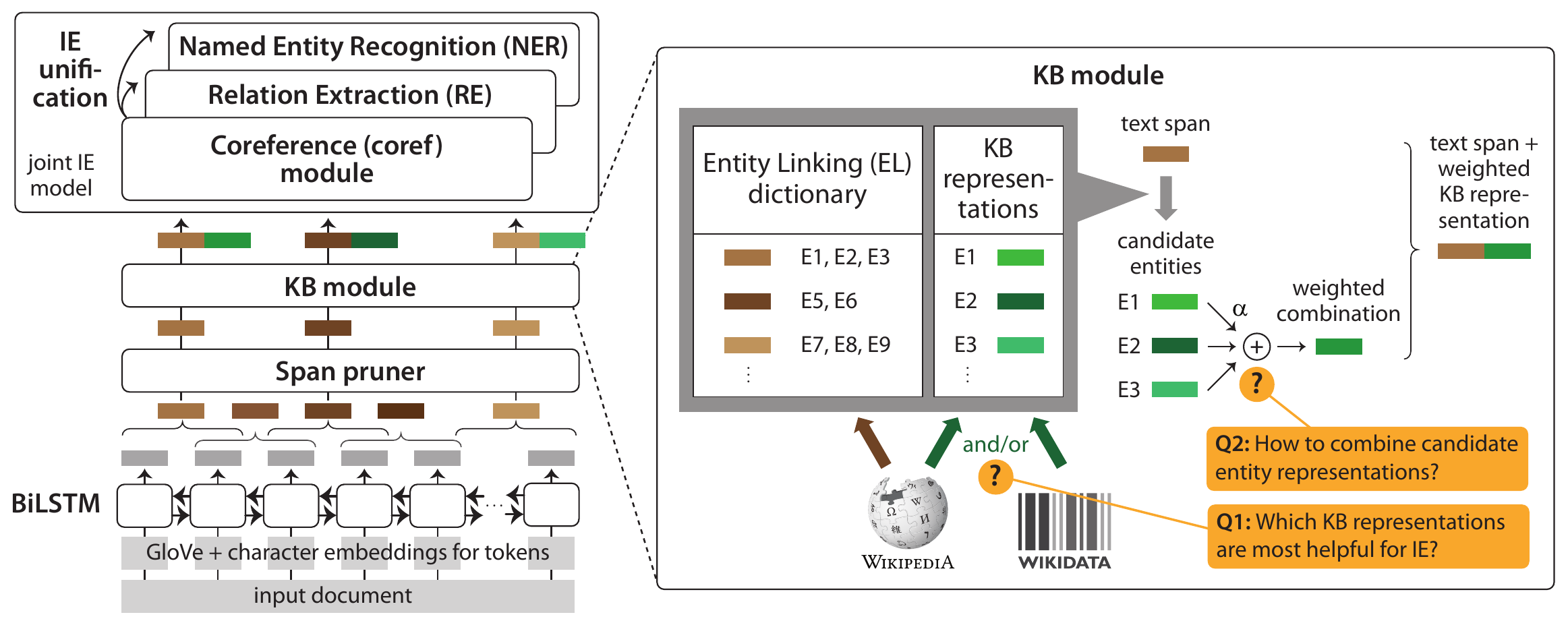}
\caption[Joint information extraction model with addition of a knowledge base module]{Joint information extraction (IE) model with addition of a knowledge base (KB) module.}
\label{ch_injecting:fig:model-overview}
\end{figure*}

\Figref{ch_injecting:fig:model-overview} illustrates our model architecture.
Input document tokens are represented using concatenated GloVe \cite{pennington2014} and character embeddings \cite{ma2016} and pushed through a BiLSTM to obtain contextualized token representations, which are combined into spans.
Similar to \cite{luan2019general,zaporojets2021dwie}, a span pruner limits the number of spans for downstream modules.
The \emph{KB module} (\secref{ch_injecting:sec:kb-module}) combines span representations with KB entity representations (\secref{ch_injecting:sec:model-embeddings}), trained either on Wikidata (\emph{KB-graph}) or Wikipedia (\emph{KB-text}).
The KB-enriched span representations then serve as input for joint predictions on downstream IE tasks (\secref{ch_injecting:sec:model-joint}). 


\subsection{Entity representations}
\label{ch_injecting:sec:model-embeddings}

We experiment with 3 possible entity representations: \emph{KB-text}, \emph{KB-graph}, and concatenating \emph{both}. 

\noindent\textbf{KB-text}: We follow \cite{yamada2016joint} 
to obtain the entity representations
using a skip-gram architecture \cite{mikolov2013efficient,mikolov2013distributed}, 
training to jointly predict
\begin{enumerate*}[(i)]
    \item the linked entities (through Wikipedia hyperlinks) given the target entity, and 
    \item the neighboring words for a given entity hyperlink. 
\end{enumerate*}

\noindent\textbf{KB-graph}: We adopt \cite{joulin2017fast} to train the entity embeddings directly on Wikidata triples \texttt{(subj,rel,obj)} by optimizing a linear classifier to predict the \texttt{obj} entity from the \texttt{subj} entity and the relation type \texttt{rel}. 

\oldignore{Unlike the KB entity embeddings, the KG representations are trained directly on the triples of Wikidata knowledge graph. We use the approach proposed by \cite{joulin2017fast} based on a linear classifier to predict the predicate entity according to the subject entity and the relation type. }

\subsection{KB module}
\label{ch_injecting:sec:kb-module}
For a span $s_i$ from token $l$ to $r$, we obtain the 
representation $\textbf{g}_i$
as input to the KB module
by concatenating the respective hidden LSTM states $\textbf{h}_l$ and $\textbf{h}_r$, and an embedding $\boldsymbol{\psi}_{r-l}$ for the corresponding span width $r-l$:
\begin{equation}
\textbf{g}_i = [\textbf{h}_{l}; \textbf{h}_{r}; \boldsymbol{\psi}_{r-l}]. \label{ch_injecting:eq:spanrepr} 
\end{equation}
We look up a given span $s_i$ in a dictionary built from Wikipedia, to determine its candidate entities set\footnote{We limit this to the 16 most frequent ones.} $C_i$, as well as the prior probability $p_{ij}$ for each $c_{ij} \in C_i$, as per \cite[\S3]{yamada2016joint}.

To combine the KB candidates $c_{ij}$, we either use
\begin{enumerate*}[(i)]
\item a uniform average (\emph{Uniform}),
\item the prior weights $p_{ij}$ (\emph{Prior}), 
\item an attention scheme (\emph{Attention}), or
\item attention with prior information (\emph{AttPrior}).
\end{enumerate*}
The unnormalized 
attention scores for \emph{Attention} and \emph{AttPrior} are:
\begin{align}
 \Phi_\textit{Attention}(s_i,c_{ij},\textsc{k}) = \mathcal{F}_\textit{A} \left( [\textbf{g}_i; \boldsymbol{\xi}_\textsc{k}(c_{ij}) ]\right) \label{ch_injecting:eq:att} \\
   \Phi_\textit{AttPrior}(s_i,c_{ij},\textsc{k}) = \mathcal{F}_\textit{AP} \big( [\textbf{g}_i; \boldsymbol{\xi}_\textsc{k}(c_{ij}); p_{ij} ]\big) \label{ch_injecting:eq:attprior}   
\end{align}
where $\textsc{k} \in \{\emph{KB-text}, \emph{KB-graph}, \emph{both}\}$ refers to the entity representations from \secref{ch_injecting:sec:model-embeddings}, $\boldsymbol{\xi}_\textsc{k}$ returns such representation for $c_{ij}$, and $\mathcal{F}_{*}$ is a feed-forward neural network (FFNN).
%
The KB representation for span $s_i$ is a weighted average of its candidates $C_i$:
\begin{equation}
    \textbf{e}_i^\textsc{k} = \sum_{c_{ij} \in C_i} 
    \mathcal{\alpha}_{ij} \cdot \boldsymbol{\xi}_\textsc{k}(c_{ij}) \label{ch_injecting:eq:end_emb}     
\end{equation}
where weights $\mathcal{\alpha}_{ij}$ either are uniform ($1/\left|C_i\right|$), the prior $p_{ij}$, or softmax-normalized attention scores (softmax over $\Phi$ from \equref{ch_injecting:eq:att} or \equref{ch_injecting:eq:attprior}).
The concatenation $[\textbf{g}_i; \textbf{e}_i^\textsc{k}]$ forms the KB-enriched representation for span $s_i$, as input for IE modules (\secref{ch_injecting:sec:model-joint}).

\subsection{Joint IE model}
\label{ch_injecting:sec:model-joint}

The joint IE model comprises 3 modules (\figref{ch_injecting:fig:model-overview}) using the same KB-enriched representations $[\textbf{g}_i; \textbf{e}_i^\textsc{k}]$, and using a weighted combination of the 3 module losses to minimize during training. Note that NER and RE are framed as multi-label classification. 

\noindent\textbf{NER module}: We use a FFNN on each span $s_i$ to produce scores $\boldsymbol{\Phi}_{\textsc{ner}} (s_i) \in \mathbb{R}^{|L_E|}$, with $L_E$ the set of possible entity types.
At inference, we accept type $l \in L_E$ for span $s_i$ if $\boldsymbol{\Phi}_{\textsc{ner}}(s_i)_l>0$. 

\noindent\textbf{Coref module}: We use the coreference scheme proposed by \cite{lee2017end}, using a FFNN to produce scores $\Phi_{\mathrm{coref}} (s_i, s_j)$: at inference time, the highest scoring antecedent of span $s_j$ is then chosen (potentially $s_j$ itself). 
Indeed, to allow for singletons we accept self-references $(s_j, s_j)$ if NER predicts the span $s_j$ to be an entity.

\noindent\textbf{RE module}: Similar to \cite{luan2019general,luan2018multi}, we use a FFNN to produce scores  $\boldsymbol{\Phi}_{\textsc{re}} (s_i, s_j) \in \mathbb{R}^{|L_R|}$ for each pair of spans $(s_i,s_j)$, with $L_R$ the set of relation types.
We accept relation $l \in L_R$ for pair $(s_i,s_j)$ if $\boldsymbol{\Phi}_{\textsc{re}} (s_i, s_j)_l > 0$. 

\noindent\textbf{IE unification}: Above modules make span level predictions. We obtain entity-centric predictions using the coref clusters, by assigning the union of predicted entity/relation types within a coref cluster to all its members, as do \cite{zaporojets2021dwie}. 


\section{Experimental setup}
\label{ch_injecting:sec:experimental_setup}

\begin{table}[t]
    \centering
    \resizebox{0.7\columnwidth}{!}
    {\begin{tabular}{lcccc}
        \toprule
         \multirow{2}{*}{Dataset} & \# Entity & \# Entity & \multirow{2}{*}{\# Relations} & \# Relation \\
          & clusters & types &  & types \\         
         \midrule
         \dwiedataset & 23,130 & 311 & 21,749 & 65 \\
         \docreddataset & 98,610 & 6 & 50,503 & 96 \\
    \bottomrule
    \end{tabular}}
    \caption[Dataset statistics]{Dataset statistics.} 
    \label{ch_injecting:tab:dataset}
\end{table}

We evaluate our proposed models\footnote{\revklim{Code and models available at \url{https://github.com/klimzaporojets/e2e-kb-ie}.}} on  entity-centric multi-task datasets, summarized in \Tabref{ch_injecting:tab:dataset}: {\dwiedataset} \cite{zaporojets2021dwie} and {\docreddataset} \cite{yao2019docred}. 
We report on coreference resolution (coref), NER and relation extraction (RE).
For coref, we report the average of 3 common F1 scores, as implemented by \cite{pradhan2014scoring}:
MUC~\cite{vilain1995model}, B$^3$~\cite{bagga1998algorithms} and {\CEAFe} \cite{luo2005coreference}.
Since we focus on entity-centric, document-level IE, for NER and RE we use \emph{hard} metrics \cite{zaporojets2021dwie} on the level of entity clusters (\ie aforementioned coref clusters): predictions are counted as correct only if
\begin{enumerate*}[(i)]
    \item all mentions (with exact boundary match) are present in the entity cluster, and 
    \item the predicted entity type (for NER) or relation type between two clusters (for RE) is correct.
\end{enumerate*} 

\oldignore{
\textbf{---------OLD TEXT----------\\}
To report the results for coreference resolution we use the average F1 scores of MUC \cite{vilain1995model}, B$^3$ \cite{bagga1998algorithms}, and {\CEAFe} \cite{luo2005coreference} metrics as implemented by \cite{pradhan2014scoring}. For the NER and RE tasks we use the \emph{hard} F1 scoring metric described in \cite{zaporojets2021dwie}. This metric only considers an entity or relation to be correctly predicted if \begin{enumerate*}[(i)]
    \item all the entity mentions (exact boundary match) are present in the entity cluster, and 
    \item the predicted entity cluster type (for NER) or relation type between two clusters (for RE) is correct
\end{enumerate*}.    
}

\oldignore{
We divide our model setup based on the used weighting schema in \begin{enumerate*}[(i)]
    \item \emph{Uniform} where the candidate entities are averaged,
    \item \emph{Attention} using the attention-based weighting schema (\equref{ch_injecting:eq:att}),
    \item \emph{AttPrior} that uses the attention+prior weighting schema (\equref{ch_injecting:eq:attprior}), and 
    \item \emph{Prior} where the prior weights are used to perform the weighted average of candidate entity embeddings. 
\end{enumerate*}
}

Our experiments address 2 main questions (see \figref{ch_injecting:fig:model-overview}):
\begin{enumerate*}[label=\textbf{(Q\arabic*)}]
   \item \label{ch_injecting:it:q-kb-type} Which type of KB representation is most helpful for IE (\emph{KB-text}, \emph{KB-graph}, or \emph{both}; see \secref{ch_injecting:sec:model-embeddings})?
    \item \label{ch_injecting:it:q-attention} Which weighting scheme to use for $\alpha$ (\emph{Uniform}, \emph{Prior}, \emph{Attention}, \emph{AttPrior}; see \secref{ch_injecting:sec:kb-module})?
\end{enumerate*}


\section{Results}
\label{ch_injecting:sec:results}

\begin{table*}
\centering
\resizebox{1.0\textwidth}{!}{\begin{tabular}{ccclllllll}
\toprule
 &  & \multicolumn{3}{c}{\dwiedataset} && \multicolumn{3}{c}{\docreddataset}\\
\cmidrule(lr){3-5}\cmidrule(lr){7-9}
 KB Source & Setup & \multicolumn{1}{c}{Coref } &    \multicolumn{1}{c}{NER} & \multicolumn{1}{c}{RE} && \multicolumn{1}{c}{Coref } &    \multicolumn{1}{c}{NER} & \multicolumn{1}{c}{RE}\\
\toprule
-- & Baseline & 90.0${\scriptstyle \pm\text{0.2}}$
& 71.7${\scriptstyle \pm\text{0.5}}$ & 47.0${\scriptstyle \pm \text{1.4}}$ && 81.9${\scriptstyle \pm \text{0.3}}$ & 68.5${\scriptstyle \pm \text{0.3}}$ & 23.5${\scriptstyle \pm \text{0.6}}$ \\
\midrule
 & Uniform & 90.7${\scriptstyle \pm\text{0.2}}$ & 73.5${\scriptstyle \pm \text{0.5}}$ & 48.5${\scriptstyle \pm \text{1.1}}$ && 82.9${\scriptstyle \pm \text{0.1}}$ & 70.7${\scriptstyle \pm \text{0.2}}$ & 24.5${\scriptstyle \pm \text{0.3}}$\\
KB-text  & Attention&  \textbf{90.7${\scriptstyle \pm \text{0.3}}$} & 73.4${\scriptstyle \pm \text{0.8}} $ & 49.0${\scriptstyle \pm \text{0.4}}$ && \textbf{83.4${\scriptstyle \pm \text{0.1}}$} & 71.2${\scriptstyle \pm \text{0.1}}$ & 24.5${\scriptstyle \pm \text{0.3}}$\\
 & AttPrior& 90.7${\scriptstyle \pm\text{0.3}} $ & 73.7${\scriptstyle \pm\text{0.6}}$ & \textbf{49.6${\scriptstyle \pm\text{0.8}}$} && 83.2${\scriptstyle \pm\text{0.2}}$ & \textbf{71.3${\scriptstyle \pm\text{0.2}}$} & 24.8${\scriptstyle \pm\text{0.4}}$\\
 & Prior& 90.7${\scriptstyle \pm\text{0.2}}$ & \textbf{73.8${\scriptstyle \pm\text{0.5}}$} & 49.4${\scriptstyle \pm\text{0.4}}$ && 82.9${\scriptstyle \pm\text{0.2}}$ & 70.9${\scriptstyle \pm\text{0.3}} $ & \textbf{25.3${\scriptstyle \pm\text{0.4}}$}\\
\midrule
  & Uniform & 91.0${\scriptstyle \pm\text{0.3}} $ & 73.6${\scriptstyle \pm\text{0.4}} $ &  48.0${\scriptstyle \pm\text{1.2}}$ && 83.3${\scriptstyle \pm\text{0.2}} $ &  71.1${\scriptstyle \pm\text{0.2}} $ & 24.9${\scriptstyle \pm\text{0.2}}$\\
KB-graph & Attention & 91.2${\scriptstyle \pm\text{0.3}}$ & 73.9${\scriptstyle \pm\text{0.5}}$ & 50.1${\scriptstyle \pm\text{1.1}}$ && \underline{\textbf{83.7${\scriptstyle \pm\text{0.1}}$}} & \textbf{71.6${\scriptstyle \pm\text{0.1}}$} & 25.0${\scriptstyle \pm\text{0.4}}$\\
  & AttPrior & \textbf{91.3${\scriptstyle \pm\text{0.2}}$} & \textbf{74.6${\scriptstyle \pm\text{0.3}}$} & \textbf{50.5${\scriptstyle \pm\text{1.0}}$} && 83.5${\scriptstyle \pm\text{0.3}}$ & 71.5${\scriptstyle \pm\text{0.2}}$ & 25.1${\scriptstyle \pm\text{0.2}}$\\
  & Prior & 90.8${\scriptstyle \pm\text{0.3}}$ & 73.6${\scriptstyle \pm\text{0.6}}$ & 49.6${\scriptstyle \pm\text{1.1}}$ && 83.4${\scriptstyle \pm\text{0.1}}$ & 71.1${\scriptstyle \pm\text{0.1}}$ & \textbf{25.2${\scriptstyle \pm\text{0.2}}$}\\
\midrule
both & Uniform & 91.1${\scriptstyle \pm\text{0.1}}$  & 74.1${\scriptstyle \pm\text{0.5}}$ & 49.3${\scriptstyle \pm\text{0.5}}$ && 83.5${\scriptstyle \pm\text{0.1}}$ & 71.3${\scriptstyle \pm\text{0.2}}$  & 24.8${\scriptstyle \pm\text{0.1}}$\\
(KB-graph + & Attention & 91.2${\scriptstyle \pm\text{0.3}} $ & 74.3${\scriptstyle \pm\text{0.6}}$ & 51.3${\scriptstyle \pm\text{1.3}}$ && 83.5${\scriptstyle \pm\text{0.2}}$ & 71.5${\scriptstyle \pm\text{0.1}}$ & 24.8${\scriptstyle \pm\text{0.3}}$\\
KB-text) & AttPrior & \underline{\textbf{91.5${\scriptstyle \pm\text{0.2}}$}} & \underline{\textbf{75.0${\scriptstyle \pm\text{0.4}}$}} & \underline{\textbf{52.1${\scriptstyle \pm\text{1.2}}$}} && \textbf{83.6${\scriptstyle \pm\text{0.2}}$} & \underline{\textbf{71.8${\scriptstyle \pm\text{0.3}}$}} & \underline{\textbf{25.7${\scriptstyle \pm\text{0.7}}$}}\\
  & Prior & 90.8${\scriptstyle \pm\text{0.1}}$ & 73.8${\scriptstyle \pm\text{0.2}}$ & 49.8${\scriptstyle \pm\text{1.2}}$ && 83.2${\scriptstyle \pm\text{0.1}}$ & 71.2${\scriptstyle \pm\text{0.1}}$ & 25.1${\scriptstyle \pm\text{0.3}}$\\
\bottomrule
\end{tabular}}
\caption[General experimental results]{Main results of the experiments in F1 scores grouped by the background KB source. We report Avg.\ F1 scores of MUC, B$^3$ and {\CEAFe} for Coref, and hard F1 metrics for NER and RE. \textbf{Bold} font indicates the best results for each of the different \emph{KB source} types. Additionally, the best overall results are \underline{underlined}.}
\label{ch_injecting:tab:overview_results}
\end{table*}

We summarize the comparison of various model choices for both {\dwiedataset} and {\docreddataset} datasets in \Tabref{ch_injecting:tab:overview_results}.
First, looking into \ref{ch_injecting:it:q-kb-type}, we note that including background information from \emph{KB-graph} and \emph{KB-text} significantly boosts performance compared to the \emph{Baseline} without any KB. \revklim{Additionally, our model outperforms the results from \cite{zaporojets2021dwie} (not listed in the table) by about 2 percentage points F1, using the same input (GloVe) representations.}
Furthermore, 
\revklim{we observe a general improvement in results when combining \emph{both} representations,}
suggesting that a (hyper)text corpus (Wikipedia) and a knowledge graph (Wikidata) embed complementary information for raising IE performance. 

Deeper analysis reveals that adding KB representations mainly benefits performance for ``rare'' entity types: \eg limiting the test set to entity types that occur $\leq$50 times in the training set for {\dwiedataset}, compared to \emph{Baseline}, F1 for NER goes up by $+$13.9 for \emph{KB-both} with \emph{AttPrior}, while the benefit gradually decreases for more frequently occurring entity types.
For RE, we note that overall we also see a clear 
performance gain from adding KB information (\eg +5.1\% F1 for \emph{both} KB sources with \emph{AttProp} compared to \emph{Baseline} for {\dwiedataset}), yet the boost is not as clear for relations with fewer training instances.
(The latter makes sense, since we inject KB representations of entities rather than explicitly also for relations; we leave studying adding relation embedding information for future work.)
\begin{figure}[t]
\centering
\includegraphics[width=0.6\columnwidth, trim={0.2cm 10.2cm 16.4cm 0.2cm},clip]{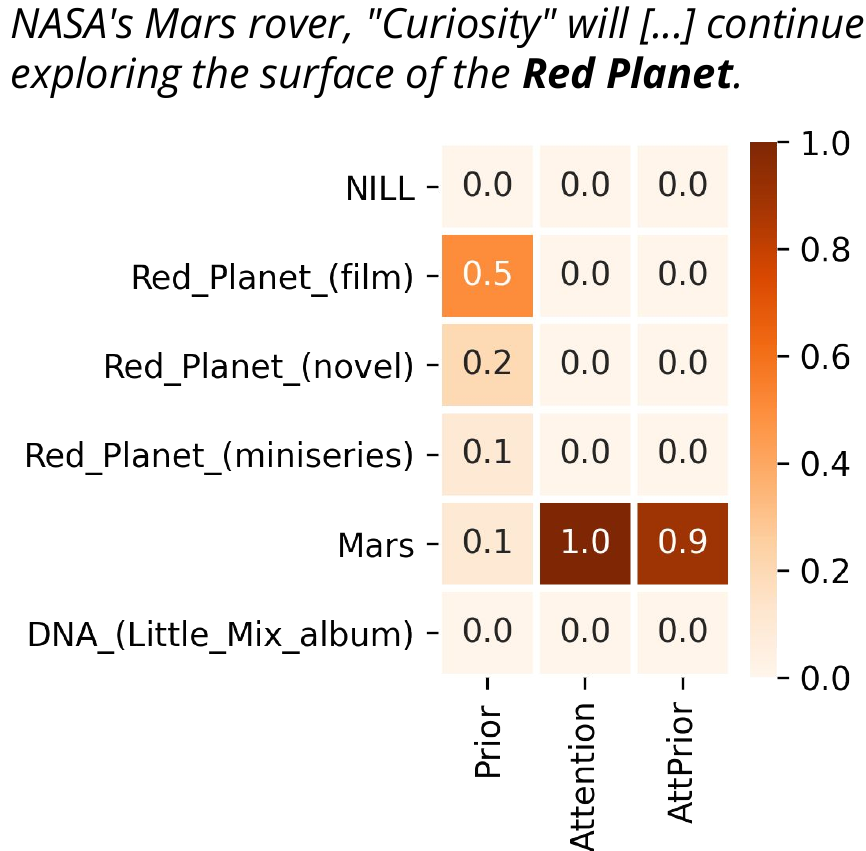}
\captionsetup{singlelinecheck=off}
\caption[Illustration of EL candidate weighting]{Illustration of EL candidate weighting: the $\alpha$ weights for top candidates for ``Red Planet'' from the example sentence at the top. 
Attention-based weighting (\emph{Attention}, \emph{AttPrior}) correctly identify the ``Mars'' entity, while the Wikipedia-based \emph{Prior} fails, as most of Wikipedia's ``Red Planet'' links refer to the film.}
\label{ch_injecting:fig:concrete_example}
\end{figure}

Second, for \ref{ch_injecting:it:q-attention}, we note that the \emph{AttPrior} scheme is the overall winner among the different EL candidate weigthing schemes. 
We observed that in terms of ranking EL candidates, \emph{Prior} performs quite well on {\dwiedataset} --- for 86.5\% of entity mentions it assigns the highest score to the correct EL candidate, while \emph{Attention} and \emph{AttPrior} achieve it for 46.2\%, resp.\ 77.2\% of the mentions --- which basically confirms that 
{\dwiedataset}
has a similar entity distribution as Wikipedia.\footnote{{\dwiedataset} is a news article corpus.
}
Yet, it seems necessary to include alternative candidates, and the attention-based schemes thus can correct EL mistakes of \emph{Prior}, as illustrated in \figref{ch_injecting:fig:concrete_example}.
This correction leads to a resulting boost for the IE tasks as reported in \Tabref{ch_injecting:tab:overview_results}.
E.g., we found that for {\dwiedataset}, looking at clusters with entity mentions for which \emph{Prior} makes wrong EL predictions, the \emph{AttPrior} weighting scheme retrieves $+$3.7\% more of the gold standard annotated named entities (as opposed to just $+$0.6\% in the clusters with correct \emph{Prior} EL candidates).
Perfecting the EL prediction would potentially boost IE performance even more.

\oldignore{
\textbf{--------OLD TEXT------}
To our surprise, the \emph{Prior} setup exhibits a rather competitive performance. We hypothesize that this can be explained by a good performance of the prior score for the correct candidate entity
(highest score for correct candidate for 86.5\% of entity mentions in comparison of 
46.2\%
of \textit{Attention} and 
70.4\%
of \textit{AttPrior} weighting schemes). 
However, from our qualitative analysis, the attention-based methods performs better in corner-cases where ambiguous entity mentions are used 
to refer to less frequently co-occurring entities.
In these cases, the attention-based methods benefit from additional contextual information given by the text surrounding the span to disambiguate. This is illustrated in \figref{ch_injecting:fig:concrete_example} where the attention-based mechanisms are able to correctly capture the planet \emph{Mars} as the most relevant entity from the candidate list, something that is missed by the prior weighting schema. 
We hypothesize that a smaller performance gain of attention-based weighting schemas with respect to the prior in \docreddataset~is explained by an even better prior performance due to the fact that \docreddataset~uses the same corpus (Wikipedia articles) that we use to calculate the candidate priors. 
%
An additional analysis on \dwiedataset~reveals that for entity clusters involving mentions with incorrect prior, attention-based methods are able to recover 3.7\% more correct entities and 5.5\% more correct relations compared to the prior. Conversely, this difference is of only 0.6\% for entities and 2.0\% for relations for clusters where the candidate entities are correctly scored by the prior.
\oldignore{An additional analysis on \dwiedataset~reveals that for entity clusters involving mentions with incorrect prior, the performance gap between attention-based and prior setups is 3.2\% F1 for NER and 5.5\% on RE tasks. Conversely, this difference is of only 0.5\% F1 for NER and 1.1\% for RE for clusters where all the candidate entities are correctly scored by the prior. }
}  

\section{Related work}
\label{ch_injecting:sec:related_work}
As stated earlier, we studied how to integrate 
\begin{enumerate*}[(i)]
\item \label{ch_injecting:it:kb-into-ie} knowledge base information into IE, and particularly
\item \label{ch_injecting:it:e2e-ie} end-to-end IE combining multiple tasks (NER, relation extraction, coreference resolution), while 
\item \label{ch_injecting:it:entity-centric-ie} taking an entity-centric perspective, \ie focus on making consistent decisions on the document level.
\end{enumerate*}
%
For \ref{ch_injecting:it:kb-into-ie}, integrating KB into IE has been applied for individual tasks: relation classification \cite{poerner2020bert, zhang2019long,yang2017leveraging}, entity typing \cite{peters2019knowledge} and NER \cite{yamada2020luke}.
For \ref{ch_injecting:it:e2e-ie}, recently span-based architectures \cite{lee2017end,luan2019general,wadden2019entity,fei2020boundaries} have been proposed.
Our work unifies the KB integration concept into such span-based IE system, in particular an entity-centric one (as per \ref{ch_injecting:it:entity-centric-ie}), building on \cite{jia2019document, zaporojets2021dwie}.
For the KB integration approach, we exploit entity representations trained on a hypertext corpus, as in \cite{yamada2016joint, 
ganea2017deep, yamada2020luke} or learnt from a knowledge graph \cite{yang2017leveraging,han2018neural,zhang2019long}.
Our results show that both offer complementary value for IE. 
\revklim{Similarly to our work, \cite{yamada2019neural} also explore using an attention-weighted combination of entity representations, but they use it to build a full document representation (with mentions having the entities as candidates) for a text classification task. In contrast, our span-based attention model
is able to ``inject'' knowledge in each of the mentions separately, for more fine-grained downstream IE tasks that are mention-dependent, \eg coreference resolution, relation extraction and NER. }

\section{Conclusion}
\label{ch_injecting:sec:conclusion}
We propose an end-to-end model for joint IE (NER + relation extraction + coreference resolution) incorporating entity representations from a background knowledge base (KB), using a span-based system.
We find that representations built from a knowledge graph and a hypertext corpus are complementary in boosting IE performance.
To combine candidate entity representations for text spans, we explore various weighting schemes: an attention-based combination is successful in combining prior frequency information from a hypertext corpus with contextual information to identify the relevant entity, and achieves highest IE performance. 

\oldignore{
\textbf{-----------OLD TEXT-----------}\\
In this paper, we propose to use knowledge graph and hyperlinked corpus-based entity representations to inject background knowledge in a 
joint 
end-to-end IE framework. 
We find that both of the information sources are complementary and give superior results when used together. 
Additionally, we explore the prior and attention weighting schemes to combine the candidate entity representations for each of the spans. 
Our experiments show the advantage of using attention-based weights, specially in the corner cases where the entity mention is used to refer to less frequently co-occurring entities.
}

\section*{Acknowledgments}
\revklim{\noindent Part of the research leading to these results has received funding from
\begin{enumerate*}[(i)]
\item the European Union's Horizon 2020 research and innovation programme under grant agreement no.\ 761488 for the CPN project,\footnote{\url{https://www.projectcpn.eu/}} and
\item the Flemish Government under the programme ``Onderzoeksprogramma Artifici\"{e}le Intelligentie (AI) Vlaanderen''.
\end{enumerate*}
}

\renewcommand{\labelenumi}{\arabic{enumi}.} 

\clearpage




\clearpage{\pagestyle{empty}\cleardoublepage}

\graphicspath{{klim_ch_temporal_el/figures/}}
\widowpenalty100000
\clubpenalty100000

\definecolor{deepcarmine}{rgb}{0.93, 0.47, 0.13}
\colorlet{negative0001}{deepcarmine!75}
\colorlet{negative001}{deepcarmine!50}
\colorlet{negative01}{deepcarmine!25}

\definecolor{darkspringgreen}{rgb}{0.0, 0.5, 1.0}
\colorlet{positive0001}{darkspringgreen!75}
\colorlet{positive001}{darkspringgreen!50}
\colorlet{positive01}{darkspringgreen!25}

\newcommand{\underlinepers}[1]{#1}

\newcommand{\revklimb}[1]{\textcolor{RoyalBlue}{#1}}
\newcommand{\revchris}[1]{#1}
\newcommand{\hideme}[1]{}
\newcommand{\lucie}[1]{#1}


\hyphenation{}


%
\chapter[Temporal Entity Linking]{{\ourdataset}: Linking Dynamically Evolving and Newly Emerging Entities}
\label{chap:temporal_el}

\renewcommand\evenpagerightmark{{\scshape\small Chapter \arabic{chapter}}}
\renewcommand\oddpageleftmark{{\scshape\small Temporal Entity Linking}}

\renewcommand{\bibname}{References}

\begin{flushright}
\end{flushright}

\noindent\emph{In this chapter, we go one step further and analyze the evolution of the entities from temporal perspective. In order to achieve this, we create a new dataset which consists of 10 yearly snapshots of Wikipedia entities from 2013 until 2022. We further study how \textit{entity linking} task is affected by \begin{enumerate*}[(i)]
    \item changes of existing entities in time, and 
    \item creation of new emerging entities.
\end{enumerate*}. Furthermore, we do not restrict our analysis to the realm of \textit{named entities}, but incorporate all existing entities and concepts defined in Wikipedia. Our analysis showcases a continual decrease of performance in time, indicating that the entities from later versions of Wikipedia are harder to disambiguate than entities from earlier versions. Additionally, we demonstrate that the decrease in performance is specially sharp for entities requiring additional new knowledge (\eg new entities related to COVID-19 pandemic) for which the model was not pre-trained.}
\begin{center}
\par{$\star\star\star$}
\end{center}
\vspace{0.15in}

\par{\noindent\large{\textbf{K.~Zaporojets, L.A.~Kaffee, J.~Deleu, T.~Demeester, C.~Develder and I. Augenstein}}}
\vspace{0.1in}
\par{\noindent\textbf{Thirty-sixth Conference on Neural Information Processing Systems
Datasets and Benchmarks Track: NeurIPS, 2022.}}
\vspace{0.15in}

\par{\noindent\bf{Abstract}}
In our continuously evolving world, entities change over time and new, previously non-existing or unknown, entities appear.
    We study how this evolutionary scenario impacts the performance 
    on
    a well established \textit{entity linking} (EL) task. 
    For that study, we
    introduce {\ourdataset}, an entity linking dataset that consists of time-stratified English Wikipedia snapshots from 2013 to 2022, 
    from which we collect both \emph{anchor mentions} of entities, and these \emph{target entities}' descriptions.
    By capturing such temporal aspects, 
    our newly introduced {\ourdataset} resource
    contrasts with 
    currently 
    existing
    entity linking 
    datasets, which are 
    composed 
    of
    fixed mentions linked to a
    single static version of a target Knowledge Base (\eg Wikipedia 2010 for CoNLL-AIDA). 
    Indeed, for each of 
    our collected temporal snapshots, {\ourdataset} contains 
    links 
    to 
    entities that are \emph{continual}, \ie occur in 
    all of 
    the years,
    as well as 
    completely \emph{new} 
    entities that appear for the first time at some point.
    Thus, we enable to quantify the performance of 
    current state-of-the-art EL models for:
    \begin{enumerate*}[(i)]
        \item entities that are subject to changes over time in  their Knowledge Base 
        descriptions
        as well as their mentions' contexts, and
        \item newly created entities that were previously non-existing (\eg at the time the EL model was trained).
    \end{enumerate*}
    Our experimental results show that in terms of temporal performance degradation, 
    \begin{enumerate*}[(i)]
        \item \emph{continual} entities suffer a decrease of up to \revklim{3.1\% }
        EL accuracy, 
        while
        \item for \emph{new} entities this accuracy drop is up to \revklim{17.9\%}.
    \end{enumerate*}
    This highlights the challenge of the introduced {\ourdataset} dataset and opens new research prospects in the area of time-evolving entity disambiguation.\footnote{\revklim{\ourdataset~dataset, code and models are made public at \url{https://github.com/klimzaporojets/TempEL}.}}

\section{Introduction}
\begin{figure}[!t]
\centering
\includegraphics[width=0.7\columnwidth]{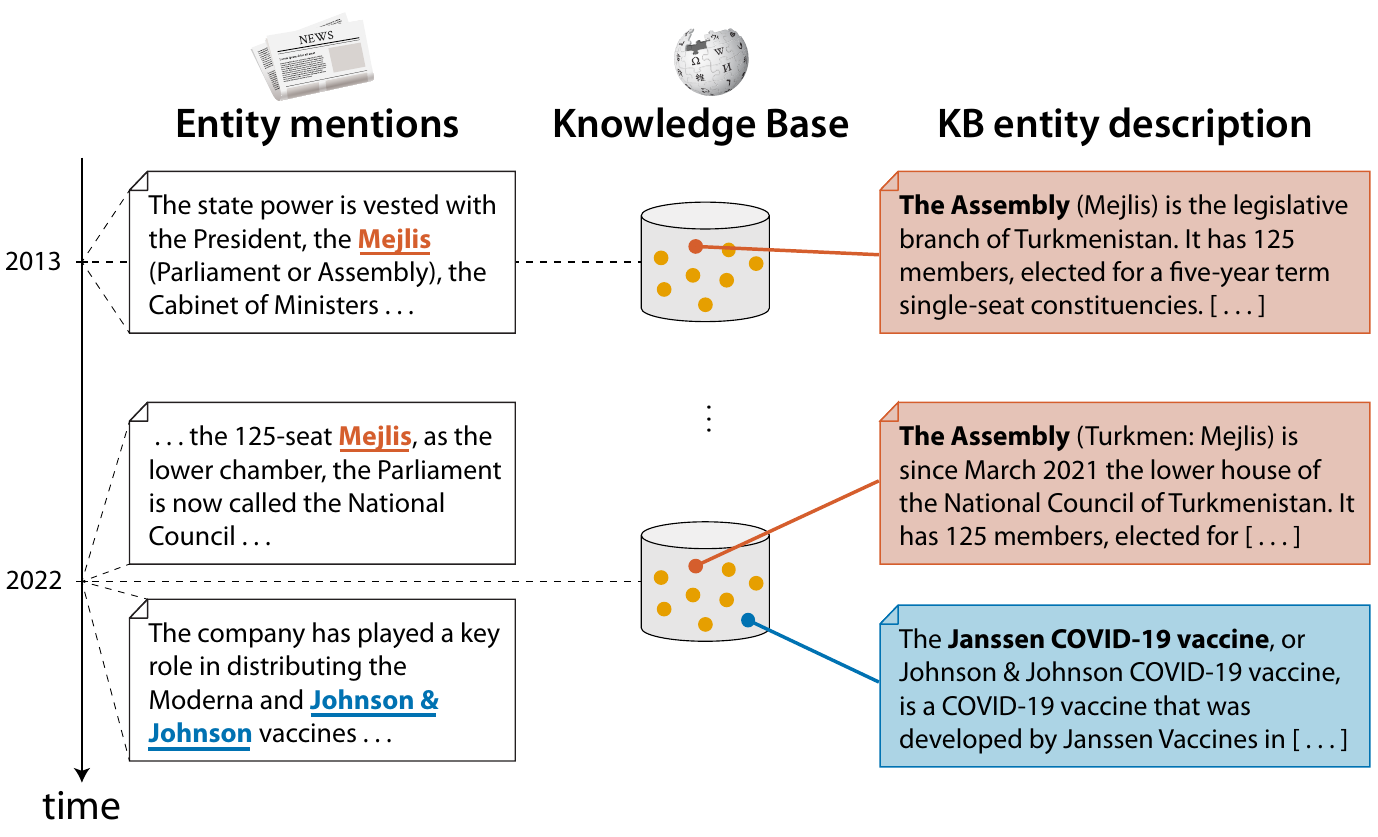}
\caption[Illustration of KB entities changing over time]{Illustration of KB entities changing over time: the ``Mejlis'' entity changes over time (both in its KB description and the contexts in which it is referenced to), while the Johnson \& Johnson vaccine is an entirely new one that did not exist before.}
\label{fig:el_introduction}
\end{figure}

Entity linking (EL) is a well-established task that is concerned with mapping anchor \textit{mentions} in text to target \textit{entities} that 
describe
them in a Knowledge Base (KB) (\eg Wikipedia).\footnote{Some of the related work 
\cite{ganea2017deep, kolitsas2018end, sevgili2020neural, zhang2021entqa, zaporojets2021consistent} 
distinguishes between \textit{entity disambiguation} and \textit{entity linking} tasks. This latter including \textit{mention detection} and \textit{disambiguation} in an end-to-end setting. In the current work, we follow a more conservative 
naming convention 
\cite{rao2013entity, wu2019zero, logeswaran2019zero, onoe2020fine, raiman2022deeptype}, and use the term \textit{entity linking} and \textit{entity disambiguation} interchangeably.}
Existing benchmark datasets for EL \cite{usbeck2015gerbil,roder2018gerbil,sevgili2020neural,petroni2020kilt} are composed of a fixed set of annotated mentions linked to a single version of a target KB.
This static setup is oblivious to the inherently non-stationary nature of the entity linking task where both target entities as well as anchor mentions change over time. The example in \figref{fig:el_introduction} illustrates this time-evolving essence of entity linking with a simple evolutionary comparison between Wikipedia 2013 and 2022. It showcases two scenarios studied in the current paper: \begin{enumerate*}[(i)]
    \item temporal evolution of existing (\textit{continual}) entities across temporal snapshots, and
    \item appearance of \emph{new}, previously non-existent entities
\end{enumerate*}. 
For instance,
the description of the \emph{continual} entity \textit{The Assembly} differs between Wikipedia 2013 and 2022. Furthermore, the context of a mention ``Mejlis'' referring to \textit{The Assembly} also changes over time. Conversely, the \emph{new} entity \textit{Janssen COVID-19 vaccine} is newly introduced in 2021 with the corresponding mentions (\eg ``Johnson \& Johnson'' in \figref{fig:el_introduction}) that are linked to it. 

In this paper 
we introduce \ourdataset, a novel dataset to study this time-evolving aspect of the entity linking task.
We therefore extract 10 equally spread 
yearly snapshots  
from English Wikipedia 
entities starting from 
January 1, 2013 until January 1, 2022.
We use each of these temporal snapshots of Wikipedia to also extract anchor mentions with the surrounding text. Thus, {\ourdataset} captures the temporal evolution not only in the target entities as they are defined in 
the Wikipedia KB,
but also in the contexts of anchor mentions linked to these entities. 
Each of the 10 temporal snapshots of our dataset 
is composed of training, test and validation sets with equal numbers of mentions and entities 
across the 
snapshots. 
Furthermore, 
\ourdataset~is designed to comprise 
mentions pointing to \emph{continual} entities across all 
the temporal snapshots, and to \emph{new} entities inside a given temporal 
snapshot. 

Finally, as a baseline, we finetune and evaluate the bi-encoder component of the BLINK model \cite{wu2019zero} on the various temporal snapshots of our newly introduced {\ourdataset} dataset. 
The bi-encoder is widely used in state-of-the-art entity linking models \cite{zhang2021entqa, wu2019zero} to retrieve the top $K$ (in this work we experiment with $K = 64$)
candidate target entities for a given anchor mention context. Furthermore, its straightforward finetuning and fast retrieval performance on millions of candidate entities \cite{johnson2019billion}, make it an ideal choice to test on \ourdataset.
Our experiments demonstrate
a 
consistent
temporal model deterioration 
for mentions linked to both
\emph{continual} 
(\revklim{3.1\% }
accuracy@64
points)
as well as \emph{new} 
(\revklim{17.9\%} accuracy@64 points)
entities. A more detailed analysis
reveals
that the maximum drop in performance is observed for \emph{new} entities that require 
fundamentally different world knowledge that was not present in the corpus originally used to pre-train BERT. 
This is e.g. the case for \emph{new} entities related to COVID-19 for which the bi-encoder model suffers additional deterioration of 
\revklim{14\%}
accuracy@64
points compared to the rest of the new entities. 

\section{Related work}
\label{sec:related}

Our work is related to multiple different, yet interconnected research areas described below. First, we explain how \ourdataset~compares to the currently widely used \textit{entity linking datasets}. Next, we relate our work to already existing \textit{temporal datasets} covering different aspects of the temporal evolution of the data. Finally, we describe the existing \textit{entity-centric} research efforts, comparing the \ourdataset~entity linking dataset to other datasets that heavily depend on the use of entities.
\paragraph{Entity linking datasets} Most current state-of-the-art EL models \cite{yamada2020global, orr2020bootleg, de2020autoregressive, zhang2021entqa, de2021highly} report on 
datasets from predominantly the news domain such as AIDA \cite{hoffart2011robust}, 
KORE50~\cite{hoffart2011robust}, AQUAINT \cite{milne2008learning}, ACE 2004, MSNBC \cite{ratinov2011local}, N$^3$ \cite{roder2014n3}, DWIE\cite{zaporojets2021dwie}, VoxEL\cite{rosales2018voxel}, and TAC-KBP 2010-2015 \cite{ji2010overview,ji2015overview}. Other frequently used datasets include the web-based IITB \cite{kulkarni2009collective} and OKE 15/16 \cite{nuzzolese2015open}, as well as the tweet-based Derczynski \cite{derczynski2015analysis}. Additionally, larger yet automatically annotated datasets such as WNED-WIKI and WNED-CWEB \cite{guo2018robust} 
have been also widely adopted. 
Finally, 
a number of
resources such as the domain-specific biomedical MedMentions \cite{mohan2018medmentions}, the zero-shot ZeShEL \cite{logeswaran2019zero}, and the multi tasking DWIE \cite{zaporojets2021dwie} and \ouraida \cite{zaporojets2021consistent} datasets have been recently introduced. 
Many of the mentioned datasets are further covered by entity linking evaluation frameworks such as GERBIL \cite{usbeck2015gerbil,roder2018gerbil} and KILT \cite{petroni2020kilt} that provide a common interface to evaluate the models. Yet, the mentioned resources are limited to static mention annotations linked to entities from a single version of a Knowledge Base. 
This contrasts with our newly introduced {\ourdataset} dataset, where the anchor mentions as well as the target entity descriptions are taken from different time periods. The datasets most closely related to our work are the recently introduced WikilinksNED \cite{eshel2017named, onoe2020fine} and ShadowLink \cite{provatorova2021robustness}. WikilinksNED contains only unseen mention-entity pairs in its test subset, thus encouraging the design of models invariant to overfitting and memorization biases. Furthermore, ShadowLink contains \textit{overshadowed entities}: entities referred to by ambiguous mentions whose most likely target entity is different, \eg the anchor mention ``Michael Jordan'' linked to the scientist instead of to the more widely referred to target entity describing the former basketball player. We incorporate the challenges presented in both of these datasets in \ourdataset~(see \secref{sec:dataset_construction} for further details).
\paragraph{Temporal datasets} Research on temporal drift in data has gained a lot of interest in  recent  years. The focus has mostly been on creating datasets to train language models on different temporal snapshots of corpora derived from scientific~\cite{lazaridou2021mind}, newswire~\cite{lazaridou2021mind,dhingra2022time}, Wikipedia~\cite{jang2022temporalwiki}, and Twitter~\cite{loureiro2022timelms} domains. 
More recently, temporal datasets have appeared to address 
tasks 
such as sentiment analysis \cite{lukes2018sentiment, ni2019justifying, agarwal2022temporal}, text classification \cite{huang2018examining,he2018time}, named entity recognition \cite{derczynski2016broad,rijhwani2020temporally}, question answering \cite{lazaridou2021mind}, and entity typing \cite{luu2021time}, among others. 
However, the creation of datasets tackling the temporal aspect of entity linking has largely been left unexplored. 
To the best of our knowledge, the dataset most closely related to {\ourdataset} is diaNED, introduced by \cite{agarwal2018dianed}. There, the authors annotate mentions 
that require additional temporal information from the context to be correctly disambiguated. Conversely, in {\ourdataset} both mentions and entities are extracted from evolving temporal snapshots.
 
\paragraph{Entity-driven datasets} 
Recent research has demonstrated the benefits of incorporating entity knowledge in various downstream tasks \cite{yang2017leveraging, peters2019knowledge,yamada2020luke,guu2020realm,verga2021adaptable,yasunaga2021qa,liu2022knowledge}.
This progress has been accompanied by the creation of entity-driven datasets for tasks such as language modeling \cite{petroni2019language,agarwal2021knowledge,kassner2021multilingual}, question answering \cite{yih2015semantic, joshi2017triviaqa, jiang2019freebaseqa,lewis2021paq,saxena2021question}, fact checking \cite{thorne2018fever, onoe2021creak, aly2021feverous} and information extraction  
\cite{yao2019docred,zaporojets2021dwie}, 
to name a few. 
Yet, recent findings \cite{runge2020exploring,fevry2020entities,lewis2020retrieval,verlinden2021injecting,heinzerling2021language,ri2021mluke}
suggest that entity \textit{representation} and \textit{identification} 
(\ie identifying the correct entity that match a given text) 
are among the main challenges 
that should be solved to further increase performance on such datasets.
We believe that {\ourdataset} can contribute to addressing these challenges by:
\begin{enumerate*}[(i)]
\item encouraging research on devising more robust methods to creating \textit{entity representations} that are invariant to temporal changes; and 
\item improving entity identification for non-trivial scenarios involving ambiguous and uncommon mentions (\eg linked to \textit{overshadowed entities} as defined above).
\end{enumerate*} 
\section{The {\ourdataset} dataset}
In this section we will provide details on how {\ourdataset} was constructed (\secref{sec:dataset_construction}), describing the main components of the creation pipeline as sketched in \figref{fig:el_pipeline}.
Furthermore, we discuss the aspects taken into account to guarantee the overall quality of our dataset (\secref{sec:quality}). Finally, we present statistics of {\ourdataset} (\secref{sec:dataset_statistics}), 
illustrating its dynamically evolving nature. 
\subsection{Dataset construction}
\label{sec:dataset_construction}

\begin{figure}[!t]
\centering
\includegraphics[width=1.0\columnwidth]{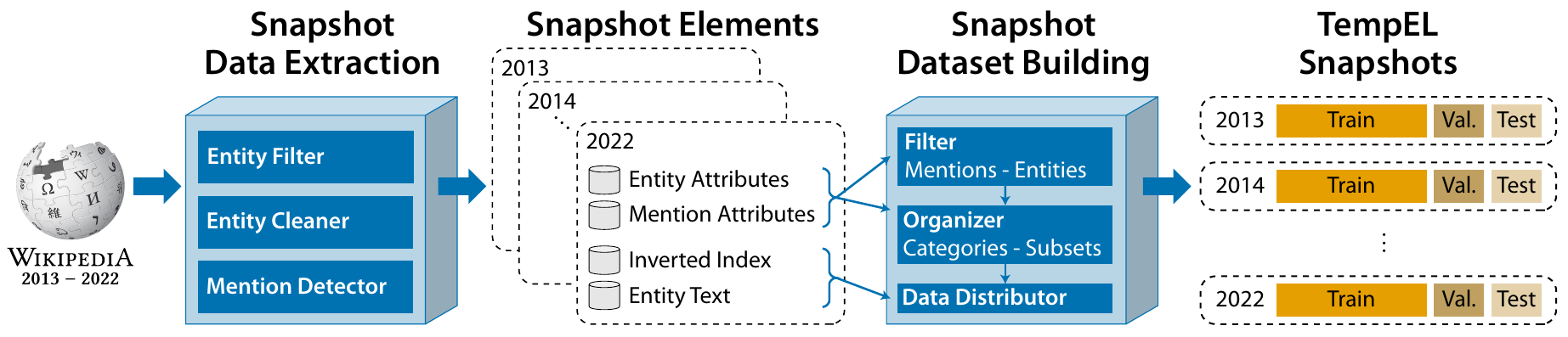}
\caption[The pipeline to create {\ourdataset} dataset]{The pipeline to create our {\ourdataset} dataset. All the components are explained in \secref{sec:dataset_construction}.}
\label{fig:el_pipeline}
\end{figure}

\paragraph{Snapshot data extraction}
As \figref{fig:el_pipeline} indicates, we start from the history log dumps from February 1, 2022 of Wikipedia itself.
We first filter these (see \textit{Entity Filter} in \figref{fig:el_pipeline})
to:
\begin{enumerate*}[(i)]
    \item 
    exclude
    pages that are irrelevant for {\ourdataset} (\ie categories, disambiguation pages, redirects and lists); and
    \item select 
    the most temporally stable 
    version of 
    a Wikipedia page from the last month of the year in order to avoid introducing 
    more volatile and 
    potentially corrupted content edits (see \secref{sec:quality} for further details).
\end{enumerate*}
Next, 
the Wikipedia pages are cleaned (see \emph{Entity Cleaner} in \figref{fig:el_pipeline}) by stripping from the Wikitext markup content.\footnote{\url{https://en.wikipedia.org/wiki/Help:Wikitext}} We use both regular expressions as well as the MediaWiki API for more difficult cases, such as the 
parsing
of some of the Wikitext templates.
Finally, we detect the mentions (see \textit{Mention Detector} in \figref{fig:el_pipeline}) in each of the Wikipedia entity pages, filtering out the ones that point to anchors (\ie subsections in Wikipedia pages), pages in languages other than English,
files, red links (\ie links pointing to not yet existing Wikipedia pages) and redirects. 

The output of the \textit{Snapshot Data Extraction} step first of all includes a set of \emph{Entity} and \emph{Mention Attributes} (\eg the last modification date of the target entity), which are detailed in the supplementary material \revklim{(see \secref{app:mention_entity_attributes})}.
These attributes form part of the final dataset, making it possible to perform additional analyses of the results. 
Furthermore, the \emph{Inverted Index} is generated to quickly access the Wikipedia pages that include a mention for a given target entity.
Finally, the \emph{Entity Text} files are extracted containing the (potentially yearly varying) textual content from the Wikipedia entity definition, as well as anchor mentions therein. These mentions of Wikipedia anchors that link to an entity will be extracted in the \textit{Snapshot Dataset Building} step described further.

\paragraph{Snapshot dataset building} 
Starting from the \emph{Snapshot Elements} produced by the \emph{Snapshot Data Extraction} process described above, the actual 
\ourdataset~dataset is now generated.
The first step is to apply an additional \textit{Filter} to both entities and mentions with the goal of creating a more challenging dataset. 
This is done by
excluding mentions
for which the correct entity it refers to has the highest prior~\cite{yamada2016joint}.
More formally, the \textit{mention prior} is calculated as follows,  
    \begin{equation}
    P(e\vert m) = \vert A_{e,m} \vert / \vert A_{*,m} \vert \label{eq:mention_prior},   
    \end{equation}
where $A_{*,m}$ is the set of all anchors that have the same mention $m$, and $A_{e,m}$ is the subset thereof that links to entity $e$.
Additionally, 
we exclude the mentions whose normalized edit distance from the target entity title is below an established threshold.\footnote{During the generation of \ourdataset, we use a 
threshold of 0.2.} 
By ignoring the mentions with the highest prior and exact match with the title, we ensure that {\ourdataset} contains non-trivial disambiguation cases where the naive approaches (\eg defaulting to the most frequently linked entity for a given mention) would fail \cite{guo2018robust, logeswaran2019zero, wu2019zero, provatorova2021robustness}. 

Furthermore, the entities are organized (see \emph{Organizer} in \figref{fig:el_pipeline}) into two \emph{categories}: \begin{enumerate*}[(i)]
    \item \emph{new}, emerging and previously non-existent entities that are introduced in a particular snapshot; and
    \item \emph{continual} entities across all the temporal snapshots. 
\end{enumerate*} 
Next, the mentions are divided in separate subsets (\ie train, validation and test), with the constraint of normalized edit distance between the mentions in different subsets referring to the same target entity be higher than 0.2.
This way, we expect to discourage potential models from memorizing the mapping between mentions and entities \cite{onoe2020fine}. 

Finally, the data is distributed equally (see \emph{Data Distributor} in \figref{fig:el_pipeline}) across all of the temporal snapshots.
This way, the difference in performance can only be attributed to temporal evolution and not to 
inconsistencies
related to dataset variability (\eg different number of training instances in each of the temporal snapshots). Concretely, we enforce that the number of \emph{continual} and \emph{new} entities as well as the number of mentions stays the same across the temporal snapshots (see \tabref{tab:dataset_detail2}). 
We achieve this by performing a random mention subsampling in 
snapshots
with higher number of mentions, weighted by the difference in the number of mentions-per-entity. 
This produces
a very similar
mention-entity distribution 
across the temporal snapshots.
Finally, the filtered anchor mentions are located in the cleaned Wikipedia pages (\ie the \textit{Entity Text} in \figref{fig:el_pipeline}) using the \textit{Inverted Index} created in the previous \emph{Snapshot Data Extraction} step. 
The context of each of the mentions is further paired with the respective content of target pages, outputting this way the final {\ourdataset} dataset.
\begin{table}[t]
\caption[Summary statistics of \ourdataset]{Summary statistics of \ourdataset. The number of entities and mentions is the same across all of the temporal snapshots.}
\label{tab:dataset_detail2}
\vspace{.75\baselineskip}
\centering
\begin{tabular}{l ccc}
        \toprule
         Statistic & \multicolumn{1}{c}{Train} & \multicolumn{1}{c}{Validation} & \multicolumn{1}{c}{Test} \\ 
         \midrule 
         Temporal Snapshots & 10 & 10 & 10 \\ 
         Continual Entities & 10,000 & 10,000 & 10,000 \\
         \ \ \ \# Anchor Mentions & 136,227 & 42,096 & 46,765 \\ 
        New Entities & 373 & 373 & 373 \\
         \ \ \ \# Anchor Mentions & 1,764 & 1,231 & 1,450 \\
        \bottomrule 
        \end{tabular}
\end{table}
\subsection{Quality control} 
\label{sec:quality}
\paragraph{Corrupted content} Wikipedia is an open resource that relies on efforts of millions of Wikipedians 
to update and extend its contents.\footnote{\url{https://en.wikipedia.org/wiki/Wikipedia:Wikipedians}} 
As such, 
that content is not always reliable, with errors due to human mistakes or intentional vandalism. 
Despite 
efforts to prevent the introduction of such erroneous 
edits
\cite{west2010detecting,dang2016quality,wang2020assessing}, 
we have detected numerous cases
of corrupted 
entity descriptions
during our preliminary tests.
As a result, we adopted a simple, yet very effective heuristic: for each of the entities of a particular yearly snapshot, we select the most \textit{stable} (\ie the version of the entity that lasted the longest before being changed) content 
of the last month of the year (December). Due to the fact that most of the corrupted content is rolled back very quickly, and even automatically by specialized bots \cite{zheng2019roles,jiang2020good}, this heuristic is very robust. We double checked the correctness of the extracted content by manually inspecting the evolution of hundred entities with lowest Jaccard vocabulary similarity between temporal snapshots and observed no obviously erroneous entries.
\paragraph{Entity relevance} 
We filter out entities that have less than 10 in-links (\ie number of mentions linking to the entity) or contain less than 10 tokens in its Wikipedia page in order to avoid including noisy content \cite{eshel2017named}. Additionally, in order to avoid evaluation bias towards mentions pointing to more popular entities 
\cite{orr2020bootleg,chen2021evaluating}, we 
limit
the number of mentions per entity to 10 for our test and validation sets. 
This way, 
we expect the accuracy scores to not be
dominated by links to 
popular target entities (\ie entities with a big number of incoming links). 

\paragraph{Content filtering} 
We only consider mentions linked to the main Wikipedia articles describing entities. The mentions pointing to anchors (subsections in a Wikipedia document), images, files, and wiki pages in other languages are filtered out in \textit{Snapshot Data Extraction} step (see \figref{fig:el_pipeline}). In this step we also ignore pages that are not Wikipedia articles (\eg files, information on Wikipedia users, etc.) as well as redirect pages. This way, the target entities as well as anchor mentions in our dataset are obtained from a cleaned list of candidate pages referring to entities that contain a meaningful textual description in Wikipedia. 

\paragraph{Dataset distribution} During the construction of \ourdataset, we constrain the subsets to be of equal size and contain 
similar
mention-per-entity distributions across all the temporal 
snapshots.
This is implemented in \textit{Data Distributor} sub-component of the dataset creation pipeline (see \secref{sec:dataset_construction}). For example, the number of mentions linked to continual entities in our training subset is 
\revklim{136,227}
across all of the 
snapshots
(see \tabref{tab:dataset_detail2} for further details). 
We argue that this setting will produce  uniform, structurally unbiased 
snapshots.
This will allow to study exclusively the temporal effect on the performance of the models for each of the different time periods.  
Our reasoning is supported by previous work demonstrating that the size alone of the training set \cite{loureiro2022timelms} as well as a different distribution of the number of mentions per entity \cite{orr2020bootleg} can significantly affect the performance of the final model. Furthermore, we do not constrain the total number of entities from the Wikipedia KB to be equal 
across the temporal snapshots
(see \figref{fig:nr_wiki_entities}), since we consider it a part of the evolutionary nature of the entity linking task (\ie the temporal evolution of the target KB) we intend to study. 
\paragraph{Flexibility and extensibility} Finally, we provide a framework that can be used to re-generate the dataset with different parameters as well as to extend it with newer temporal snapshots. 
This includes the option to generate a new dataset with a 
customized
number of temporal snapshots (\eg quarterly instead of yearly spaced), different mention attributes (\eg filtering by mention prior values), entity popularity (\eg filtering out entities that have more than a certain number of in-links), among others 
(see \revklim{\secref{app:hyperparameters}} of the supplementary material for a complete list). 
\begin{figure}[t]
\begin{subfigure}{.31\textwidth}
  \centering
  \includegraphics[width=1.0\linewidth, trim={0.0cm 0.0cm 0.0cm 1.00cm},clip]{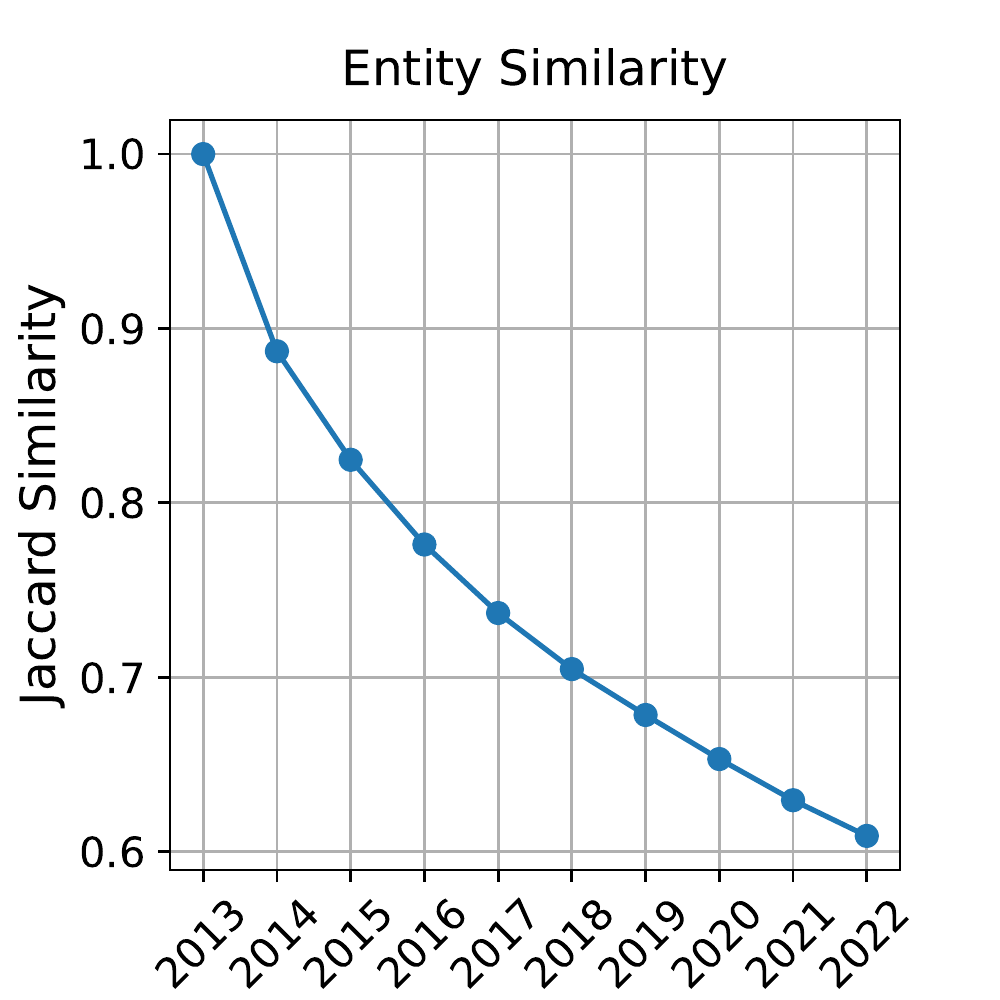}  
  \caption{Evolution of entities in terms of Jaccard vocabulary similarity.}
  \label{fig:evolution_entities_jaccard}
\end{subfigure} \hfill
\begin{subfigure}{.31\textwidth}
  \centering
  \includegraphics[width=1.0\linewidth, trim={0.0cm 0.0cm 0.0cm 1.00cm},clip]{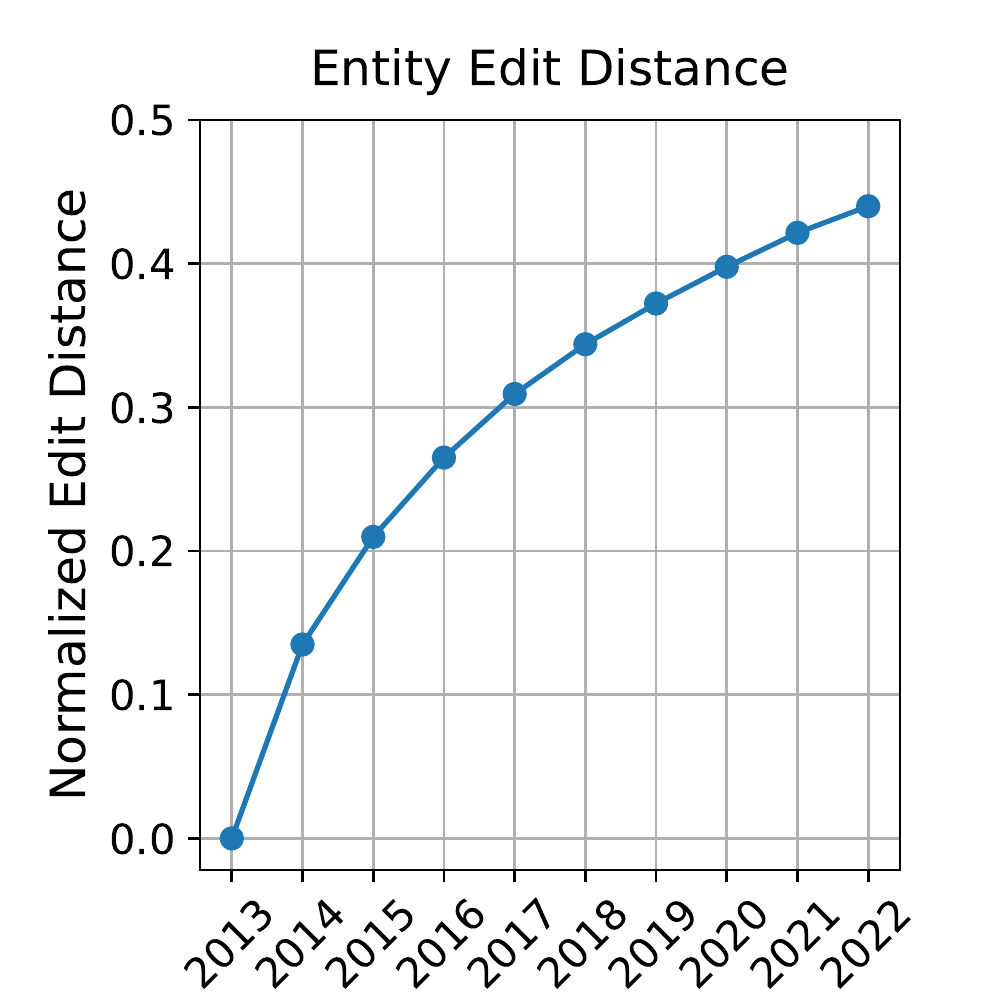}  
  \caption{Evolution of entities in terms of edit distance of the content.}
  \label{fig:evolution_entities_edistance}
\end{subfigure} \hfill
\begin{subfigure}{.31\textwidth}
  \centering
  \includegraphics[width=1.0\linewidth, trim={0.0cm 0.0cm 0.0cm 1.00cm},clip]{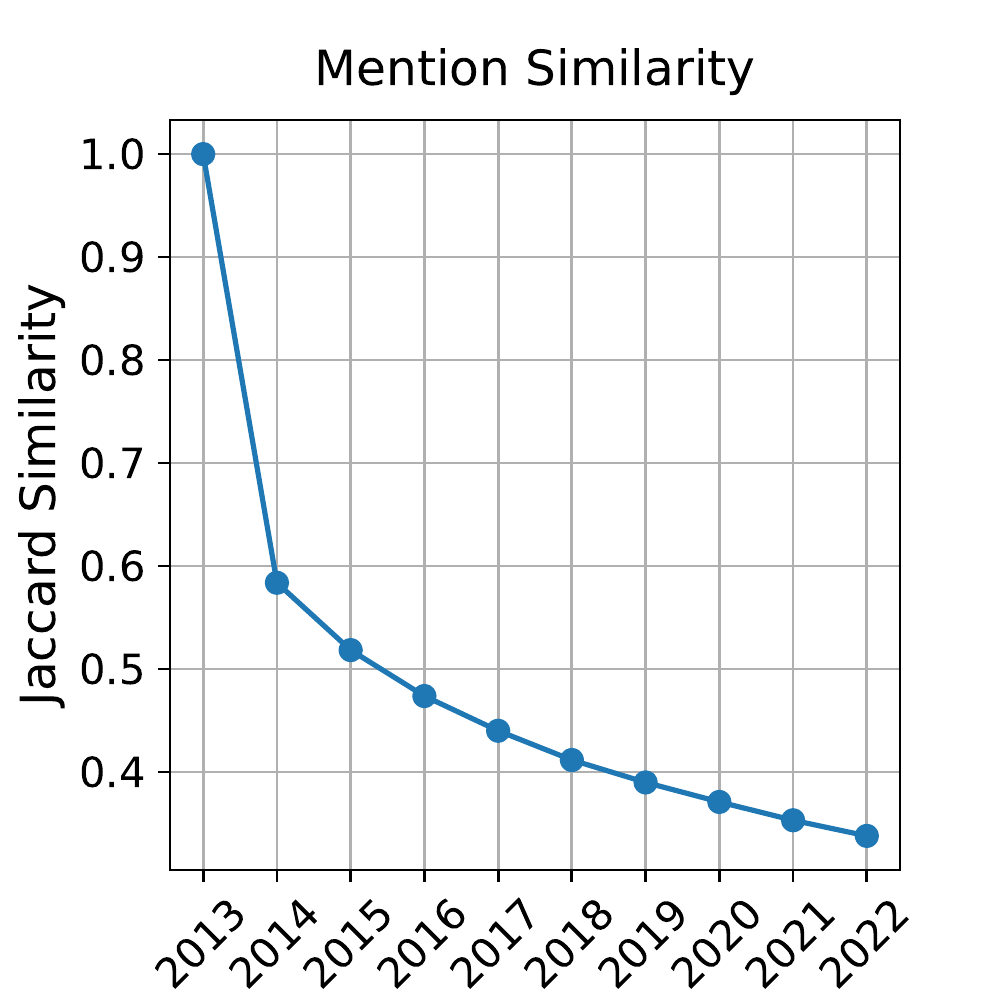}  
  \caption{Evolution of context around the mentions (Jaccard similarity).}
  \label{fig:evolution_mentions}
\end{subfigure}
\caption[Temporal change of textual content]{Change of textual content of entities and context around mentions across temporal yearly snapshots (x-axis). }
\label{fig:fig}
\end{figure}
\subsection{Dataset statistics}
\label{sec:dataset_statistics}
\tabref{tab:dataset_detail2} summarizes the dataset statistics. 
We divide each of the temporal snapshots into train, 
validation and test subsets containing an equal number of \emph{continual} and \emph{new} entities. The number of mentions differs between the subsets since we
limit the number of mentions per entity to 10 in both validation and test sets
(see \textit{entity relevance} in \secref{sec:quality} for 
further
details). 

Additionally, we collect statistics related to temporal drift in content for both the target entities (Figs.~\ref{fig:evolution_entities_jaccard} and \ref{fig:evolution_entities_edistance}) as well as the context around the anchor mentions (\figref{fig:evolution_mentions}). Concretely, \figref{fig:evolution_entities_jaccard} visualizes Jaccard vocabulary similarity between the textual description of \emph{continual} entities in 2013 and that of posterior yearly snapshots
in \ourdataset. We observe a continual decrease, indicating that on average, the content of the entity description in Wikipedia is constantly evolving in terms of the used vocabulary. This is also supported by the graph in \figref{fig:evolution_entities_edistance}, which 
showcases a continuous
temporal increase of the average value of normalized edit distance across \emph{continual} entities. Finally, \figref{fig:evolution_mentions}  
illustrates
the temporal drift in the vocabulary (\ie Jaccard vocabulary similarity) of the context around the mentions pointing to 
the same entity.
We find it experiences a more significant change compared to the Jaccard similarity of entity content illustrated in \figref{fig:evolution_entities_jaccard}. This suggests that the context around the anchor mentions 
is subject to
a higher degree of temporal 
transformation
compared to that of target entities, making it an interesting 
item of future work. 
\section{Experiments}
Our final {\ourdataset} comprises 10 different yearly snapshots and we evaluate entity linking (EL) performance on each of them individually. This evaluation setup allows us to 
study the effect of temporal corpus changes and assess the impact of increasing time lapses between the data used for model training and that on which the EL model is deployed
\cite{he2018time, agarwal2022temporal, luu2021time}. 
We 
train 
a bi-encoder baseline EL model (detailed in \secref{sec:baseline}) 
on the temporal snapshots from 2014 to 2022 separately and
then evaluate EL performance using the test sets of both past and future snapshots.

More specifically, our experiments
aim
to answer the following research questions:
\begin{enumerate*}[label=\textbf{(Q\arabic*)}]
    \item \label{it:q1-temp-degradation}
    Does a fixed entity linking (EL) model's performance degrade when applied to newer content?
    \item \label{it:q2-finetuning} 
    How does finetuning an EL model on more recent training data affect its performance on both old and newer content?
    \item \label{it:q3-emerging} 
    How does EL performance differ for resolving \emph{new} versus \emph{continual} entities?
\end{enumerate*}

\subsection{Baseline}
\label{sec:baseline}
\begin{figure}[t]
\begin{subfigure}{.32\textwidth}
  \centering
  \includegraphics[width=1.0\linewidth, trim={0.0cm 0.0cm 0.0cm 1.00cm},clip]{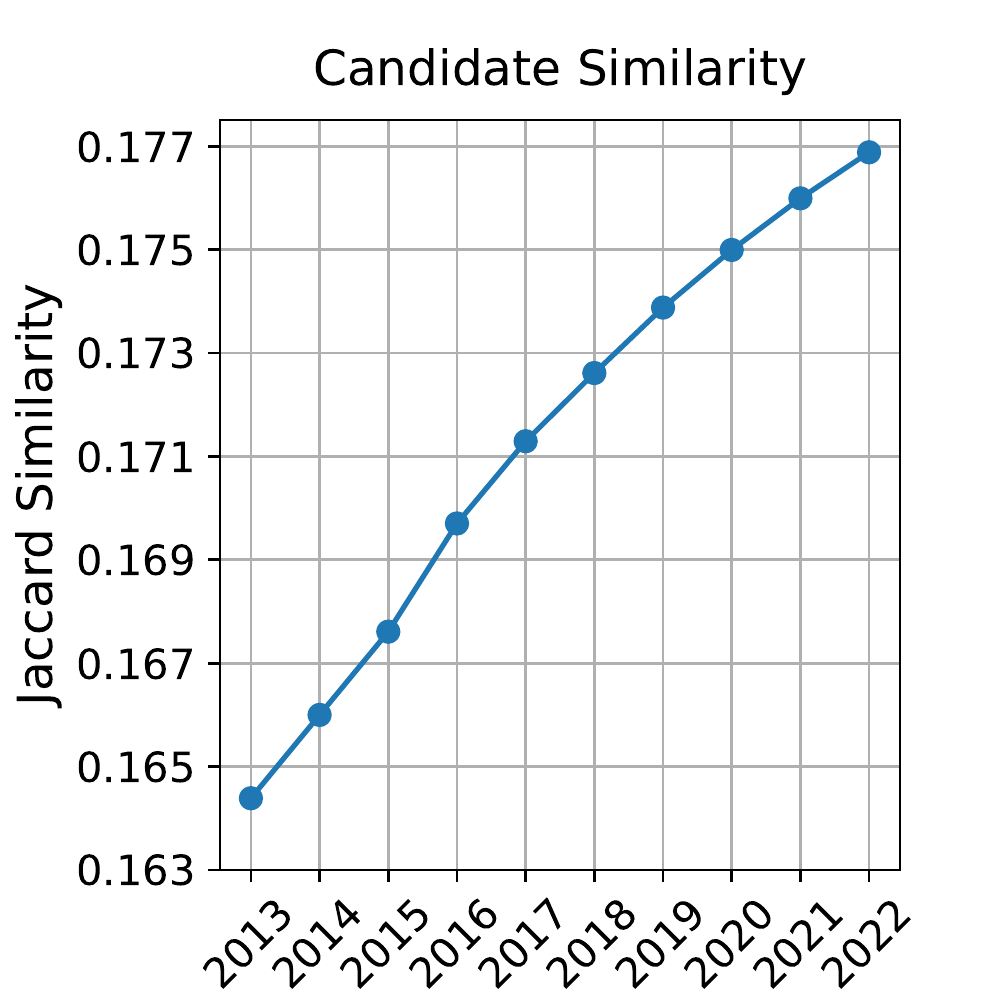}  
  \caption{Similarity between candidates returned by the bi-encoder baseline.}
  \label{fig:candidate_similarity}
\end{subfigure}\hfill
\begin{subfigure}{.32\textwidth}
  \centering
  \includegraphics[width=1.0\linewidth, trim={0.0cm 0.0cm 0.0cm 1.00cm},clip]{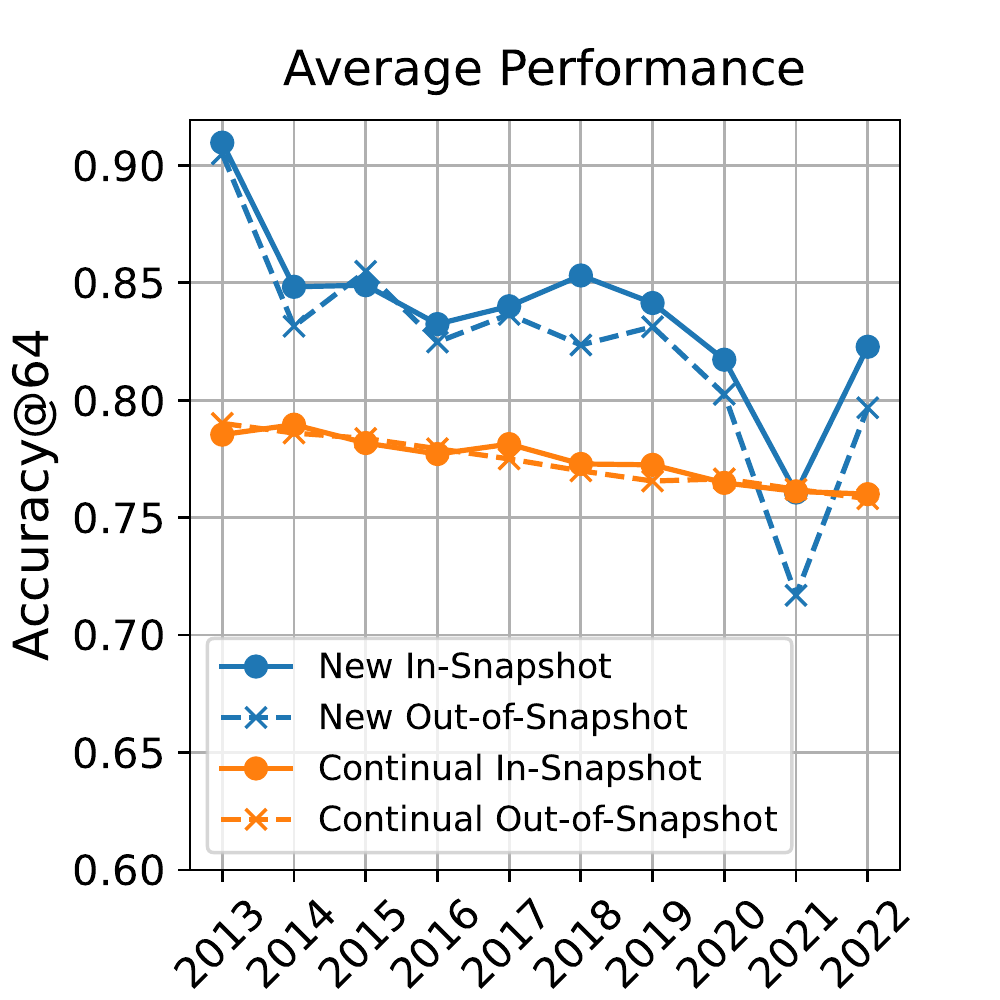}
  \caption{Difference in performance between \emph{new} and \emph{continual} entities.}
  \label{fig:avg_performance}
\end{subfigure}\hfill
\begin{subfigure}{.32\textwidth}
  \centering
  \includegraphics[width=1.0\linewidth, trim={0.0cm 0.0cm 0.0cm 1.00cm},clip]{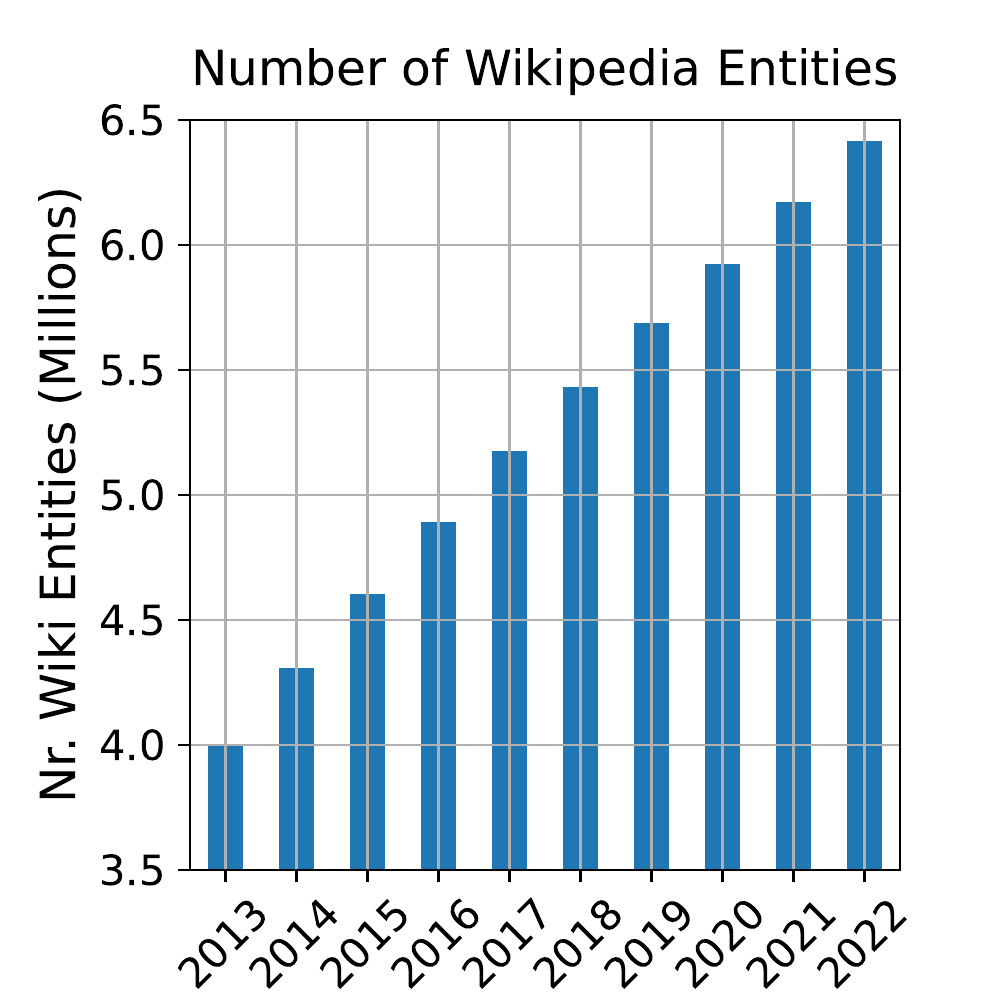}  
  \caption{Evolution of the number of entities in the Wikipedia KB.}
  \label{fig:nr_wiki_entities}
\end{subfigure}
\caption[Statistics related to the analysis of the results]{Statistics related to the analysis of the results (\secref{sec:results_and_analysis}) across the temporal snapshots (x-axis).}
\label{fig:candidates_and_performance}
\end{figure}
We experiment with the bi-encoder \cite{mazare2018training,dinan2018wizard} baseline introduced in the BLINK model \cite{wu2019zero}. This method independently encodes the mention contexts 
from 
the entity descriptions, and then performs the retrieval in a dense space \cite{karpukhin2020dense} by matching the context of each mention with the closest candidate entities.  
For the entity description, we concatenate the title to the content of the page describing a particular entity. Both mention context as well as entity descriptions are truncated to 128 BERT tokens as per BLINK model \cite{wu2019zero}.
Similarly to \cite{agarwal2022temporal, loureiro2022timelms},
we start from a pre-trained BERT model,\footnote{We use BERT-large, which is trained on a Wikipedia snapshot from 2018 \cite{joshi2019bert}.} which we finetune using our {\ourdataset} snapshots' training data --- rather than fully re-training the BERT language model on the respective year's full Wikipedia corpus.
We leave the latter full-fledged BERT (re-)training approach for future work.

\subsection{Results and analysis}
\label{sec:results_and_analysis}
\begin{table}[t] 
\caption[Accuracy@64 for \emph{continual} and \emph{new} entities]{Accuracy@64 for \emph{continual} (top) and \emph{new} (bottom) entities. The intensity of colors is set on a row-by-row basis and indicates whether performance is \textcolor{darkspringgreen!100}{\textbf{better}} or \textcolor{deepcarmine!100}{\textbf{worse}} compared to the year the model was finetuned on (\ie the values that form the white diagonal).} 
\label{tab:general_table_shared}
\vspace{.75\baselineskip}
\centering 
\resizebox{\columnwidth}{!} 
{\begin{tabular}{c cccccccccc} 
	 \toprule 
	 \multicolumn{11}{c}{Continual Entities} \\ 
	 \midrule 
	 \backslashbox{\textbf{Train}}{\textbf{Test}}& 2013 & 2014 & 2015 & 2016 & 2017 & 2018 & 2019 & 2020 & 2021 & 2022 \\ 
	\midrule 
2013  & 0.785 & \cellcolor{deepcarmine!16}0.782 & \cellcolor{deepcarmine!23}0.778 & \cellcolor{deepcarmine!34}0.772 & \cellcolor{deepcarmine!40}0.769 & \cellcolor{deepcarmine!52}0.762 & \cellcolor{deepcarmine!60}0.758 & \cellcolor{deepcarmine!60}0.758 & \cellcolor{deepcarmine!68}0.754 & \cellcolor{deepcarmine!75}0.750 \\ 
2014  & \cellcolor{darkspringgreen!16}0.792 & 0.790 & \cellcolor{deepcarmine!20}0.785 & \cellcolor{deepcarmine!28}0.781 & \cellcolor{deepcarmine!37}0.777 & \cellcolor{deepcarmine!49}0.771 & \cellcolor{deepcarmine!59}0.767 & \cellcolor{deepcarmine!58}0.767 & \cellcolor{deepcarmine!67}0.763 & \cellcolor{deepcarmine!75}0.760 \\ 
2015  & \cellcolor{darkspringgreen!20}0.786 & \cellcolor{darkspringgreen!14}0.784 & 0.782 & \cellcolor{deepcarmine!23}0.777 & \cellcolor{deepcarmine!32}0.773 & \cellcolor{deepcarmine!44}0.769 & \cellcolor{deepcarmine!53}0.765 & \cellcolor{deepcarmine!57}0.764 & \cellcolor{deepcarmine!67}0.760 & \cellcolor{deepcarmine!75}0.757 \\ 
2016  & \cellcolor{darkspringgreen!43}0.789 & \cellcolor{darkspringgreen!29}0.784 & \cellcolor{darkspringgreen!22}0.781 & 0.777 & \cellcolor{deepcarmine!21}0.773 & \cellcolor{deepcarmine!36}0.768 & \cellcolor{deepcarmine!49}0.763 & \cellcolor{deepcarmine!49}0.763 & \cellcolor{deepcarmine!64}0.758 & \cellcolor{deepcarmine!75}0.755 \\ 
2017  & \cellcolor{darkspringgreen!54}0.794 & \cellcolor{darkspringgreen!43}0.791 & \cellcolor{darkspringgreen!35}0.788 & \cellcolor{darkspringgreen!22}0.785 & 0.781 & \cellcolor{deepcarmine!32}0.775 & \cellcolor{deepcarmine!45}0.771 & \cellcolor{deepcarmine!43}0.772 & \cellcolor{deepcarmine!59}0.768 & \cellcolor{deepcarmine!75}0.763 \\ 
2018  & \cellcolor{darkspringgreen!75}0.791 & \cellcolor{darkspringgreen!63}0.788 & \cellcolor{darkspringgreen!54}0.786 & \cellcolor{darkspringgreen!42}0.782 & \cellcolor{darkspringgreen!29}0.778 & 0.773 & \cellcolor{deepcarmine!24}0.769 & \cellcolor{deepcarmine!23}0.769 & \cellcolor{deepcarmine!41}0.764 & \cellcolor{deepcarmine!54}0.760 \\ 
2019  & \cellcolor{darkspringgreen!75}0.795 & \cellcolor{darkspringgreen!65}0.792 & \cellcolor{darkspringgreen!56}0.789 & \cellcolor{darkspringgreen!43}0.784 & \cellcolor{darkspringgreen!34}0.781 & \cellcolor{darkspringgreen!20}0.776 & 0.772 & \cellcolor{darkspringgreen!11}0.773 & \cellcolor{deepcarmine!24}0.767 & \cellcolor{deepcarmine!32}0.765 \\ 
2020  & \cellcolor{darkspringgreen!75}0.787 & \cellcolor{darkspringgreen!63}0.783 & \cellcolor{darkspringgreen!61}0.782 & \cellcolor{darkspringgreen!46}0.777 & \cellcolor{darkspringgreen!37}0.774 & \cellcolor{darkspringgreen!20}0.768 & \cellcolor{deepcarmine!10}0.765 & 0.765 & \cellcolor{deepcarmine!21}0.761 & \cellcolor{deepcarmine!35}0.756 \\ 
2021  & \cellcolor{darkspringgreen!75}0.788 & \cellcolor{darkspringgreen!67}0.785 & \cellcolor{darkspringgreen!59}0.782 & \cellcolor{darkspringgreen!47}0.777 & \cellcolor{darkspringgreen!38}0.773 & \cellcolor{darkspringgreen!27}0.769 & \cellcolor{darkspringgreen!17}0.764 & \cellcolor{darkspringgreen!17}0.764 & 0.761 & \cellcolor{deepcarmine!20}0.757 \\ 
2022  & \cellcolor{darkspringgreen!75}0.790 & \cellcolor{darkspringgreen!67}0.787 & \cellcolor{darkspringgreen!60}0.783 & \cellcolor{darkspringgreen!51}0.779 & \cellcolor{darkspringgreen!44}0.776 & \cellcolor{darkspringgreen!34}0.771 & \cellcolor{darkspringgreen!26}0.768 & \cellcolor{darkspringgreen!27}0.768 & \cellcolor{darkspringgreen!19}0.764 & 0.760 \\ 
	\midrule 
	 \multicolumn{11}{c}{New Entities} \\ 
	 \midrule 
	 \backslashbox{\textbf{Train}}{\textbf{Test}}& 2013 & 2014 & 2015 & 2016 & 2017 & 2018 & 2019 & 2020 & 2021 & 2022 \\ 
	\midrule 
2013  & 0.910 & \cellcolor{deepcarmine!36}0.819 & \cellcolor{deepcarmine!26}0.853 & \cellcolor{deepcarmine!34}0.826 & \cellcolor{deepcarmine!30}0.841 & \cellcolor{deepcarmine!38}0.812 & \cellcolor{deepcarmine!36}0.819 & \cellcolor{deepcarmine!44}0.791 & \cellcolor{deepcarmine!75}0.688 & \cellcolor{deepcarmine!49}0.774 \\ 
2014  & \cellcolor{darkspringgreen!36}0.908 & 0.848 & \cellcolor{darkspringgreen!16}0.862 & \cellcolor{deepcarmine!19}0.827 & \cellcolor{deepcarmine!12}0.843 & \cellcolor{deepcarmine!17}0.832 & \cellcolor{deepcarmine!12}0.842 & \cellcolor{deepcarmine!25}0.814 & \cellcolor{deepcarmine!75}0.704 & \cellcolor{deepcarmine!35}0.791 \\ 
2015  & \cellcolor{darkspringgreen!32}0.898 & \cellcolor{deepcarmine!21}0.823 & 0.849 & \cellcolor{deepcarmine!22}0.822 & \cellcolor{deepcarmine!28}0.808 & \cellcolor{deepcarmine!26}0.813 & \cellcolor{deepcarmine!17}0.832 & \cellcolor{deepcarmine!37}0.788 & \cellcolor{deepcarmine!75}0.706 & \cellcolor{deepcarmine!41}0.781 \\ 
2016  & \cellcolor{darkspringgreen!46}0.897 & \cellcolor{deepcarmine!10}0.832 & \cellcolor{darkspringgreen!26}0.862 & 0.832 & \cellcolor{darkspringgreen!13}0.839 & \cellcolor{deepcarmine!15}0.823 & \cellcolor{deepcarmine!15}0.823 & \cellcolor{deepcarmine!27}0.802 & \cellcolor{deepcarmine!75}0.718 & \cellcolor{deepcarmine!33}0.791 \\ 
2017  & \cellcolor{darkspringgreen!43}0.906 & \cellcolor{deepcarmine!13}0.832 & \cellcolor{darkspringgreen!18}0.857 & \cellcolor{deepcarmine!22}0.817 & 0.840 & \cellcolor{deepcarmine!18}0.824 & \cellcolor{deepcarmine!12}0.835 & \cellcolor{deepcarmine!35}0.791 & \cellcolor{deepcarmine!75}0.714 & \cellcolor{deepcarmine!26}0.808 \\ 
2018  & \cellcolor{darkspringgreen!38}0.908 & \cellcolor{deepcarmine!19}0.835 & \cellcolor{darkspringgreen!12}0.858 & \cellcolor{deepcarmine!21}0.830 & \cellcolor{deepcarmine!13}0.846 & 0.853 & \cellcolor{deepcarmine!19}0.835 & \cellcolor{deepcarmine!34}0.806 & \cellcolor{deepcarmine!75}0.728 & \cellcolor{deepcarmine!35}0.803 \\ 
2019  & \cellcolor{darkspringgreen!51}0.910 & \cellcolor{darkspringgreen!10}0.842 & \cellcolor{darkspringgreen!17}0.853 & \cellcolor{deepcarmine!22}0.821 & \cellcolor{darkspringgreen!10}0.842 & \cellcolor{darkspringgreen!10}0.843 & 0.841 & \cellcolor{deepcarmine!29}0.810 & \cellcolor{deepcarmine!75}0.734 & \cellcolor{deepcarmine!35}0.799 \\ 
2020  & \cellcolor{darkspringgreen!72}0.903 & \cellcolor{darkspringgreen!18}0.828 & \cellcolor{darkspringgreen!29}0.844 & \cellcolor{darkspringgreen!23}0.835 & \cellcolor{darkspringgreen!28}0.843 & \cellcolor{darkspringgreen!11}0.819 & \cellcolor{darkspringgreen!21}0.833 & 0.817 & \cellcolor{deepcarmine!75}0.728 & \cellcolor{deepcarmine!14}0.811 \\ 
2021  & \cellcolor{darkspringgreen!75}0.910 & \cellcolor{darkspringgreen!37}0.825 & \cellcolor{darkspringgreen!49}0.852 & \cellcolor{darkspringgreen!37}0.825 & \cellcolor{darkspringgreen!43}0.837 & \cellcolor{darkspringgreen!34}0.817 & \cellcolor{darkspringgreen!40}0.830 & \cellcolor{darkspringgreen!33}0.814 & 0.761 & \cellcolor{darkspringgreen!32}0.812 \\ 
2022  & \cellcolor{darkspringgreen!68}0.905 & \cellcolor{darkspringgreen!26}0.846 & \cellcolor{darkspringgreen!31}0.852 & \cellcolor{deepcarmine!11}0.820 & \cellcolor{darkspringgreen!14}0.830 & \cellcolor{darkspringgreen!14}0.830 & \cellcolor{darkspringgreen!16}0.832 & \cellcolor{deepcarmine!20}0.808 & \cellcolor{deepcarmine!75}0.732 & 0.823 \\ 
 
        \bottomrule 
        \end{tabular}} 
\end{table}
The results for \emph{continual} and \emph{new} entities are shown in \tabref{tab:general_table_shared}.
The rows thereof represent the snapshots
whose train set we used to finetune the bi-encoder model, while the columns indicate the snapshots test data each of the finetuned models was tested on.
The used metric is 
accuracy@64, 
which amounts to the fraction of anchor mentions in the test set for which the top-64 candidate entity list from the EL model includes the correct target.
We observe a consistent temporal decrease in performance for \emph{continual} entities \ref{it:q1-temp-degradation}. This is also 
reflected
in \figref{fig:avg_performance}, which 
illustrates
the average temporal degradation across all the finetuned models. 
We hypothesize that this degradation over time is because, as time evolves, the relative ``semantic distance'' between the ever growing number of entities shrinks: entities become harder to distinguish from one another.
In order to demonstrate this, we calculate the \textit{Jaccard Similarity} between consecutive descriptions of the top 64 candidate entities returned by the bi-encoder. We observe a consistent increase in this similarity metric illustrated in \figref{fig:candidate_similarity}. This growth in more similar entities is accompanied with a general increase in the number of entities in the Wikipedia KB (see \figref{fig:nr_wiki_entities}).
Consequently, the model is given an ever-increasing number of candidate target entities, which can potentially impact its performance.

\begin{figure}[t]
\begin{subfigure}{.32\textwidth}
  \centering
  \includegraphics[width=1.0\linewidth]{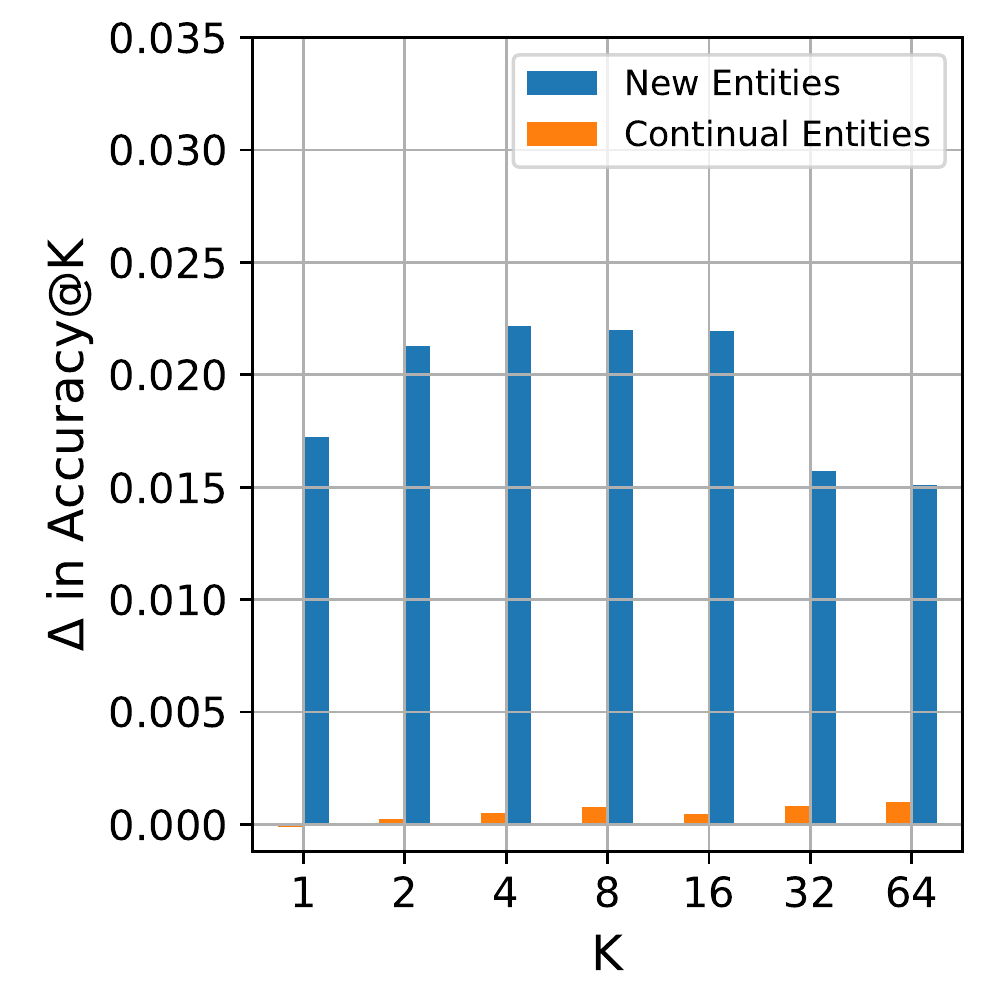}  
  \caption{\revklim{Effect of in-snapshot finetuning (y-axis) across different accuracy thresholds $K$.}}
  \label{fig:in-snapshot-k}
\end{subfigure}\hfill
\begin{subfigure}{.32\textwidth}
  \centering
  \includegraphics[width=1.0\linewidth]{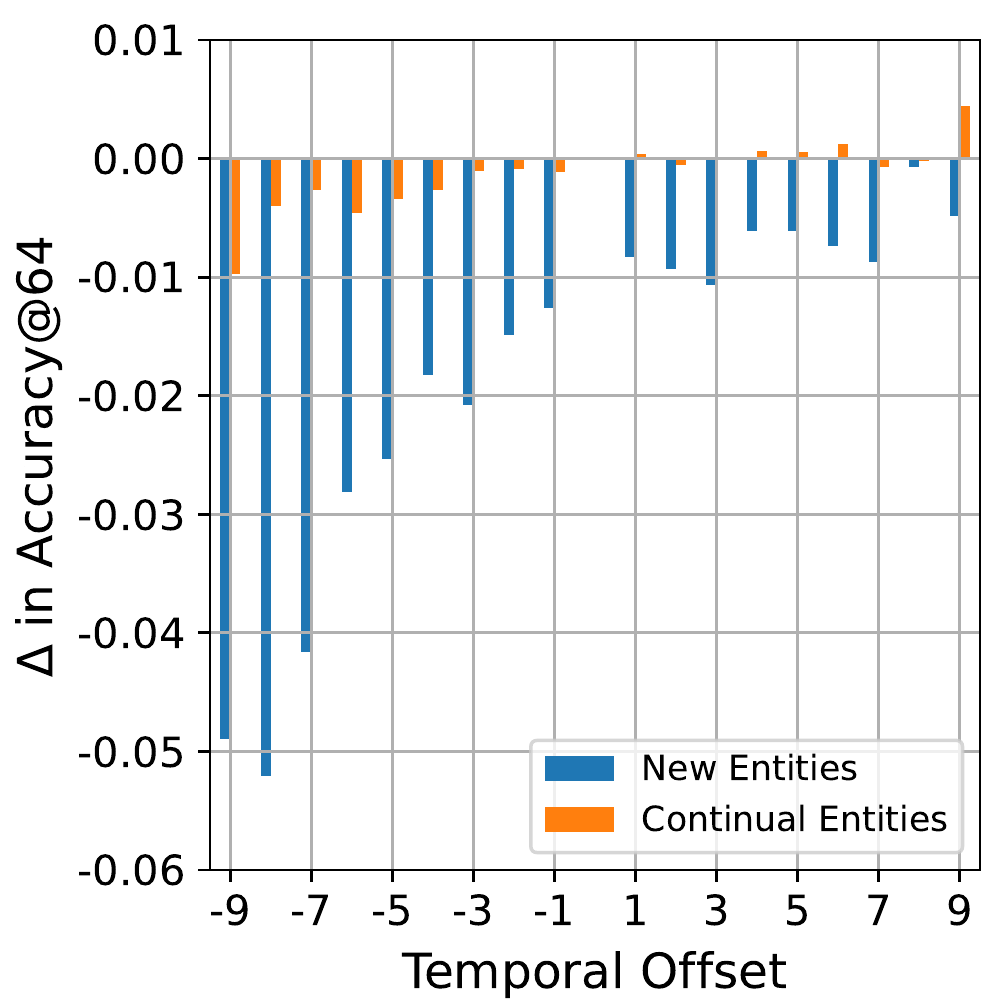}
  \caption{\revklim{In-snapshot finetuning (offset 0) compared to finetuning on past and future snapshots ($-$ and $+$ offsets).}}
  \label{fig:in-snapshot-offset}
\end{subfigure}\hfill
\begin{subfigure}{.32\textwidth}
  \centering
  \includegraphics[width=1.0\linewidth]{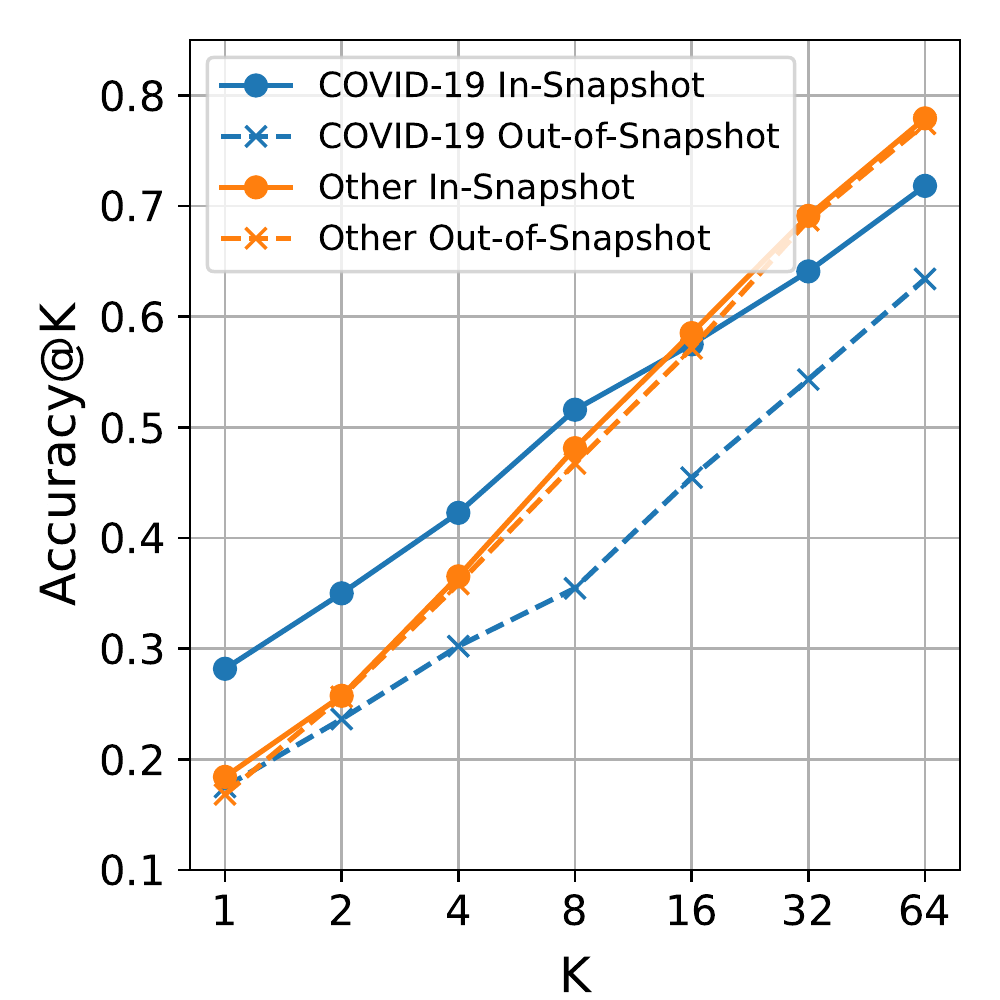}  
  \caption{\revklim{In-snapshot finetuning effect on COVID-19 related and other \textit{new entities} from 2021 snapshot.}}
  \label{fig:in-snapshot-covid}
\end{subfigure}
\caption[Impact of finetuning]{\revklim{Impact of finetuning and evaluating on the same snapshot 
(\emph{in-snapshot}) 
compared to finetuning and evaluating on different snapshots
(\emph{out-of-snapshot}). 
We observe: 
\begin{enumerate*}[(a)]
    \item a superior impact of in-snapshot finetuning on \textit{new} entities compared to \textit{continual} ones,
    \item a decrease in performance when finetuning on increasingly older spanshots, 
    and 
    \item dominant effect of in-snapshot finetuning on entities that require fundamentally new knowledge (\eg COVID-19 related entities). 
\end{enumerate*}
}}
\label{fig:in_snapshot}
\end{figure}

\revklim{Furthermore, we analyze the impact finetuning on different snapshots has on the performance of the model \ref{it:q2-finetuning}. 
To this end, we distinguish between \textit{in-snapshot} and \textit{out-of-snapshot} 
finetuning setups. 
In \textit{in-snapshot} setup, 
the bi-encoder model is finetuned and evaluated
on the same snapshot. 
Conversely, in \textit{out-of-snapshot} setting, the model is evaluated on a different snapshot than the one used for its finetuning.
\Figref{fig:in-snapshot-k} illustrates the difference in performance between the in-snapshot and out-of-snapshot predictions for new and continual entities. 
We observe a general increase in performance for in-snapshot finetuning with a marginal gain for \textit{continual} entities compared to the \textit{new} ones.\footnote{\revklim{We analyze more in detail the difference in performance between \emph{new} and \textit{continual} entities in next paragraphs when addressing \ref{it:q3-emerging}.}} 
This general lower impact of in-snapshot finetuning on \textit{continual} entities, 
leads us to hypothesize that the actual knowledge needed to disambiguate 
most of these entities in {\ourdataset} changes very little with time. In order to 
verify this hypothesis, 
we randomly selected 100 
continual entity-mention pairs, 
and compared the difference in both mention contexts and entity descriptions between the years 2013 and 2022.
We found that in most cases (>95\%), while the textual description of the continual entity is changed (supported by \figsref{fig:evolution_entities_jaccard}{fig:evolution_entities_edistance}), its meaning remains the same.
}

\revklim{Moreover, we address the second part of \textbf{Q2}
targeting
the effect of timespan between the snapshot used for finetuning and the one used for evaluation.
To accomplish this, in \figref{fig:in-snapshot-offset} 
 we showcase 
the impact of in-snapshot finetuning relative to the \textit{temporal offset} between the snapshot the model was tested and the snapshot the model was finetuned on. For negative temporal offset,\footnote{\revklim{Evaluation snapshot comes from later time period than the snapshot the model was finetuned on.}} we observe a decrease in the performance difference 
between in-snapshot and out-of-snapshot setups
as the offset approaches to zero. This indicates that the model can benefit more from recent snapshots than from snapshots further in the past. 
Curiously, we 
observe a 
slight increase in performance for out-of-snapshot \textit{continual} entities trained on future snapshots (positive temporal offsets in \figref{fig:in-snapshot-offset}). 
This suggests that the changes in continual entities are \textit{accumulative} in Wikipedia,
with later versions of entity descriptions also including the information from the past. 
For instance,
we have observed that for entities describing people, the newly added information on the occupation (\eg soccer coach) is appended to the occupation description a person had in the past (\eg soccer player).}

Next, we analyze the EL performance on \emph{new} entities and whether they are differently affected than the \emph{continual} ones \ref{it:q3-emerging}.
\revklim{We plot the in-snapshot and out-of-snapshot average temporal change in accuracy@64 scores across all finetuned models for both types of entities in \figref{fig:avg_performance}}.
\revklim{We observe that, in general, the performance on \emph{new} entities is superior to that on \emph{continual} ones. Furthermore, as observed above, the performance gain from in-snapshot finetuning on new entities is superior compared to that on continual ones (supported by \figref{fig:avg_performance} and \figsref{fig:in-snapshot-k}{fig:in-snapshot-offset}). This difference suggests that new entities require a higher degree of additional snapshot-specific knowledge to be correctly disambiguated.
Additionally, the graph in \figref{fig:avg_performance} reveals that this delta in performance is 
larger
for more recent years (starting from 2018). We hypothesize 
that this behaviour is due to the fact that the used original BERT model\cite{devlin2019bert} has not been exposed to more recent new entities during pre-training. It also suggests a complementary effect between task-specific finetuning on \ourdataset~dataset and language model pre-training on larger corpora.}  

\revklim{Furthermore, 
to better understand the superior performance on new entities, 
we manually analyze 100 randomly selected
\emph{new} entities from our dataset.} 
\revklim{We found that a large 
majority ($\sim$90\%) of entities 
were either events that are 
recurrent in nature (\eg ``2018 BNP Paribas Open'') ($\sim$68\%) or extracts of already existing pages ($\sim$22\%).}
We conjecture\footnote{\revklim{See \secref{app:additional_results} of the supplementary material for further details on the performance on these different new entity types.}} 
that these entities require little additional knowledge
to be disambiguated, since either they already exist (as part of the content of other entities) or are very similar to already existing entities in Wikipedia. 
This contrasts sharply with the performance drop observed for \emph{new} entities in the temporal snapshot 2021, as exhibited in both \figref{fig:avg_performance} and \tabref{tab:general_table_shared}.
This decrease is mostly driven by COVID-19 related entities, which constitute 
\revklim{24\%}
of the new entities, which are linked to by 
\revklim{30\%}
of the mentions \revklim{in this spanshot}. 
The disambiguation of these cases requires completely new and 
fundamentally
different, previously non-existent knowledge.
\revklim{Since this knowledge is not present in the original corpus used to pre-train the BERT encoder nor in any of the previous snapshots,
our EL model based on it struggles.}

\revklim{Finally, we 
analyze the impact of new entities finetuning \ref{it:q2-finetuning} on the temporal snapshot 2021, 
for which our model exhibits the lowest temporal performance
 driven by COVID-19 disambiguation instances (see above)}. \revklim{\Figref{fig:in-snapshot-covid} showcases the 
impact of in- and out-of-snapshot finetuning
on the performance on
COVID-19 related entities compared to \textit{other} new entities 
for different thresholds $K$ of the accuracy@$K$ metric. 
We observe a large difference in performance 
(up to 14\% accuracy@64 points)
between COVID-19 related and the rest of the instances for out-of-snapshot finetuning.
This difference 
is significantly decreased when finetuning on the 2021 snapshot (in-snapshot finetuning), 
achieving
superior accuracy on COVID-19 related entities for lower values of $K$ compared to \textit{other} entities. In contrast, the difference between out- and in-snapshot performance on these non-COVID-19 related entities (\textit{other} entities in \figref{fig:in-snapshot-covid}) is marginal. 
This suggests that in-snapshot finetuning 
has dominant
impact on new entities that require fundamentally new, previously non-existent knowledge in Wikipedia. 
}

\section{Limitations and future work}
A number of dataset and model-related aspects were left unexplored in the current work.
Our clarifications thereof below may help the community to understand the limitations and 
potential future research directions to extend our efforts. 
\paragraph{Effect of pre-training on new corpora} Recent work has demonstrated the benefits of pre-training language models on more recent corpora (\eg the latest Wikipedia versions) when applied on downstream tasks \cite{agarwal2022temporal, loureiro2022timelms}. We hypothesize 
that this pre-training may also improve EL performance for our {\ourdataset},
especially for \emph{new} entities that require new 
world knowledge.
\paragraph{Changes in mention context} Our work focused mostly on changes in target entities, 
leaving the effect of changes in mention context on EL performance unexplored.
For example,
\figref{fig:evolution_mentions} 
shows a notable temporal drop in Jaccard vocabulary similarity of the context 
surrounding mentions.
This suggests that 
mentions, as well as the text surrounding them, are 
quite volatile and 
evolve over time, making them an interesting subject for future research. 

\paragraph{Cross-lingual time evolution} Our dataset is limited to English Wikipedia. Yet, since recent work \cite{botha2020entity, de2021multilingual} has shown the benefits of training EL models in a cross-lingual setting, 
studying cross-lingual temporal evolution of entity linking task 
may also
be an interesting future research direction. Furthermore, it will complement the recent growing interest in creating entity linking datasets for a number of low-resourced languages \cite{hennig2021mobie, ogrodniczuk2020wikipedia, caillaut2022automated,rosales2021towards}.

\section{Conclusion}
This paper introduced {\ourdataset}, a new large-scale 
temporal
entity linking 
dataset 
composed of 10 yearly snapshots of Wikipedia target entities linked to by anchor mentions. 
In our dataset creation pipeline, we put
special focus on the quality assurance and future extensibility of \ourdataset.
Furthermore, we 
established baseline entity linking results across different years,
which revealed a noticeable performance deterioration on test data more recent than the training data.
We further examined the most challenging cases, 
suggesting the need for updating the pre-trained language model of our EL model, at least to perform well on newly appearing entities that require new world knowledge (\eg in case of COVID-19).
Finally, we described limitations of our work and discussed potential future research directions.
\newpage
\section*{Acknowledgements}
\revklim{\noindent Part of the research leading to these results has received funding from
\begin{enumerate*}[(i)]
\item the European Union's Horizon 2020 research and innovation programme under grant agreement no.\ 761488 for the CPN project,\footnote{\url{https://www.projectcpn.eu/}}
\item the Flemish Government under the programme ``Onderzoeksprogramma Artifici\"{e}le Intelligentie (AI) Vlaanderen'', 
 \item the Research Foundation -- Flanders grant no.\ V412922N for Long Stay Abroad at Copenhagen University, and
 \item DFF Sapere Aude grant No 0171-00034B ‘Learning to Explain Attitudes on Social Media (EXPANSE)’.
\end{enumerate*}}
{
\small
}

\clearpage
\section*{Checklist}

The checklist follows the references.  Please
read the checklist guidelines carefully for information on how to answer these
questions.  For each question, change the default TODO to Yes,
No, or N/A.  You are strongly encouraged to include a {\bf
justification to your answer}, either by referencing the appropriate section of
your paper or providing a brief inline description.  For example:
\begin{itemize}
  \item Did you include the license to the code and datasets? Yes. See the supplementary materials.
\end{itemize}
Please do not modify the questions and only use the provided macros for your
answers.  Note that the Checklist section does not count towards the page
limit.  In your paper, please delete this instructions block and only keep the
Checklist section heading above along with the questions/answers below.

\begin{enumerate}

\item For all authors...
\begin{enumerate}
  \item Do the main claims made in the abstract and introduction accurately reflect the paper's contributions and scope?
    Yes
  \item Did you describe the limitations of your work?
    Yes
  \item Did you discuss any potential negative societal impacts of your work?
    N/A
  \item Have you read the ethics review guidelines and ensured that your paper conforms to them?
    Yes
\end{enumerate}

\item If you are including theoretical results...
\begin{enumerate}
  \item Did you state the full set of assumptions of all theoretical results?
    N/A
	\item Did you include complete proofs of all theoretical results?
    N/A
\end{enumerate}

\item If you ran experiments (e.g. for benchmarks)...
\begin{enumerate}
  \item Did you include the code, data, and instructions needed to reproduce the main experimental results (either in the supplemental material or as a URL)?
    Yes. The link to the dataset will be shared as part of the supplementary material.
  \item Did you specify all the training details (e.g., data splits, hyperparameters, how they were chosen)?
    Yes. See the supplementary material.
	\item Did you report error bars (e.g., with respect to the random seed after running experiments multiple times)?
    No. No additional computational resources for this, yet the results across multiple temporal snapshots used to finetune are consistent. 
	\item Did you include the total amount of compute and the type of resources used (e.g., type of GPUs, internal cluster, or cloud provider)?
    Yes. See the supplementary material.
\end{enumerate}

\item If you are using existing assets (e.g., code, data, models) or curating/releasing new assets...
\begin{enumerate}
  \item If your work uses existing assets, did you cite the creators?
    Yes
  \item Did you mention the license of the assets?
    Yes. See supplementary material. 
  \item Did you include any new assets either in the supplemental material or as a URL?
    No
  \item Did you discuss whether and how consent was obtained from people whose data you're using/curating?
    N/A
  \item Did you discuss whether the data you are using/curating contains personally identifiable information or offensive content?
    N/A
\end{enumerate}

\item If you used crowdsourcing or conducted research with human subjects...
\begin{enumerate}
  \item Did you include the full text of instructions given to participants and screenshots, if applicable?
    N/A
  \item Did you describe any potential participant risks, with links to Institutional Review Board (IRB) approvals, if applicable?
    N/A
  \item Did you include the estimated hourly wage paid to participants and the total amount spent on participant compensation?
    N/A
\end{enumerate}

\end{enumerate}
\clearpage
\newcounter{savechap}
\setcounter{savechap}{\value{chapter}}
\setcounter{section}{0}
\renewcommand*{\thesection}{\thesavechap.\Alph{section}}

\section{Supplementary material}
\subsection{Dataset and code distribution}
\paragraph{Link to the dataset} The reviewers can access the dataset using the following link: \url{https://cloud.ilabt.imec.be/index.php/s/RinXy8NgqdW58RW}. The dataset and the baseline code will be made publicly available in a dedicated GitHub repository upon acceptance.
\paragraph{License} \ourdataset~is distributed under Creative Commons Attribution-ShareAlike 4.0 International license (CC BY-SA 4.0).\footnote{\url{https://creativecommons.org/licenses/by-sa/4.0/}} 
\paragraph{Maintenance} The maintenance and extension to further temporal snapshots of \ourdataset~will be carried out by the authors of the paper. Additionally, we will make the code public to create potential new variations and extensions of \ourdataset~using a number of hyperparameters (see Sections \ref{app:hyperparameters} and \ref{app:dataset_extension} for further details). 

\subsection{Datasheet for \ourdataset}
In this section we provide a more detailed documentation of the dataset with the intended uses. We base ourselves on the datasheet proposed by \cite{gebru2021datasheets}. 
\subsubsection{Motivation}
\paragraph{For what purpose was the dataset created?}
The \ourdataset~dataset was created to evaluate how the temporal 
change
of anchor mentions and that of target Knowledge Base (KB; \ie modification or creation of new entities) affects the \textit{entity linking} (EL) task. This contrasts with the currently existing datasets \cite{usbeck2015gerbil,roder2018gerbil,sevgili2020neural,petroni2020kilt}, which are associated with a single version of the target KB such as the Wikipedia 2010 for the widely adopted CoNLL-AIDA\cite{hoffart2011robust} dataset. We expect that \ourdataset~will encourage research in devising new models and architectures that are robust to temporal changes both in mentions as well as in the target KBs. 

\paragraph{Who created the dataset and on behalf of which entity?}
The dataset is the result of joint effort involving researchers from the University of Copenhagen and Ghent University. 

\paragraph{Who funded the creation of the dataset?}
The creation of \ourdataset~was funded by the following grants: 
\begin{enumerate}
    \item FWO (Fonds voor Wetenschappelijk Onderzoek) long-stay abroad grant V412922N.
    \item The Flemish Government fund under the programme ``Onderzoeksprogramma Artifici\"{e}le Intelligentie (AI) Vlaanderen''.
\end{enumerate}

\subsubsection{Composition}
\paragraph{What do the instances that comprise the dataset represent?}
Each of the instances consists of a mention in Wikipedia linked to target entity, i.e., a Wikipedia page, with a set of attributes. 
The dataset is organized in 10 yearly temporal snapshots starting from 
January 1, 2013 until January 1, 2022. See \secref{app:mention_entity_attributes} for further details on the attributes associated with each of the instances of our \ourdataset~dataset. 

\paragraph{How many instances are there in total?}
\Tabref{tab:dataset_detail2} of the main manuscript summarizes the number of instances (\# Anchor Mentions) of each of the entity categories (\textit{continual} and \textit{new}) in \ourdataset. See \secref{app:mention_per_entity_distribution} for additional statistics on mention per entity distribution. 
\begin{figure}[!t]
\centering
\includegraphics[width=0.6\columnwidth]{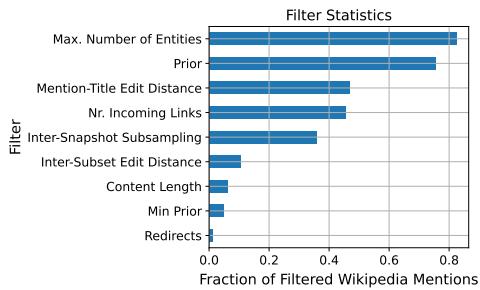}
\caption[Fraction of filtered Wikipedia mentions]{Figure showcasing the fraction of filtered Wikipedia mentions by each of the filters executed during \ourdataset~generation. }
\label{fig:filter_effect}
\end{figure}
\paragraph{Does the dataset contain all possible instances or is it a sample
(not necessarily random) of instances from a larger set?}
\ourdataset~contains a sample of all the possible anchor mentions linked to target entities from Wikipedia. The following are the filters applied to obtain the instances in the final \ourdataset~dataset whose effect is also summarized in \figref{fig:filter_effect}:
\begin{enumerate}
    \item \textbf{Prior-based filtering}: we exclude all the mentions for which the correct entity it refers to has the highest \textit{prior}~\cite{yamada2016joint} as calculated in \equref{eq:mention_prior} of the manuscript. This filtering is done with the goal of creating a more challenging dataset. 

    \emph{Value to create \ourdataset}: mentions with mention prior rank $>$ 1 among other mentions referring to the same entity. 
    
    \emph{Percentage of filtered out instances}: between 74.20\% and 76.28\%, depending on the temporal snapshot. 
    
    \emph{Hyperparameter name:} \texttt{min\_men\_prior\_rank} (see \tabref{tab:hyperparameters} in \secref{app:hyperparameters}).
    
    \item \textbf{Entity relevance filtering}: we impose the restriction for target entity of having at least 10 incoming links (\ie at least 10 mentions linking to it) in order to be included in \ourdataset. Additionally, we filter out target entities whose description contains less than 10 tokens. This is done in order to avoid introducing potentially noisy and irrelevant entities that have not been sufficiently established by the Wikipedia community.  
    
    \emph{Value to create \ourdataset}: 10 for minimum number of incoming links and 10 for minimum content length (in number of tokens) of target entity.
    
    \emph{Percentage of filtered out instances}: 
    \begin{itemize}
    	\item Minimum number of incoming links: between 42.66\% and 48.32\%, depending on the temporal snapshot. 
    	\item Minimum content length: between 0.06\% and 0.95\% depending on the temporal snapshot. 
    \end{itemize}

    \emph{Hyperparameter names:} \texttt{min\_nr\_inlinks} for minimum number of incoming links and \texttt{min\_len\_target\_ent} for minimum number of content length tokens (see \tabref{tab:hyperparameters} in \secref{app:hyperparameters}).

    \item \textbf{Min prior subsampling}: the mentions with very low mention prior are filtered out from \ourdataset. This way, we avoid introducing too infrequent and potentially erroneous mentions to refer to a particular entity. 
    
    \emph{Value to create \ourdataset}: 0.0001
    
    \emph{Percentage of filtered out instances}: between 0.37\% and 0.61\%, depending on the snapshot. 
    
    \emph{Hyperparameter name:} \texttt{min\_men\_prior} (see \tabref{tab:hyperparameters} in \secref{app:hyperparameters}).
    \item \textbf{Minimum mentions per entity}: has similar effect as previously explained \textit{min prior subsampling} (see above) filter. We do not use it in the creation of \ourdataset, relying completely on the \textit{min prior subsampling} filter.
        
    \emph{Value to create \ourdataset}: 1
    
    \textit{Percentage of filtered out instances}: 0\%
    
    \emph{Hyperparameter name:} \texttt{min\_mens\_per\_ent} (see \tabref{tab:hyperparameters} in \secref{app:hyperparameters}).
    
    \item \textbf{Edit distance mention title}: filters out the anchor mentions that are very similar to target entity page. This way, we expect to reduce the trivial cases where the entity linking can be simply predicted by mapping the mention to the title of the target entity.
    
    \emph{Value to create \ourdataset}: 0.2 (normalized edit distance). 
    
    \textit{Percentage of filtered out instances}: between 44.85\% and 48.99\%, depending on the snapshot. 
    
    \emph{Hyperparameter name:} \texttt{ed\_men\_title} (see \tabref{tab:hyperparameters} in \secref{app:hyperparameters}).
    
    \item \textbf{Redirect filtering}: we filter out anchor mentions that point to redirect pages (pages without content redirecting to other pages in Wikipedia). 
    
    \textit{Percentage of filtered out instances}: between 1.02\% and 1.47\%, depending on the snapshot. 
    
    \item \textbf{Inter-subset filtering}: 
    we enforce normalized edit distance between the mentions in different subsets referring to the same target entity to be higher than 0.2. This entails that the entities in \ourdataset~are linked to at least by 3 mentions with different surface form. The main goal of this filter is to avoid mention-entity tuple memorization by the models \cite{onoe2020fine}. 
    
    \emph{Value to create \ourdataset}: 0.2 normalized edit distance between mentions in different subsets. 
    
    \textit{Percentage of filtered out instances}: 10\%.
    
    \emph{Hyperparameter name:} \texttt{ed\_men\_subsets} (see \tabref{tab:hyperparameters} in \secref{app:hyperparameters}).
    
    \item \textbf{Maximum number of entities}: we restrict the number of target entities to 10,000 for \emph{continual} instances. The reason behind this is to build a dataset of manageable size with a reasonable number of target entities to experiment with. 

    \emph{Value to create \ourdataset}: 10,000 for \emph{continual} entities.

    \emph{Percentage of filtered out instances}: 82\%. 

    \emph{Hyperparameter name:} \texttt{nr\_ct\_ents\_per\_cut} (see \tabref{tab:hyperparameters} in \secref{app:hyperparameters})
    
    \item \textbf{Maximum number of mentions per entity}: this filtering limits the number of mentions per entity in order for the dataset to not be dominated by most popular entities. Particularly, for test and evaluation subsets we limit the number of mentions per entity to 10. This way, we expect the accuracy scores to not be dominated by links to popular target entities (\ie entities with a big number of incoming links). The limit for training set is higher (500), since we want it to be representative of the real mention per entity distribution in Wikipedia. The effect of imposing this limits can be observed in \figref{fig:men_entity_distribution} for both \emph{continual} as well as \emph{new} entities represented by a significant leap in the mentions-per-entity curve, particularly noticeable for validation and test subsets. 

	\emph{Value to create \ourdataset}: 10 for validation and test subsets, 500 for the train subset.

    \emph{Percentage of filtered out instances}: for \emph{continual} instances, 84\% for validation and test subsets and 28\% for the train subset. For \emph{new} instances, 45\% for validation and test subsets and 0.3\% for the train subset. 

    \emph{Hyperparameter name:} \texttt{max\_mens\_per\_ent} (see \tabref{tab:hyperparameters} in \secref{app:hyperparameters}).
    
    \item \textbf{Inter-snapshot subsampling}: 
    finally, we enforce that the number of continual and new entities as well as the number of mentions stays the same across the temporal snapshots (see \tabref{tab:dataset_detail2}). We achieve this by performing a random mention subsampling in snapshots with higher number of mentions, weighted by the difference in the number of mentions-per-entity. This produces a very similar mention-entity distribution across the temporal snapshots (see \secref{app:mention_per_entity_distribution} for further details). 
    
    \emph{Percentage of filtered out instances}: between 5\% and 35\%, it increases for more recent temporal snapshots as they have more instances in Wikipedia.     
\end{enumerate}

We do not filter on any attribute that could potentially produce evident biases in \ourdataset~(\eg gender, geographic location of the entities, etc.).

\paragraph{What data does each instance consist of?}
Each instance of a snapshot consists of: 
\begin{enumerate}
    \item Cleaned contextual text surrounding the anchor mention from the Wikipedia snapshot. Furthermore, we include the bert-tokenized version of the text used in our baseline. 
    \item Cleaned textual description of the target entity taken from the Wikipedia snapshot. Furthermore, we include the bert-tokenized version of the text used in our baseline. 
    \item A set of additional attributes defining the anchor mention and target entity. 
\end{enumerate}
For more details about the attributes, see \secref{app:mention_entity_attributes}. Furthermore, concrete examples 
of \ourdataset's instances are showcased in \secref{app:examples}. 

\paragraph{Is there a label or target associated with each instance? }
Yes, the target entity is represented by the Wikipedia page id. Furthermore, we also pair it with Wikidata QID of the corresponding Wikidata entity. These targets correspond to the attributes \texttt{target\_page\_id} and \texttt{target\_qid} described in \tabref{tab:attributes} (see \secref{app:mention_entity_attributes} for further details). 

\paragraph{Is any information missing from individual instances? } No, all the instances should have a complete information corresponding to the content as well as to the attributes. 

\paragraph{Are relationships between individual instances made explicit? } Yes, the relations between each of the instances and the target entity are made explicit by means of \texttt{target\_page\_id} and \texttt{target\_qid} attributes (see \secref{app:mention_entity_attributes} for further details), which uniquely identify the id of the Wikipedia page describing a particular entity and the Wikidata entity respectively.

\paragraph{Are there recommended data splits (e.g., training, development/validation,
testing)? } Yes, the dataset is divided in train, validation and test subsets (see \tabref{tab:dataset_detail2} for the distribution). 

\paragraph{Are there any errors, sources of noise, or redundancies in the dataset?}
We have taken multiple measures to build a high quality dataset, minimizing the number of noise or other errors (see \secref{sec:quality} of the main manuscript). Yet, \ourdataset~is not 100\% error free, and contains a few errors mostly due to erroneous Wikitext edits by the Wikipedia users. 

\paragraph{Is the dataset self-contained, or does it link to or otherwise rely on external resources?}
Yes, the dataset is self contained and consists of: \begin{enumerate}
    \item Instances divided in train, validation and test subsets (see \tabref{tab:dataset_detail2}).
    \item A description of all the entities of each of the Wikipedia snapshots. These entities form the complete candidate pool used by the models to predict the correct target entity. \Figref{fig:nr_wiki_entities} of the main manuscript illustrates the temporal evolution in size of the number of candidate entities. 
\end{enumerate}

\paragraph{Does the dataset contain data that might be considered confidential?} No, Wikipedia is a public resource. 

\paragraph{Does the dataset contain data that, if viewed directly, might be offensive, insulting, threatening, or might otherwise cause anxiety?} No, we haven't detected instances of such characteristics in \ourdataset. 

\paragraph{Does the dataset identify any subpopulations (e.g., by age, gender)?}  
While there are articles on different subpopulations on Wikipedia, there is no emphasis of the dataset on identifying or annotating those. 

\paragraph{Is it possible to identify individuals (i.e., one or more natural persons), either directly or indirectly (i.e., in combination with other data) from the dataset? } Only based on their Wikipedia article, no editor information is retained.

\paragraph{Does the dataset contain data that might be considered sensitive in any way? } Wikipedia is overall a resource aiming to be factual, therefore we can exclude this concern for most instances of \ourdataset.

\subsubsection{Collection process}
\paragraph{How was the data associated with each instance acquired? } The textual data of the context of anchor mention and that of the description of the target entity is directly taken from the Wikipedia snapshots. Conversely, the attributes associated with each of the instances are calculated (see \secref{app:mention_entity_attributes} for further details).

\paragraph{What mechanisms or procedures were used to collect the data
(e.g., hardware apparatuses or sensors, manual human curation,
software programs, software APIs)? }
The dataset was collected using the Wikipedia dumps from February of 2022. We detail further on the aspects related to the preprocessing, cleaning and labeling of \ourdataset~instances in \secref{app:sheet_preprocessing_cleaning} of the datasheet.  

\paragraph{Who was involved in the data collection process (e.g., students, crowdworkers, contractors) and how were they compensated (e.g., how much were crowdworkers paid)?} The dataset was automatically generated based on existing Wikipedia articles. Therefore, no human intervention was needed for the dataset generation.

\paragraph{Over what timeframe was the data collected? }
The \ourdataset~dataset was collected from 10 yearly snapshots of Wikipedia starting from 
January 1, 2013 until January 1, 2022.

\paragraph{Were any ethical review processes conducted (e.g., by an institutional review board)?} N/A 

\subsubsection{Preprocessing/cleaning/labeling}
\label{app:sheet_preprocessing_cleaning}
\paragraph{Was any preprocessing/cleaning/labeling of the data done (e.g.,
discretization or bucketing, tokenization, part-of-speech tagging,
SIFT feature extraction, removal of instances, processing of missing values)?}
The Wikipedia history logs content is available exclusively in Wikitext markup format.\footnote{\url{https://en.wikipedia.org/wiki/Help:Wikitext}} In order to obtain cleaned text we proceed as follows: 
\begin{enumerate}
    \item We use MediaWiki API to process the templates which can not be parsed using regular expressions. For example, this is the case of the Wikitext template \texttt{Convert}, where the markup like ``\texttt{\{\{convert|37|mm|in|abbr=on\}\}}'' is converted to ``\texttt{1.5 in}''.  
    \item We use regular expressions to extract mentions and links. While this can also be done using online Wikitext parsing tools, we found that these did not account for all the corner cases of mention parsing such as the ones involving the \textit{pipe trick}.\footnote{\url{https://en.wikipedia.org/wiki/Help:Pipe_trick}}
    \item Finally, we use \texttt{mwparserfromhell}\footnote{\url{https://github.com/earwig/mwparserfromhell}} tool for parsing the rest of the Wikitext content. 
\end{enumerate}
Furthermore, our dataset files also contain BERT tokenization of the context around the mentions as well as the textual content of entities.

\paragraph{Was the ``raw'' data saved in addition to the preprocessed/cleaned/labeled
data (e.g., to support unanticipated future uses)? } 
Yes, the raw data containing the Wikipedia history logs was saved on our cloud server in the following link: \url{https://cloud.ilabt.imec.be/index.php/s/BF9SkmQG2Tdjw8o}.

\paragraph{Is the software that was used to preprocess/clean/label the data
available?} Yes, the software will be made public upon acceptance. 

\subsubsection{Uses}
\paragraph{Has the dataset been used for any tasks already? } Yes, in our submitted manuscript we describe a retriever bi-encoder baseline \cite{wu2019zero} (see \secref{sec:results_and_analysis}). 

\paragraph{Is there a repository that links to any or all papers or systems that use the dataset? } N/A

\paragraph{What (other) tasks could the dataset be used for?} The covered task is temporally evolving entity linking. 

\paragraph{Is there anything about the composition of the dataset or the way it was collected and preprocessed/cleaned/labeled that might impact future uses? } N/A

\paragraph{ Are there tasks for which the dataset should not be used? } N/A

\paragraph{Will the dataset be distributed to third parties outside of the entity (e.g., company, institution, organization) on behalf of which
the dataset was created?} Yes, the dataset is of public access. 

\paragraph{How will the dataset be distributed (e.g., tarball on website,
API, GitHub)? } The \ourdataset~dataset will be made public on a GitHub repository together with the code to generate it. The baseline code and models will also be made public on the same repository. Due to the size, the dataset files will be hosted on the cloud server that belongs to Internet Technology and Data Science Lab (IDLab) at Ghent University (\url{https://cloud.ilabt.imec.be/index.php/s/RinXy8NgqdW58RW}). 

\paragraph{When will the dataset be distributed?} The dataset will be publicly distributed upon the submission of the camera ready version of our manuscript. 

\paragraph{Will the dataset be distributed under a copyright or other intellectual property (IP) license, and/or under applicable terms of use (ToU)? } The \ourdataset~dataset will be distributed under Creative Commons Attribution-ShareAlike 4.0 International license (CC BY-SA 4.0). 

\paragraph{Have any third parties imposed IP-based or other restrictions on the data associated with the instances?} N/A

\paragraph{Do any export controls or other regulatory restrictions apply to the dataset or to individual instances? } N/A

\subsubsection{Maintenance}
\paragraph{Who will be supporting/hosting/maintaining the dataset?}
The maintenance and extension of \ourdataset~will be carried out by the authors of the paper. Additionally, we will make the code publicly available to create potential new variations of \ourdataset~using a number of hyperparameters (see
\secref{app:hyperparameters} and \secref{app:dataset_extension} for further details). 

The dataset files will be hosted on the cloud server that belongs to Internet Technology and Data Science Lab (IDLab) at Ghent University (\url{https://cloud.ilabt.imec.be/index.php/s/RinXy8NgqdW58RW}).

\paragraph{How can the owner/curator/manager of the dataset be contacted (e.g., email address)?}
The owners of the dataset can be contacted at the following e-mail address: \url{klim.zaporojets@ugent.be}. 

\paragraph{Is there an erratum?} No, there is no erratum yet. 

\paragraph{Will the dataset be updated (e.g., to correct labeling errors, add new instances, delete instances)?}
The \ourdataset~will be regularly updated with newer snapshots (see \secref{app:dataset_extension}). 
In circumstances such as labeling errors, we will release the fixed version of the dataset with the respective version number. The introduction of the new version will be communicated using the \ourdataset~GitHub repository. 

\paragraph{If the dataset relates to people, are there applicable limits on the
retention of the data associated with the instances (e.g., were the
individuals in question told that their data would be retained for
a fixed period of time and then deleted)?} N/A

\paragraph{Will older versions of the dataset continue to be supported/hosted/maintained?}
Yes, the older version of the dataset will continue to be supported and hosted. All the versions will be numbered and we will provide the link to access each of these versions on our cloud storage server. 

\paragraph{If others want to extend/augment/build on/contribute to the
dataset, is there a mechanism for them to do so?}
Yes, we provide the code and functionality to re-generate and extend the dataset with new temporal snapshots (see Sections \ref{app:hyperparameters} and \ref{app:dataset_extension}). Yet, it is the responsibility of the users to provide hosting and maintenance to the newly generated dataset variations. 

\subsection{Mentions per entity distribution}
\label{app:mention_per_entity_distribution}
\begin{figure}[!t]
\centering
\includegraphics[width=1.0\columnwidth, trim={0.0cm 0.0cm 0.0cm 0.0cm},clip]{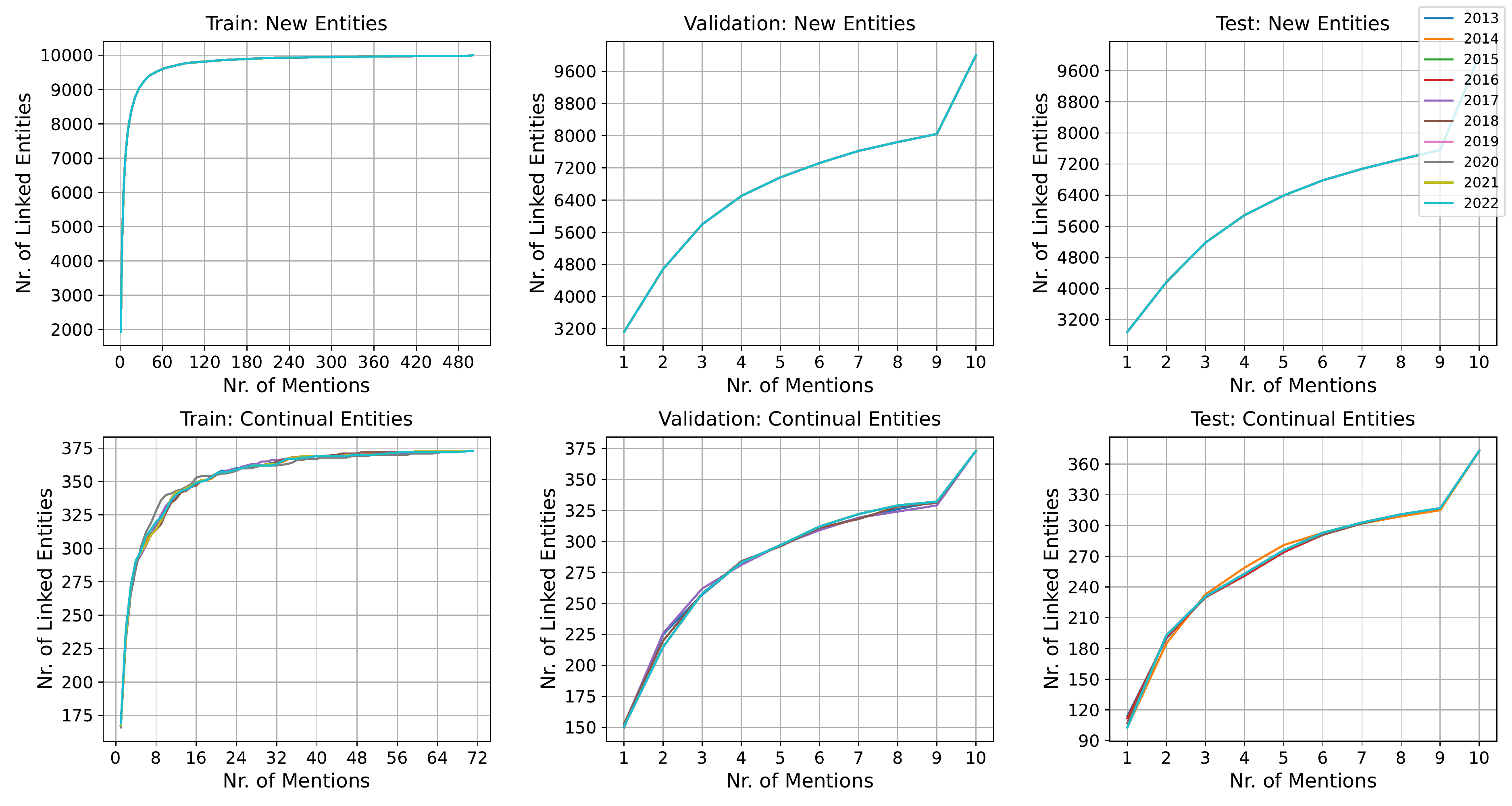}
\caption[Distribution of the data across the temporal snapshots]{Similar distribution of the data across the temporal snapshots (number of mentions per entity). 
This structurally unbiased setting enable to study exclusively the temporal effect
on the performance of the models for each of the different time periods. }
\label{fig:men_entity_distribution}
\end{figure}

\Figref{fig:men_entity_distribution} illustrates the similarity of mention per entity distribution across the temporal snapshots. This is achieved using weighted random subsampling so all the snapshots have equal number of instances (see \textit{Data Distributor} component description in \secref{sec:dataset_construction}). 
By enforcing this similarity between temporal snapshots, we ensure that the potential difference in the results is independent of cross-snapshot dataset distributional variations and only influenced by the dynamic temporal evolution of the content in \ourdataset. 
\subsection{Dataset creation hyperparameters}
\label{app:hyperparameters}
\begin{table}[t!]\small
\vspace{.75\baselineskip}
\centering
\caption[Hyperparameters of {\ourdataset} dataset creation.]{Hyperparameters that can be tuned during {\ourdataset} dataset creation. }
\vspace{.75\baselineskip}
\label{tab:hyperparameters}
{\renewcommand{\arraystretch}{1.2}
\begin{tabular}{p{0.24\linewidth}  p{0.55\linewidth} P{0.11\linewidth}}
        \toprule
         Hyperparamter & \multicolumn{1}{c}{Description} & \multicolumn{1}{c}{\ourdataset} \\ 
         \midrule 
         \texttt{snapshots} & Details (\eg timestamps) of the temporal snapshots to be generated. & 10 years \\
         \texttt{nr\_ct\_ents\_per\_cut} & Number of \textit{continual} entities per snapshot. & 10,000 \\
         \texttt{min\_mens\_per\_ent} & Minimum number of links a particular mention needs to have to target entity in order to be considered to be added in \ourdataset. & 1 \\
         \texttt{min\_men\_prior} & Minimum mention prior (see 
         \equref{eq:mention_prior} 
         in the main manuscript). & 0.0001 \\
         \texttt{max\_men\_prior} & Maximum mention prior. & 0.5 \\         
         \texttt{min\_men\_prior\_rank} & Minimum rank of mention prior among all the mentions pointing to a specific entity. & 2 \\
         \texttt{min\_ent\_prior} & Minimum entity prior as defined in \cite{yamada2016joint}: the ratio of links to the entity with respect to all of the links in the Wikipedia snapshot. & 0.0 \\
         \texttt{max\_ent\_prior} & Maximum entity prior. & 1.0 \\
         \texttt{min\_nr\_inlinks} & Minimum number of incoming links per entity. & 10 \\
         \texttt{min\_len\_target\_ent} & Minimum length of target entity page (in tokens). & 10 \\
         \texttt{max\_mens\_per\_ent} & Maximum number of mentions per entity. & 500/10/10\footnotemark \\
         \texttt{ed\_men\_title} & Minimum normalized edit distance between the mentions and the title of the target page they are linked to. & 0.2 \\
         \texttt{ed\_men\_subsets} & Minimum normalized edit distance between the mentions in different subsets linked to the same target entity.  & 0.2 \\
         \texttt{stable\_interval} & In seconds, the interval of time before the end of each snapshot from which the most stable version of Wikipedia has to be taken (see \secref{sec:quality} for further details).  & 2,592,000 (30 days) \\
         \revklim{\texttt{equal\_snapshots}} & \revklim{Whether the number of instances and the number of mentions per entity distribution is the same across the snapshots (see \secref{sec:quality} for further details). Equal cross-snapshot mention per entity distribution in \figref{fig:men_entity_distribution} is the result of setting this hyperparameter in True.}  & \revklim{True} \\         
         \bottomrule 
        \end{tabular}}
\end{table}
\footnotetext[5]{For train, validation and test sets respectively.}
\Tabref{tab:hyperparameters} summarizes the hyperparameters that can be tuned in order to automatically create the \ourdataset~dataset. This way, it is possible for the user to create different variation of the \ourdataset. The most relevant hyperparameter is \texttt{snapshots} that is used to specify the temporal intervals to create the snapshots. Below we detail two possible options we provide to specify such intervals. 
\paragraph{Option 1 - explicit snapshot specification} The user is expected to provide a list of timestamps in the format of \texttt{YYYY-MM-DDTHH:MM:SSZ}, each one defining a different snapshot. 

\paragraph{Option 2 - time span and interval} This option enables the user to define start and end dates of the time span from which the snapshots should be extracted. Furthermore, the interval value (\ie by using keywords such as ``weekly'' or specifying the interval in seconds)  has to also be specified. 

\subsection{Dataset extension}
\label{app:dataset_extension}
Additionally, we provide the option to extend the already existing dataset with new snapshots. Similarly as in the creation of new dataset (see \secref{app:hyperparameters} above), the \texttt{snapshots} hyperparameter is used to specify new snapshots which are then added 
to already existing \ourdataset~dataset. 

\subsection{Mention and entity attributes}
\label{app:mention_entity_attributes}
\clearpage
\begin{table}[t!]\small
\centering
\caption[Mention-entity attributes in {\ourdataset}]{Attributes associated to each of the mention-entity pairs for each of the temporal snapshots in \ourdataset. }
\vspace{.75\baselineskip}
\label{tab:attributes}
{\renewcommand{\arraystretch}{1.2}
\begin{tabular}{p{0.26\linewidth}  p{0.68\linewidth}}
        \toprule
         Attribute & \multicolumn{1}{c}{Description} \\ 
         \midrule 
         \texttt{subset} & The name of current subset (\ie train, validation or test).\\         
         \texttt{target\_page\_id} & The unique Wikipedia page id of the target entity. \\
        \texttt{target\_qid} & The unique Wikidata QID of the target entity. \\
         \texttt{snapshot} & The timestamp of the temporal snapshot from which the anchor mention and target entity attributes were extracted. \\
         \texttt{target} & The textual content of the target entity Wikipedia page. \\
         \texttt{target\_len} & The length in tokens of target Wikipedia page. \\         
         \texttt{target\_title} & The title of target entity Wikipedia page. \\
         \texttt{category} & Category of the target entity (\textit{new} or \textit{continual}). \\         
         \texttt{mention} & The text of the mention. \\
         \texttt{context\_left} & The textual context to the left of the mention. \\
         \texttt{context\_right} & The textual context to the right of the mention. \\
         \texttt{anchor\_len} & The length in tokens of the Wikipedia page where the anchor mention is located. \\
         \texttt{ed\_men\_title} & Normalized edit distance between the anchor mention and the title of the target Wikipedia page.  \\
         \texttt{overlap\_type} & Overlap type between the anchor mention and the target title as defined by \cite{logeswaran2019zero}. \\
         \texttt{men\_prior} & The mention prior (see \equref{eq:mention_prior} of the main manuscript). \\
         \texttt{men\_prior\_rank} & The rank of the current anchor mention compared to other mentions in Wikipedia pointing to target entity. \\
         \texttt{avg\_men\_prior} & The average value of prior of the mentions linked to the target entity in Wikipedia for \texttt{snapshot}. \\
         \texttt{ent\_prior} & Entity prior as defined in \cite{yamada2016joint}: the ratio of links to the entity with respect to all of the links in the Wikipedia snapshot. \\
         \texttt{nr\_inlinks} & Total number of incoming links to target entity. \\
         \texttt{nr\_dist\_mens} & Number of distinct (\ie with different surface form) mentions linked to target entity. \\
         \texttt{nr\_mens\_per\_ent} & Number of times the current mention appears in Wikipedia linked to target entity. \\ 
         \texttt{nr\_mens\_extracted} & Number of anchor mentions per current target entity in the \texttt{subset}. \\
         \texttt{anchor\_creation\_date} & The creation date (timestamp) of Wikipedia page where the anchor mention is located. \\
         \texttt{anchor\_revision\_date} & The timestamp of when the anchor Wikipedia page was last revised. \\
         \texttt{target\_creation\_date} & The timestamp of when the target Wikipedia entity page was created.  \\
         \texttt{target\_revision\_date} & The timestamp of when the target Wikipedia entity page was last modified. \\
         \bottomrule 
        \end{tabular}}
\end{table}
\clearpage
\Tabref{tab:attributes} describes the anchor mention and target entity related attributes present in \ourdataset. These attributes can be used to perform more in-depth analysis of the results.

\subsection{Baseline implementation details}
\label{app:training_hyperparameters}
We base our bi-encoder baseline model on the publicly available BLINK code.\footnote{\url{https://github.com/facebookresearch/BLINK}} We train all the models for 10 epochs with the learning rate of 1e-04 and the batch size of 64. We use AdamW optimizer with 10\% of warmup steps. Finally, we rely on \texttt{transformers} library \cite{wolf2020transformers} to get the pre-trained BERT-large representations.    All the experiments were run on NVIDIA V100 GPU with the following execution times: 
\begin{enumerate}
    \item \emph{Training}: 
    36 hours to train for 10 epochs per single snapshot.
    \item \emph{All Wikipedia entity encoding}: 7 days
    per finetuned model (on all the 10 Wikipedia snapshots) running on a single V100 GPU.  
    \item \emph{Evaluation}: 30 seconds per finetuned model per snapshot using FAISS \cite{johnson2019billion} library on GPU. 
\end{enumerate}

\subsection{Total amount of compute and the type of resources used to create \ourdataset}
In this section we provide the details on the computational resources used in each of the processing steps (see \secref{sec:dataset_construction} and  \figref{fig:el_pipeline} for further details) to create the \ourdataset~dataset: 

\begin{enumerate}
    \item \emph{Snapshot Data Extraction}: this processing step is responsible for creating the snapshots from the Wikipedia log files from February 1, 2022. This is a multi-processing step that is executed on a cluster with 80 CPUs and 110 GB of RAM and takes 5 days and 8 hours to complete. 
    \item \emph{Snapshot Dataset Building}: this is a multi-processing step that is executed on a cluster with 30 CPUs and 250 GB of RAM and takes 5 hours to complete. 
\end{enumerate}

\subsection{License of the assets}
We base the implementation of our baseline bi-encoder model on the publicly available BLINK \cite{wu2019zero} code. This asset is made available under MIT License (\url{https://opensource.org/licenses/MIT}).
\subsection{Examples}
\label{app:examples}
This section presents two illustrative examples of instances in \ourdataset. The first example contains the anchor mention linked to \textit{continual} entity, while the second one is the example of a link to \textit{new} entity. Both of the examples were taken from the snapshot of January 1, 2021. Furthermore, we trim the content length (\eg \texttt{target} attribute value) to only a few tokens for space reasons.

\subsubsection{Example 1: continual target entity}
\label{app:example1_continual}
\Tabref{tab:example1_continual} illustrates an example of the link to \textit{continual} target entity \textit{Sacramental\_bread}. It is worth noting that the creation date of this entity in Wikipedia (\texttt{target\_creation\_date} attribute) is of January 3, 2005. Yet, the version saved in the snapshot (\texttt{target\_revision\_date} attribute) is from December 30, 2020. \\ 
\clearpage
\begin{table}[t!]
\centering
\caption[\textit{Continual} entity example from {\ourdataset}]{Example of the instance corresponding to mention link to \textit{continual} entity (Sacramental\_bread created in 2005-01-03) in \ourdataset. }
\vspace{.75\baselineskip}
\label{tab:example1_continual}
{\renewcommand{\arraystretch}{1.2}
\begin{tabular}{p{0.26\linewidth}  p{0.68\linewidth}}
        \toprule
         Attribute & \multicolumn{1}{c}{Value} \\ 
         \midrule 
         \texttt{subset} & train \\
         \texttt{target\_page\_id} & 1359030 \\
         \texttt{target\_qid} & Q207104 \\
         \texttt{snapshot} & 2021-01-01T00:00:00Z \\
         \texttt{target} & ``Sacramental bread, sometimes called altar bread, Communion ...'' \\
         \texttt{target\_len} & 7,568 \\         
         \texttt{target\_title} & ``Sacramental\_bread''. \\
         \texttt{category} & continual \\         
         \texttt{mention} & ``host'' \\
         \texttt{context\_left} & ``... devotional image, portrait or other religious symbol (such as the'' \\
         \texttt{context\_right} & ``). Garland paintings were typically collaborations between a ...'' \\
         \texttt{anchor\_len} & 6,519 \\
         \texttt{ed\_men\_title} & 0.9411  \\
         \texttt{overlap\_type} & LOW\_OVERLAP \\
         \texttt{men\_prior} & 0.0750 \\
         \texttt{men\_prior\_rank} & 7 \\
         \texttt{avg\_men\_prior} & 0.6864 \\
         \texttt{ent\_prior} & 1.7790e-6 \\
         \texttt{nr\_inlinks} & 225 \\
         \texttt{nr\_dist\_mens} & 13 \\
         \texttt{nr\_mens\_per\_ent} & 79 \\ 
         \texttt{nr\_mens\_extracted} & 58 \\
         \texttt{anchor\_creation\_date} & 2009-09-25T21:09:07Z \\
         \texttt{anchor\_revision\_date} & 2020-10-04T16:15:13Z\\
         \texttt{target\_creation\_date} & 2005-01-03T17:41:14Z \\
         \texttt{target\_revision\_date} & 2020-12-30T12:38:50Z \\
         \bottomrule 
        \end{tabular}}
\end{table}
\clearpage
\subsubsection{Example 2: new target entity}
\label{app:example2_new_entity}
\Tabref{tab:example2_new} illustrates an example of the link to \textit{new} target entity \textit{COVID-19\_pandemic\_in\_Portland,\_Oregon}. It is worth noting that the creation date of this entity in Wikipedia (\texttt{target\_creation\_date} attribute) is of March 23, 2020, which belongs to the interval of the considered snapshot: from January 1, 2020 until January 1, 2021. \\ 
\begin{table}[t!]
\centering
\caption[\textit{New} entity example from {\ourdataset}]{Example of the instance corresponding to mention link to \textit{new} entity (COVID-19\_pandemic\_in\_Portland,\_Oregon created in 2020-03-23) in \ourdataset. }
\vspace{.75\baselineskip}
\label{tab:example2_new}
{\renewcommand{\arraystretch}{1.2}
\begin{tabular}{p{0.26\linewidth}  p{0.68\linewidth}}
        \toprule
         Attribute & \multicolumn{1}{c}{Value} \\ 
         \midrule 
         \texttt{subset} & train \\
         \texttt{target\_page\_id} & 63449958 \\
         \texttt{target\_qid} & Q88484856 \\
         \texttt{snapshot} & 2021-01-01T00:00:00Z \\
         \texttt{target} & ``The COVID-19 pandemic was confirmed to have reached ...'' \\
         \texttt{target\_len} & 26,432 \\         
         \texttt{target\_title} & ``COVID-19\_pandemic\_in\_Portland,\_Oregon'' \\
         \texttt{category} & new \\         
         \texttt{mention} & ``COVID-19 pandemic'' \\
         \texttt{context\_left} & ``Xico Xico and Xica both offered pickup service during the'' \\
         \texttt{context\_right} & ``, as of May 2020. '' \\
         \texttt{anchor\_len} & 2,437 \\
         \texttt{ed\_men\_title} & 0.5405  \\
         \texttt{overlap\_type} & AMBIGUOUS\_SUBSTRING \\
         \texttt{men\_prior} & 0.0009 \\
         \texttt{men\_prior\_rank} & 4 \\
         \texttt{avg\_men\_prior} & 0.2548 \\
         \texttt{ent\_prior} & 2.9255e-7 \\
         \texttt{nr\_inlinks} & 37 \\
         \texttt{nr\_dist\_mens} & 3 \\
         \texttt{nr\_mens\_per\_ent} & 23 \\ 
         \texttt{nr\_mens\_extracted} & 18 \\
         \texttt{anchor\_creation\_date} & 2020-12-08T00:23:50Z \\
         \texttt{anchor\_revision\_date} & 2020-12-09T15:41:18Z\\
         \texttt{target\_creation\_date} & 2020-03-23T04:22:55Z \\
         \texttt{target\_revision\_date} & 2020-11-16T03:59:06Z \\
         \bottomrule 
        \end{tabular}}
\end{table}

\subsection{Additional results}
\label{app:additional_results}
\begin{figure}[!t]
\centering
\includegraphics[width=0.6\columnwidth]{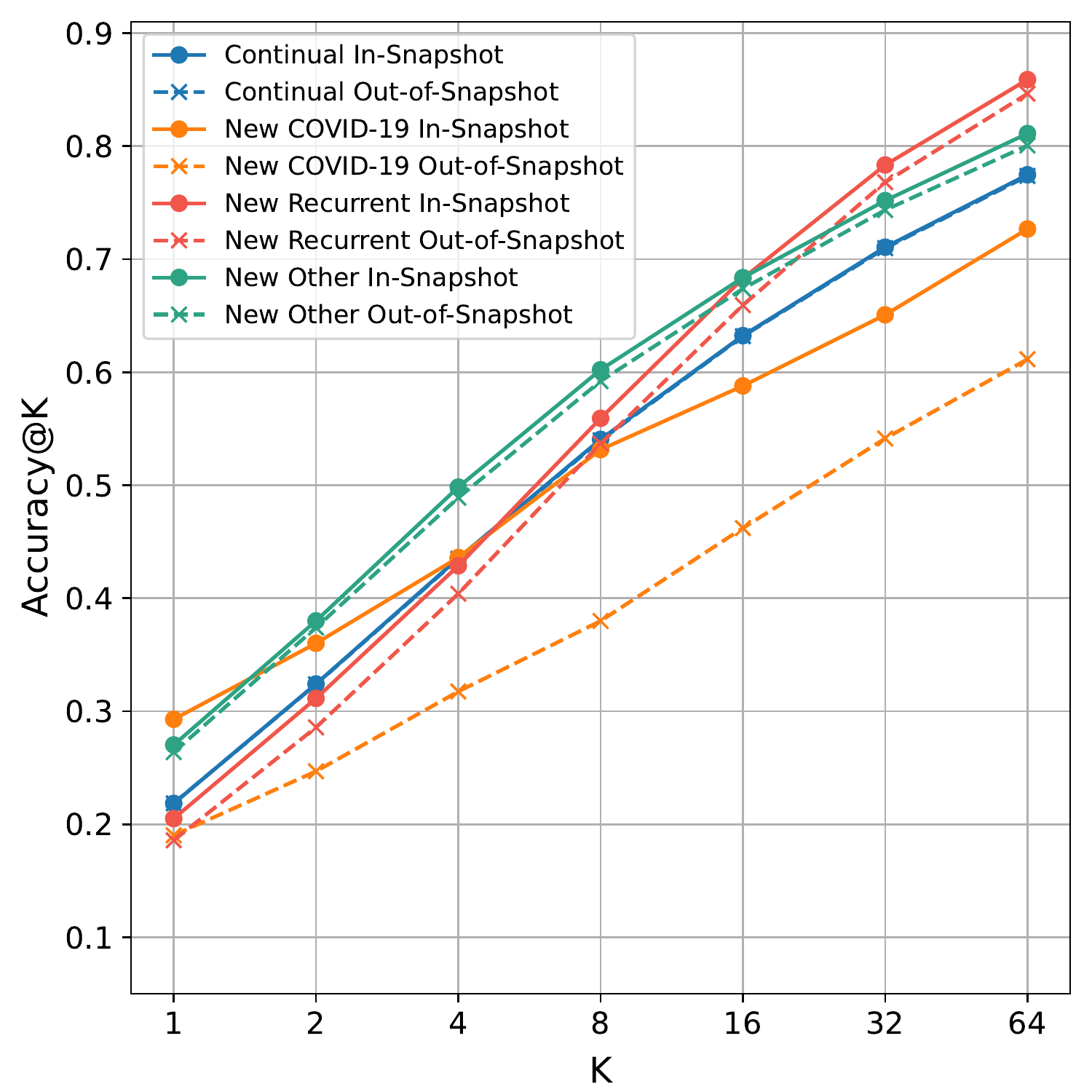}
\caption[Accuraccy@$K$ for different types of new entities]{\revklim{Accuraccy@$K$ for different values of $K \in \{1,2,4,8,16,32,64\}$. The results are grouped in four main categories: \begin{enumerate*}[(i)]
    \item mentions linked to \textit{continual} entities that exist in all of the \ourdataset~snapshots, 
    \item mentions linked to \textit{COVID-related} new entities (\ie with keywords such as ``COVID'' in target entity title), 
    \item mentions linked to \textit{recurrent} new entities (\ie entities representing events occurring periodically such as ``2018 BNP Paribas Open''), and 
    \item mentions linked to \textit{other} new entities. 
\end{enumerate*}}}
\label{fig:new_entity_types}
\end{figure}
Tables \ref{tab:general_table_shared_k_in_1}-\ref{tab:general_table_shared_k_in_64} present the results for different accuracy@$K$ for $K \in \{1,2,4,8,16,32,64\}$. 
\revklim{Furthermore, \figref{fig:new_entity_types} 
illustrates
the mean in- and out-of-snapshot (see \secref{sec:results_and_analysis} of the main manuscript) accuracy@$K$ performance across temporal snapshots on the following four target entity categories:}
\begin{enumerate}
    \item \revklim{\textit{Continual}: all the target \textit{continual} entities (\ie the entities that exist across all the temporal snapshots in \ourdataset~dataset).} 
    \item \revklim{\textit{COVID-19}: target \textit{new} entities that have COVID-related (\eg ``COVID'', ``coronavirus'', etc.) terms in the target entity title.}
    \item \revklim{\textit{Recurrent}: target \textit{new} entities whose titles contain the year and some of the keywords (\eg ``league'', ``election'', ``cup'', etc.) that indicate that an entity \revklim{is a repetitive event}
    (\eg ``2018 BNP Paribas Open'' which is part of \textit{yearly} BNB Paribas Open competitions).}
    \item \revklim{\textit{Other}: all the other target \textit{new} entities.}
\end{enumerate}
\revklim{The following are the main conclusions that can be drawn from the graph in \figref{fig:new_entity_types} that support or complement the findings described in \secref{sec:results_and_analysis} of the main manuscript:} 
\begin{enumerate}
    \item \revklim{New entities that require fundamentally new, previously non-existent knowledge to be disambiguated tend to have the lowest out-of-snapshot performance. This is the case of COVID-19 related disambiguation instances. These instances also experience the highest boost in performance when evaluated on in-snapshot setting (\ie the model is evaluated and finetuned on the same temporal snapshot).} 
    \item \revklim{The difference between in- and out-of-snapshot performances on \textit{continual} entities is the lowest. This is also supported by \figref{fig:avg_performance} and \figsref{fig:in-snapshot-k}{fig:in-snapshot-offset} in the main manuscript. This suggests that the actual knowledge needed to disambiguate most of the \textit{continual} entities in {\ourdataset} changes very little with time.}
    \item \revklim{The model has the highest accuracy@64 performance on \textit{recurrent} new entities. Yet, the performance on these entities drops sharply for lower values of $K$. We hypothesize that predicting the correct recurrent event gets more challenging as $K$ decreases because 
    of the large number of
    very similar candidates to pick from (\eg many ``BNP Paribas Open'' championships that only differ in very few details such as the date).}
    \item \revklim{The difference between in- and out-of-snapshot performance for \textit{other} new entities is lower than for \textit{recurrent} and \textit{COVID-19} related ones. This is driven by new entities that are derived from existing entities in Wikipedia (\ie their content is a copy of already 
    established
    entities). We hypothesize that the model requires little additional knowledge to disambiguate these entities. 
    Still,
    it is part of future work to study \textit{other} new entities more in detail in order to find cases that represent intrinsically new knowledge similar to the identified COVID-19 entity cluster.}
\end{enumerate}

\clearpage
\begin{table}[t] 
\caption[Accuracy@1 results]{\textbf{Accuracy@1} for \emph{continual} (top) and \emph{new} (bottom) entities. The intensity of colors is set on a row-by-row basis and indicates whether performance is \textcolor{darkspringgreen!100}{\textbf{better}} or \textcolor{deepcarmine!100}{\textbf{worse}} compared to the year the model was finetuned on (\ie the values that form the white diagonal).} 
\vspace{.75\baselineskip}
\label{tab:general_table_shared_k_in_1} 
\centering 
\resizebox{\columnwidth}{!} 
{
\begin{tabular}{c cccccccccc} 
         \toprule 
         \multicolumn{11}{c}{Continual Entities} \\ 
         \midrule 
         \backslashbox{\textbf{Train}}{\textbf{Test}}& 2013 & 2014 & 2015 & 2016 & 2017 & 2018 & 2019 & 2020 & 2021 & 2022 \\ 
        \midrule 
2013  & 0.225 & \cellcolor{deepcarmine!21}0.219 & \cellcolor{deepcarmine!29}0.215 & \cellcolor{deepcarmine!26}0.217 & \cellcolor{deepcarmine!35}0.212 & \cellcolor{deepcarmine!46}0.206 & \cellcolor{deepcarmine!54}0.203 & \cellcolor{deepcarmine!54}0.203 & \cellcolor{deepcarmine!66}0.197 & \cellcolor{deepcarmine!75}0.192 \\ 
2014  & \cellcolor{darkspringgreen!17}0.229 & 0.226 & \cellcolor{deepcarmine!23}0.220 & \cellcolor{deepcarmine!19}0.221 & \cellcolor{deepcarmine!30}0.217 & \cellcolor{deepcarmine!40}0.212 & \cellcolor{deepcarmine!43}0.211 & \cellcolor{deepcarmine!52}0.207 & \cellcolor{deepcarmine!62}0.203 & \cellcolor{deepcarmine!75}0.197 \\ 
2015  & \cellcolor{darkspringgreen!37}0.228 & \cellcolor{darkspringgreen!22}0.223 & 0.219 & \cellcolor{darkspringgreen!10}0.219 & \cellcolor{deepcarmine!17}0.216 & \cellcolor{deepcarmine!32}0.211 & \cellcolor{deepcarmine!42}0.208 & \cellcolor{deepcarmine!46}0.206 & \cellcolor{deepcarmine!53}0.204 & \cellcolor{deepcarmine!75}0.196 \\ 
2016  & \cellcolor{darkspringgreen!37}0.230 & \cellcolor{darkspringgreen!26}0.227 & \cellcolor{darkspringgreen!13}0.222 & 0.221 & \cellcolor{deepcarmine!18}0.218 & \cellcolor{deepcarmine!30}0.214 & \cellcolor{deepcarmine!39}0.211 & \cellcolor{deepcarmine!48}0.208 & \cellcolor{deepcarmine!56}0.205 & \cellcolor{deepcarmine!75}0.199 \\ 
2017  & \cellcolor{darkspringgreen!58}0.240 & \cellcolor{darkspringgreen!46}0.237 & \cellcolor{darkspringgreen!21}0.229 & \cellcolor{darkspringgreen!20}0.229 & 0.226 & \cellcolor{deepcarmine!26}0.221 & \cellcolor{deepcarmine!34}0.219 & \cellcolor{deepcarmine!43}0.216 & \cellcolor{deepcarmine!60}0.211 & \cellcolor{deepcarmine!75}0.207 \\
\bottomrule 
        \end{tabular}} 
\end{table}
\begin{table}[t] 
\caption[Accuracy@2 results]{\textbf{Accuracy@2} for \emph{continual} (top) and \emph{new} (bottom) entities. The intensity of colors is set on a row-by-row basis and indicates whether performance is \textcolor{darkspringgreen!100}{\textbf{better}} or \textcolor{deepcarmine!100}{\textbf{worse}} compared to the year the model was finetuned on (\ie the values that form the white diagonal).} 
\vspace{.75\baselineskip}
\label{tab:general_table_shared_k_in_2} 
\centering 
\resizebox{\columnwidth}{!} 
{\begin{tabular}{c cccccccccc} 
         \toprule 
         \multicolumn{11}{c}{Continual Entities} \\ 
         \midrule 
         \backslashbox{\textbf{Train}}{\textbf{Test}}& 2013 & 2014 & 2015 & 2016 & 2017 & 2018 & 2019 & 2020 & 2021 & 2022 \\ 
        \midrule 
2013  & 0.337 & \cellcolor{deepcarmine!20}0.330 & \cellcolor{deepcarmine!28}0.324 & \cellcolor{deepcarmine!31}0.322 & \cellcolor{deepcarmine!39}0.317 & \cellcolor{deepcarmine!49}0.311 & \cellcolor{deepcarmine!56}0.306 & \cellcolor{deepcarmine!61}0.302 & \cellcolor{deepcarmine!63}0.301 & \cellcolor{deepcarmine!75}0.293 \\ 
2014  & \cellcolor{darkspringgreen!17}0.339 & 0.335 & \cellcolor{deepcarmine!20}0.329 & \cellcolor{deepcarmine!21}0.328 & \cellcolor{deepcarmine!32}0.322 & \cellcolor{deepcarmine!41}0.317 & \cellcolor{deepcarmine!47}0.314 & \cellcolor{deepcarmine!54}0.310 & \cellcolor{deepcarmine!62}0.306 & \cellcolor{deepcarmine!75}0.299 \\ 
2015  & \cellcolor{darkspringgreen!36}0.339 & \cellcolor{darkspringgreen!23}0.333 & 0.327 & \cellcolor{deepcarmine!14}0.325 & \cellcolor{deepcarmine!20}0.323 & \cellcolor{deepcarmine!34}0.317 & \cellcolor{deepcarmine!46}0.312 & \cellcolor{deepcarmine!52}0.309 & \cellcolor{deepcarmine!60}0.305 & \cellcolor{deepcarmine!75}0.299 \\ 
2016  & \cellcolor{darkspringgreen!50}0.341 & \cellcolor{darkspringgreen!33}0.334 & \cellcolor{darkspringgreen!15}0.328 & 0.326 & \cellcolor{deepcarmine!19}0.322 & \cellcolor{deepcarmine!35}0.316 & \cellcolor{deepcarmine!40}0.314 & \cellcolor{deepcarmine!49}0.310 & \cellcolor{deepcarmine!60}0.306 & \cellcolor{deepcarmine!75}0.301 \\ 
2017  & \cellcolor{darkspringgreen!62}0.351 & \cellcolor{darkspringgreen!48}0.346 & \cellcolor{darkspringgreen!28}0.338 & \cellcolor{darkspringgreen!26}0.338 & 0.332 & \cellcolor{deepcarmine!21}0.328 & \cellcolor{deepcarmine!32}0.324 & \cellcolor{deepcarmine!43}0.320 & \cellcolor{deepcarmine!55}0.316 & \cellcolor{deepcarmine!75}0.309 \\
\bottomrule 
        \end{tabular}} 
\end{table}
\begin{table}[t] 
\caption[Accuracy@4 results]{\textbf{Accuracy@4} for \emph{continual} (top) and \emph{new} (bottom) entities. The intensity of colors is set on a row-by-row basis and indicates whether performance is \textcolor{darkspringgreen!100}{\textbf{better}} or \textcolor{deepcarmine!100}{\textbf{worse}} compared to the year the model was finetuned on (\ie the values that form the white diagonal).} 
\vspace{.75\baselineskip}
\label{tab:general_table_shared_k_in_4} 
\centering 
\resizebox{\columnwidth}{!} 
{\begin{tabular}{c cccccccccc} 
         \toprule 
         \multicolumn{11}{c}{Continual Entities} \\ 
         \midrule 
         \backslashbox{\textbf{Train}}{\textbf{Test}}& 2013 & 2014 & 2015 & 2016 & 2017 & 2018 & 2019 & 2020 & 2021 & 2022 \\ 
        \midrule 
2013  & 0.449 & \cellcolor{deepcarmine!19}0.442 & \cellcolor{deepcarmine!24}0.439 & \cellcolor{deepcarmine!33}0.433 & \cellcolor{deepcarmine!41}0.428 & \cellcolor{deepcarmine!49}0.422 & \cellcolor{deepcarmine!57}0.417 & \cellcolor{deepcarmine!57}0.417 & \cellcolor{deepcarmine!66}0.411 & \cellcolor{deepcarmine!75}0.405 \\ 
2014  & \cellcolor{darkspringgreen!21}0.455 & 0.448 & \cellcolor{deepcarmine!19}0.443 & \cellcolor{deepcarmine!25}0.439 & \cellcolor{deepcarmine!36}0.433 & \cellcolor{deepcarmine!43}0.428 & \cellcolor{deepcarmine!51}0.424 & \cellcolor{deepcarmine!52}0.423 & \cellcolor{deepcarmine!63}0.416 & \cellcolor{deepcarmine!75}0.410 \\ 
2015  & \cellcolor{darkspringgreen!29}0.455 & \cellcolor{darkspringgreen!14}0.446 & 0.444 & \cellcolor{deepcarmine!21}0.438 & \cellcolor{deepcarmine!29}0.434 & \cellcolor{deepcarmine!41}0.427 & \cellcolor{deepcarmine!50}0.422 & \cellcolor{deepcarmine!50}0.422 & \cellcolor{deepcarmine!63}0.415 & \cellcolor{deepcarmine!75}0.409 \\ 
2016  & \cellcolor{darkspringgreen!46}0.453 & \cellcolor{darkspringgreen!30}0.446 & \cellcolor{darkspringgreen!21}0.442 & 0.437 & \cellcolor{deepcarmine!21}0.432 & \cellcolor{deepcarmine!35}0.426 & \cellcolor{deepcarmine!43}0.422 & \cellcolor{deepcarmine!44}0.422 & \cellcolor{deepcarmine!59}0.415 & \cellcolor{deepcarmine!75}0.408 \\ 
2017  & \cellcolor{darkspringgreen!75}0.464 & \cellcolor{darkspringgreen!58}0.458 & \cellcolor{darkspringgreen!46}0.454 & \cellcolor{darkspringgreen!26}0.448 & 0.443 & \cellcolor{deepcarmine!25}0.438 & \cellcolor{deepcarmine!38}0.434 & \cellcolor{deepcarmine!40}0.433 & \cellcolor{deepcarmine!55}0.428 & \cellcolor{deepcarmine!72}0.423 \\
\bottomrule 
        \end{tabular}} 
\end{table}
\begin{table}[t] 
\caption[Accuracy@8 results]{\textbf{Accuracy@8} for \emph{continual} (top) and \emph{new} (bottom) entities. The intensity of colors is set on a row-by-row basis and indicates whether performance is \textcolor{darkspringgreen!100}{\textbf{better}} or \textcolor{deepcarmine!100}{\textbf{worse}} compared to the year the model was finetuned on (\ie the values that form the white diagonal).} 
\vspace{.75\baselineskip}
\label{tab:general_table_shared_k_in_8} 
\centering 
\resizebox{\columnwidth}{!} 
{\begin{tabular}{c cccccccccc} 
         \toprule 
         \multicolumn{11}{c}{Continual Entities} \\ 
         \midrule 
         \backslashbox{\textbf{Train}}{\textbf{Test}}& 2013 & 2014 & 2015 & 2016 & 2017 & 2018 & 2019 & 2020 & 2021 & 2022 \\ 
        \midrule 
2013  & 0.556 & \cellcolor{deepcarmine!17}0.551 & \cellcolor{deepcarmine!23}0.546 & \cellcolor{deepcarmine!32}0.539 & \cellcolor{deepcarmine!41}0.532 & \cellcolor{deepcarmine!49}0.526 & \cellcolor{deepcarmine!56}0.520 & \cellcolor{deepcarmine!57}0.520 & \cellcolor{deepcarmine!66}0.513 & \cellcolor{deepcarmine!75}0.507 \\ 
2014  & \cellcolor{darkspringgreen!16}0.563 & 0.559 & \cellcolor{deepcarmine!18}0.553 & \cellcolor{deepcarmine!29}0.546 & \cellcolor{deepcarmine!36}0.540 & \cellcolor{deepcarmine!48}0.532 & \cellcolor{deepcarmine!56}0.527 & \cellcolor{deepcarmine!58}0.526 & \cellcolor{deepcarmine!65}0.520 & \cellcolor{deepcarmine!75}0.514 \\ 
2015  & \cellcolor{darkspringgreen!26}0.561 & \cellcolor{darkspringgreen!16}0.555 & 0.552 & \cellcolor{deepcarmine!24}0.543 & \cellcolor{deepcarmine!29}0.540 & \cellcolor{deepcarmine!42}0.533 & \cellcolor{deepcarmine!54}0.526 & \cellcolor{deepcarmine!54}0.526 & \cellcolor{deepcarmine!63}0.520 & \cellcolor{deepcarmine!75}0.514 \\ 
2016  & \cellcolor{darkspringgreen!47}0.559 & \cellcolor{darkspringgreen!35}0.554 & \cellcolor{darkspringgreen!28}0.550 & 0.542 & \cellcolor{deepcarmine!20}0.537 & \cellcolor{deepcarmine!34}0.531 & \cellcolor{deepcarmine!48}0.524 & \cellcolor{deepcarmine!48}0.524 & \cellcolor{deepcarmine!61}0.518 & \cellcolor{deepcarmine!75}0.511 \\ 
2017  & \cellcolor{darkspringgreen!61}0.569 & \cellcolor{darkspringgreen!51}0.565 & \cellcolor{darkspringgreen!42}0.562 & \cellcolor{darkspringgreen!23}0.555 & 0.549 & \cellcolor{deepcarmine!30}0.542 & \cellcolor{deepcarmine!43}0.537 & \cellcolor{deepcarmine!43}0.537 & \cellcolor{deepcarmine!60}0.530 & \cellcolor{deepcarmine!75}0.525 \\
\bottomrule 
        \end{tabular}} 
\end{table}
\begin{table}[t] 
\caption[Accuracy@16 results]{\textbf{Accuracy@16} for \emph{continual} (top) and \emph{new} (bottom) entities. The intensity of colors is set on a row-by-row basis and indicates whether performance is \textcolor{darkspringgreen!100}{\textbf{better}} or \textcolor{deepcarmine!100}{\textbf{worse}} compared to the year the model was finetuned on (\ie the values that form the white diagonal).} 
\vspace{.75\baselineskip}
\label{tab:general_table_shared_k_in_16} 
\centering 
\resizebox{\columnwidth}{!} 
{\begin{tabular}{c cccccccccc} 
         \toprule 
         \multicolumn{11}{c}{Continual Entities} \\ 
         \midrule 
         \backslashbox{\textbf{Train}}{\textbf{Test}}& 2013 & 2014 & 2015 & 2016 & 2017 & 2018 & 2019 & 2020 & 2021 & 2022 \\ 
        \midrule 
2013  & 0.648 & \cellcolor{deepcarmine!17}0.643 & \cellcolor{deepcarmine!21}0.639 & \cellcolor{deepcarmine!32}0.632 & \cellcolor{deepcarmine!40}0.626 & \cellcolor{deepcarmine!51}0.617 & \cellcolor{deepcarmine!58}0.613 & \cellcolor{deepcarmine!58}0.613 & \cellcolor{deepcarmine!68}0.605 & \cellcolor{deepcarmine!75}0.600 \\ 
2014  & \cellcolor{darkspringgreen!19}0.657 & 0.650 & \cellcolor{deepcarmine!15}0.647 & \cellcolor{deepcarmine!26}0.639 & \cellcolor{deepcarmine!32}0.635 & \cellcolor{deepcarmine!46}0.627 & \cellcolor{deepcarmine!53}0.622 & \cellcolor{deepcarmine!55}0.620 & \cellcolor{deepcarmine!67}0.613 & \cellcolor{deepcarmine!75}0.608 \\ 
2015  & \cellcolor{darkspringgreen!25}0.651 & \cellcolor{darkspringgreen!15}0.645 & 0.642 & \cellcolor{deepcarmine!21}0.636 & \cellcolor{deepcarmine!28}0.633 & \cellcolor{deepcarmine!45}0.624 & \cellcolor{deepcarmine!53}0.619 & \cellcolor{deepcarmine!54}0.619 & \cellcolor{deepcarmine!66}0.612 & \cellcolor{deepcarmine!75}0.608 \\ 
2016  & \cellcolor{darkspringgreen!41}0.652 & \cellcolor{darkspringgreen!30}0.646 & \cellcolor{darkspringgreen!23}0.643 & 0.637 & \cellcolor{deepcarmine!22}0.631 & \cellcolor{deepcarmine!41}0.621 & \cellcolor{deepcarmine!52}0.616 & \cellcolor{deepcarmine!54}0.615 & \cellcolor{deepcarmine!64}0.610 & \cellcolor{deepcarmine!75}0.605 \\ 
2017  & \cellcolor{darkspringgreen!71}0.660 & \cellcolor{darkspringgreen!54}0.655 & \cellcolor{darkspringgreen!46}0.652 & \cellcolor{darkspringgreen!28}0.646 & 0.640 & \cellcolor{deepcarmine!30}0.633 & \cellcolor{deepcarmine!45}0.628 & \cellcolor{deepcarmine!47}0.628 & \cellcolor{deepcarmine!66}0.621 & \cellcolor{deepcarmine!75}0.618 \\
\bottomrule 
        \end{tabular}} 
\end{table}
\begin{table}[t] 
\caption[Accuracy@32 results]{\textbf{Accuracy@32} for \emph{continual} (top) and \emph{new} (bottom) entities. The intensity of colors is set on a row-by-row basis and indicates whether performance is \textcolor{darkspringgreen!100}{\textbf{better}} or \textcolor{deepcarmine!100}{\textbf{worse}} compared to the year the model was finetuned on (\ie the values that form the white diagonal).} 
\vspace{.75\baselineskip}
\label{tab:general_table_shared_k_in_32} 
\centering 
\resizebox{\columnwidth}{!} 
{\begin{tabular}{c cccccccccc} 
         \toprule 
         \multicolumn{11}{c}{Continual Entities} \\ 
         \midrule 
         \backslashbox{\textbf{Train}}{\textbf{Test}}& 2013 & 2014 & 2015 & 2016 & 2017 & 2018 & 2019 & 2020 & 2021 & 2022 \\ 
        \midrule 
2013  & 0.723 & \cellcolor{deepcarmine!16}0.719 & \cellcolor{deepcarmine!20}0.716 & \cellcolor{deepcarmine!31}0.710 & \cellcolor{deepcarmine!38}0.705 & \cellcolor{deepcarmine!50}0.697 & \cellcolor{deepcarmine!57}0.693 & \cellcolor{deepcarmine!59}0.692 & \cellcolor{deepcarmine!66}0.687 & \cellcolor{deepcarmine!75}0.682 \\ 
2014  & \cellcolor{darkspringgreen!18}0.731 & 0.727 & \cellcolor{deepcarmine!17}0.723 & \cellcolor{deepcarmine!26}0.717 & \cellcolor{deepcarmine!32}0.714 & \cellcolor{deepcarmine!46}0.706 & \cellcolor{deepcarmine!54}0.702 & \cellcolor{deepcarmine!55}0.702 & \cellcolor{deepcarmine!67}0.695 & \cellcolor{deepcarmine!75}0.690 \\ 
2015  & \cellcolor{darkspringgreen!22}0.727 & \cellcolor{darkspringgreen!14}0.723 & 0.721 & \cellcolor{deepcarmine!23}0.714 & \cellcolor{deepcarmine!31}0.710 & \cellcolor{deepcarmine!45}0.703 & \cellcolor{deepcarmine!52}0.700 & \cellcolor{deepcarmine!53}0.699 & \cellcolor{deepcarmine!65}0.693 & \cellcolor{deepcarmine!75}0.688 \\ 
2016  & \cellcolor{darkspringgreen!42}0.726 & \cellcolor{darkspringgreen!29}0.721 & \cellcolor{darkspringgreen!25}0.719 & 0.713 & \cellcolor{deepcarmine!20}0.709 & \cellcolor{deepcarmine!42}0.700 & \cellcolor{deepcarmine!52}0.696 & \cellcolor{deepcarmine!52}0.696 & \cellcolor{deepcarmine!63}0.692 & \cellcolor{deepcarmine!75}0.687 \\ 
2017  & \cellcolor{darkspringgreen!59}0.734 & \cellcolor{darkspringgreen!45}0.730 & \cellcolor{darkspringgreen!36}0.726 & \cellcolor{darkspringgreen!21}0.722 & 0.718 & \cellcolor{deepcarmine!32}0.710 & \cellcolor{deepcarmine!45}0.706 & \cellcolor{deepcarmine!46}0.706 & \cellcolor{deepcarmine!60}0.701 & \cellcolor{deepcarmine!75}0.696 \\
\bottomrule 
        \end{tabular}} 
\end{table}
\begin{table}[t] 
\caption[Accuracy@64 results]{\textbf{Accuracy@64} for \emph{continual} (top) and \emph{new} (bottom) entities. The intensity of colors is set on a row-by-row basis and indicates whether performance is \textcolor{darkspringgreen!100}{\textbf{better}} or \textcolor{deepcarmine!100}{\textbf{worse}} compared to the year the model was finetuned on (\ie the values that form the white diagonal).} 
\vspace{.75\baselineskip}
\label{tab:general_table_shared_k_in_64} 
\centering 
\resizebox{\columnwidth}{!} 
{\begin{tabular}{c cccccccccc} 
         \toprule 
         \multicolumn{11}{c}{Continual Entities} \\ 
         \midrule 
         \backslashbox{\textbf{Train}}{\textbf{Test}}& 2013 & 2014 & 2015 & 2016 & 2017 & 2018 & 2019 & 2020 & 2021 & 2022 \\ 
        \midrule 
2013  & 0.785 & \cellcolor{deepcarmine!16}0.782 & \cellcolor{deepcarmine!23}0.778 & \cellcolor{deepcarmine!34}0.772 & \cellcolor{deepcarmine!40}0.769 & \cellcolor{deepcarmine!52}0.762 & \cellcolor{deepcarmine!60}0.758 & \cellcolor{deepcarmine!60}0.758 & \cellcolor{deepcarmine!68}0.754 & \cellcolor{deepcarmine!75}0.750 \\ 
2014  & \cellcolor{darkspringgreen!16}0.792 & 0.790 & \cellcolor{deepcarmine!20}0.785 & \cellcolor{deepcarmine!28}0.781 & \cellcolor{deepcarmine!37}0.777 & \cellcolor{deepcarmine!49}0.771 & \cellcolor{deepcarmine!59}0.767 & \cellcolor{deepcarmine!58}0.767 & \cellcolor{deepcarmine!67}0.763 & \cellcolor{deepcarmine!75}0.760 \\ 
2015  & \cellcolor{darkspringgreen!20}0.786 & \cellcolor{darkspringgreen!14}0.784 & 0.782 & \cellcolor{deepcarmine!23}0.777 & \cellcolor{deepcarmine!32}0.773 & \cellcolor{deepcarmine!44}0.769 & \cellcolor{deepcarmine!53}0.765 & \cellcolor{deepcarmine!57}0.764 & \cellcolor{deepcarmine!67}0.760 & \cellcolor{deepcarmine!75}0.757 \\ 
2016  & \cellcolor{darkspringgreen!43}0.789 & \cellcolor{darkspringgreen!29}0.784 & \cellcolor{darkspringgreen!22}0.781 & 0.777 & \cellcolor{deepcarmine!21}0.773 & \cellcolor{deepcarmine!36}0.768 & \cellcolor{deepcarmine!49}0.763 & \cellcolor{deepcarmine!49}0.763 & \cellcolor{deepcarmine!64}0.758 & \cellcolor{deepcarmine!75}0.755 \\ 
2017  & \cellcolor{darkspringgreen!54}0.794 & \cellcolor{darkspringgreen!43}0.791 & \cellcolor{darkspringgreen!35}0.788 & \cellcolor{darkspringgreen!22}0.785 & 0.781 & \cellcolor{deepcarmine!32}0.775 & \cellcolor{deepcarmine!45}0.771 & \cellcolor{deepcarmine!43}0.772 & \cellcolor{deepcarmine!59}0.768 & \cellcolor{deepcarmine!75}0.763 \\
\bottomrule 
        \end{tabular}} 
\end{table}

\clearpage
{
\small
}

\renewcommand{\labelenumi}{\arabic{enumi}.} 

\clearpage

\renewcommand*{\thesection}{\thechapter.\arabic{section}}       

\clearpage{\pagestyle{empty}\cleardoublepage}

\graphicspath{{klim_ch_conclusion/fig/}}

\widowpenalty100000
\clubpenalty100000

\hyphenation{}


\chapter[Conclusions and Future Research Directions]{Conclusions and Future Research}
\label{chap:conclusion}

\renewcommand\evenpagerightmark{{\scshape\small Chapter \arabic{chapter}}}

\renewcommand\oddpageleftmark{{\scshape\small Conclusions and Future Research}}

\renewcommand{\bibname}{References}
\begin{flushright}
\end{flushright}

\noindent%
\emph{%
We outline the main conclusions for each of the presented chapters in the current thesis. Additionally, we discuss possible future research directions that can address some of the limitations of the current work. 
}
\begin{center}
\par{$\star\star\star$}
\end{center}
\vspace{0.15in}

\section{Conclusions}
\subsection{DWIE: an entity-centric dataset for multi-task document-level information extraction}
\label{ch_conclusion:sec:dwie}
In \chapref{chap:dwie} we introduce \datasetname, a manually annotated multi-task dataset that comprises named entity recognition, coreference resolution, relation extraction and entity linking as the main tasks. We highlight how \datasetname~differs from the mainstream datasets by focusing on document-level and entity-centric annotations. This also makes the predictions on this dataset more challenging by having not only to consider explicit, but also implicit document-level interactions between entities. 
Furthermore, we show how Graph Neural Networks can help to tackle this issue by propagating local contextual mention span information on a document level for a single task as well as across the tasks on the \datasetname~dataset. 
We experiment with known graph propagation techniques driven by the scores of the coreference resolution (\propformat{CorefProp}) and relation extraction (\propformat{RelProp}) components, as well as introduce a new latent task-independent attention-based graph propagation method (\propformat{AttProp}). We demonstrate that, without relying on the task-specific scorers, \propformat{AttProp} can boost the performance of single-task as well as joint models, performing on par and even outperforming significantly in some scenarios the \propformat{RelProp} and \propformat{CorefProp} graph propagations. Furthermore, our experimental results show complementarity between some of the evaluated IE tasks, with superior performance when using \textit{joint} model compared to independently trained \textit{single} models.


\subsection{Towards consistent document-level entity linking: joint models for entity linking and coreference resolution}
\label{ch_conclusion:sec:coreflinker}
In \chapref{chap:coreflinker}, we propose two end-to-end models to solve entity linking and coreference resolution tasks in a joint setting. 
Both of our joint architectures are characterized by formulating EL+coref as a single, structurally constrained task.  This contrasts with previous attempts to join coref+EL tasks \cite{hajishirzi2013joint,dutta2015c3el,angell2021clustering}, where both of the models are trained separately and additional logic is required to merge the predictions of coref and EL tasks.
It further contrasts with the joint architecture proposed in \chapref{chap:dwie}, where the loss function is composed of a weighted linear combination of multiple losses, each one corresponding to a particular task. This multi-task approach presents additional challenges when defining the weights in order to normalize and avoid interference between each of the task's gradients \cite{chen2018gradnorm,vandenhende2021multi,zhang2021survey, yu2020gradient,chen2020just,liu2021conflict}.
Conversely, both of our proposed models allow to efficiently compute the \textit{exact}
log-likelihood loss of the joint EL+coref target by marginalization over all possible configurations (\eg all possible spanning trees of our \textit{global} model). 
Our joint architectures achieve superior performance compared to the standalone counterparts on both coreference and entity linking tasks. 
Further analysis reveals that this boost in performance is driven by more coherent predictions on the level of mention clusters (linking to the same entity) and extended candidate entity coverage.

\subsection{Injecting knowledge base information into end-to-end joint entity and relation extraction and coreference resolution}
\label{ch_conclusion:sec:injecting}
In \chapref{chap:injecting_knowledge}, we propose an end-to-end model for joint IE (NER + relation extraction + coreference resolution) incorporating entity representations from a background knowledge base (KB) in a span-based model.
We find that representations built from a knowledge graph and a hypertext corpus are complementary in boosting IE performance. Concretely, when using both entity embeddings from the textual Wikipedia \cite{yamada2016joint} and entity representations derived from the Wikidata Knowledge Graph \cite{joulin2017fast}, we observe an improvement of performance compared to when incorporating each of these representations separately. 
To combine these candidate entity representations for text spans, we explore various weighting schemes: 
\begin{enumerate*}[(i)]
\item a uniform average of candidate entities (\emph{Uniform}),
\item the prior weights of each of the candidate entities (\emph{Prior}), 
\item an attention scheme (\emph{Attention}), or
\item attention with prior information (\emph{AttPrior}).
\end{enumerate*} Our experimental results show a superior performance when applying the \textit{AttPrior} scheme, showcasing the complementary effect of combining prior frequency information from a hypertext corpus with contextual information to identify the relevant entities.



\subsection{Temporal entity linking}
\label{ch_conclusion:sec:temporal_el}
In \chapref{chap:temporal_el} we introduced \ourdataset, a first large-scale 
temporal
entity linking 
dataset 
composed of 10 yearly snapshots of Wikipedia target entities linked to by anchor mentions.
We divided this dataset into mentions linked to \textit{continual} (existing in all the temporal snapshots) and \textit{new} (new to a particular snapshot) target entities.
We described the dataset creation pipeline, putting special focus on the quality assurance and future extensibility of \ourdataset. Our preliminary analysis of the \ourdataset~showcases a decrease in Jaccard similarity of entity definition as well as the context of mentions linked to a specific entity. This demonstrates the dynamic and evolving nature of \ourdataset, affected by changes in \begin{enumerate*}[(i)]
    \item target entity definitions, and 
    \item the anchor mentions linked to those target entities
\end{enumerate*}. Furthermore, we experiment with the bi-encoder baseline model and showcase a consistent temporal deterioration in performance of entity linking task. We conclude that such a decrease in performance is particularly affected by the temporally increasing number of (ever more granular) candidate entities in Wikipedia. We further examined the most challenging cases for this model, concluding the critical aspect of new knowledge acquisition during the pre-training phase in order to successfully disambiguate \textit{new} entities. 

\section{Future directions} 
\label{ch_conclusion:sec:future_directions}
\vspace{2mm}
Even though the presented work has extended the knowledge in multiple areas of information extraction, we have only scratched the surface of 
potential research avenues to be explored. In this section we aim to provide a brief description of such future research directions that could complement or extend the methodologies introduced in this thesis.
\\
\noindent\textbf{On extending coreference annotations:}
 One future direction consists in extending the coreference annotations to include nominal and anaphoric expressions. This will challenge and open new perspective into studying the complementary relation between entity linking and coreference component described in \chapref{chap:coreflinker}. Furthermore, we expect that 
 including these diverse mention types (whose initial span embedding representation can be different from coreferenced named entities), will allow to investigate further the potential benefits of using joint entity-centric models such as the ones explored in \chapref{chap:dwie} and \chapref{chap:coreflinker}.
 
\noindent \textbf{Effect of pre-training on new corpora:} 
Recent work has demonstrated the benefits of pre-training 
language models on more recent corpora (\eg the latest 
Wikipedia versions) when applied on the downstream tasks 
\cite{agarwal2021temporal,loureiro2022timelms}. Yet, in 
our work described in 
 \chapref{chap:temporal_el}, 
we fine-tune existing pre-trained BERT based models on 
downstream task of entity linking. We hypothesize that 
pre-training BERT from scratch on newer versions of 
Wikipedia can boost the performance on \ourdataset~introduced in \chapref{chap:temporal_el}, 
specially on \textit{new} entities that often require 
additional knowledge to be correctly disambiguated. 



\noindent \textbf{Temporal changes in mention context:} the work in \chapref{chap:temporal_el} focused mostly on changes in target entities, leaving unexplored a study on how changes in mention context affect EL task. However, \figref{fig:evolution_mentions} in \chapref{chap:temporal_el} showcases a big drop in Jaccard vocabulary similarity of context around mentions, specially compared to the immediately preceding year. This suggests that the mentions, as well as the text surrounding them, are highly volatile and evolving with time, making 
the temporal mention evolution
an interesting subject for future research. Concretely, our hypothesis is that the temporal performance drift in entity linking task is not only affected by changes in the target entities, but also by changes in the mentions linked to these entities. 

\noindent \textbf{Cross-lingual entity information:} the work described in this thesis is limited to entity definitions in English Wikipedia. Yet, recent research \cite{botha2020entity, de2021multilingual} has shown the benefits of training entity linking models in a cross-lingual setting. We hypothesize that this setting can also be applied 
to study the evolution of entity linking task introduced in \chapref{chap:temporal_el} of this thesis. This can be achieved by adapting the \ourdataset~creation framework introduced in \chapref{chap:temporal_el} to process the  history of Wikipedias in other languages than English and extract the respective anchor mentions and target entities. 


\noindent\textbf{KB-driven entity linking:} in the current thesis we have covered entity linking 
constrained to anchor mentions with specific characteristics. For example, in Chapters~\ref{chap:dwie}--\ref{chap:injecting_knowledge}, we focused on linking proper nouns (\ie named entities). 
Yet, we envision a complete entity linking approach driven exclusively by the entities in a particular Knowledge Base and not by some characteristics of anchor mentions. 
We argue this setup would 
enrich
the connection between the text and KB, allowing to study in greater detail the effects of entity-centric reasoning and transfer of knowledge
in models such as the ones developed in Chapters \ref{chap:dwie}--\ref{chap:injecting_knowledge} of this thesis. 
Recent work in this direction has been limited to domain-specific entity linking datasets such as MedMentions \cite{mohan2018medmentions} in the biomedical domain. Each of the possible textual mention spans in MedMentions was carefully examined by professionals with experience in biomedical content to find the corresponding match in UMLS \cite{bodenreider2004unified} 
biomedical
 knowledge base. As a result, MedMentions presents densely annotated documents with entity links driven by the content of the target UMLS KB. More recently, the authors of \cite{rosales2020fine} propose a fine-grained entity linking annotation scheme, including mentions in grammatical categories such as adjectives, verbs and adverbs. The authors further extend three entity linking datasets (VoxEL \cite{rosales2018voxel}, KORE50 \cite{hoffart2012kore}, and ACE2004 \cite{ratinov2011local}) using the proposed annotation scheme. Experimental results on these datasets show very low recall of state-of-the-art models, indicating their limitation in identifying the set of all mentions to be linked. Yet, despite these interesting findings, the annotated datasets are small (ranging between 1 and 20 documents each), which makes them impractical to train models. This research gap is exacerbated by a lack of KB-driven entity linking annotations on \textit{multi-task} IE datasets such as DWIE. 

\noindent\textbf{Cross-document IE:} in Chapters \ref{chap:dwie} -- \ref{chap:injecting_knowledge} we focused in document-level entity-centric annotations. We think that the next natural step is to extend entity-centric information extraction (IE) approach to cross-document setting. This setting is characterized by IE annotations, such as 
coreferent mentions for coreference resolution task,
created on a set of multiple related documents (\ie cross-document). This contrasts with \datasetname~introduced in \chapref{chap:dwie}, where all the structured annotations are made on a single document, which represents a specific news article. 
Most of the related work 
on cross-document IE 
has focused on 
\textit{coreference resolution} task \cite{bagga1998entity, logan2021benchmarking,hsu2022contrastive,cattan2021realistic,cattan2021cross,cattan2021scico,cattan2020streamlining,gooi2004cross,singh2011large,barhom2019revisiting,caciularu2021cross,ravenscroft2021cd2cr}. This task consists in identifying coreferent mentions on a set of documents 
given as input. 
More recently, there has been an ever-growing interest in other 
cross-document tasks such as entity linking \cite{agarwal2021entity} and relation extraction \cite{yao2021codred}. Yet, to the best of our knowledge, there is still a research gap in integrating the different cross-document IE annotations in a single multi-task dataset. Such a dataset 
would  
encourage the research in creating joint IE models to efficiently process the information in multiple documents. 
This setting involving multiple documents as input can be challenging to tackle with current state-of-the-art BERT-driven IE models that can only process a very limited context.  
Fortunately, recent work in efficient transformers such as Longformer \cite{beltagy2020longformer} and BigBird \cite{zaheer2020big}, among others \cite{treviso2022efficient}, is promising to be applied to tackle this problem. This is the case of the recently introduced Longformer-based cross-document language model (CDLM) \cite{caciularu2021cdlm} that achieves state-of-the-art results in the cross-document coreference resolution IE task.  
\noindent\textbf{Harnessing and editing knowledge in language models}, 
recent work using BERT-based language models (LMs) \cite{devlin2019bert,lewis2020bart,brown2020language} has advanced the state-of-the-art in information extraction (IE). For instance, span-based IE models \cite{wadden2019entity,eberts2020span,zhong2021frustratingly,lin2020joint,wang2020two} use contextualized BERT embeddings as input to generate span representations. 
Furthermore, recent dense passage retrieval models \cite{karpukhin2020dense} have successfully used BERT encoders as retriever components in Information Extraction tasks such as entity linking \cite{wu2019zero,zhang2021entqa,li2020efficient} and slot filling \cite{petroni2020kilt}. 
More recently, autoregressive and generative models \cite{de2020autoregressive,petroni2020context,petroni2020kilt,rongali2020don,nogueira2020document,lewis2020bart} have advanced the state-of-the-art by generating directly the information stored in LMs for IE tasks such as entity linking \cite{de2020autoregressive,de2021multilingual,de2021highly,mrini2022detection,yuan2022generative}, relation extraction \cite{cabot2021rebel,josifoski2021genie,saxena2022sequence}, event prediction \cite{jin2019recurrent}, argument extraction \cite{huang2022multilingual} and slot filling \cite{glass2022re2g,lu2022unified}. 
Finally, recent advances in prompt engineering \cite{liu2021pre} go one step further by explicitly probing the language models for answers. Such prompt-driven models have been applied in a few-shot setting for a number of IE tasks such as named entity recognition \cite{ma2021template,liu2022qaner}, relation extraction \cite{zhang2022prompt,chen2022knowprompt,yeh2022decorate,gong2021prompt,chia2022relationprompt} 
and event argument extraction \cite{liu2022dynamic,ma2022prompt}.
Yet, the performance of these language models is highly dependent on the knowledge stored in their internal structure \cite{petroni2019language,alkhamissi2022review,yin2022survey}. 
This is also suggested by the drop in performance in \chapref{chap:temporal_el} for our BERT-based bi-encoder on new entities that require additional knowledge related to COVID-19 pandemic. In order to tackle this issue, 
future work should focus on exploring efficient mechanisms, ideally working with one-shot exposure to new knowledge in KBs in order to inject new facts in the pre-trained language models. 
Recent work \cite{loureiro2022timelms,wang2020k,de2021editing,sinitsin2020editable,cossu2022continual,jin2022lifelong,zhu2020modifying,hospedales2021meta} demonstrates that knowledge injection methods such as continual pre-training and hypernetworks can indeed improve the performance on various IE tasks. Yet, such mechanisms can be further improved, particularly in the number of training examples the model has to be exposed to in order to be able to effectively incorporate (or learn to incorporate) new knowledge. 

\noindent\textbf{Entity-centric semantic frames}, 
we use the concept \textit{semantic frames} 
to refer to preliminarly defined structure interconnecting multiple entities such as lexically driven FrameNet \cite{baker1998berkeley} as well as more general \textit{n-ary relations} \cite{jia2019document,peng2017cross,song2018n,giunti2021representing,lentschat2022new,lai2020bert,lehmberg2019synthesizing} and \textit{events} \cite{xiang2019survey,hogenboom2016survey,zhan2019survey}. For example, we can envision a semantic frame \textit{marriage} interconnecting entities related to both \textit{partners}, \textit{date}, \textit{place}, etc. Unlike relations, the number of linked entities in semantic frames is not restricted to 2 (head and tail).  
The current structure of most of these frames is mention-driven. We believe the design and annotation of entity-centric semantic frames (\ie frames where each slot is connected to conceptual entities, instead of entity mentions), would allow to abstract from individual mentions and focus on entities, some of which might even not be explicitly mentioned in a document. Following this direction, one of the possible future works could consist in annotating such semantic frames as an additional task in the introduced multi-task \datasetname~dataset (see \chapref{chap:dwie}). 
\\

\clearpage
\clearpage{\pagestyle{empty}\cleardoublepage}

\appendix
\graphicspath{{klim_ch_mwp/figures/}}
\newcommand{\yesmarker}{\cmark}
\newcommand{\nomarker}{$-$}

\widowpenalty100000
\clubpenalty100000
\renewcommand*{\thesection}{A.\arabic{section}}

\hyphenation{}

\chapter[Solving Arithmetic Word Problems by Scoring Equations with Recursive Neural Networks]{Solving Arithmetic Word Problems by Scoring Equations with Recursive Neural Networks}
\label{chap:mwp}

\renewcommand\evenpagerightmark{{\scshape\small Chapter A}}
\renewcommand\oddpageleftmark{{\scshape\small Solving Arithmetic Word Problems by Scoring Equations with Recursive Neural Networks}}

\renewcommand{\bibname}{References}

\begin{flushright}
\end{flushright}

\noindent\emph{In this chapter we propose to use recursive neural networks to mimic the structure on equation trees to solve mathematical world problems. We showcase a significant improvement using our approach.}

\begin{center}
\par{$\star\star\star$}
\end{center}
\vspace{0.15in}

\par{\noindent\large{\textbf{K.~Zaporojets, G.~Bekoulis, J.~Deleu, C.~Develder and T.~Demeester}}}
\vspace{0.1in}
\par{\noindent\textbf{Expert Systems with Applications, 2021.}}
\vspace{0.15in}

\par{\noindent\bf{Abstract}} Solving arithmetic word problems is a cornerstone task in assessing language understanding and reasoning capabilities in NLP systems. Recent works 
  use automatic extraction and ranking of candidate solution equations providing the answer to arithmetic word problems. In this work, we explore novel approaches to score such candidate solution equations using tree-structured
  recursive neural network (Tree-RNN) configurations. 
  The advantage of this Tree-RNN approach over using more established sequential representations, is that it can naturally capture the structure of the equations. Our proposed method consists of transforming the mathematical expression of the equation into an expression tree. Further, we encode this tree into a Tree-RNN by using different \TreeLSTM{} architectures.  
   Experimental results show that our proposed method 
   \begin{enumerate*}[(i)]
    \item  improves overall performance with more than 3\% accuracy points compared to previous state-of-the-art, and with over 15\% points on a subset of problems that require more complex reasoning, and \item outperforms sequential LSTMs by 4\% accuracy points on such more complex problems. 
\end{enumerate*}

\section{Introduction}
\label{ch_mwp:sec:intro}

\begin{figure}[b]
\centering
\includegraphics[width=0.7\columnwidth]{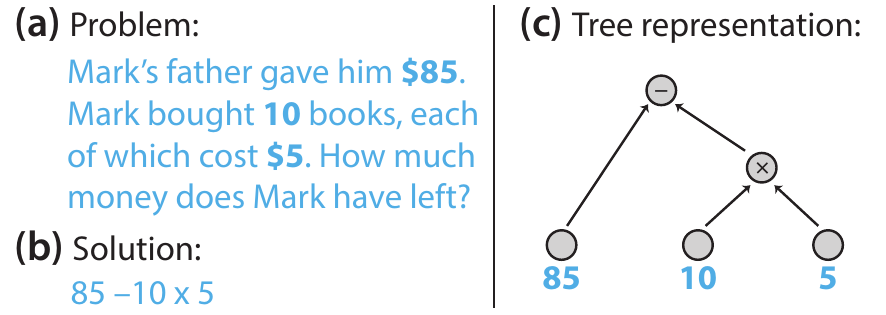}
\captionsetup{singlelinecheck=off}
\caption[An example of arithmetic word problem from the SingleEQ dataset.]{
An example of arithmetic word problem from the SingleEQ dataset. It illustrates the 
\protect\begin{enumerate*}[(a)]
    \item\label{ch_mwp:fig1a} \textit{an arithmetic word problem} statement,
    \item\label{ch_mwp:fig1b} the respective \textit{solution} formula, and
    \item\label{ch_mwp:fig1c} the \textit{expression tree} representing the solution.
\protect\end{enumerate*}
}
\label{ch_mwp:fig:example}
\end{figure}

\noindent Natural language understanding often requires the ability to comprehend and reason with expressions involving numbers. This has produced a recent rise in interest to build applications to automatically solve math word problems~\cite{kushman2014learning,koncel2015parsing,mitra2016learning,wang2018mathdqn,zhang2019gap}. These math problems consist of a textual description comprising numbers with a question that will guide the reasoning process to get the numerical solution (see \figref{ch_mwp:fig:example} for an example). 
This is a complex task because of
\begin{enumerate*}[label=(\roman*)]
    \item \label{ch_mwp:problem1} the large output space of the possible equations representing a given math problem, and
    \item \label{ch_mwp:problem2} reasoning required to understand the problem.
\end{enumerate*} 

The research community has focused in solving mainly two types of mathematical word problems: \textit{arithmetic word problems} 
\cite{hosseini2014learning, mitra2016learning, wang2017deep, li2019modeling, chiang2019semantically} 
and \textit{algebraic word problems} \cite{kushman2014learning, shi2015automatically, ling2017program, amini2019mathqa}. Arithmetic word problems 
can be solved using basic mathematical operations ($+, -, \times, \div$) and 
involve a single unknown variable. Algebraic word problems, on the other hand, 
involve more complex operators such as square root, exponential and logarithm with multiple unknown variables. 
In this work, we focus on solving \textit{arithmetic word problems} such as the one illustrated in \figref{ch_mwp:fig:example}. This figure illustrates \begin{enumerate*}[label=(\alph*)]
    \item \textit{arithmetic word problem} statement, 
    \item the arithmetical formula of the \textit{solution} to the problem, and 
    \item the \textit{expression tree} representation of the solution formula where the leaves are connected to quantities and internal nodes represent operations
\end{enumerate*}. 

The main idea of this paper is to explore the use of tree-based Recursive Neural Networks (Tree-RNNs) to encode and score the expression tree (illustrated in \figref{ch_mwp:fig:example}\ref{ch_mwp:fig1c} that represents a candidate arithmetic expression of a specific arithmetic word problem). 
This contrasts with predominantly sequential neural representations \cite{wang2017deep, wang2018translating, chiang2019semantically} that encode the problem statement from left to right or vice versa.
By using Tree-RNN architectures, we can naturally embed the equation inside a tree structure such that the link structure directly reflects the various mathematical operations between operands selected from the sequential textual input. We hypothesize that this structured approach can efficiently capture the semantic representations of the candidate equations to solve more complex arithmetic problems involving multiple and/or non-commutative operators. To test our results, we use the recently introduced SingleEQ dataset \cite{koncel2015parsing}. It contains a collection of 508 arithmetic word problems with varying degrees of complexity. This allows us to track the performance of the evaluated systems on subsets that require different reasoning capabilities. More concretely, we subdivide the initial dataset into different subsets of varying reasoning complexity (\ie based on the number of operators, commutative (symmetric) or non-commutative (asymmetric) operations), to investigate whether  
the performance of the proposed architecture remains consistent across problems of increasing complexity.

\begin{figure}[!ht]
\centering
\includegraphics[width=1.0\columnwidth,trim={0.5cm 7.0cm 7.7cm 0.2cm},clip]{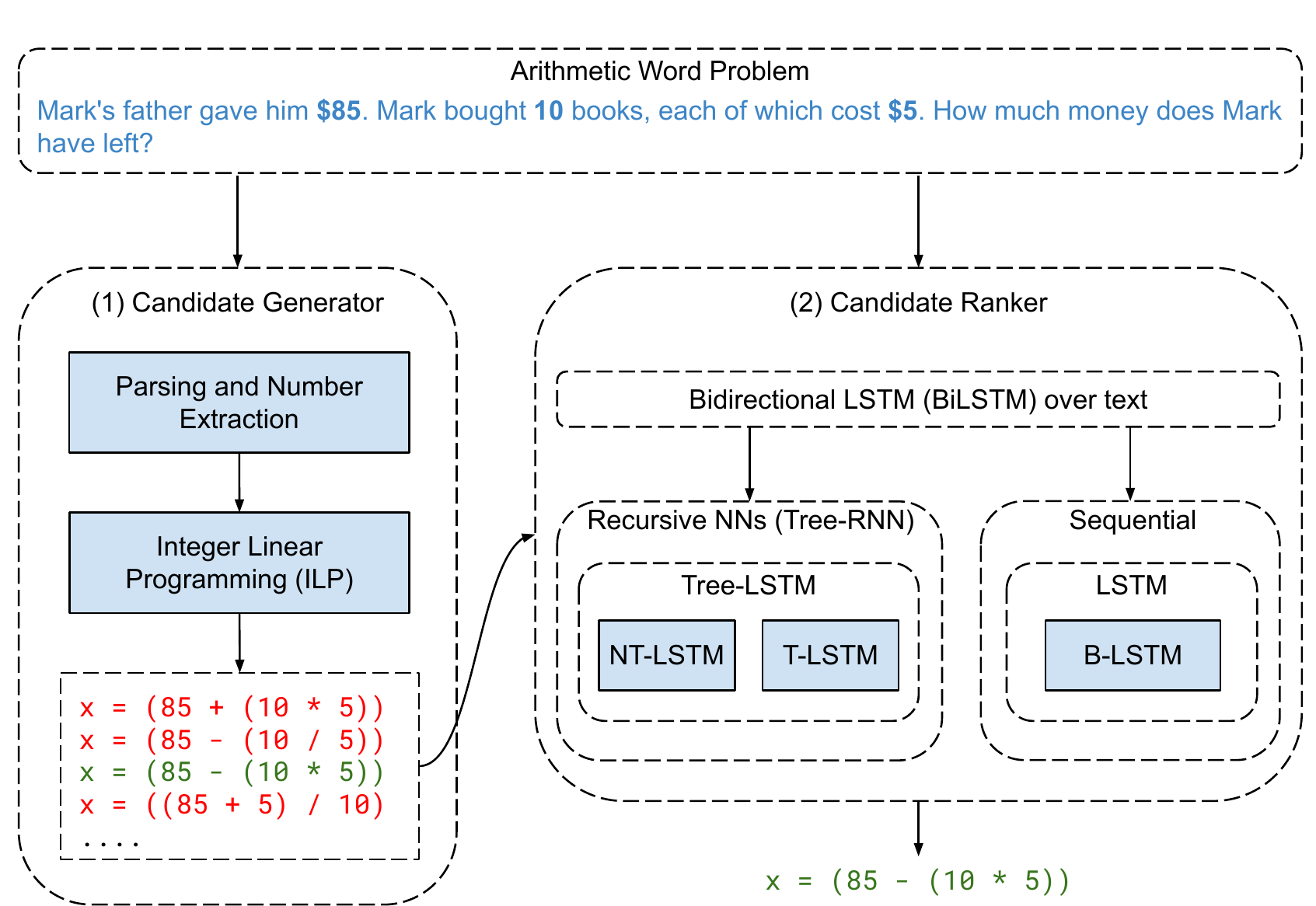}
\captionsetup{singlelinecheck=off}
\caption[High-level conceptual view of the arithmetic word problem architecture]{
High-level conceptual view of the arithmetic word problem architecture used throughout the paper. It consists of two main components:
\protect\begin{enumerate*}[(1)]
    \item\label{ch_mwp:fig2a} \textit{candidate generator} responsible for generating candidate equations to solve a particular \textit{arithmetic word problem}, and
    \item\label{ch_mwp:fig2b} \textit{candidate ranker}, for selecting the best candidate from the list provided by \textit{candidate generator}, using the models \NTLSTM{}, \TLSTM{}, or \BLSTM{}. 
\protect\end{enumerate*}
}
\label{ch_mwp:fig:conceptualview}
\end{figure}

\Figref{ch_mwp:fig:conceptualview} provides a high-level conceptual view of the interconnection between the main components of our proposed system. The processing flow consists of two main steps. In the first step, we use the \textit{candidate generator} to generate a list of potential candidate equations for solving a particular \textit{arithmetic word problem}. To achieve this, we employ the Integer Linear Programming (ILP) constraint optimization component proposed by \mbox{\cite{koncel2015parsing}} (see \secref{ch_mwp:sec:candidate_generator}). In the second step, the candidate equations are ranked by the \textit{candidate ranker}, and the equation with the highest score is chosen as the solution to the processed \textit{arithmetic word problem} (see \secref{ch_mwp:sec:model}). 
In this paper, we focus on this second step by exploring the impact of structural Tree-RNN-based and sequential Long Short Term Memory-based (LSTM; \cite{hochreiter1997long}) candidate equation encoding methods.
More specifically, we define two Tree-RNN models inspired by the work of \cite{tai2015improved} on \TreeLSTM{} models:
\begin{enumerate*}[label=(\roman*)]
    \item \TLSTM{} (Child-Sum \TreeLSTM{}), and 
    \item \NTLSTM{} (N-ary \TreeLSTM{})
\end{enumerate*}. In the rest of the manuscript we refer to the general tree-structured architecture of these models as \TreeLSTM{}.
The main difference between the two is that, while in \TLSTM{} the child node representations 
are summed up, in \NTLSTM{} they are concatenated. Unlike the representation used in \cite{tai2015improved}, where the input is given by the word embeddings, our Tree-LSTM models also take as input the operation embeddings (in inner nodes) that represent each of the arithmetic operators ($-$, $+$, $\div$, $\times$). This allows our architecture to distinguish between different operators that are contained in a particular expression tree. We show that \NTLSTM{} is more suitable to deal with equations that involve non-commutative operators because this architecture is able to capture the order of the operands.
We also compare our \TreeLSTM{} models with a sequential LSTM model which we call \BLSTM{}. 
All the models (\TLSTM{}, \NTLSTM{}, and \BLSTM{}) take as input the contextualized representation of the numbers in text produced by a bidirectional \LSTM{} layer (\biLSTM{}) (see \secref{ch_mwp:sec:model} for details).
After conducting a thorough multi-fold
experimentation phase involving multiple random weight re-initializations in order to ensure the validity of our results, we will show that the main added value of our \TreeLSTM{}-based models compared to state-of-the-art methods lays in an increased performance for 
more complex arithmetic word problems.

More concretely, 
our contribution is three-fold:
\begin{enumerate*}[(i)]
\item we propose using \TreeLSTMs{} for solving arithmetic word problems, to embed structural information of the equation,
\item we compare it against a strong neural baseline model (\BLSTM{}) that relies on sequential LSTMs,
and
\item we perform an extensive experimental study on the SingleEQ dataset, showing that our \TreeLSTM{} model achieves an overall accuracy improvement of 3\%, including an increase 
$>$15\% for more complex problems (\ie requiring multiple and non-commutative operations), compared to previous state-of-the-art results. 
\end{enumerate*}

\section{Related work}
\label{ch_mwp:sec:related}

\noindent Over the last few years,
there has been an increasing interest in building systems to solve \textit{arithmetic word problems}. The adopted approaches 
can be grouped in three main categories: \begin{enumerate*}[label=(\roman*)]
    \item Rule-based systems,
    \item Statistical systems, and
    \item Neural network systems
\end{enumerate*}. 

\noindent\textbf{Rule-based systems}: 
The first attempts to solve
arithmetic problems date back to the 1960s 
with the work by \cite{bobrow1964natural}, who proposed and implemented 
STUDENT, a rule-based parsing system to extract numbers 
and operations between them by using pattern matching techniques. 
 \cite{charniak1968carps, charniak1969computer} extended STUDENT by including basic coreference resolution and capability to work with rate expressions (\eg ``kms per hour"). 
On the other hand, \cite{fletcher1985understanding}
designed and implemented
a system that given a propositional representation of a math problem\footnote{With propositions such as \textit{GIVE Y X P9}, where entity Y gives to entity X the object defined in P9. This proposition in particular can be linked to the first sentence of example in \figref{ch_mwp:fig:example}: ``Mark's father gave him \$85", where Y represents ``Mark's father", X represents ``him" which is coreferenced to ``Mark", and P9 represents ``\$85" that are being given. }, applies a set of rules to calculate the final solution. 
The disadvantage of this system is that it needs a 
parsed propositional 
representation of a problem as input and cannot operate directly on raw text. 
This issue was tackled
by \cite{bakman2007robust}, who developed 
a schema-based system that consisted of six main reasoning schemas, each one with slots to fill in.  
After instantiating the schemas for a particular math problem using lexical verb-based rules, the system could derive the corresponding mathematical equation to solve the problem.  

The main disadvantages of such rule-based approaches are that they \begin{enumerate*}[label=(\roman*)]
    \item rely on hard-coded lexico-grammar rules, and 
    \item lack an integrated view of the problem to be solved, extracting operations one by one
\end{enumerate*}. We address these issues by proposing a model that integrates the mathematical representation of a problem in a single structured expression tree. This way, we are able to capture the operator-operator and number-operator relations involved in a particular mathematical expression in a unified manner. Furthermore, we 
avoid
the use of lexico-grammar hard-coded rules (\eg the use of pattern-based matching) when connecting numbers with the operators, replacing them by composition-semantic representations that link the arithmetic operations with parameters (numbers or other operations) in a recursive tree.
Consequently, our solution is more generalizable by not depending on explicit hand-crafted logic. 

\noindent\textbf{Statistical systems}: 
Recently, there has been a shift towards statistical feature-driven systems that automatically produce models by capturing patterns present in arithmetic word problem datasets.  
For example, \cite{hosseini2014learning}
 presented an 
 inductive model that links specific lexicon-based features (\eg verb categories) to equation operators. The mathematical solution to the problem is built sequentially using state transitions related to operators that are triggered by different verb categories found in the problem statement. On the other hand, \cite{mitra2016learning} connected carefully designed features to equation templates in order to solve specific problem types.  
While these techniques produced competitive results, they were limited to addition ($+$) and subtraction ($-$) operations 
on a very narrow problem set domain. 
In order to solve more diverse types of problems that also involve multiplication and division operators,
the community shifted towards more integrated approaches involving 
tree structure representations.
\cite{koncel2015parsing}  
proposed to rank candidate expression trees by training jointly a \textit{local} model to link spans of text with operator tree nodes, and a \textit{global} model that is used to score the consistency of an entire tree. The list of candidates to these two models is generated by an ILP constraint optimization component that, given a set of extracted numbers from a arithmetic word problem text as input, produces a set of candidate solution equations. 
Conversely, \cite{roy2015solving, roy2017unit} introduced the concept of \textit{monotonic expression tree} to generate candidates. It defines a set of conditions (\eg two division and subtraction nodes cannot be connected to each other) that considerably restricts the expression tree search space. The authors propose to score the resulting monotonic expression trees jointly by summing up the scores of different classifiers related to a specific expression tree (\eg 
the mathematical operator between two numbers in the tree, 
whether a particular number is related to a rate such as ``kms per hour", etc).
Recently, the same authors \cite{roy2018mapping} included additional latent declarative rules (\eg $[\mathrm{Verb1} \in \mathrm{HAVE}] \land [\mathrm{Verb2} \in \mathrm{GIVE}] \land [\mathrm{Coref(Subj1,Subj2)}] \implies \mathrm{Subtraction}$)
to link textual expression patterns (derived from preliminary
dependency parsing) to specific operations. While these statistical approaches rely on tree structures to evaluate the mathematical expressions, 
on one hand, they require high manual effort to engineer the features and, on the other hand, it is hard to scale the features to capture operations between more than two numbers. This makes it challenging to apply such models to more complex equations that involve multiple operators. 
We tackle this problem by defining a single Tree-RNN structure that evaluates an entire mathematical expression at once. This is done by recursively combining the information from the child nodes in the expression tree and then using a backpropagation mechanism to correspondingly adjust the weights of our model. Furthermore, our equation ranking architecture does not depend on hand-crafted features and parsing-dependent rules, making it more effective in generalizing across different domains. 

\noindent\textbf{Neural network systems}: 
Recently, as in all sub-domains of natural language processing, neural network architectures have been applied to tackle math word problems. The first contribution was made by 
\cite{wang2017deep}, who introduced a model trained to map problem statements to equation templates.
Their model
was expanded upon by \cite{huang2018neural}, who introduced an attention-based copy mechanism for tokens representing numbers. They used  
a reinforcement learning setting, where positive 
rewards
were assigned when the predicted mathematical expression resulted in a correct answer.
Recently, \cite{chiang2019semantically} used stack structures inside a sequential encoder-decoder setting where the encoder captures the semantics of a math word problem in a vector that is used by decoder to generate the equation to solve the problem. 
Moreover, \cite{wang2018mathdqn} proposed the
use of Q-Networks in order to generate expression trees, by giving positive reward 
whenever the 
operator between two numbers is correct. 
The aforementioned studies, while showing promising results, 
were not designed to naturally capture the structural form of mathematical expressions when multiple operators are involved (\eg $1+(2/3)$ \vs $(1+2)/3$). 
We propose encoding equations with \TreeLSTMs{}~\cite{tai2015improved}, which are recursive neural sequence models, thus allowing to naturally reflect the execution order of operations in an expression tree by recursively combining the children nodes' semantic representations. 

\renewcommand\baselinestretch{1}{\begin{table}[t]
\centering
\resizebox{0.95\textwidth}{!}{
\begin{tabular}{lccccc}
\toprule
\textbf{Method} & 
\textbf{Rules} & \textbf{Features} & \textbf{N-Nets} & \textbf{Tree-Based} & \textbf{Tree-Based}   \\
 & & & & \textbf{Representation} & \textbf{Encoding}  \\
 \midrule
 \cite{bobrow1964natural} & \yesmarker & \nomarker & \nomarker & \nomarker & \nomarker \\
 \cite{charniak1968carps, charniak1969computer} & \yesmarker & \nomarker & \nomarker & \nomarker & \nomarker \\
 \cite{fletcher1985understanding} & \yesmarker & \nomarker & \nomarker & \nomarker & \nomarker \\
 \cite{bakman2007robust} & \yesmarker & \nomarker & \nomarker & \nomarker & \nomarker \\
 \cite{hosseini2014learning} & \nomarker & \yesmarker & \nomarker & \nomarker & \nomarker \\
  \cite{koncel2015parsing} & \yesmarker & \yesmarker & \nomarker & \yesmarker & \nomarker \\ 
 \cite{mitra2016learning} & \nomarker & \yesmarker & \nomarker & \nomarker & \nomarker \\
 \cite{roy2015solving, roy2017unit} & \nomarker & \yesmarker & \nomarker & \yesmarker & \nomarker \\ 
 \cite{wang2017deep} & \nomarker & \nomarker & \yesmarker & \nomarker & \nomarker \\    
 \cite{roy2018mapping} & \yesmarker & \yesmarker & \nomarker & \yesmarker & \nomarker \\ 
 \cite{huang2018neural} & \nomarker & \yesmarker & \yesmarker & \nomarker & \nomarker \\ 
 \cite{wang2018mathdqn} & \nomarker & \yesmarker & \yesmarker & \yesmarker & \nomarker \\
 \cite{chiang2019semantically} & \nomarker & \nomarker & \yesmarker & \nomarker & \nomarker \\
 \cite{li2019modeling} & \nomarker & \nomarker & \yesmarker & \nomarker & \nomarker \\
 Our Approach (\TLSTM{} \& \NTLSTM{}) & \nomarker & \nomarker & \yesmarker & \yesmarker & \yesmarker \\   
\midrule
\end{tabular}
}
    \caption[Comparison of the various architectures explored in related work]{Comparison of the various architectures explored in related work. We focus on the following five characteristics: \protect\begin{enumerate*}[(i)] 
	 \item \textit{Rules} indicates whether a rule-based approach is used or not, 
	 \item \textit{Features} specifies whether the architecture relies on manually engineered features, 
	 \item \textit{N-Nets} indicates whether artificial neural networks are used or not, 
	 \item \textit{Tree-Based Representation} groups 
	 the models that incorporate information coming from tree structures (\eg by using trees for feature engineering), and
	 \item \textit{Tree-Based Encoding} indicates whether the tree structures are used as encoders in a neural network model.
	 \end{enumerate*} The {\yesmarker} indicates the presence of a particular characteristic.}
    \label{ch_mwp:tab:comparative_related_work}
\end{table}}

\tabref{ch_mwp:tab:comparative_related_work} compares our approach (the use of \TreeLSTM{}-based \TLSTM{} and \NTLSTM{} models) with the rest of the methods described in this Section.
The main difference of our architecture is that we explore the impact of using tree-based neural encoding (\ie by means of Tree-LSTM models). We hypothesize that this approach allows
to better capture the arithmetic equation structure than the currently predominant neural sequential models \cite{wang2017deep,wang2018translating, chiang2019semantically}. Furthermore, the independence from feature-based and rule-based methods makes our solution more generalizable. This is because our model does not depend on 
hand-crafted rules or features to capture the patterns of a particular dataset.
This aspect will be explored further when comparing the performance of our model to the current feature-based state-of-the-art system \cite{koncel2015parsing} in \secref{ch_mwp:sec:results}. 

\textbf{Tree-RNN} models \cite{socher2011parsing} have been shown to perform better for modeling data on tasks that have an inherently hierarchical structure. For example, \cite{socher2011parsing} proposed to use recursive models in order to model the compositional structure of scene images (\eg a scene image of a house can be split in composing regions such as doors, windows, walls, etc.). The authors show that a Tree-RNN-based architecture outperforms previous methods in prediction of hierarchical structure of scene images and in scene image classification. Later, \cite{socher2013parsing} also showed how recursive structures can be used to encode the inherently hierarchical phrase structural grammar (\eg the sentence ``riding a bike" can be decomposed in the verb ``riding" and the noun phrase ``a bike", which itself can be decomposed into determiner ``a" and the noun ``bike"). This way, the authors achieved state-of-the-art performance in grammatical parsing of the sentences. More recently, \cite{tai2015improved} and \cite{chen2017enhanced} showed how encoding the syntactic parsing trees of the sentence with 
Tree-LSTM
models can improve the performance in tasks such as sentiment classification and semantic relatedness (\eg natural language inference). Similarly, we propose to take advantage of the inherently hierarchical representation of mathematical expression trees by encoding them using 
Tree-LSTM
architectures. Our experiments demonstrate that this representation can be helpful in capturing the semantic relations between operators needed in order to solve more complex arithmetic problems consisting of multiple and/or non-commutative operations.

\section{Proposed architecture}
\label{ch_mwp:sec:architecture}
\noindent 
Shortly stated, our task at hand is to identify the correct arithmetic equation, corresponding to an arithmetic problem expressed in natural language text.
We follow a two-step approach 
similar to the work of
\cite{koncel2015parsing}, which formalizes solving multi-sentence arithmetic word problems as 
\begin{enumerate*}[(i)]
\item the generation and 
\item ranking
\end{enumerate*}
of expression trees.
The {first step} consists of generating candidate equations using the ILP optimization solver proposed in \cite{koncel2015parsing} (\textit{candidate generator} component in \figref{ch_mwp:fig:conceptualview}). 
The {second step} ranks these candidates and selects the top ranked one as the final answer to the arithmetic word problem (\textit{candidate ranker} component in \figref{ch_mwp:fig:conceptualview}). We use the rest of this section to provide more insights into the \textit{candidate generator} component in~\secref{ch_mwp:sec:candidate_generator}, and to describe in detail our proposed \textit{candidate ranker} model in~\secref{ch_mwp:sec:model}.

\subsection{Candidate generator}
\label{ch_mwp:sec:candidate_generator}

\noindent This component is responsible for
generating possible candidate equations 
to solve a given arithmetic word problem. 
A straightforward solution would be to perform an exhaustive search on all the possible arithmetic expression trees given $n$ extracted numbers from a particular problem.
 However, the resulting search space would grow exponentially with $n$, which makes this approach not scalable.
In order to deal with this exponential growth in the number of candidates, we re-use the Integer Linear Programming (ILP) solver proposed by \cite{koncel2015parsing}. This solver takes as input the extracted numeric quantities with extra attributes derived from syntactic parsing\footnote{Stanford Dependency Parser in CoreNLP 3.4 is used.}, and generates the most promising candidate equations using two types of constraints:  
\begin{enumerate}
    \item \textit{Hard Constraints:} such as the maximum equation length and syntactic validity of equations (\eg only one unknown allowed, no division by 0, etc.). As a post-processing step, the ILP solver also removes the arithmetic expressions that produce negative or fractional results.
    \item \textit{Soft Constraints:} these constraints assign additional weight to candidate equations whose related entity types (extracted from dependency parse tree) are consistent. For example, in the problem of \figref{ch_mwp:fig:example}, the sum (85 + 5) will be prioritized over the sum (5 + 10), because both 85 and 5 refer to the same entity type (``\$''), while 10 refers to entity type ``books''.
\end{enumerate}
\noindent To provide a fair comparison between the \textit{candidate ranker} model of ALGES proposed by \cite{koncel2015parsing} and our approach (see \secref{ch_mwp:sec:model}), we use both
the same constraint configuration, and
also consider only the top 100 equations produced by the candidate generator.
As in ALGES, we report the coverage as \textit{\MaxILP{}} in our results section (see \secref{ch_mwp:sec:results}). 
Additionally, we include in our result tables the performance of the \textit{\TopILP{}} approach, which consists of selecting the highest scored candidate by the ILP solver. 
This score allows us to estimate the impact of the \textit{candidate ranker} component.

\subsection{Candidate ranker}
\label{ch_mwp:sec:model}
\begin{figure}[!ht]
\centering
\includegraphics[width=.60\columnwidth,clip]{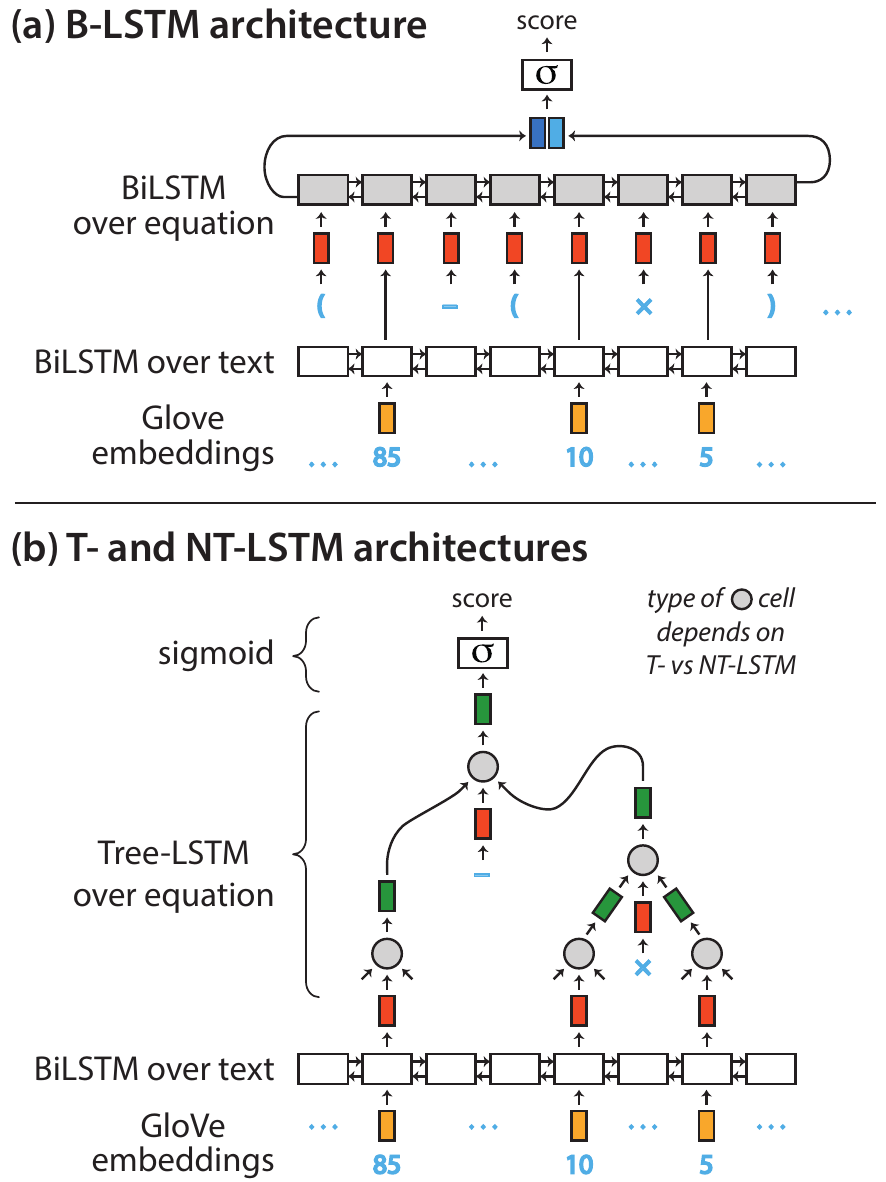}
\captionsetup{singlelinecheck=off}
\caption[Models for scoring equations]{Models for scoring equations, taking the text and the equation from \figref{ch_mwp:fig:example} to score (\eg $85 - (10 \times 5)$) as input:
    \protect\begin{enumerate*}[(i)]
    \item a word embedding layer at the bottom,
    \item a \biLSTM{} layer over the text, and
    \item a top layer that encodes the equation.
    \protect\end{enumerate*}
 For the latter we consider either
  \protect\begin{enumerate*}[label=\textbf{(\alph*)}]
     \item \label{ch_mwp:arch:bilstm} a sequential \biLSTM{} (\BLSTM{} architecture), 
     or
     \item \label{ch_mwp:arch:treelstm} a structured \TreeLSTM{} (\TLSTM{} and \NTLSTM{} architectures).
  \protect\end{enumerate*}
}
\label{ch_mwp:fig:model}
\end{figure}

\noindent Our proposed candidate ranker model architecture is sketched in~\figref{ch_mwp:fig:model} and comprises: 
\begin{enumerate*}[(i)]
    \item a word embedding layer,
    \item a bidirectional \LSTM{} layer (\biLSTM{}) over the text, and
    \item an additional layer that encodes the equation, using either \biLSTM{} (\BLSTM{} model) or \TreeLSTM{} (\TLSTM{} and \NTLSTM{} models) based approaches, detailed below.
\end{enumerate*}

The input to our model is a \textcolor{wordcolor}{\textbf{sequence of tokens}} of length $N$, $W=\{w_1,..., w_N\}$ of the arithmetic word problem, which we pass through an \textcolor{embeddingcolor}{\textbf{embedding layer}} to obtain embedded representations $X=\{x_1,..., x_N\}$ where 
$x_t \in \mathbb{R}^{d_1}$.
We adopt a \biLSTM{} to obtain \textcolor{biLSTMcolor}{\textbf{contextual representations}} of the tokens. The following is the formal representation of the first \LSTM{} \cite{hochreiter1997long} layer used to produce the representation referred to as ``\textit{\biLSTM{} over text}'' in \figref{ch_mwp:fig:model}: 

\begin{align}
    i_t = \sigma \left(W_i x_t + U_i h_{t-1} + b_i \right) \label{ch_mwp:lstm:1} \\ 
    o_t = \sigma \left(W_o x_t + U_o h_{t-1} + b_o \right) \label{ch_mwp:lstm:2} \\ 
    f_t = \sigma \left(W_f x_t + U_f h_{t-1} + b_f \right) \label{ch_mwp:lstm:3} \\ 
    u_t = \mathrm{tanh} \left(W_u x_t + U_u h_{t-1} + b_u \right) \label{ch_mwp:lstm:4} \\
    c_t = f_t \odot c_{t-1} + i_t \odot u_t  \label{ch_mwp:lstm:5}
    \\
    h_t = o_t \odot \mathrm{tanh}(c_t)  \label{ch_mwp:lstm:6}
\end{align}
\noindent where $t \in \{1,..., N\}$ represents a particular recursive execution time step and
$h_t \in \mathbb{R}^{d_2}$ is the \LSTM{} hidden state. The advantage of using the \LSTM{}-based structure instead of a simpler recursive formulation, such as $h_t = \tanh (W x_t + Uh_{t-1} + b)$, is that an \LSTM{} model avoids the problems of exploding or vanishing gradients during the training process discussed in \cite{hochreiter1997long,bengio1994learning}. This is achieved by using additional weight matrices 
and \textit{gates} $\sigma$ in \equsrefrange{ch_mwp:lstm:1}{ch_mwp:lstm:3} in order to regulate the amount of information from previous execution steps $h_{t-1}$ and current input $x_t$ that affect the current state $h_t$.\footnote{For a more detailed description of the \LSTM{} architecture please refer to \cite{hochreiter1997long}.}
More concretely, $W_i$, $W_f$, $W_o$, $W_c \in \mathbb{R}^{d_2\times d_1}$ and $U_i$, $U_f$, $U_o$, $U_c \in \mathbb{R}^{d_2\times d_2}$ are the weight matrices related to different LSTM gates, and $b_i$, $b_f$, $b_o$, $b_c \in \mathbb{R}^{d_2}$ are the respective biases. 
In our experiments we initialize $x_t$ with GloVe word embeddings \cite{pennington:14} and keep them static during training. These \textit{GloVe embeddings} 
are depicted at the bottom of graphs \textbf{(a)} and \textbf{(b)} in \figref{ch_mwp:fig:model}.
In order to obtain the \biLSTM{} representation (\textit{``\biLSTM{} over text''} in \figref{ch_mwp:fig:model}), we run two \LSTMs{} in different directions and concatenate the respective hidden states. This results in $N$ hidden state representations $H=\{h_1^{(b)}, ..., h_N^{(b)}\}$ where $h_i^{(b)} \in \mathbb{R}^{d_3}$ and $d_3 = 2 \cdot d_2$. Using the input in $H$, we propose two different models to encode the candidate equations referred to as \textbf{(a)} and \textbf{(b)} in \figref{ch_mwp:fig:model}, and explained below: 

\noindent \textbf{\ref{ch_mwp:arch:bilstm} Sequential \BLSTM{}}: We perform an in-order traversal of the expression tree to obtain a sequential representation of the equation (\eg $\left(85-(10\times5)\right))$
that is encoded using a second \biLSTM{} (see \textit{``BiLSTM over equation''} in \figref{ch_mwp:fig:model}\ref{ch_mwp:arch:bilstm}). 
We use as input the hidden state representations $H$ calculated above for the numbers and (trainable) embeddings 
$O \ =\{o_{-}, o_{+}, o_{\div}, o_{\times}, o_{(}, o_{)} \}$ for the operators ($-, +, \div, \times$) and opening/closing parentheses. More formally, the input to \biLSTM{} is represented by 
$X^E = \{x^e_1, ..., x^e_K\}$ where $x^e_t \in \{ H \cup O \}$ , $x^e_t \in \mathbb{R}^{d_3}$
and $K$ is the number of tokens in the equation, including parentheses and operations. E.g., the equation $(85-(10 \times 5))$ contains $9$ tokens. In terms of the formal notation of LSTM in \equsrefrange{ch_mwp:lstm:1}{ch_mwp:lstm:6}, each $x^e_t$ corresponds to input vector $x_t$. In order to obtain a score for ranking the equation, we concatenate the last (left and right) hidden states of the \biLSTM{} producing a vector of dimensionality $d_4$, and then apply a linear transformation followed by a \textit{sigmoid} function.  

\noindent \textbf{\ref{ch_mwp:arch:treelstm} \TreeLSTM{}}: 
We base our implementation on the \TreeLSTM{} architecture proposed by \cite{tai2015improved}.
This architecture is based on the LSTM formulation described in \equsrefrange{ch_mwp:lstm:1}{ch_mwp:lstm:6}, but instead of being linearly linked, the input to a particular \LSTM{} cell can come from different child step \LSTM{} executions. More formally, we can describe the \TLSTM{} structure as follows: 

\begin{align}
    \tilde{h}_t &= \sum_{k\in \{L, R\}}{h_{t-1}^{k}} \label{ch_mwp:tlstm:0} \\
    i_t &= \sigma \left(W_i x_t + U_i \tilde{h}_t + b_i \right) \label{ch_mwp:tlstm:1} \\ 
    o_t &= \sigma \left(W_o x_t + U_o \tilde{h}_t + b_o \right) \label{ch_mwp:tlstm:2} \\ 
    f_{t}^{k} &= \sigma \left(W_f x_t + U_f h_{t-1}^k + b_f \right) \label{ch_mwp:tlstm:3} \\ 
    u_t &= \mathrm{tanh} \left(W_u x_t + U_u \tilde{h}_t + b_u \right) \label{ch_mwp:tlstm:4} \\
    c_t &= i_t \odot u_t + \sum_{k\in \{L,R\}}{f_{t}^k} \odot c_{t-1}^k \label{ch_mwp:tlstm:5} \\
    h_t &= o_t \odot \mathrm{tanh}\left(c_t\right) \label{ch_mwp:tlstm:6}
\end{align}

\noindent where $\{L,R\}$ is the set that consists of left ($L$) and right ($R$) child nodes for the current execution node at step $t$. More specifically, a particular execution step $t$ corresponds to the respective arithmetic operation in the expression tree (see \figref{ch_mwp:fig:example}\ref{ch_mwp:fig1c}). This step takes as input the cell ($c$) and hidden ($h$) states of previous execution step ($t-1$) for each of the child nodes ($\{L,R\}$) that correspond to left and right operands in the expression tree. This execution process is recursive: each of the execution steps produces as output a hidden state $h_t$ (\equref{ch_mwp:tlstm:6}) which is used by the parent execution step recursively in \equref{ch_mwp:tlstm:0} either as left ($h_{t-1}^L$) or right ($h_{t-1}^R$) child. Additionally, a \textit{cell state} $c_t$ is passed across the execution steps, and contains a summarized historic information of the tree traversal\footnote{Post-order traversal is used, since it reflects the order of operator execution in an arithmetic equation to obtain the final result.} operations performed so far. Similarly as with LSTM, a \textit{forget} gate $f_{t}^k$, \textit{input} ($i_t$) and \textit{update} ($u_t$) gates are used to determine which historic information is kept (forget gate) and which new information is added (input/update gates) to the cell state.
 $W_i$, $W_o$, $W_f$, $W_u \in \mathbb{R}^{d_4 \times d_3}$ together with $U_i$, $U_o$, $U_f$, $U_u \in \mathbb{R}^{d_4 \times d_4}$ are the weight matrices that transform the inputs $x_t \in \mathbb{R}^{d_3}$, the current hidden state $\tilde{h}_t \in \mathbb{R}^{d_4}$ and the children's hidden states 
 $h_{t-1}^k \in \mathbb{R}^{d_4}$, by means of
 the \TreeLSTM{} gate representations. As depicted in \figref{ch_mwp:fig:model}\ref{ch_mwp:arch:treelstm}, the inputs $x_t$ 
 to the leaf nodes are the hidden state representations in $H$ (coming from \textit{``BiLSTM over text''} in \figref{ch_mwp:fig:model}\ref{ch_mwp:arch:treelstm}) on the positions where the numbers occur in the problem statement. The input $x_t$ to the inner nodes, on the other hand, are one of the randomly initialized operation embeddings $O = \{o_{-}, o_{+}, o_{\div}, o_{\times} \}$ 
 depending on the operation represented by the node. 
 This contrasts with the original setup proposed in \cite{tai2015improved} where the input $x_t$ always comes from the word representation in the sentence. By using a separate operation embeddings set $O$ as input, we expect our model to be able to capture a semantic representation for each of the different operations $o \in O$.
 The \TreeLSTM{} model finally outputs the hidden state for the root of the expression tree (\ie the last executed operation), which is then passed through a sigmoid to deliver the score for a particular candidate arithmetic expression.

While \TLSTM{} allows to encode the equation information in a tree structure, it is symmetric in its child nodes. This is because 
the hidden states of the children are first summed up in \equref{ch_mwp:tlstm:0}
 before applying the linear transformation and the gate activation functions. This could be problematic for non-commutative operations ($-$ and $\div$) where the result depends on the order of the operands. The reason for this is that \equref{ch_mwp:tlstm:0} is commutative with respect to child nodes. Thus, given two child nodes $k \in \{L,R\}$ we have that $\tilde{h}_t = h_{t-1}^L + h_{t-1}^R = h_{t-1}^R + h_{t-1}^L$. As a consequence, the affine transformations $U_i$, $U_o$, and $U_u$ in \eqs~\ref{ch_mwp:tlstm:1}, \ref{ch_mwp:tlstm:2} and \ref{ch_mwp:tlstm:4} cannot capture the order of the states of the input nodes. Furthermore, since there is only one weight matrix $U_f$ for both $h_{t-1}^L$ and $h_{t-1}^R$ in \equref{ch_mwp:tlstm:3}, it can not apply a different affine transformation for left and right child nodes. This makes the \TLSTM{} model indifferent to the order of the arguments of the operations in a particular expression tree. 
 Therefore, we introduce a second model, called \NTLSTM{}, that uses distinct weight matrices to transform each of the children's hidden states. More formally, the gate definition in \NTLSTM{} is as follows: 

\begin{align}
    i_t &= \sigma \left(W_i x_t + \sum_{k\in\{L,R\}}{U_i^{k} h_{t-1}^{k}} + b_i \right) \label{ch_mwp:ntlstm:1} \\ 
    o_t &= \sigma \left(W_o x_t + \sum_{k\in \{L,R\}}{U_o^{k} h_{t-1}^{k}} + b_o \right) \label{ch_mwp:ntlstm:2} \\ 
    f_{t}^k &= \sigma \left(W_f x_t + \sum_{l \in \{L,R\}}{U_f^{kl} h_{t-1}^{l}} + b_f \right) \label{ch_mwp:ntlstm:3} \\ 
    u_t &= \mathrm{tanh} \left(W_u x_t + \sum_{k \in \{L,R\}}{U_u^{k} h_{t-1}^{k}} + b_u \right) \label{ch_mwp:ntlstm:4} \\
    c_t &= i_t \odot u_t + \sum_{k \in \{L,R\}}{f_{t}^k \odot c_{t-1}^k} \label{ch_mwp:ntlstm:5} \\
    h_t &= o_t \odot \mathrm{tanh}\left(c_t\right) \label{ch_mwp:ntlstm:6}
\end{align} 

\noindent where, similarly as for \TLSTM{}, $\{L,R\}$ is the set of child nodes. By introducing different weights $U$ for each of the child node states $h_{t-1}^{k}$, we make sure that the model can differentiate between the order of the operands. This is because now each of the affine transformations $U_i^{(l)}$, $U_o^{(l)}$ 
and $U_t^{(l)}$ is different for each input child hidden state $h_{t-1}^{l}$ in \eqs~\ref{ch_mwp:ntlstm:1}, \ref{ch_mwp:ntlstm:2} and \ref{ch_mwp:ntlstm:4}. Similarly, each of the children's ($k \in \{L,R\}$) forget gates $f_t^k$ contains now two affine transformations $U_f^{kl}$ ($l \in \{L,R\}$), one for each child. This way, the model can prioritize (components of $f_t^k$ close to $1$) or inhibit (components of $f_t^k$ close to $0$) separately the input of a particular child $k$ based on the state of another child $l$ ($k \neq l$). This can be useful when the state of one of the operands (\eg influenced by the words that surround a particular number in text) has a strong indication of some operation, while the state of the other has very little evidence.
As we will show in~\secref{ch_mwp:sec:results}, the use of \NTLSTM{} makes a big difference compared to the performance of \TLSTM{} for equations involving non-commutative operations. 

\section{Experimental setup}
\label{ch_mwp:sec:experimental_setup}
\noindent We evaluate the proposed models 
(code publicly available\footnote{\url{https://github.com/klimzaporojets/arithmetic-word-problems}})
on the SingleEQ dataset introduced by~\cite{koncel2015parsing}. SingleEQ consists of 1,117 sentences and 15,292 words, and includes 508 arithmetic problems of varying complexity (\ie equations with single or multiple operators). Each of the word problems is mapped to a single correct equation with one unknown. These equations include one or more of the following operators: multiplication ($\times$), division ($\div$), subtraction ($-$), and addition ($+$). The data was gathered from the following grade-school websites: \url{http://math-aids.com}, \url{http://k5learning.com}, and \url{http://ixl.com} as well as from a subset of problems from \cite{kushman2014learning}.
To obtain results comparable to previous work, we perform 5-fold cross-validation using the original splits defined in~\cite{koncel2015parsing}. Similar to the work of \cite{koncel2015parsing} and \cite{wang2018mathdqn}, we report performance using the overall accuracy metric. 
The training/testing process is run for 5 different splits, in each one a separate fold is left as test set. This way, our results are reported on the whole SingleEQ dataset by concatenating the predictions of \textit{test} folds across the splits. In total, we train 25 models with different seeds (5 for each split) and report average and standard deviation in Tables~\ref{ch_mwp:tab:results_full}--\ref{ch_mwp:tab:results_complex} and \ref{ch_mwp:tab:results_asym} in \secref{ch_mwp:sec:results}. Furthermore, we tune the neural net hyperparameters independently for each of the splits on the validation set that consists of  20\% randomly selected arithmetic problems in each of the train folds. Due to limited resources that prevented us to perform a complete grid search, we conduct the hyperparameter tuning in steps. More specifically,
in each step we perform a grid search on two hyperparameters that we identified as most correlated with each other. \Tabref{ch_mwp:tab:hyper_search_space} summarizes our hyperparameter search space for each of the sequential tuning steps. Besides the usual hyperparameters (\ie learning rate, batch size and dropout) tuning, we also adjust the dimensionalities $d_3$ (Dim LSTM) of the first BiLSTM layer (indicated as ``\textit{\biLSTM{} over text}'' in \figref{ch_mwp:fig:model}), and $d_4$ (Dim Encoder) of either the sequential \biLSTM{}  (``\textit{\biLSTM{} over equation}'' in \figref{ch_mwp:fig:model}) or the tree-based \NTLSTM{} models' encoder layers (``\textit{\TreeLSTM{}}'' in \figref{ch_mwp:fig:model}). The best hyperparameters are chosen after training for 75 epochs for each of the cross-validation splits independently.
\begin{table*}[!ht]
\centering
\resizebox{0.9\textwidth}{!}{%
\begin{tabular}{c@{\hspace{.7cm}}ccccc} 
 \toprule
 \multirow{2}{*}{Step} & 
 \multicolumn{5}{c}{Hyperparameters}  \\
\cline{2-6}
  & \multicolumn{1}{c}{Learning Rate} & \multicolumn{1}{c}{Batch Size} & Dim LSTM & Dim Encoder & Dropout \\
 \midrule
1 & \{\num{3e-4}, \num{1e-4}\} & \{\num{64}, \num{128}\} & - & - & - \\
2 & - & - & \{\num{256}, \num{512}\} & - &  \{\num{0.3}, \num{0.4}\} \\
3 & - & - & - & \{\num{256}, \num{512}\} & \{\num{0.3}, \num{0.4}\} \\
\bottomrule
\end{tabular}
}
\caption[The range of the hyperparameter search space]{The range of the hyperparameter search space for each of the hyperparameter tuning steps for each of the cross-validation splits of SingleEQ dataset.}
\label{ch_mwp:tab:hyper_search_space}
\end{table*}

Furthermore, we partition the dataset into several subsets to investigate the effect of varying problem complexity on the models' performances.
These different subsets are characterized in \tabref{ch_mwp:tab:subsets}. We form three main categories:
\begin{enumerate*}[(i)]
\item \textbf{Full:} the whole dataset is included in this setting,
    \item \textbf{Complexity:} two subsets (\ie Single, Multi) are formed based on the number of operators in the solution's equation, and
    \item \textbf{Symmetry:} four main subsets, namely Single$_{\textrm{sym}}$, Single$_{\textrm{asym}}$, Multi$_{\textrm{sym}}$, and Multi$_{\textrm{asym}}$ are formed to indicate whether the solution's equation contains single/multiple symmetric ($\times$ and $+$) or asymmetric ($\div$ and $-$) operations.
\end{enumerate*}

\begin{table}[t]
    \centering
    \resizebox{0.7\columnwidth}{!}{
    \begin{tabular}{lcc}
    \toprule
    {\textbf{Subset}} & \textbf{Equation types} & \textbf{\# Problems} \\
    \midrule
    Full & All operators & 508 \\ 
    \midrule
    Single & Single operator & 390 \\ 
    Multi & Multiple operators & 118 \\ 
    \midrule
    Single$_{\text{sym}}$ & Single symmetric operators & 208 \\ 
    Multi$_{\text{sym}}$ & Multiple symmetric operators  & 68 \\ 
    Single$_{\text{asym}}$ & Single asymmetric operators & 182 \\ 
    Multi$_{\text{asym}}$ & Multiple asymmetric operators  & 50 \\ 
   \bottomrule
    \end{tabular}
    }
    \caption{The defined subsets of the SingleEQ dataset with varying degrees of complexity.}
    \label{ch_mwp:tab:subsets}
\end{table}

We hypothesize that our \TreeLSTM{} models will exhibit stronger performance on subsets involving multiple and/or non-commutative operations (Multi, Multi$_\text{sym}$, Multi$_\text{asym}$), since they should be able to better capture the semantic relationships between operator nodes encoded in a tree structure. We also expect a significant difference between \TLSTM{} and \NTLSTM{} architectures on subsets involving non-commutative operations (Single$_\textrm{asym}$ and Multi$_\textrm{asym}$).
By using different weight matrices to transform each of the children's states (see Eqs.~\ref{ch_mwp:ntlstm:1}--\ref{ch_mwp:ntlstm:4} of the \NTLSTM{} in \secref{ch_mwp:sec:model} for more details), the \NTLSTM{} model should be able to capture the order of the operands and link 
the resulting structural information of a particular
non-commutative
mathematical expression
to the semantic representation of the problem statement. 

We obtain the top-$100$ equation-trees using the ILP solver of~\cite{koncel2015parsing}, which we rank using scores provided by our proposed model (see~\secref{ch_mwp:sec:model}). Training of our model is performed using the Adam optimizer~\cite{kingma:14}. As a bottom token representation layer, we use pre-trained 100-dimensional ($d_1 = 100$) GloVe embeddings~\cite{pennington:14}\footnote{ \url{https://nlp.stanford.edu/projects/glove/}} which we keep static during the training process. 
\section{Results}
\label{ch_mwp:sec:results}

\noindent In this section, we evaluate the performance of our proposed models on the SingleEQ dataset. Besides the performance on the full dataset, we are particularly interested in evaluating how each architecture behaves when evaluated on arithmetic problems of varying complexity. We assume that the problems become more complex
\begin{enumerate*}[(i)]
    \item as the number of needed mathematical operators grows, and
    \item when the used operators are non-commutative (asymmetric).
\end{enumerate*}
We hypothesize that our structured \TreeLSTM{}-based approach is better suited to solve 
the aforementioned
complex problems. 
In order to demonstrate this, we perform an extensive evaluation (Tables~\ref{ch_mwp:tab:results_full}--\ref{ch_mwp:tab:results_complex} and \ref{ch_mwp:tab:results_asym}) of our models on subsets of different degree of complexity as defined in \tabref{ch_mwp:tab:subsets}. Furthermore, in all of the result tables we include the potential maximum accuracy that can be achieved when using the candidates from the ILP \textit{candidate generator} (\MaxILP{}). This allows us to estimate how much  improvement can still be achieved by \textit{candidate ranker}. Conversely, in order to evaluate the impact of \textit{candidate ranker} models, we also report the accuracy achieved when picking the top-weighted candidate by ILP solver (\TopILP{}).

\noindent\textbf{Comparison on the Full dataset}:~\tabref{ch_mwp:tab:results_full} shows the results of the evaluated systems on the Full 
SingleEQ dataset. The proposed models are the
\begin{enumerate*}[(i)]
    \item \BLSTM{},
    \item \TLSTM{}, and  
    \item \NTLSTM{} as presented in~\secref{ch_mwp:sec:model}. 
\end{enumerate*}
 Clearly, all newly proposed architectures outperform previous methods. 
 \begin{table}[t]
\centering
\resizebox{0.7\columnwidth}{!}{%
\begin{tabular}{@{\extracolsep{4pt}}ccccccccc@{}} 
 \toprule
 \multicolumn{1}{c}{Model} & \multicolumn{1}{c}{Features} & \multicolumn{1}{c}{Trees}  & \multicolumn{1}{c}{Accuracy ($\%$)}    \\
 \midrule
\cite{hosseini2014learning} &\cmark&\xmark& 48.00   \\ 
\cite{wang2018mathdqn} &\cmark&\cmark& 52.96  \\ 
\cite{roy2015solving} &\cmark&\cmark & 66.38  \\ 
\cite{roy2017unit} &\cmark&\cmark & 72.25  \\ 
ALGES &\cmark&\cmark& 72.39 \\
\toprule
\MaxILP{} & - & -  &  91.34  \\
\TopILP{}  & - & - &  52.56  \\
\toprule
\BLSTM{}  &\xmark&\xmark  &  74.88$\pm$0.64  \\
\TLSTM{}&\xmark&\cmark & 74.88$\pm$1.06  \\ 
\NTLSTM{} &\xmark&\cmark& \textbf{75.47$\pm$0.62} \\
\bottomrule
\end{tabular}
}
\caption[Accuracy attained by the proposed and state-of-the-art methods on SingleEQ]{Accuracy attained by the proposed and state-of-the-art methods on the \emph{Full} 
SingleEQ dataset. The \cmark{} and \xmark{} symbols indicate whether or not a model adopts hand-crafted features (`Features') or tree-structured encoding of the equations (`Trees'). The best result is typeset in \textbf{bold}. 
}
\label{ch_mwp:tab:results_full}
\end{table}
Concretely, our methods are able to outperform strong baselines on the task, reporting an accuracy improvement of more than 3\% without relying on hand-crafted features \cite{hosseini2014learning, koncel2015parsing, roy2015solving, roy2017unit}. As detailed later on in this section (see analysis of \tabref{ch_mwp:tab:results_complex} and \tabref{ch_mwp:tab:results_asym}), most of this improvement with respect to the current state-of-the-art \cite{koncel2015parsing} comes from an increased performance on the more complex arithmetic word problems that involve non-commutative and multiple operations. This supports our original hypothesis that tree-based architectures are superior in representing mathematical operations between operands, specially when the mathematical expressions involve multiple operations. The hand-crafted features, used in previous works, are usually related to terms indicating specific operations and thus if they are not detected in the data, the system cannot generalize well on out-of-domain mathematical descriptions. This also applies to recent neural-based methods (see, \eg \cite{wang2018mathdqn}) where explicitly defined features are encoded in the neural structure. Furthermore, in order to ensure the validity of the differences between our proposed approaches, we carry out a bootstrap significance analysis \cite{efron1994introduction} by sampling with replacement the results of \BLSTM{}, \TLSTM{}, and \NTLSTM{} models 10,000 times. We compare the performance with respect to the \NTLSTM{} model in \tabref{ch_mwp:tab:results_full}. 
We observe that, while our \NTLSTM{} model seems to outperform \TLSTM{} and \BLSTM{} models, this difference in performance is not significant.

\noindent \textbf{Comparison for different problem complexity}: \tabref{ch_mwp:tab:results_complex} compares our models with ALGES \cite{koncel2015parsing} (\ie the best performing state-of-the-art model of \tabref{ch_mwp:tab:results_full}), for subsets of different complexity levels (defined in~\tabref{ch_mwp:tab:subsets}).~We use bootstrap significance testing to estimate the degree of certainty between the lower performing models and the best performing one in each of the subsets. We indicate significant differences with p-values below the $1\%$, $5\%$, and $10\%$ level (respectively denoted with $\ddag$, $\dag$, and $\star$) in order to identify models performing significantly different from the best performing model in each of the subsets.

\begin{table*}[t]
\centering
\resizebox{1.0\textwidth}{!}{%
\begin{tabular}{c@{\hspace{.7cm}}cc@{\hspace{.7cm}}cc@{\hspace{.7cm}}cc} 
 \toprule
 \multirow{2}{*}{Model} & 
 \multicolumn{2}{c}{Complexity}  & \multicolumn{2}{c}{Symmetric} & \multicolumn{2}{c}{Asymmetric}  \\
\cline{2-3}
\cline{4-5}
\cline{6-7}
  & \multicolumn{1}{c}{Single} & \multicolumn{1}{c}{Multi} &Single$_{\text{sym}}$ & Multi$_{\text{sym}}$  & Single$_{\text{asym}}$ & Multi$_{\text{asym}}$   \\
 \midrule
\MaxILP{}  & 93.33 & 84.75 & 94.71 & 83.82 & 91.76 & 86.00 \\
\TopILP{}  & 56.41 & 39.83 & 53.85 & 69.12 & 59.34 & 0.00 \\
\midrule
ALGES  & 77.69$^\ddag$ & 54.70$^\ddag$ & \textbf{89.90} & 72.06 & 63.74$^\ddag$ & 30.64$^\ddag$ \\
\BLSTM{}  & 79.59$\pm$0.72 & 59.32$\pm$2.34 & 80.87$\pm$0.64$^\ddag$ & 69.12$\pm$2.08$^\ddag$ & 78.13$\pm$1.36 & \textbf{46.00$\pm$4.38} \\
\TLSTM{}  & 79.59$\pm$1.24 & 59.32$\pm$1.61 & 81.35$\pm$0.98$^\ddag$ & \textbf{72.35$\pm$1.44} & 77.58$\pm$2.72 & 41.60$\pm$2.33$^\star$ \\ 
\NTLSTM{} & \textbf{80.21$\pm$0.95} & \textbf{59.83$\pm$1.75} & 81.35$\pm$1.44$^\ddag$ & 71.17$\pm$2.20 & \textbf{78.90$\pm$2.13} & 44.40$\pm$4.96 \\
\bottomrule
\end{tabular}
}
\caption[Accuracy attained by the proposed and state-of-the-art methods on the defined subsets of SingleEQ]{Comparison of the proposed methods with the state-of-the-art on the SingleEQ dataset in terms of accuracy. \textbf{Bold} font indicates the best results for each subset of SingleEQ (see~\tabref{ch_mwp:tab:subsets}). The markers $\star$, $\dag$, $\ddag$ respectively indicate the achieved bootstrap significance levels $\alpha$ \textless 0.1, \textless 0.05 and \textless 0.01 with respect to the best performing model in each of the subsets.}
\label{ch_mwp:tab:results_complex}
\end{table*}
We observe that our newly proposed models do not significantly differ among each other for solving problems involving single (Single, Single$_\mathrm{sym}$, and Single$_\mathrm{asym}$ subsets) operations. Conversely, on the problem subset requiring multiple commutative operations in their solution (Multi$_\mathrm{sym}$), our tree-based \TLSTM{} significantly outperforms  the sequential \BLSTM{} model, suggesting a potential benefit in using tree-based models to solve the problems involving multiple operations. For the subset involving multiple non-commutative operations (Multi$_\mathrm{asym}$) the \BLSTM{} and \NTLSTM{} models outperform the \TLSTM{} model, indicating a potential limitation of the latter in dealing with non-commutative operations, due to its symmetrical structure in its child nodes (a single weight matrix is used on the sum of children's states $\tilde{h}_t$ as described in \secref{ch_mwp:sec:model}). We were surprised by an overall good performance of our sequential \BLSTM{} model, specially on Multi$_\mathrm{asym}$ subset, where it performs on par with the potentially more expressive \NTLSTM{} model. This fact also motivated us to explore the robustness of our models against additional asymmetric noise (see further analysis in the next paragraphs corresponding to the results in \tabref{ch_mwp:tab:results_asym}). 
 
 The results in \tabref{ch_mwp:tab:results_complex} further show that the feature-based ALGES model has competitive performance on problems requiring single and/or non-commutative operators in the solution equations. In fact, it significantly outperforms all our models on the Single$_{\textrm{sym}}$ dataset and is only marginally outperformed by our tree-based \TLSTM{} model on Multi$_{\mathrm{sym}}$.
 This suggests that the feature-based ALGES is able to explicitly capture symmetric operations by focusing on carefully engineered features. However, we observe a large drop in performance of ALGES on problems that require non-commutative (asymmetric) operations to be solved. 
 This is showcased by a difference of more than 15\% accuracy points on Single$_{\mathrm{asym}}$ and Multi$_{\mathrm{asym}}$ subsets in \tabref{ch_mwp:tab:results_complex}. This validates our initial intuition that feature-based models fall short to capture the reasoning necessary to address
 problems that require more complex (non-commutative and multiple) operators. 

\begin{table*}[t]
\centering
\resizebox{1.0\textwidth}{!}{%
\begin{tabular}{ccccccccc} 
 \toprule
\multirow{2}{*}{Candidates} & \multirow{2}{*}{Metric} &
 \multicolumn{7}{c}{Subsets}  \\
\cline{3-9}
    &  & \multicolumn{1}{c}{Full} & \multicolumn{1}{c}{Single} & \multicolumn{1}{c}{Multi} &Single$_{\text{sym}}$ & Multi$_{\text{sym}}$  & Single$_{\text{asym}}$ & Multi$_{\text{asym}}$   \\
 \midrule
\multirow{2}{*}{ILP}  & Correct & 2.53 & 1.44 & 6.13 & 1.89 & 7.72 & 0.92 & 3.96 \\
     & Incorrect & 12 & 2.9 & 42.08 & 2.48 & 28.43 & 3.38 & 60.64 \\
    \midrule 
\multirow{2}{*}{ILP + Asym}  & Correct & 2.41 & 1.44 & 5.62 & 1.89 & 7.66 & 0.92 & 2.84 \\ 
  & Incorrect & 15.08 & 4.06 & 51.5 & 3.57 & 35.43 & 4.62 & 73.36 \\
    \midrule 
  & $\Delta$ Correct & $-$4.74\% & 0.00\% & $-$8.32\% & 0.00\% & $-$0.78\% & 0.00\% & $-$28.28\% \\ 
  & $\Delta$ Incorrect & 25.67\% & 40.00\% & 22.39\% & 43.95\% & 24.62\% & 36.69\% & 20.98\% \\
\bottomrule
\end{tabular}
}
\caption[Statistics of SingleEQ with additional equations with asymmetric operators]{This table illustrates the difference in average number of \textit{Correct} and \textit{Incorrect} candidate equations per problem between the original \textit{ILP} candidate generation process and the one obtained by adding noisy equations with asymmetric operators (\textit{ILP + Asym}).}
\label{ch_mwp:tab:asymmetric_noise}
\end{table*}

  \noindent \textbf{Robustness against asymmetric noise}: The results analyzed so far are based on scoring the candidates generated by the ILP component introduced in \cite{koncel2015parsing}. However, this component already significantly reduces the number of incorrect candidates, particularly those involving asymmetric operators (\eg by removing candidate equations that produce negative or fractional results as described in \secref{ch_mwp:sec:candidate_generator}). In order to evaluate the robustness of the proposed models, we train and evaluate them on a noisy asymmetric candidate set where we add all possible permutations to the equations involving non-commutative operators. For example, if a particular candidate equation is $x=8/2$, we would also add $x=2/8$ to the candidate set. \Tabref{ch_mwp:tab:asymmetric_noise} shows the statistics of the noisy dataset (ILP + Asym) with respective deltas that indicate the percentage points (\%) of increase/decrease in the average number of correct/incorrect candidate equations per problem with respect to the original ILP-generated candidate set. We observe a significant increase in the number of incorrect candidates for all subsets, as well as a drop in average number of correct equations for the subsets involving asymmetric operations (Multi and Multi$_{\mathrm{asym}}$). This is because, similarly as in the original \textit{ILP} setup, we only consider the first 100 generated candidates, which in \textit{ILP + Asym} include more incorrect equations, leaving many correct ones out. This results in a lower correct/incorrect ratio that makes it more challenging for the evaluated models to find the right mathematical expression to solve a particular problem. 
\begin{table*}[t]
\centering
\resizebox{1.0\textwidth}{!}{
\begin{tabular}{cc@{\hspace{.7cm}}cc@{\hspace{.7cm}}cc@{\hspace{.7cm}}cc} 
 \toprule
  \multirow{2}{*}{Model} & \multirow{2}{*}{Full} &
 \multicolumn{2}{c}{Complexity}  & \multicolumn{2}{c}{Symmetric} & \multicolumn{2}{c}{Asymmetric}  \\
\cline{3-4}
\cline{5-6}
\cline{7-8}
  & & \multicolumn{1}{c}{Single} & \multicolumn{1}{c}{Multi} &Single$_{\text{sym}}$ & Multi$_{\text{sym}}$  & Single$_{\text{asym}}$ & Multi$_{\text{asym}}$   \\
 \midrule
\MaxILP{}  & 91.14 & 93.33 & 83.90 & 94.71 & 83.82 & 91.76 & 84.00 \\
\TopILP{} & 52.56 & 56.41 & 39.83 & 53.85 & 69.12 & 59.34 & 0.00 \\
\midrule 
ALGES  & 68.44$^\ddag$ & 75.90$^\dag$ & 43.59$^\ddag$ & \textbf{85.58} & 61.76$^\ddag$ & 64.83$^\ddag$ & 18.36$^\ddag$ \\
\BLSTM{}     & 72.99$\pm$1.14 & \textbf{78.21$\pm$0.97} & 55.76$\pm$2.10$^\ddag$ & 83.36$\pm$1.20$^\dag$ & 71.76$\pm$3.40$^\ddag$ & 72.30$\pm$2.37 & 34.00$\pm$2.19$^\star$ \\
\TLSTM{}  & 57.95$\pm$1.34$^\ddag$ & 61.69$\pm$1.49$^\ddag$ & 45.59$\pm$1.25$^\ddag$ & 80.58$\pm$2.44$^\ddag$ & 72.65$\pm$2.20$^\ddag$ & 40.11$\pm$0.92$^\ddag$ & 8.80$\pm$0.98$^\ddag$ \\ 
\NTLSTM{}  & \textbf{73.19}$\pm$\textbf{0.93} & 76.97$\pm$1.02$^\dag$ & \textbf{60.67}$\pm$\textbf{1.15} & 80.76$\pm$2.37$^\ddag$ & \textbf{76.47$\pm$0.93} & \textbf{72.63}$\pm$\textbf{1.61} & \textbf{39.20}$\pm$\textbf{2.40} \\
\bottomrule
\end{tabular}
}
\caption[Impact of additional equations with asymmetric operators on accuracy]{Comparison of the proposed methods with the state-of-the-art model (\ie ALGES) on the SingleEQ dataset in terms of accuracy evaluated on candidate equations generated using \textit{ILP + Asym} procedure (see \tabref{ch_mwp:tab:asymmetric_noise}). \textbf{Bold} font indicates the best results for each subset of SingleEQ (see~\tabref{ch_mwp:tab:subsets}). The markers $\star$, $\dag$, $\ddag$ respectively indicate the achieved bootstrap significance levels $\alpha$ \textless 0.1, \textless 0.05 and \textless 0.01 with respect to the best performing model in each of the subsets.}
\label{ch_mwp:tab:results_asym}
\end{table*}    
\Tabref{ch_mwp:tab:results_asym} compares our models with the best performing state-of-the-art model (\ie ALGES) on candidates generated in the \textit{ILP + Asym} setting. Compared to the results presented in \tabref{ch_mwp:tab:results_complex}, we observe a sharp decrease in performance of the ALGES model on subsets involving multiple operations (Multi, Multi$_{\mathrm{sym}}$ and Multi$_{\mathrm{asym}}$). This demonstrates once more the weakness of this feature-based model in capturing the reasoning necessary to distinguish the order of the operands involved in equations containing multiple and non-commutative operators. 
Furthermore, we observe that the sequential \BLSTM{} model is now significantly outperformed by the tree-based \NTLSTM{} on subsets involving multiple operations to be solved (Multi, Multi$_{\mathrm{sym}}$ and Multi$_{\mathrm{asym}}$). 
This again supports our initial hypothesis that tree-structured approach is better suited to capture more complex reasoning which is necessary to solve arithmetic problems. In the \textit{ILP + Asym} candidate generation setting this is even more important because of the additional noise introduced with the incorrect candidates that involve multiple and asymmetric operations. 
Conversely, for arithmetic problems involving single operations to be solved (Single, Single$_\mathrm{sym}$, and Single$_\mathrm{asym}$ subsets), the \BLSTM{} model shows a competitive performance, surpassing the tree-based \NTLSTM{} model on problems requiring single commutative operations (Single$_\mathrm{sym}$).
Additionally, we observe an important drop in performance of \TLSTM{} model which is mainly influenced by low accuracy scores on asymmetric subsets (Single$_{\mathrm{asym}}$ and Multi$_{\mathrm{asym}}$). This is in line with our initial intuition 
  that by using a single weight matrices $U_i$, $U_o$, $U_f$, $U_u$ to transform either the sum of the children's states $\tilde{h}_t$ (see \equsrefrange{ch_mwp:tlstm:0}{ch_mwp:tlstm:2} and \ref{ch_mwp:tlstm:4}) or the individual children states $h_k$ (\equref{ch_mwp:tlstm:3}), the \TLSTM{} model is unable to distinguish the order of the operands involved in asymmetric equations. This difference is less evident in \tabref{ch_mwp:tab:results_complex} because most of the incorrect candidates involving non-commutative operations are already filtered out by the ILP component. However, in our \textit{ILP + Asym} candidate generation setup, we make sure that for each candidate involving non-commutative operation, we also include noisy candidates with all the possible asymmetric permutations. This makes it necessary not only to detect the right operation, but also to distinguish the order of the operands, where the \TLSTM{} model fails. Finally, we observe that overall (on Full dataset) our tree-based \NTLSTM{} model 
  exhibits less variance among the different bootstrap results, 
  compared to the sequential \BLSTM{} model. This indicates that \NTLSTM{} model is less susceptible to different seed initialization during the training process, making it more robust than other proposed models (\TLSTM{} and \BLSTM{}). 
\begin{table}[t]
    \centering
    \resizebox{0.9\textwidth}{!}{
    \begin{tabular}{p{25mm}p{65mm}p{20mm}}    
    \toprule
    \textbf{Type} & \textbf{Problem Text} & \textbf{\NTLSTM{}} \\
    \midrule
    \multirow{1}{25mm}{Complex reasoning (57\%)} & Seth bought 20 cartons of ice cream and 2 cartons of yogurt. Each carton of ice cream cost \$6 and each carton of yogurt cost \$1. How much more did Seth spend on ice cream than on yogurt? & $20/2-1\times6$ \\ 
    \midrule
    \multirow{1}{25mm}{Parsing and counting (22\%)} 
    & Jane's dad brought home 24 marble potatoes. If Jane's mom made potato salad for lunch and served an equal amount of potatoes to Jane, herself and her husband, how many potatoes did each of them have? & n/a \\ 
    \midrule
    \multirow{2}{25mm}{World Knowledge (21\%)} & Bert runs 2 miles every day. How many miles will Bert run in 3 weeks? & $3 \times 2$ \\ 
    \cmidrule{2-3}
    & The sum of three consecutive odd numbers is 69. What is the smallest of the three numbers? & n/a \\ 
    \bottomrule
    \end{tabular}} 
    \caption[Examples of problems where our \NTLSTM{} model fails]{Examples of problems where our \NTLSTM{} model fails. }
    \label{ch_mwp:tab:errors}
\end{table}
\begin{table}[t]
    \small
    \centering
    \resizebox{0.9\textwidth}{!}{
    \begin{tabular}{p{50mm}p{30mm}p{30mm}}
    \toprule
    \textbf{Problem Text} & \textbf{ALGES} & \textbf{\NTLSTM{}} \\
    \midrule
   Nancy bought 615 crayons that came in packs of 15.  How many packs of crayons did Nancy buy? & $615-15$ & $615/15$ \\ 
    \midrule
    Carrie's mom gave her \$91 to go shopping. She bought a sweater for \$24, a T-shirt for \$6, and a pair of shoes for \$11. How much money does Carrie have left? & $91 + 24 + 6 + 11$ & $91 - (24 + 6 + 11)$ \\
    \midrule 
    Melanie had 19 dimes in her bank. Her dad gave her 39 dimes and her mother gave her 25 dimes. How many dimes does Melanie have now ? & $19 - 39 + 25$ & $19+39+25$ \\
    \midrule
    On Saturday, Sara spent \$10.62 each on 2 tickets to a movie theater. Sara also rented a movie for \$1.59, and bought a movie for \$13.95. How much money in total did Sara spend on movies? & $10.62 + 2\times1.59 + 13.95$ & $10.62 \times 2 + 13.95 + 1.59$ \\ 
    \bottomrule
    \end{tabular}}
    \caption{Examples of problems that \NTLSTM{} provides a correct solution, but current state-of-the-art ALGES~\cite{koncel2015parsing} fails. }
    \label{ch_mwp:tab:model_correct}
\end{table}

\noindent\textbf{Error Analysis}: In order to understand our system's weaknesses, we manually analyzed the errors that it consistently makes across different training seed instances. We grouped them into three main categories represented in \tabref{ch_mwp:tab:errors}: \textit{complex reasoning}, \textit{parsing and counting}, and \textit{world knowledge} errors. We observe that more than half (57\%) of our system's errors are due to problems requiring \textit{complex reasoning} while the numbers have been correctly extracted from the text. This reflects the results from Tables \ref{ch_mwp:tab:results_complex} and \ref{ch_mwp:tab:results_asym}  that show lower performance of our models on problems requiring multiple and/or non-commutative operations. As future work to alleviate this type of problems we can complement the tree-structures using additional information such as the entities inside the sentence. For instance, in the first example illustrated in \tabref{ch_mwp:tab:errors}, if the system would know that ``ice cream" from the second sentence represents the same concept as in the first one, it would be easier to link numbers 6 and 20. A second consistent type of error is related to \textit{parsing and counting}. It mainly happens when there are several entities involved in a problem statement and the system has to count them correctly. For instance, in the second example presented in \tabref{ch_mwp:tab:errors}, our current system is unable to produce the correct candidate mathematical expression since it can only extract the number 24 from text. Further work in improving aspects related to parsing and entity identification in the problem statement should significantly reduce this kind of mistakes. Finally, the \textit{world knowledge} related errors account for 21\% of the total mistakes. Most of these errors are due to the fact that the system is unable to capture the units correctly (\ie there are 7 days in a week, or one dime equals 0.1 dollars). However, as in the second example, some of the problems require a more advanced conceptual world understanding, such as the notion of odd numbers. Future work can be directed towards methods that are able to capture and represent this kind of world knowledge.  

\noindent\textbf{Limitations of the current state-of-the-art:} We performed an empirical study on the predicted results to understand better where our proposed model outperforms the current state-of-the art model, ALGES \cite{koncel2015parsing}. \tabref{ch_mwp:tab:model_correct} illustrates some examples of the problems where our model gets consistently correct predictions on different training initialization weights (\secref{ch_mwp:sec:experimental_setup}). Most of the gains came from improving on problems requiring multiple and/or asymmetric operations, corroborating our previous findings. 
\begin{table}[t]
    \small
    \centering
    \resizebox{0.9\textwidth}{!}{
    \begin{tabular}{p{50mm}p{30mm}p{30mm}}
    \toprule
    \textbf{Problem Text} & \textbf{ALGES} & \textbf{\NTLSTM{}} \\
    \midrule
  Diane is a beekeeper. Last year, she harvested 2,479 \textbf{pounds} of honey. This year, she bought some new hives \textbf{and} increased her honey harvest by 6,085 \textbf{pounds}. How many pounds of honey did Diane harvest this year? & \textcolor{darkspringgreen}{$6,085 + 2,479$} & \textcolor{deepcarmine}{$6,085 - 2,479$} \\ 
  \midrule 
  Jack has a section filled with short story booklets. If each booklet has 9 \textbf{pages} and there are 49 \textbf{booklets} in the short story section, how many pages will Jack need to go through if he plans to read them all? & \textcolor{darkspringgreen}{$9 \times 49$} & \textcolor{deepcarmine}{$9 + 49$} \\ 
    \midrule
    Benny received 67 \textbf{dollars} for his birthday. He went to a sporting goods store \textbf{and} bought a baseball glove, baseball, \textbf{and} bat. He had 33 \textbf{dollars} left over. How much did he spent on the baseball gear? & \textcolor{deepcarmine}{$67 + 33$} & \textcolor{darkspringgreen}{$67 - 33$} \\
    \midrule
    Jane’s mom picked cherry tomatoes from their backyard. If she gathered 56 \textbf{cherry tomatoes} \textbf{and} is about to place them in small jars which can contain 8 \textbf{cherry tomatoes} at a time, how many jars will she need? & \textcolor{deepcarmine}{$56 + 8$} & \textcolor{darkspringgreen}{$56 / 8$} \\
    \bottomrule
    \end{tabular}}
    \caption[Examples of problems that require a single operation to be solved]{Examples of problems that require a single operation to be solved. The first two involve commutative operations ($+$ and $\times$ respectively) where our \NTLSTM{} model fails compared to the feature-based model (ALGES; \cite{koncel2015parsing}). The rest of the examples illustrate cases where ALGES fails and \NTLSTM{} returns the correct answer. The words that represent features used in ALGES that are highly correlated with the predicted operation (\textit{entity match} and the word \textit{``and''}) are highlighted.} 
    \label{ch_mwp:tab:model_single_eq_correct}
\end{table}

\noindent\textbf{Strengths of the current state-of-the-art and limitations of our approach:} Tables~\ref{ch_mwp:tab:results_complex} and \ref{ch_mwp:tab:results_asym} illustrate that in the case of single symmetric operations (Single$_{\textrm{sym}}$), the ALGES method outperforms the proposed architectures (\ie \BLSTM{}, \TLSTM{}, and \NTLSTM{}).
We hypothesize that the main reason for this is the use of carefully hand-engineered features, many of which depend on third-party tools (\eg dependency parsing).
\tabref{ch_mwp:tab:model_single_eq_correct} illustrates four examples whose solution requires mathematical expressions with a single operator. In the first two cases our \NTLSTM{} model is outperformed by the current state-of-the-art ALGES which  correctly predicts the commutative operators ($+$ in the first example and $\times$ in the second one). We have found that these correctly predicted commutative cases are highly correlated with the \textit{entity match} feature (\ie when the noun phrase connected to the number such as ``pounds'' in the first example is the same in two numbers). This feature has high positive correlation with addition and negative correlation with multiplication operations, which is illustrated in the first and second examples respectively. It also requires an additional dependency parsing which, in case of ALGES, is performed using Stanford Dependency Parser\footnote{More concretely, the Stanford Dependency Parser in CoreNLP 3.4 is used.}. Other word-based features are also highly correlated with some operations. For example, the presence of the word ``and'' in the description of the problem is correlated with addition. However, while these features may be a strong indicators of some operators, their application is limited to problems where the underlying patterns appear. This is illustrated in the last two examples that contain two features highly correlated with the addition (\ie \textit{entity match} and ``and'' word), but that require a different (non-commutative) operation in their solutions. In both cases, biased by the most likely feature-based operation, the answer given by ALGES is incorrect. This contrasts with our feature-independent \NTLSTM{} model which manages to predict the correct equation.
This is reflected in Tables \ref{ch_mwp:tab:results_complex} and \ref{ch_mwp:tab:results_asym}, where the features-based approach falls short in capturing the more intricate nature of solutions involving non-commutative operations (Single$_{\mathrm{asym}}$ and Multi$_{\mathrm{asym}}$). In these cases, our tree-based \NTLSTM{} model exhibits superior performance. 

\section{Conclusion}
\label{ch_mwp:sec:conclusion}

\noindent In this work we addressed the reasoning component involved in solving arithmetic word problems. We proposed a recursive tree architecture to encode the underlying equations for solving arithmetic word problems. More concretely, we proposed to use two different \TreeLSTM{} architectures for the task of scoring candidate equations. We performed an extensive experimental study on the SingleEQ dataset and demonstrated consistent effectiveness (\ie more than 3\% increase in accuracy on the Full dataset and more than 15\% for a subset of complex reasoning tasks) of our models compared to current state-of-the-art. 

We observed that, while very strong on simple instances involving single operations, the current feature-based state-of-the-art model exhibits a significant gap in performance for mathematical problems whose solution comprises non-commutative and/or multiple operations. 
This reveals the weakness of this method to capture the intricate nature of reasoning necessary to solve more complex arithmetic problems. Furthermore, our experiments show that, while a traditional sequential approach based on recurrent encoding implemented using \biLSTMs{} over the equation proves to be a robust baseline, it is outperformed by our recursive \TreeLSTM{} architecture to encode the candidate solution equation on more complicated problems that require multiple operations to be solved. 
This difference in performance becomes more significant as we introduce additional noise in our set of candidates by adding incorrect equations that contain non-commutative operations. 

\section*{Acknowledgments}
\noindent Part of the research leading to these results has received funding from
\begin{enumerate*}[(i)]
\item the European Union's Horizon 2020 research and innovation programme for the CPN project under grant agreement no.\ 761488, and
\item the Flemish Government under the ``Onderzoeksprogramma Artifici\"{e}le Intelligentie (AI) Vlaanderen'' programme.
\end{enumerate*}



\graphicspath{{klim_ch_clpsych/}}

\def \figname {Fig.}
\def \Figname {Figure}
\def \secname {Section}
\def \Secname {Section}
\newcommand{\Secref}[1]{\secname~\ref{#1}}
\def \tabname {Table}
\def \eg {e.g., }
\def \ie {i.e., }
\def \cf {cf.\ }
\def \vs {vs.\ }
\def \etal {\ \textit{et al.\ }}
\def \Fone {{$\text{F}_1$}}

\makeatletter
\def\Url@twoslashes{\mathchar`\/\@ifnextchar/{\kern-.2em}{}}
\g@addto@macro\UrlSpecials{\do\/{\Url@twoslashes}}
\makeatother

\def\num#1{\numx#1}\def\numx#1e#2{{#1}\mathrm{e}{#2}}

\widowpenalty100000
\clubpenalty100000
\renewcommand*{\thesection}{B.\arabic{section}}

\hyphenation{}

\appendixB[Predicting Psychological Health from Childhood Essays The UGent-IDLab CLPsych 2018 Shared Task System
]{Predicting Psychological Health from Childhood Essays The UGent-IDLab CLPsych 2018 Shared Task System}

\label{chap:clpsych}

\renewcommand\evenpagerightmark{{\scshape\small Chapter B}}
\renewcommand\oddpageleftmark{{\scshape\small Predicting Psychological Health}}


\renewcommand{\bibname}{References}

\begin{flushright}
\end{flushright}

\noindent\emph{In this chapter we describe our contribution to CLPsych 2018 shared task where we achieve competitive results using an ensemble consisting of multiple models to predict depression and anxiety in textual surveys.  
}

\begin{center}
\par{$\star\star\star$}
\end{center}
\vspace{0.15in}

\par{\noindent\large{\textbf{K.~Zaporojets, L.~Sterckx\textsuperscript{$*$}, J.~Deleu, T.~Demeester and C.~Develder}}}
\vspace{0.1in}
\par{\noindent\textbf{5th Annual Workshop on Computer Linguistics and Clinical Psychology (CLPsych 2018) at NAACL-HLT 2018}}
\vspace{0.15in}

\par{\noindent\bf{Abstract}}
  This paper describes the IDLab system submitted to Task A of the CLPsych 2018 shared task. 
  The goal of this task is predicting psychological health of children based on language used in hand-written essays and socio-demographic control variables. 
  Our entry uses word- and character-based features as well as lexicon-based features and features derived from the essays such as the quality of the language. We apply linear models, gradient boosting as well as neural-network based regressors (feed-forward, CNNs and RNNs) to predict scores. 
  We then make ensembles of our best performing models using a weighted average.

\section{Introduction}
\label{sec:introduction}
The goal of the CLPsych 2018 shared task is to predict the psychological health of children based on essays and socio-demographic control variables. 
The provided data stems from the National Child Development Study (NCDS) which followed a number of people born in a single week of March 1958 in the UK \cite{power2005cohort}. 
The psychological health of this group of individuals was monitored in intervals of several years. 
At the age of 11, participants were asked to write an essay describing where they saw themselves at age 25. 
Simultaneously, their psychological health was evaluated by their teachers based on metrics defined by the Bristo Social Adjustment Guides (BSAG) \cite{shepherd2013bristol}. 

Given the written essays and social control variables (gender and social class), CLPsych participants are to predict three types of BSAG scores: 
\begin{enumerate*}[label=(\roman*)]
\item {\tt total} BSAG score, 
\item the {\tt depression} BSAG score, and
\item the {\tt anxiety} BSAG score. 
\end{enumerate*}
In order to predict these scores, participants are allowed to use the social control variables next to the features extracted from the essays themselves.

Our system uses several types of features: bag-of-word and bag-of-character features, features derived from lexicons and term lists, and features based on text statistics (see Section \ref{sec:des_features} for more details). 
Using these features, we apply several types of regressors: linear models, gradient boosting and neural-network based models.
For each of the regressors, we explore different combinations of features to predict each of the BSAG scores. 
Subsequently, these models are combined using weighted average ensembling.
Two sets of predictions were made: the first one is based on the single best models,  a second uses an ensemble of models for each of the three scores ({\tt depression}, {\tt anxiety} and {\tt total} BSAG scores). 

Our ensemble of models gives a competitive result, positioning our system on the second place with only 0.01 points under the winner of this shared task. We think that this good performance is mostly due to the different nature of our individual models which complement each other when ensembled. 

The remainder of this paper is organized as follows: \secref{sec:task_and_data} describes the shared task in more detail. 
\Secref{sec:features} presents the features used by the regressors. 
\Secref{sec:model_description} describes regressors and the general methodology of our approach. 
\Secref{sec:results} describes results we obtained during development on our internal validation set and on the real test set. 
Finally, we summarize our findings and present future directions in \secref{sec:conclusion_future_work}. 

\section{Task and data}
\label{sec:task_and_data}

Input for task A consists of essays written by 11-year-old children describing where they see themselves at age 25, as well as several social control variables: 
\begin{enumerate}
    \item \textbf{Gender}: gender of the participant child. 
    \item \textbf{Social Class}: the job hierarchy of the father of the participant child. The domain comprises 6 values representing different job categories: starting with professional and managerial occupations and ending with unskilled occupations. 
    \item \textbf{Essay}: content of the essay written by the participant child. Originally, the essays were hand-written and later transcribed in digital format. The average length of the essays is 225 characters. 
\end{enumerate}

The goal of shared task A is to predict the current psychological health of the children. 
Psychological health is measured using scores assigned by teachers of the children following metrics defined in the BSAG. 
These guides score the total psychological health using 12 different syndromes (depression, anxiety, hostility, etc.). 
CLPsych shared task A requires participants to predict three scores: 
\begin{enumerate}
    \item \textbf{Total}: the sum of all the BSAG scores of all the different syndromes. 
    \item \textbf{Depression}: the BSAG score related to the depression syndrome. 
    \item \textbf{Anxiety}: the BSAG score related to the anxiety syndrome. 
\end{enumerate}

Participants are given a training set consisting of essays from 9,217 children with corresponding input variables and BSAG scores.  

\section{Features}
\label{sec:features}
In this section, we present features used by our models, and experiment with a number of different categories of feature extraction.

\subsection{Lexical features} 
\label{sec:lex_features}
We use bag-of-n-gram features both on word- and character-level. 
The latter provides robustness to the spelling variation found in children's writing.
For word-level we experiment with n-grams for $n$ ranging from 1 to 4. 
At character-level, we experiment with 3- up to 6-grams. 
These one-hot encodings are weighted using TF-IDF. 

\subsection{Feature engineering}
\label{sec:des_features}
Next to the sparse bag-of-n-grams representations of the essays, we apply several manually designed features.

\noindent \textbf{Social control features}
These features are given as input in the data and consist of the {\em gender\/} and {\em social class\/} of the participants. 
In order to be used in regressors, we encode these features as one-hot vectors. 

\noindent \textbf{Lexicon-based features} 
We experiment with features based on two lexicons: the Linguistic Inquiry Word Count (LIWC) described in \cite{pennebaker2015development} and the DepecheMood \cite{staiano2014depechemood}. 
The LIWC is a psycholinguistic lexicon that allows to measure the emotional health of individuals by providing a set of term categories related to different mental states. 
In our experiments we use all 73 (partly overlapping) psychological word categories found in the LIWC dictionary. 

Similarly, DepecheMood is a lexicon consisting of 37k different words (verbs, nouns, adjectives and adverbs). 
Each of the words has weights associated to the following 8 mental states: afraid, amused, angry, annoyed, don't care, happy, inspired and sad. 
In our experiments, we calculate the average of TF-IDF weights for these categories. 
These TF-IDF weights are already given inside DepecheMood lexicon and are originally calculated on articles from {\tt rappler.com} based on Rappler's {\em Mood Meter} crowdsourcing.

\noindent \textbf{Textual statistics features}
We extract a number of features describing several characteristics of the essays:
\begin{itemize}
    \item Total number of words 
    \item Average sentence length 
    \item Average word length 
    \item Ratio of spelling mistakes
    \item Ratio of different words 
    \item Number of words not recognized (illegible) when transcribing the essays from hand-written to digital form. 
\end{itemize}

\noindent \textbf{Sentiment features}
We reason that the participants' psychological health can partially be detected by evaluating the essay in a positive-negative sentiment spectrum. 
We use the pretrained sentiment classifier from \cite{cagan2014generating}.\footnote{The python library can be found at: \url{https://pypi.python.org/pypi/sentiment\_classifier}}
We hypothesize that individuals with good psychological health will tend to use more positive expressions than individuals with high scores in any of BSAG syndromes. 

\noindent \textbf{Language model features}
Coming from the intuition that mental state may be related to the development of language skills, we include two language model features.
Our primary language model feature is the average perplexity of the essays, as it is an often used metric to score the general language quality and coherence of the texts. 
As a secondary feature, we include the fraction of out-of-vocabulary tokens over the entire essay, with respect to the Penn Treebank data.
We use the word-level \texttt{AWD\_LSTM} language model trained on the Penn Treebank, presented by \cite{merity2017}.

\section{Models description} 
\label{sec:model_description}
We train a variety of different regression models predicting the three aforementioned BSAG scores. 
We include simple linear models as well as gradient boosted trees and neural network-based models. 
Our best performing models are subsequently combined using ensembling. 
As a general rule, we try to select different model function types in order to achieve lower correlation between predictions from the different types of models. 

\subsection{Linear models}
We experiment with two types of linear regressors: support vector machines (SVMs) and ridge regression. 
Linear models are trained on two sets of features.

\begin{enumerate}
    \item {\em Lexical features} based purely on the text of the essays (see \secref{sec:lex_features}). Here we use TF-IDF weighted bag-of-word features as well as character features. 
    \item {\em Designed features} through feature engineering (see \secref{sec:des_features}).
\end{enumerate}

To avoid overfitting, we tune the regularization parameter $\alpha$ on a validation set. For SVM models this parameter corresponds to squared L2 penalty. For ridge models, it corresponds to the strength of L2 regularization term. We experiment with selecting models based on lowest RMSE error as well as the ones with highest disattenuated Pearson correlation score. 

\subsection{Gradient boosting}
We apply gradient boosted tree regressors using XGBoost \cite{chen2016xgboost} trained on the \emph{designed features} (see  \secref{sec:des_features}). 
To train XGBoost models, we use early stopping by evaluating on a validation set with 10,000 estimators and a logarithmic scale grid search of learning rate from $\num{10e-5}$ to $\num{10e+5}$. We experiment with RMSE as well as disattenuated Pearson correlation scores as criterion to perform early stopping. 

\subsection{Feed-forward neural networks}
As a second type of non-linear models, we use feed-forward neural networks (FFNNs). 
We train FFNNs on our {\em designed features} (see Section \ref{sec:des_features}) expecting that the introduced non-linearity will complement the results of previous models. 
Our FFNN architecture consists of 3 hidden layers with $\tanh$ activation units. 
We apply dropout regularization of 0.5 between each of the layers. 
The network has a total of 223 input features in the first layer and 256 neurons in  each of the three intermediate hidden layers. 
We experiment with optimizing for three loss functions: 
\begin{enumerate}
    \item \textbf{Mean squared error (MSE)}: this is our default choice used for most of the regressors. 
    \item \textbf{Huber}: Huber loss is less sensitive to outliers which are present in BSAG scores (high BSAG scores for few individuals).
    \item \textbf{Pearson correlation}: we experiment with correlation loss because it is directly related to the metric used to evaluate the model performance by organizers of shared task A. 
\end{enumerate}

\subsection{Neural sequence encoders}
\label{sec:cnn}
We include two types of models based on neural networks which encode the essays to a low dimensional representation, after which a score is predicted using a feed-forward layer. 
Essays are encoded using two of the most prevalent neural network architectures for modeling of sequences, convolutional neural networks (CNN) and recurrent neural networks (RNN).

\noindent \textbf{Pretrained embeddings} 
The first layer of NN architectures embeds the one-hot token representations into a vector space of lower dimensionality, which it then fine-tuned through back-propagation.
We initialize the embedding layer using embeddings from dedicated word embedding techniques Word2Vec \cite{mikolov2013distributed} and Glove \cite{pennington2014glove}.
This proved to be essential for good performance of the neural sequence models.

\noindent \textbf{CNNs}  We apply the architecture proposed by \cite{kim2014convolutional} which consists of a single convolutional layer with multiple filter sizes, followed by one feed-forward layer over the three-dimensional score vector.
We use filters of size 3, 4, 5, 6 and 7 and vary the amount from 64 to 512 filters for each size.

\noindent \textbf{RNNs} We experiment with two types of RNNs to encode the essays, long short-term memory networks (LSTM) \cite{hochreiter1997long} and gated recurrent units (GRU) \cite{cho2014learning}. After encoding the essay in forward and backward direction, we use the concatenated sequences of hidden states to predict scores. 
To reduce the dimensionality of this representation, we use max-pooling and self-attention to obtain the final essay encodings \cite{lin2017structured}.
We experiment with single-layer bidirectional RNNs with hidden state vectors of 64, 128 and 256 dimensions. A fully connected layer of 32 and 64 nodes is used to predict scores.

\subsection{Model ensembling}
\label{sec:mod_ensemble}
To produce weighted averages of predictions, we use the {\em forward model selection} algorithm that greedily selects the combination of models that maximizes the disattenuated Pearson correlation on the evaluation set. 
We use 100 iterations and choose the best model if there is no improvement after 30 iterations on the evaluation set.

\begin{table*}[t]
	\centering
    \small
\resizebox{1.0\textwidth}{!}{    
  \begin{tabular}{@{}lcccrcccrccc@{}}
  \toprule
  & \multicolumn{3}{c}{\textbf{Anxiety}} && \multicolumn{3}{c}{\textbf{Depression}}&& \multicolumn{3}{c}{\textbf{Total}}\\
  \cmidrule{2-4} \cmidrule{6-8} \cmidrule{10-12}
&  {RMSE}&{MAE}&{Diss. R}&&{RMSE}&{MAE}&{Diss. R} &&{RMSE}&{MAE}&{Diss. R}\\
\midrule
\textbf{Development}   \\
Ridge RMSE (lex. feat.) & 1.222 & 0.784 & 0.2100 && 1.460 & 1.076 & 0.3493 && 8.356 & 6.472 & 0.4532 \\
\ \ +Diss. R (lex. feat.) & 1.225 & 0.782 & 0.2160 && 1.497 & 1.138 & 0.4046 && 8.643 & 7.043 & 0.4783 \\
\ \ +RMSE (des. feat.) & 1.218 & 0.773 & 0.2136 && 1.446 & 1.073 & 0.3781 && 8.272 & 6.280 & 0.4719 \\
\ \ +Diss. R (des. feat.) & 1.218 & 0.773 & 0.2136 && 1.446 & 1.073 & 0.3781 && 8.272 & 6.280 & 0.4719 \\
\midrule
SVM RMSE (lex. feat.) & 1.260 & 0.690 & 0.1129 && 1.517 & 1.046 & 0.2542 && 8.643 & 5.940 & 0.4526 \\
\ \ +Diss. R (lex. feat.) & 1.360 & \textbf{0.573} & 0.1220 && 1.811 & 1.007 & 0.4094 && 9.047 & 6.091 & 0.4624 \\
\ \ +RMSE (des. feat.) & 1.241 & 0.723 & 0.1227 && 1.470 & 1.005 & 0.3736 && 8.683 & 6.920 & 0.3418 \\
\ \ +Diss. R (des. feat.) &1.352 & \textbf{0.573} & 0.1026 && 1.897 & 1.694 & 0.3508 && 8.449 & 6.019 & 0.4473 \\
\midrule 
XGBoost RMSE (des. feat.) & 1.221 & 0.769 & 0.1982 && 1.452 & 1.081 & 0.3624 && 8.302 & 6.257 & 0.4600 \\
\ \ +Diss. R (des. feat.) & 1.225 & 0.768 & 0.1997 && 1.458 & 1.073 & 0.3579 && 8.312 & 6.343 & 0.4557 \\
\midrule
CNN RMSE loss & 1.221 & 0.772 & 0.2053 && 1.473 & 1.128 & 0.3863 && 8.390 & 6.488 & 0.4556 \\
RNN RMSE loss & 1.228 & 0.769 & 0.1630 && 1.444 & 1.070 & 0.3938 && 8.271 & 6.206 & 0.4805 \\
\midrule
FFNN MSE loss (des. feat.) & \textbf{1.216} & 0.775 & 0.2253 && 1.445 & 1.073 & 0.3837 && \textbf{8.219} & 6.310 & 0.4945 \\
\ \ +Huber loss (des. feat.) & 1.246 & 0.697 & 0.2294  && 1.483 & 0.997 & 0.3921 && 8.486 & \textbf{5.884} & 0.5000 \\
\ \ +Diss. R loss (des. feat.) & 1.288 & 0.616 & 0.2010 && 1.675 & \textbf{0.959} & 0.3488 && 11.556 & 7.743 & 0.4290 \\
\midrule
Ensemble & 1.223 & 0.743 & \textbf{0.2660} && \textbf{1.435} & 1.035 & \textbf{0.4246} && 8.252 & 6.047 & \textbf{0.5191} \\
\midrule
\textbf{Test Runs}   \\
Submission 1 (Ensemble) &1.119 & \textbf{0.476}  & \textbf{0.1946} && \textbf{1.393} & \textbf{1.004} & \textbf{0.4536} && \textbf{7.843} & 5.691 & \textbf{0.5667} \\
Submission 2 (Single Model) & \textbf{1.022} & 0.697  & 0.1760 && 1.403 & 1.019 & 0.4192 && 8.134 & \textbf{5.688} & 0.5140 \\
\bottomrule
\end{tabular}}
\caption[Results on internal evaluation set]{ Results on internal evaluation set for best individual models; ``lex. feat." refers to the lexical features (see section \ref{sec:lex_features}), whereas ``des. feat." are the designed features (see section \ref{sec:des_features}).}
\label{tab:scores_internal}
\end{table*}

\begin{table}[t]
	\centering
    \scriptsize
  \begin{tabular}{@{}lccc@{}}
  \toprule
  & \multicolumn{1}{c}{\textbf{Anxiety}} & \multicolumn{1}{c}{\textbf{Depression}}& \multicolumn{1}{c}{\textbf{Total}}\\
\midrule
Ridge RMSE (lex. feat.) & 0.2698 & 0.0625 & 0.1825 \\
Ridge RMSE (des. feat.) & - & - & - \\
\midrule
SVM RMSE (lex. feat.) & - & - & - \\
SVM RMSE (des. feat.) & 0.0688 & 0.1563 & 0.0584 \\
\midrule 
XGBoost RMSE (des. feat.) & 0.2646 & 0.0469 & 0.0949 \\
\midrule
CNN RMSE loss & 0.0423 & 0.1250 & - \\
RNN RMSE loss & - & 0.3281 & 0.2993 \\
\midrule
FFNN MSE loss (des. feat.) & - & 0.2813 & 0.0365 \\
FFNN Huber loss (des. feat.) & 0.3545 & - & 0.3285 \\
FFNN Diss. R loss (des. feat.) & - & - & - \\
\bottomrule
\end{tabular}
\caption[Weights of the ensemble components]{Weights of the ensemble components.}
\label{tab:ensemble_combinations}
\end{table}
\section{Experiments}
\label{sec:results}

\subsection{Training details}
\label{sec:train_details}

We divide the training set of 9,217 individual evaluations into two parts:
\begin{enumerate*}[label=(\roman*)]
\item a \emph{train set} consisting of 7,835 examples, and
\item an \emph{evaluation set} consisting of the rest (1,382 examples).
\end{enumerate*}
For SVM, Ridge and XGBoost models, we select the best models on our evaluation set using two metrics:
\begin{enumerate*}[label=(\roman*)]
\item models with the lowers RMSE score, and
\item models with the highest disattenuated Pearson correlation score. 
\end{enumerate*}
For feed-forward neural nets we experiment with three loss functions: \begin{enumerate*}[label=(\roman*)]
\item MSE, \item Huber, and \item disattenuated Pearson correlation.
\end{enumerate*}
Finally, for neural sequence encoders, we use MSE as a loss function. 
In order to build an ensemble of models, we further subdivide our evaluation set in two equal parts: 
\begin{enumerate}
    \item \textbf{Validation set}: the validation set is used to choose the best combination of models using forward model selection (see \secref{sec:mod_ensemble}). 
    \item \textbf{Test set}: the test set is used to verify that a given model combination does not overfit the evaluation set. 
\end{enumerate}

Before extracting features from the text of input essays, we perform basic text preprocessing functions: lowercasing, removal of punctuation and extra spaces. For TF-IDF and embedding lexical features we also remove the stop words. Additionally, we use TextBlob (\url{https://textblob.readthedocs.io/}) in order to correct the spelling mistakes.

Feed-forward neural networks are trained for 100 epochs with learning rate of $\num{1e-5}$. We also apply a weight decay (L2 penalty) of $\num{1e-6}$ on the Adam optimizer. Most of the models converge after training approximately for 20 epochs with a batch size of 8. 

CNN and RNN models are trained with Adam and early stopping based on disattenuated Pearson correlation. Models converge after training for approximately 10 epochs, with batch size 32.
For RNN models we apply a dropout with probability 0.3 on the embedding layer and the output layer. For both CNN and RNN models we apply dropout on the fully connected layer with probability 0.15.

\subsection{Results}
\tabref{tab:scores_internal} summarizes results for different models on our validation set. 
For linear models, we notice that SVM models are sensitive to optimizing towards RMSE or disattenuated correlation score. 
We also observe that SVM models have lower disattenuated correlation scores for the anxiety BSAG metric. 
For feed-forward neural nets, use of the Huber loss obtains the best performance.
We speculate that this is because this method is not as influenced by outliers as other loss functions. 
The rest of the models has approximately similar performance. 

A large boost in performance is observed when creating ensembles of models. 
We gain between 0.02 and 0.04 points on our validation set for the disattenuated correlation metric. 
We don't see this improvement on RMSE and MAE metrics since our ensemble is greedily built to optimize for Pearson correlation between predicted and ground truth results. 

\tabref{tab:ensemble_combinations} shows the weight combinations of our ensemble for all three objectives to predict. 
We only add best RMSE models for Ridge, SVM and XGBoost regressors. 
The reason is that adding models that had the best performance on Pearson disattenuated correlation score decreased significantly the RMSE and MAE scores of the ensemble. 
How these models can still be added without producing this drop in performance is left for future work. 


The bottom rows of \tabref{tab:scores_internal} show the results of our two submissions on the official CLPsych test collection. 
We obtain a considerable improvement using ensembles of models with respect to our single best model submission, resulting in the overall second best submission. 
We speculate that this is because of different score distributions produced by dissimilar models used in this work. 
This generates low correlation of individual model predictions, which results in better ensembles. 
We were surprised to see that disattenuated correlation score was several points higher in depression and total BSAG predictions than on our internal validation set. 
The anxiety score, on the other hand, is considerably lower. 
Further analysis is needed to understand these differences, and to investigate the impact of the individual types of hand-designed features.

\section{Conclusion and future work}
\label{sec:conclusion_future_work}
In this paper we briefly described the Ghent University -- IDLab submission to the CLPsych 2018 shared task A. 
We found that linear models, gradient boosting as well as neural network based models perform similarly but produce different models that, when combined, can increase the performance on the test set considerably. 

For future work, we plan to conduct a careful error analysis (e.g. ablation tests) and examine the best ways to design our train-validation splits in order to decrease the score difference between the validation and test sets. 
We also plan to experiment with more sophisticated ways of ensembling and stacking techniques.

We consider that in the end, most of the success of this task comes down to designing a good set of features. 
In particular, one of the features we didn't explore is topic modeling. 
Additional features can be obtained from topic model distributions as they provide positive results on similar tasks described in \cite{resnik2015university} and \cite{cohan2016triaging}. 

Finally, another direction we want to explore consists of using word and phrase embeddings, pre-trained on a corpus of individuals with psychological disorders. Some work has already been done to gather this kind of corpus from online resources (Twitter and Reddit in particular) \cite{yates2017depression} and \cite{coppersmith2015clpsych}. We hypothesize that we can get a significant improvement by initializing our CNN and RNN models with these embeddings. 

\section*{Acknowledgments}
We are grateful to Giannis Bekoulis for fruitful discussions on model cross-validation and for providing resources, support and encouragement. 

\section*{Human Subjects Review}
This study was evaluated by the Ethics Committee of the faculty of Psychology and Educational Sciences of Ghent University, which concluded that ethical approval was not needed for the research conducted for this manuscript.


\renewcommand*{\thesection}{\thechapter.\arabic{section}}       



\bibliographystyle{phdbib}
\bibliography{references}

\begin{thebibliography}{100}

\bibitem{koncel2015parsing}
R.~Koncel-Kedziorski, H.~Hajishirzi, A.~Sabharwal, O.~Etzioni, and S.~D. Ang.
\newblock {\em Parsing algebraic word problems into equations}.
\newblock Transactions of the Association for Computational Linguistics,
  3:585--597, 2015.

\bibitem{martinez2020information}
J.~L. Martinez-Rodriguez, A.~Hogan, and I.~Lopez-Arevalo.
\newblock {\em Information extraction meets the semantic web: a survey}.
\newblock Semantic Web, 11(2):255--335, 2020.

\bibitem{wang2018clinical}
Y.~Wang, L.~Wang, M.~Rastegar-Mojarad, S.~Moon, F.~Shen, N.~Afzal, S.~Liu,
  Y.~Zeng, S.~Mehrabi, S.~Sohn, et~al.
\newblock {\em Clinical information extraction applications: a literature
  review}.
\newblock Journal of biomedical informatics, 77:34--49, 2018.

\bibitem{zaporojets2021dwie}
K.~Zaporojets, J.~Deleu, C.~Develder, and T.~Demeester.
\newblock {\em {DWIE}: An entity-centric dataset for multi-task document-level
  information extraction}.
\newblock Information Processing \& Management, 58(4):102563, 2021.
\newblock Available from: \url{https://doi.org/10.1016/j.ipm.2021.102563}.

\bibitem{corcoglioniti2016knowledge}
F.~Corcoglioniti, M.~Dragoni, M.~Rospocher, and A.~P. Aprosio.
\newblock {\em Knowledge Extraction for Information Retrieval}.
\newblock In Proceedings of the 13th International Conference on The Semantic
  Web. Latest Advances and New Domains-Volume 9678, pages 317--333, 2016.

\bibitem{hussain2019survey}
S.~Hussain, O.~Ameri~Sianaki, and N.~Ababneh.
\newblock {\em A survey on conversational agents/chatbots classification and
  design techniques}.
\newblock In Workshops of the International Conference on Advanced Information
  Networking and Applications, pages 946--956. Springer, 2019.

\bibitem{laranjo2018conversational}
L.~Laranjo, A.~G. Dunn, H.~L. Tong, A.~B. Kocaballi, J.~Chen, R.~Bashir,
  D.~Surian, B.~Gallego, F.~Magrabi, A.~Y. Lau, et~al.
\newblock {\em Conversational agents in healthcare: a systematic review}.
\newblock Journal of the American Medical Informatics Association,
  25(9):1248--1258, 2018.

\bibitem{bavaresco2020conversational}
R.~Bavaresco, D.~Silveira, E.~Reis, J.~Barbosa, R.~Righi, C.~Costa, R.~Antunes,
  M.~Gomes, C.~Gatti, M.~Vanzin, et~al.
\newblock {\em Conversational agents in business: A systematic literature
  review and future research directions}.
\newblock Computer Science Review, 36:100239, 2020.

\bibitem{jiang2020recipe}
Y.~Jiang, K.~Zaporojets, J.~Deleu, T.~Demeester, and C.~Develder.
\newblock {\em Recipe instruction semantics corpus (RISeC): resolving semantic
  structure and zero anaphora in recipes}.
\newblock In AACL-IJCNLP 2020, the 1st Conference of the Asia-Pacific Chapter
  of the Association Computational Linguistics and 10th International Joint
  Conference on Natural Language Processing, pages 821--826. Association for
  Computational Linguistics (ACL), 2020.

\bibitem{jiang2022cookdial}
Y.~Jiang, K.~Zaporojets, J.~Deleu, T.~Demeester, and C.~Develder.
\newblock {\em CookDial: a dataset for task-oriented dialogs grounded in
  procedural documents}.
\newblock Applied Intelligence, pages 1--19, 2022.

\bibitem{zaporojets2018predicting}
K.~Zaporojets, L.~Sterckx, J.~Deleu, T.~Demeester, and C.~Develder.
\newblock {\em Predicting psychological health from childhood essays: the
  UGent-IDLab CLPsych 2018 shared task system.}
\newblock In 5th Annual Workshop on Computer Linguistics and Clinical
  Psychology (CLPsych 2018) at NAACL-HLT 2018, pages 119--125. Association for
  Computational Linguistics, 2018.

\bibitem{trifan2020understanding}
A.~Trifan, R.~Antunes, S.~Matos, and J.~L. Oliveira.
\newblock {\em Understanding depression from psycholinguistic patterns in
  social media texts}.
\newblock In European Conference on Information Retrieval, pages 402--409.
  Springer, 2020.

\bibitem{jacobson2021deep}
N.~C. Jacobson, D.~Lekkas, R.~Huang, and N.~Thomas.
\newblock {\em Deep learning paired with wearable passive sensing data predicts
  deterioration in anxiety disorder symptoms across 17--18 years}.
\newblock Journal of Affective Disorders, 282:104--111, 2021.

\bibitem{tadesse2019detection}
M.~M. Tadesse, H.~Lin, B.~Xu, and L.~Yang.
\newblock {\em Detection of suicide ideation in social media forums using deep
  learning}.
\newblock Algorithms, 13(1):7, 2019.

\bibitem{bitew2019predicting}
S.~K. Bitew, I.~Bekoulis, J.~Deleu, L.~Sterckx, K.~Zaporojets, T.~Demeester,
  and C.~Develder.
\newblock {\em Predicting suicide risk from online postings in Reddit: the
  UGent-IDLab submission to the CLPysch 2019 Shared Task A}.
\newblock In CLPsych2019, the 6th Annual Workshop on Computational Linguistics
  and Clinical Psychology at NAACL-HLT 2019, pages 158--161. Association for
  Computational Linguistics (ACL), 2019.

\bibitem{thorne2018fever}
J.~Thorne, A.~Vlachos, C.~Christodoulopoulos, and A.~Mittal.
\newblock {\em {FEVER}: a Large-scale Dataset for {F}act {E}xtraction and
  {VER}ification}.
\newblock In Proceedings of the 2018 Conference of the North American Chapter
  of the Association for Computational Linguistics: Human Language Technologies
  (NAACL-HLT 2018), pages 809--819, 2018.
\newblock Available from: \url{https://aclanthology.org/N18-1074}.

\bibitem{aly2021feverous}
R.~Aly, Z.~Guo, M.~Schlichtkrull, J.~Thorne, A.~Vlachos, C.~Christodoulopoulos,
  O.~Cocarascu, and A.~Mittal.
\newblock {\em {FEVEROUS}: Fact Extraction and VERification Over Unstructured
  and Structured information}.
\newblock In Proceedings of the 2021 Conference on Neural Information
  Processing Systems Datasets and Benchmarks Track (NeurIPS 2021), 2021.
\newblock Available from:
  \url{https://datasets-benchmarks-proceedings.neurips.cc/paper/2021/hash/68d30a9594728bc39aa24be94b319d21-Abstract-round1.html}.

\bibitem{augenstein2016stance}
I.~Augenstein, T.~Rockt{\"a}schel, A.~Vlachos, and K.~Bontcheva.
\newblock {\em Stance Detection with Bidirectional Conditional Encoding}.
\newblock In Proceedings of the 2016 Conference on Empirical Methods in Natural
  Language Processing, pages 876--885, 2016.

\bibitem{riedel2017simple}
B.~Riedel, I.~Augenstein, G.~P. Spithourakis, and S.~Riedel.
\newblock {\em A simple but tough-to-beat baseline for the Fake News Challenge
  stance detection task}.
\newblock arXiv preprint arXiv:1707.03264, 2017.

\bibitem{zaporojets2021consistent}
K.~Zaporojets, J.~Deleu, T.~Demeester, and C.~Develder.
\newblock {\em Towards Consistent Document-level Entity Linking: Joint Models
  for Entity Linking and Coreference Resolution}.
\newblock In Proceedings of the 2022 Annual Meeting of the Association for
  Computational Linguistics (ACL 2022), pages 778--784, 2022.
\newblock Available from: \url{https://aclanthology.org/2022.acl-short.88}.

\bibitem{verlinden2021injecting}
S.~Verlinden, K.~Zaporojets, J.~Deleu, T.~Demeester, and C.~Develder.
\newblock {\em Injecting Knowledge Base Information into End-to-End Joint
  Entity and Relation Extraction and Coreference Resolution}.
\newblock In Findings of the 2021 Association for Computational Linguistics and
  the International Joint Conference on Natural Language Processing (ACL-IJCNLP
  2021), pages 1952--1957, 2021.
\newblock Available from:
  \url{https://doi.org/10.18653/v1/2021.findings-acl.171}.

\bibitem{zaporojets2022tempel}
K.~Zaporojets, L.-A. Kaffee, J.~Deleu, T.~Demeester, C.~Develder, and
  I.~Augenstein.
\newblock {\em {TempEL}: Linking Dynamically Evolving and Newly Emerging
  Entities}.
\newblock In 2022 Conference on Neural Information Processing Systems Datasets
  and Benchmarks Track (NeurIPS 2022), 2022.

\bibitem{jurafsky2000speech}
D.~Jurafsky.
\newblock {\em Speech \& language processing}.
\newblock Pearson Education India, 2000.

\bibitem{niklaus2018survey}
C.~Niklaus, M.~Cetto, A.~Freitas, and S.~Handschuh.
\newblock {\em A Survey on Open Information Extraction}.
\newblock In Proceedings of the 27th International Conference on Computational
  Linguistics, pages 3866--3878, 2018.

\bibitem{nasar2018information}
Z.~Nasar, S.~W. Jaffry, and M.~K. Malik.
\newblock {\em Information extraction from scientific articles: a survey}.
\newblock Scientometrics, 117(3):1931--1990, 2018.

\bibitem{grishman2015information}
R.~Grishman.
\newblock {\em Information extraction}.
\newblock IEEE Intelligent Systems, 30(5):8--15, 2015.

\bibitem{sarawagi2008information}
S.~Sarawagi.
\newblock {\em Information extraction}.
\newblock Now Publishers Inc, 2008.

\bibitem{nadeau2007survey}
D.~Nadeau and S.~Sekine.
\newblock {\em A survey of named entity recognition and classification}.
\newblock Lingvisticae Investigationes, 30(1):3--26, 2007.

\bibitem{yadav2018survey}
V.~Yadav and S.~Bethard.
\newblock {\em A Survey on Recent Advances in Named Entity Recognition from
  Deep Learning models}.
\newblock In Proceedings of the 27th International Conference on Computational
  Linguistics, pages 2145--2158, 2018.

\bibitem{li2020survey}
J.~Li, A.~Sun, J.~Han, and C.~Li.
\newblock {\em A survey on deep learning for named entity recognition}.
\newblock IEEE Transactions on Knowledge and Data Engineering, 2020.

\bibitem{sukthanker2020anaphora}
R.~Sukthanker, S.~Poria, E.~Cambria, and R.~Thirunavukarasu.
\newblock {\em Anaphora and coreference resolution: A review}.
\newblock Information Fusion, 59:139--162, 2020.

\bibitem{stylianou2021neural}
N.~Stylianou and I.~Vlahavas.
\newblock {\em A neural entity coreference resolution review}.
\newblock Expert Systems with Applications, 168:114466, 2021.

\bibitem{pawar2017relation}
S.~Pawar, G.~K. Palshikar, and P.~Bhattacharyya.
\newblock {\em Relation extraction: A survey}.
\newblock arXiv preprint arXiv:1712.05191, 2017.

\bibitem{kumar2017survey}
S.~Kumar.
\newblock {\em A survey of deep learning methods for relation extraction}.
\newblock arXiv preprint arXiv:1705.03645, 2017.

\bibitem{jurafsky2018speech}
D.~Jurafsky.
\newblock {\em Speech and Language Processing}.
\newblock 2018.

\bibitem{sang2003introduction}
E.~F. T.~K. Sang and F.~De~Meulder.
\newblock {\em Introduction to the CoNLL-2003 Shared Task: Language-Independent
  Named Entity Recognition}.
\newblock In Proceedings of the 2003 Conference of the North American Chapter
  of the Association for Computational Linguistics: Human Language
  Technologies, pages 142--147, 2003.

\bibitem{derczynski2017results}
L.~Derczynski, E.~Nichols, M.~van Erp, and N.~Limsopatham.
\newblock {\em Results of the WNUT2017 shared task on novel and emerging entity
  recognition}.
\newblock In Proceedings of the 3rd Workshop on Noisy User-generated Text,
  pages 140--147, 2017.

\bibitem{weischedel2013ontonotes}
R.~Weischedel, M.~Palmer, M.~Marcus, E.~Hovy, S.~Pradhan, L.~Ramshaw, N.~Xue,
  A.~Taylor, J.~Kaufman, M.~Franchini, et~al.
\newblock {\em Ontonotes release 5.0 ldc2013t19}.
\newblock Linguistic Data Consortium, Philadelphia, PA, 23, 2013.

\bibitem{ding2021few}
N.~Ding, G.~Xu, Y.~Chen, X.~Wang, X.~Han, P.~Xie, H.~Zheng, and Z.~Liu.
\newblock {\em Few-NERD: A Few-shot Named Entity Recognition Dataset}.
\newblock In Proceedings of the 59th Annual Meeting of the Association for
  Computational Linguistics and the 11th International Joint Conference on
  Natural Language Processing (Volume 1: Long Papers), pages 3198--3213, 2021.

\bibitem{rahman2009supervised}
A.~Rahman and V.~Ng.
\newblock {\em Supervised models for coreference resolution}.
\newblock In Proceedings of the 2009 conference on empirical methods in natural
  language processing, pages 968--977, 2009.

\bibitem{pradhan2012conll}
S.~Pradhan, A.~Moschitti, N.~Xue, O.~Uryupina, and Y.~Zhang.
\newblock {\em CoNLL-2012 shared task: Modeling multilingual unrestricted
  coreference in OntoNotes}.
\newblock In Proceedings of the 2012 Conference on Computational Natural
  Language Learning, pages 1--40, 2012.

\bibitem{recasens2010semeval}
M.~Recasens, L.~M{\`a}rquez, E.~Sapena, M.~A. Mart{\'\i}, M.~Taul{\'e},
  V.~Hoste, M.~Poesio, and Y.~Versley.
\newblock {\em Semeval-2010 task 1: Coreference resolution in multiple
  languages}.
\newblock In Proceedings of the 5th International Workshop on Semantic
  Evaluation, pages 1--8, 2010.

\bibitem{webster2018mind}
K.~Webster, M.~Recasens, V.~Axelrod, and J.~Baldridge.
\newblock {\em Mind the GAP: A Balanced Corpus of Gendered Ambiguous Pronouns}.
\newblock Transactions of the Association for Computational Linguistics,
  6:605--617, 2018.

\bibitem{zhang2017position}
Y.~Zhang, V.~Zhong, D.~Chen, G.~Angeli, and C.~D. Manning.
\newblock {\em Position-aware attention and supervised data improve slot
  filling}.
\newblock In Proceedings of the 2017 Conference on Empirical Methods in Natural
  Language Processing, pages 35--45, 2017.

\bibitem{alt2020tacred}
C.~Alt, A.~Gabryszak, and L.~Hennig.
\newblock {\em TACRED Revisited: A Thorough Evaluation of the TACRED Relation
  Extraction Task}.
\newblock arXiv preprint arXiv:2004.14855, 2020.

\bibitem{li2016biocreative}
J.~Li, Y.~Sun, R.~J. Johnson, D.~Sciaky, C.-H. Wei, R.~Leaman, A.~P. Davis,
  C.~J. Mattingly, T.~C. Wiegers, and Z.~Lu.
\newblock {\em BioCreative V CDR task corpus: a resource for chemical disease
  relation extraction}.
\newblock Database, 2016, 2016.

\bibitem{wei2015overview}
C.-H. Wei, Y.~Peng, R.~Leaman, A.~P. Davis, C.~J. Mattingly, J.~Li, T.~C.
  Wiegers, and Z.~Lu.
\newblock {\em Overview of the BioCreative V chemical disease relation (CDR)
  task}.
\newblock In Proceedings of the 5th BioCreative Challenge Evaluation Workshop,
  2015.

\bibitem{yao2019docred}
Y.~Yao, D.~Ye, P.~Li, X.~Han, Y.~Lin, Z.~Liu, Z.~Liu, L.~Huang, J.~Zhou, and
  M.~Sun.
\newblock {\em {DocRED}: A Large-Scale Document-Level Relation Extraction
  Dataset}.
\newblock In Proceedings of the 2019 Annual Meeting of the Association for
  Computational Linguistics (ACL 2019), pages 764--777, 2019.
\newblock Available from: \url{https://aclanthology.org/P19-1074}.

\bibitem{doddington2004automatic}
G.~R. Doddington, A.~Mitchell, M.~A. Przybocki, L.~A. Ramshaw, S.~M. Strassel,
  and R.~M. Weischedel.
\newblock {\em The Automatic Content Extraction (ACE) Program - Tasks, Data,
  and Evaluation}.
\newblock In Proceedings of the 2004 International Conference on Language
  Resources and Evaluation Workshop on Linguistics, pages 837--840, 2004.

\bibitem{walker2006ace}
C.~Walker, S.~Strassel, J.~Medero, and K.~Maeda.
\newblock {\em ACE 2005 multilingual training corpus}.
\newblock Linguistic Data Consortium, Philadelphia, 57, 2006.

\bibitem{roth2004linear}
D.~Roth and W.-t. Yih.
\newblock {\em A linear programming formulation for global inference in natural
  language tasks}.
\newblock Technical report, Illinois Univ at Urbana-Champaign Dept of Computer
  Science, 2004.

\bibitem{hendrickx2010semeval}
I.~Hendrickx, S.~N. Kim, Z.~Kozareva, P.~Nakov, D.~{\'O}. S{\'e}aghdha,
  S.~Pad{\'o}, M.~Pennacchiotti, L.~Romano, and S.~Szpakowicz.
\newblock {\em SemEval-2010 Task 8: Multi-Way Classification of Semantic
  Relations between Pairs of Nominals}.
\newblock In Proceedings of the 5th International Workshop on Semantic
  Evaluation, pages 33--38, 2010.

\bibitem{luan2018multi}
Y.~Luan, L.~He, M.~Ostendorf, and H.~Hajishirzi.
\newblock {\em Multi-Task Identification of Entities, Relations, and
  Coreference for Scientific Knowledge Graph Construction}.
\newblock In Proceedings of the 2018 Conference on Empirical Methods in Natural
  Language Processing, pages 3219--3232, 2018.

\bibitem{ji2010overview}
H.~Ji, R.~Grishman, H.~T. Dang, K.~Griffitt, and J.~Ellis.
\newblock {\em Overview of the TAC 2010 knowledge base population track}.
\newblock In Proceedings of the 2010 Text Analysis Conference, pages 1--25,
  2010.
\newblock Available from:
  \url{https://blender.cs.illinois.edu/paper/kbp2010overview.pdf}.

\bibitem{ji2015overview}
H.~Ji, J.~Nothman, B.~Hachey, and R.~Florian.
\newblock {\em Overview of {TAC-KBP 2015} Tri-lingual Entity Discovery and
  Linking.}
\newblock In Proceedings of the 2015 Text Analysis Conference (TAC-KBP 2015),
  2015.
\newblock Available from:
  \url{https://tac.nist.gov/publications/2015/additional.papers/TAC2015.KBP\_Trilingual\_Entity\_Discovery\_and\_Linking\_overview.proceedings.pdf}.

\bibitem{ji2017overview}
H.~Ji, X.~Pan, B.~Zhang, J.~Nothman, J.~Mayfield, P.~McNamee, C.~Costello, and
  S.~I. Hub.
\newblock {\em Overview of TAC-KBP2017 13 Languages Entity Discovery and
  Linking.}
\newblock In Proceedings of the 2017 Text Analysis Conference, 2017.

\bibitem{sevgili2020neural}
{\"{O}}.~Sevgili, A.~Shelmanov, M.~Y. Arkhipov, A.~Panchenko, and C.~Biemann.
\newblock {\em Neural entity linking: {A} survey of models based on deep
  learning}.
\newblock Semantic Web, 13(3):527--570, 2022.
\newblock Available from: \url{https://doi.org/10.3233/SW-222986}.

\bibitem{ganea2017deep}
O.-E. Ganea and T.~Hofmann.
\newblock {\em Deep Joint Entity Disambiguation with Local Neural Attention}.
\newblock In Proceedings of the 2017 Conference on Empirical Methods in Natural
  Language Processing (EMNLP 2017), pages 2619--2629, 2017.
\newblock Available from: \url{https://www.aclweb.org/anthology/D17-1277}.

\bibitem{kolitsas2018end}
N.~Kolitsas, O.-E. Ganea, and T.~Hofmann.
\newblock {\em End-to-End Neural Entity Linking}.
\newblock In Proceedings of the 2018 Conference on Computational Natural
  Language Learning (CoNLL 2018), pages 519--529, 2018.
\newblock Available from: \url{https://www.aclweb.org/anthology/K18-1050/}.

\bibitem{zhang2021entqa}
W.~Zhang, W.~Hua, and K.~Stratos.
\newblock {\em {EntQA}: Entity Linking as Question Answering}.
\newblock In Proceedings of the 2022 International Conference on Learning
  Representations (ICLR 2022), 2022.
\newblock Available from: \url{https://arxiv.org/abs/2110.02369}.

\bibitem{rao2013entity}
D.~Rao, P.~McNamee, and M.~Dredze.
\newblock {\em Entity linking: Finding extracted entities in a knowledge base}.
\newblock In Multi-Source, Multilingual Information Extraction and
  Summarization, pages 93--115. Springer, 2013.
\newblock Available from: \url{https://doi.org/10.1007/978-3-642-28569-1\_5}.

\bibitem{wu2019zero}
L.~Wu, F.~Petroni, M.~Josifoski, S.~Riedel, and L.~Zettlemoyer.
\newblock {\em Zero-shot entity linking with dense entity retrieval}.
\newblock In Proceedings of the 2020 Conference on Empirical Methods in Natural
  Language Processing (EMNLP 2020), pages 6397--6407, 2020.
\newblock Available from: \url{https://aclanthology.org/2020.emnlp-main.519}.

\bibitem{logeswaran2019zero}
L.~Logeswaran, M.-W. Chang, K.~Lee, K.~Toutanova, J.~Devlin, and H.~Lee.
\newblock {\em Zero-Shot Entity Linking by Reading Entity Descriptions}.
\newblock In Proceedings of the 2019 Annual Meeting of the Association for
  Computational Linguistics (ACL 2019), pages 3449--3460, 2019.
\newblock Available from: \url{https://aclanthology.org/P19-1335}.

\bibitem{onoe2020fine}
Y.~Onoe and G.~Durrett.
\newblock {\em Fine-Grained Entity Typing for Domain Independent Entity
  Linking.}
\newblock In Proceedings of the 2020 Conference on Artificial Intelligence
  (AAAI 2020), pages 8576--8583, 2020.
\newblock Available from:
  \url{https://ojs.aaai.org/index.php/AAAI/article/view/6380}.

\bibitem{raiman2022deeptype}
J.~Raiman.
\newblock {\em {DeepType} 2: Superhuman Entity Linking All You Need Is Type
  Interactions}.
\newblock In Proceedings of the 2022 Conference on Artificial Intelligence
  ({AAAI} 2022), 2022.
\newblock Available from:
  \url{https://www.aaai.org/AAAI22Papers/AAAI-2612.RaimanJ.pdf}.

\bibitem{mihalcea2007wikify}
R.~Mihalcea and A.~Csomai.
\newblock {\em Wikify! Linking documents to encyclopedic knowledge}.
\newblock In Proceedings of the sixteenth ACM conference on Conference on
  information and knowledge management, pages 233--242, 2007.

\bibitem{yamada2020global}
I.~Yamada, K.~Washio, H.~Shindo, and Y.~Matsumoto.
\newblock {\em Global Entity Disambiguation with Pretrained Contextualized
  Embeddings of Words and Entities}.
\newblock CoRR, 2020.
\newblock Available from: \url{https://arxiv.org/abs/1909.00426}.

\bibitem{orr2020bootleg}
L.~Orr, M.~Leszczynski, S.~Arora, S.~Wu, N.~Guha, X.~Ling, and C.~Re.
\newblock {\em Bootleg: Chasing the tail with self-supervised named entity
  disambiguation}.
\newblock In Proceedings of the 2021 Conference on Innovative Data Systems
  Research (CIDR 2021), 2021.
\newblock Available from:
  \url{http://cidrdb.org/cidr2021/papers/cidr2021\_paper13.pdf}.

\bibitem{de2020autoregressive}
N.~De~Cao, G.~Izacard, S.~Riedel, and F.~Petroni.
\newblock {\em Autoregressive Entity Retrieval}.
\newblock In Proceedings of the 2021 International Conference on Learning
  Representations (ICLR 2021), 2021.
\newblock Available from: \url{https://openreview.net/forum?id=5k8F6UU39V}.

\bibitem{de2021highly}
N.~De~Cao, W.~Aziz, and I.~Titov.
\newblock {\em Highly Parallel Autoregressive Entity Linking with
  Discriminative Correction}.
\newblock In Proceedings of the 2021 Conference on Empirical Methods in Natural
  Language Processing (EMNLP 2021), pages 7662--7669, 2021.
\newblock Available from: \url{https://aclanthology.org/2021.emnlp-main.604},
  doi:10.18653/v1/2021.emnlp-main.604.

\bibitem{hoffart2011robust}
J.~Hoffart, M.~A. Yosef, I.~Bordino, H.~F{\"u}rstenau, M.~Pinkal, M.~Spaniol,
  B.~Taneva, S.~Thater, and G.~Weikum.
\newblock {\em Robust disambiguation of named entities in text}.
\newblock In Proceedings of the 2011 Conference on Empirical Methods in Natural
  Language Processing (EMNLP 2011), pages 782--792, 2011.
\newblock Available from: \url{https://www.aclweb.org/anthology/D11-1072/}.

\bibitem{milne2008learning}
D.~Milne and I.~H. Witten.
\newblock {\em Learning to link with wikipedia}.
\newblock In Proceedings of the 2008 ACM conference on Information and
  knowledge management ({CIKM} 2008), pages 509--518, 2008.
\newblock Available from: \url{https://doi.org/10.1145/1458082.1458150}.

\bibitem{ratinov2011local}
L.~Ratinov, D.~Roth, D.~Downey, and M.~Anderson.
\newblock {\em Local and global algorithms for disambiguation to {W}ikipedia}.
\newblock In Proceedings of the 2011 Annual Meeting of the Association for
  Computational Linguistics (ACL 2011), pages 1375--1384, 2011.
\newblock Available from: \url{https://aclanthology.org/P11-1138/}.

\bibitem{roder2014n3}
M.~R{\"o}der, R.~Usbeck, S.~Hellmann, D.~Gerber, and A.~Both.
\newblock {\em N$^3$-A Collection of Datasets for Named Entity Recognition and
  Disambiguation in the NLP Interchange Format}.
\newblock In Proceedings of the 2014 International Conference on Language
  Resources and Evaluation (LREC 2014), pages 3529--3533, 2014.
\newblock Available from:
  \url{http://www.lrec-conf.org/proceedings/lrec2014/pdf/856_Paper.pdf}.

\bibitem{rosales2018voxel}
H.~Rosales-M{\'e}ndez, A.~Hogan, and B.~Poblete.
\newblock {\em VoxEL: a benchmark dataset for multilingual entity linking}.
\newblock In Proceedings of the 2018 International Semantic Web Conference
  (ISWC 2018), pages 170--186, 2018.
\newblock Available from: \url{https://doi.org/10.1007/978-3-030-00668-6\_11}.

\bibitem{kulkarni2009collective}
S.~Kulkarni, A.~Singh, G.~Ramakrishnan, and S.~Chakrabarti.
\newblock {\em Collective annotation of {W}ikipedia entities in web text}.
\newblock In Proceedings of the 2009 ACM International Conference on Knowledge
  Discovery and Data Mining (SIGKDD 2009), pages 457--466, 2009.
\newblock Available from: \url{https://doi.org/10.1145/1557019.1557073}.

\bibitem{nuzzolese2015open}
A.~G. Nuzzolese, A.~L. Gentile, V.~Presutti, A.~Gangemi, D.~Garigliotti, and
  R.~Navigli.
\newblock {\em Open knowledge extraction challenge}.
\newblock In Proceedings of the 2015 Semantic Web Evaluation Challenges
  (SemWebEval@ESWC 2015), pages 3--15, 2015.
\newblock Available from: \url{https://doi.org/10.1007/978-3-319-25518-7\_1}.

\bibitem{derczynski2015analysis}
L.~Derczynski, D.~Maynard, G.~Rizzo, M.~Van~Erp, G.~Gorrell, R.~Troncy,
  J.~Petrak, and K.~Bontcheva.
\newblock {\em Analysis of named entity recognition and linking for tweets}.
\newblock Information Processing \& Management, 51(2):32--49, 2015.
\newblock Available from: \url{https://doi.org/10.1016/j.ipm.2014.10.006}.

\bibitem{guo2018robust}
Z.~Guo and D.~Barbosa.
\newblock {\em Robust Named Entity Disambiguation with Random Walks}.
\newblock Semantic Web, 9(4):459--479, 2018.
\newblock Available from: \url{https://doi.org/10.3233/SW-170273}.

\bibitem{mohan2018medmentions}
S.~Mohan and D.~Li.
\newblock {\em MedMentions: A Large Biomedical Corpus Annotated with UMLS
  Concepts}.
\newblock In Proceedings of the 2018 Automated Knowledge Base Construction
  (AKBC 2018), 2018.
\newblock Available from: \url{https://doi.org/10.24432/C5G59C}.

\bibitem{usbeck2015gerbil}
R.~Usbeck, M.~R{\"o}der, A.-C. Ngonga~Ngomo, C.~Baron, A.~Both, M.~Br{\"u}mmer,
  D.~Ceccarelli, M.~Cornolti, D.~Cherix, B.~Eickmann, et~al.
\newblock {\em {GERBIL}: general entity annotator benchmarking framework}.
\newblock In Proceedings of the 2015 International Conference on World Wide Web
  ({WWW} 2015), pages 1133--1143, 2015.
\newblock Available from: \url{https://doi.org/10.1145/2736277.2741626}.

\bibitem{roder2018gerbil}
M.~R{\"o}der, R.~Usbeck, and A.-C. Ngonga~Ngomo.
\newblock {\em {GERBIL}--benchmarking named entity recognition and linking
  consistently}.
\newblock Semantic Web, 9(5):605--625, 2018.
\newblock Available from: \url{https://doi.org/10.3233/SW-170286}.

\bibitem{petroni2020kilt}
F.~Petroni, A.~Piktus, A.~Fan, P.~Lewis, M.~Yazdani, N.~De~Cao, J.~Thorne,
  Y.~Jernite, V.~Karpukhin, J.~Maillard, et~al.
\newblock {\em {KILT}: a benchmark for knowledge intensive language tasks}.
\newblock In Proceedings of the 2021 Conference of the North American Chapter
  of the Association for Computational Linguistics: Human Language Technologies
  (NAACL-HLT 2021), 2021.
\newblock Available from: \url{https://aclanthology.org/2021.naacl-main.200}.

\bibitem{ayoola2022refined}
T.~Ayoola, S.~Tyagi, J.~Fisher, C.~Christodoulopoulos, and A.~Pierleoni.
\newblock {\em ReFinED: An Efficient Zero-shot-capable Approach to End-to-End
  Entity Linking}.
\newblock arXiv preprint arXiv:2207.04108, 2022.

\bibitem{lee2017end}
K.~Lee, L.~He, M.~Lewis, and L.~Zettlemoyer.
\newblock {\em End-to-end Neural Coreference Resolution}.
\newblock In Proceedings of the 2017 Conference on Empirical Methods in Natural
  Language Processing (EMNLP 2017), pages 188--197, 2017.
\newblock Available from: \url{https://www.aclweb.org/anthology/D17-1018}.

\bibitem{mccallum2003early}
A.~McCallum and W.~Li.
\newblock {\em Early results for Named Entity Recognition with Conditional
  Random Fields, Feature Induction and Web-Enhanced Lexicons}.
\newblock In Proceedings of the Seventh Conference on Natural Language Learning
  at HLT-NAACL 2003, pages 188--191, 2003.

\bibitem{sutton2012introduction}
C.~Sutton, A.~McCallum, et~al.
\newblock {\em An introduction to conditional random fields}.
\newblock Foundations and Trends{\textregistered} in Machine Learning,
  4(4):267--373, 2012.

\bibitem{bekoulis2018joint}
G.~Bekoulis, J.~Deleu, T.~Demeester, and C.~Develder.
\newblock {\em Joint entity recognition and relation extraction as a multi-head
  selection problem}.
\newblock Expert Systems with Applications, 114:34--45, 2018.

\bibitem{lee2018higher}
K.~Lee, L.~He, and L.~Zettlemoyer.
\newblock {\em Higher-Order Coreference Resolution with Coarse-to-Fine
  Inference}.
\newblock In Proceedings of the 2018 Conference of the North American Chapter
  of the Association for Computational Linguistics: Human Language Technologies
  (NAACL-HLT 2018), pages 687--692, 2018.
\newblock Available from: \url{https://www.aclweb.org/anthology/N18-2108}.

\bibitem{luan2019general}
Y.~Luan, D.~Wadden, L.~He, A.~Shah, M.~Ostendorf, and H.~Hajishirzi.
\newblock {\em A general framework for information extraction using dynamic
  span graphs}.
\newblock In Proceedings of the 2019 Conference of the North American Chapter
  of the Association for Computational Linguistics: Human Language Technologies
  (NAACL-HLT 2019), pages 3036--3046, 2019.
\newblock Available from: \url{https://www.aclweb.org/anthology/N19-1308/}.

\bibitem{wadden2019entity}
D.~Wadden, U.~Wennberg, Y.~Luan, and H.~Hajishirzi.
\newblock {\em Entity, Relation, and Event Extraction with Contextualized Span
  Representations}.
\newblock In Proceedings of the 2019 Conference on Empirical Methods in Natural
  Language Processing and International Joint Conference on Natural Language
  Processing, pages 5788--5793, 2019.

\bibitem{he2018jointly}
L.~He, K.~Lee, O.~Levy, and L.~Zettlemoyer.
\newblock {\em Jointly Predicting Predicates and Arguments in Neural Semantic
  Role Labeling}.
\newblock In Proceedings of the 2018 Annual Meeting of the Association for
  Computational Linguistics, pages 364--369, 2018.

\bibitem{joshi2020spanbert}
M.~Joshi, D.~Chen, Y.~Liu, D.~S. Weld, L.~Zettlemoyer, and O.~Levy.
\newblock {\em {S}pan{BERT}: Improving pre-training by representing and
  predicting spans}.
\newblock Transactions of the Association for Computational Linguistics (TACL
  2020), 8:64--77, 2020.
\newblock Available from: \url{https://www.aclweb.org/anthology/2020.tacl-1.5}.

\bibitem{xu2020revealing}
L.~Xu and J.~D. Choi.
\newblock {\em Revealing the Myth of Higher-Order Inference in Coreference
  Resolution}.
\newblock In Proceedings of the 2020 Conference on Empirical Methods in Natural
  Language Processing (EMNLP 2020), pages 8527--8533, 2020.
\newblock Available from:
  \url{https://doi.org/10.18653/v1/2020.emnlp-main.686}.

\bibitem{xu2022modeling}
L.~Xu and J.~D. Choi.
\newblock {\em Modeling Task Interactions in Document-Level Joint Entity and
  Relation Extraction}.
\newblock arXiv preprint arXiv:2205.01909, 2022.

\bibitem{wu2020corefqa}
W.~Wu, F.~Wang, A.~Yuan, F.~Wu, and J.~Li.
\newblock {\em Coref{QA}: Coreference Resolution as Query-based Span
  Prediction}.
\newblock In Proceedings of the 2020 Annual Meeting of the Association for
  Computational Linguistics (ACL 2020), pages 6953--6963, 2020.
\newblock Available from:
  \url{https://www.aclweb.org/anthology/2020.acl-main.622}.

\bibitem{khandelwal2018sharp}
U.~Khandelwal, H.~He, P.~Qi, and D.~Jurafsky.
\newblock {\em Sharp Nearby, Fuzzy Far Away: How Neural Language Models Use
  Context}.
\newblock In Proceedings of the 56th Annual Meeting of the Association for
  Computational Linguistics (Volume 1: Long Papers), pages 284--294, 2018.

\bibitem{wu2020comprehensive}
Z.~Wu, S.~Pan, F.~Chen, G.~Long, C.~Zhang, and S.~Y. Philip.
\newblock {\em A comprehensive survey on graph neural networks}.
\newblock IEEE Transactions on Neural Networks and Learning Systems, pages
  1--21, 2020.

\bibitem{zhou2020graph}
J.~Zhou, G.~Cui, S.~Hu, Z.~Zhang, C.~Yang, Z.~Liu, L.~Wang, C.~Li, and M.~Sun.
\newblock {\em Graph neural networks: A review of methods and applications}.
\newblock AI Open, 1:57--81, 2020.

\bibitem{peters2018deep}
M.~E. Peters, M.~Neumann, M.~Iyyer, M.~Gardner, C.~Clark, K.~Lee, and
  L.~Zettlemoyer.
\newblock {\em Deep contextualized word representations}.
\newblock In Proceedings of NAACL-HLT, pages 2227--2237, 2018.

\bibitem{hochreiter1997long}
S.~Hochreiter and J.~Schmidhuber.
\newblock {\em Long short-term memory}.
\newblock Neural computation, 9(8):1735--1780, 1997.

\bibitem{karpukhin2020dense}
V.~Karpukhin, B.~Oguz, S.~Min, P.~Lewis, L.~Wu, S.~Edunov, D.~Chen, and W.-t.
  Yih.
\newblock {\em Dense Passage Retrieval for Open-Domain Question Answering}.
\newblock In Proceedings of the 2020 Conference on Empirical Methods in Natural
  Language Processing (EMNLP 2020), pages 6769--6781, 2020.
\newblock Available from: \url{https://aclanthology.org/2020.emnlp-main.550}.

\bibitem{ji2021survey}
S.~Ji, S.~Pan, E.~Cambria, P.~Marttinen, and S.~Y. Philip.
\newblock {\em A survey on knowledge graphs: Representation, acquisition, and
  applications}.
\newblock IEEE Transactions on Neural Networks and Learning Systems, 2021.

\bibitem{gutierrez2021knowledge}
C.~Guti{\'e}rrez and J.~F. Sequeda.
\newblock {\em Knowledge graphs}.
\newblock Communications of the ACM, 64(3):96--104, 2021.

\bibitem{yan2018retrospective}
J.~Yan, C.~Wang, W.~Cheng, M.~Gao, and A.~Zhou.
\newblock {\em A retrospective of knowledge graphs}.
\newblock Frontiers of Computer Science, 12(1):55--74, 2018.

\bibitem{vrandevcic2014wikidata}
D.~Vrande{\v{c}}i{\'c} and M.~Kr{\"o}tzsch.
\newblock {\em Wikidata: a free collaborative knowledgebase}.
\newblock Communications of the ACM, 57(10):78--85, 2014.

\bibitem{ellis2014overview}
J.~Ellis, J.~Getman, and S.~M. Strassel.
\newblock {\em Overview of linguistic resources for the tac kbp 2014
  evaluations: Planning, execution, and results}.
\newblock In Proceedings of TAC KBP 2014 Workshop, National Institute of
  Standards and Technology, pages 17--18, 2014.

\bibitem{ellis2015overview}
J.~Ellis, J.~Getman, D.~Fore, N.~Kuster, Z.~Song, A.~Bies, and S.~M. Strassel.
\newblock {\em Overview of Linguistic Resources for the {TAC} {KBP} 2015
  Evaluations: Methodologies and Results.}
\newblock In Proceedings of the 2015 Text Analysis Conference, 2015.

\bibitem{song2015light}
Z.~Song, A.~Bies, S.~Strassel, T.~Riese, J.~Mott, J.~Ellis, J.~Wright,
  S.~Kulick, N.~Ryant, and X.~Ma.
\newblock {\em From light to rich ere: annotation of entities, relations, and
  events}.
\newblock In Proceedings of the the 3rd Workshop on EVENTS: Definition,
  Detection, Coreference, and Representation, pages 89--98, 2015.

\bibitem{hoffart2012kore}
J.~Hoffart, S.~Seufert, D.~B. Nguyen, M.~Theobald, and G.~Weikum.
\newblock {\em KORE: keyphrase overlap relatedness for entity disambiguation}.
\newblock In Proceedings of the 21st ACM international conference on
  Information and knowledge management, pages 545--554, 2012.

\bibitem{ceccarelli2013dexter}
D.~Ceccarelli, C.~Lucchese, S.~Orlando, R.~Perego, and S.~Trani.
\newblock {\em Dexter: an open source framework for entity linking}.
\newblock In Proceedings of the sixth international workshop on Exploiting
  semantic annotations in information retrieval, pages 17--20, 2013.

\bibitem{van2013learning}
M.~Van~Erp, G.~Rizzo, and R.~Troncy.
\newblock {\em Learning with the Web: Spotting Named Entities on the
  Intersection of NERD and Machine Learning.}
\newblock In \# MSM, pages 27--30. Citeseer, 2013.

\bibitem{piccinno2014tagme}
F.~Piccinno and P.~Ferragina.
\newblock {\em From TagME to WAT: a new entity annotator}.
\newblock In Proceedings of the first international workshop on Entity
  recognition \& disambiguation, pages 55--62, 2014.

\bibitem{provatorova2020named}
V.~Provatorova, S.~Vakulenko, E.~Kanoulas, K.~Dercksen, and J.~M. van Hulst.
\newblock {\em Named Entity Recognition and Linking on Historical Newspapers:
  UvA. ILPS \& REL at CLEF HIPE 2020.}
\newblock In CLEF (Working Notes), 2020.

\bibitem{eshel2017named}
Y.~Eshel, N.~Cohen, K.~Radinsky, S.~Markovitch, I.~Yamada, and O.~Levy.
\newblock {\em Named Entity Disambiguation for Noisy Text}.
\newblock In Proceedings of the 2017 Conference on Computational Natural
  Language Learning (CoNLL 2017), pages 58--68, 2017.
\newblock Available from: \url{https://aclanthology.org/K17-1008}.

\bibitem{broscheit2019investigating}
S.~Broscheit.
\newblock {\em Investigating Entity Knowledge in {BERT} with Simple Neural
  End-To-End Entity Linking}.
\newblock In Proceedings of the 23rd Conference on Computational Natural
  Language Learning (CoNLL 2019), pages 677--685, 2019.
\newblock Available from: \url{https://www.aclweb.org/anthology/K19-1063}.

\bibitem{barba2022extend}
E.~Barba, L.~Procopio, and R.~Navigli.
\newblock {\em ExtEnD: Extractive Entity Disambiguation}.
\newblock In Proceedings of the 60th Annual Meeting of the Association for
  Computational Linguistics (Volume 1: Long Papers), pages 2478--2488, 2022.

\bibitem{martins2019joint}
P.~H. Martins, Z.~Marinho, and A.~F. Martins.
\newblock {\em Joint Learning of Named Entity Recognition and Entity Linking}.
\newblock In Proceedings of the 57th Annual Meeting of the Association for
  Computational Linguistics: Student Research Workshop, pages 190--196, 2019.

\bibitem{bekoulis2018adversarial}
G.~Bekoulis, J.~Deleu, T.~Demeester, and C.~Develder.
\newblock {\em Adversarial training for multi-context joint entity and relation
  extraction}.
\newblock In Proceedings of the 2018 Conference on Empirical Methods in Natural
  Language Processing, pages 2830--2836, 2018.

\bibitem{ruder2019latent}
S.~Ruder, J.~Bingel, I.~Augenstein, and A.~S{\o}gaard.
\newblock {\em Latent multi-task architecture learning}.
\newblock In Proceedings of the AAAI Conference on Artificial Intelligence,
  volume~33, pages 4822--4829, 2019.

\bibitem{collobert2011natural}
R.~Collobert, J.~Weston, L.~Bottou, M.~Karlen, K.~Kavukcuoglu, and P.~Kuksa.
\newblock {\em Natural language processing (almost) from scratch}.
\newblock Journal of machine learning research, 12(ARTICLE):2493--2537, 2011.

\bibitem{augenstein2018multi}
I.~Augenstein, S.~Ruder, and A.~S{\o}gaard.
\newblock {\em Multi-Task Learning of Pairwise Sequence Classification Tasks
  over Disparate Label Spaces}.
\newblock In Proceedings of the 2018 Conference of the North American Chapter
  of the Association for Computational Linguistics: Human Language
  Technologies, Volume 1 (Long Papers), pages 1896--1906, 2018.

\bibitem{joshi2019bert}
M.~Joshi, O.~Levy, L.~Zettlemoyer, and D.~S. Weld.
\newblock {\em {BERT} for Coreference Resolution: Baselines and Analysis}.
\newblock In Proceedings of the 2019 Conference on Empirical Methods in Natural
  Language Processing and the International Joint Conference on Natural
  Language Processing (EMNLP-IJCNLP 2019), pages 5807--5812, 2019.
\newblock Available from: \url{https://www.aclweb.org/anthology/D19-1588}.

\bibitem{yasunaga2021qa}
M.~Yasunaga, H.~Ren, A.~Bosselut, P.~Liang, and J.~Leskovec.
\newblock {\em {QA-GNN}: Reasoning with Language Models and Knowledge Graphs
  for Question Answering}.
\newblock In Proceedings of the 2021 Conference of the North American Chapter
  of the Association for Computational Linguistics: Human Language Technologies
  (NAACL-HLT 2021), pages 535--546, 2021.
\newblock Available from: \url{https://aclanthology.org/2021.naacl-main.45}.

\bibitem{zhang2021document}
N.~Zhang, X.~Chen, X.~Xie, S.~Deng, C.~Tan, M.~Chen, F.~Huang, L.~Si, and
  H.~Chen.
\newblock {\em Document-level Relation Extraction as Semantic Segmentation}.
\newblock In IJCAI, 2021.

\bibitem{zaporojets2021solving}
K.~Zaporojets, G.~Bekoulis, J.~Deleu, T.~Demeester, and C.~Develder.
\newblock {\em Solving arithmetic word problems by scoring equations with
  recursive neural networks}.
\newblock Expert Systems with Applications, 174:114704, 2021.

\bibitem{hajishirzi2013joint}
H.~Hajishirzi, L.~Zilles, D.~S. Weld, and L.~Zettlemoyer.
\newblock {\em Joint coreference resolution and named-entity linking with
  multi-pass sieves}.
\newblock In Proceedings of the 2013 Conference on Empirical Methods in Natural
  Language Processing (EMNLP 2013), pages 289--299, 2013.
\newblock Available from: \url{https://aclanthology.org/D13-1029/}.

\bibitem{dutta2015c3el}
S.~Dutta and G.~Weikum.
\newblock {\em {C3EL}: A joint model for cross-document co-reference resolution
  and entity linking}.
\newblock In Proceedings of the 2020 Conference on Empirical Methods in Natural
  Language Processing (EMNLP 2015), pages 846--856, 2015.
\newblock Available from: \url{https://doi.org/10.18653/v1/d15-1101}.

\bibitem{angell2021clustering}
R.~Angell, N.~Monath, S.~Mohan, N.~Yadav, and A.~McCallum.
\newblock {\em Clustering-based Inference for Biomedical Entity Linking}.
\newblock In Proceedings of the 2021 Conference of the North American Chapter
  of the Association for Computational Linguistics: Human Language Technologies
  (NAACL-HLT 2021), pages 2598--2608, 2021.
\newblock Available from:
  \url{https://doi.org/10.18653/v1/2021.naacl-main.205}.

\bibitem{zaporojets2021towards}
K.~Zaporojets, J.~Deleu, T.~Demeester, and C.~Develder.
\newblock {\em Towards Consistent Document-level Entity Linking: Joint Models
  for Entity Linking and Coreference Resolution}.
\newblock arXiv preprint arXiv:2108.13530, 2021.

\bibitem{yu2017improved}
M.~Yu, W.~Yin, K.~S. Hasan, C.~dos Santos, B.~Xiang, and B.~Zhou.
\newblock {\em Improved Neural Relation Detection for Knowledge Base Question
  Answering}.
\newblock In Proceedings of the 55th Annual Meeting of the Association for
  Computational Linguistics (Volume 1: Long Papers), pages 571--581, 2017.

\bibitem{hu2019language}
R.~Hu, A.~Rohrbach, T.~Darrell, and K.~Saenko.
\newblock {\em Language-conditioned graph networks for relational reasoning}.
\newblock In Proceedings of the IEEE International Conference on Computer
  Vision, pages 10294--10303, 2019.

\bibitem{gao2019interconnected}
Y.~Gao, P.~Li, I.~King, and M.~R. Lyu.
\newblock {\em Interconnected Question Generation with Coreference Alignment
  and Conversation Flow Modeling}.
\newblock In Proceedings of the 57th Annual Meeting of the Association for
  Computational Linguistics, pages 4853--4862, 2019.

\bibitem{bhattacharjee2020investigating}
S.~Bhattacharjee, R.~Haque, G.~M. de~Buy~Wenniger, and A.~Way.
\newblock {\em Investigating query expansion and coreference resolution in
  question answering on BERT}.
\newblock In International Conference on Applications of Natural Language to
  Information Systems, pages 47--59. Springer, 2020.

\bibitem{molla2006named}
D.~Molla, M.~van Zaanen, and D.~Smith.
\newblock {\em Named Entity Recognition for Question Answering}.
\newblock In Proceedings of the Australasian Language Technology Workshop 2006,
  pages 51--58, 2006.

\bibitem{singh2018reinvent}
K.~Singh, A.~S. Radhakrishna, A.~Both, S.~Shekarpour, I.~Lytra, R.~Usbeck,
  A.~Vyas, A.~Khikmatullaev, D.~Punjani, C.~Lange, et~al.
\newblock {\em Why reinvent the wheel: Let's build question answering systems
  together}.
\newblock In Proceedings of the 2018 World Wide Web Conference, pages
  1247--1256, 2018.

\bibitem{chen2017reading}
D.~Chen, A.~Fisch, J.~Weston, and A.~Bordes.
\newblock {\em Reading wikipedia to answer open-domain questions}.
\newblock arXiv preprint arXiv:1704.00051, 2017.

\bibitem{wang2018dkn}
H.~Wang, F.~Zhang, X.~Xie, and M.~Guo.
\newblock {\em DKN: Deep knowledge-aware network for news recommendation}.
\newblock In Proceedings of the 2018 world wide web conference, pages
  1835--1844, 2018.

\bibitem{karimi2018news}
M.~Karimi, D.~Jannach, and M.~Jugovac.
\newblock {\em News recommender systems--Survey and roads ahead}.
\newblock Information Processing \& Management, 54(6):1203--1227, 2018.

\bibitem{wang2019multi}
Z.~Wang, P.~Ng, X.~Ma, R.~Nallapati, and B.~Xiang.
\newblock {\em Multi-passage BERT: A Globally Normalized BERT Model for
  Open-domain Question Answering}.
\newblock In Proceedings of the 2019 Conference on Empirical Methods in Natural
  Language Processing and the 9th International Joint Conference on Natural
  Language Processing (EMNLP-IJCNLP), pages 5878--5882, 2019.

\bibitem{thorne2018automated}
J.~Thorne and A.~Vlachos.
\newblock {\em Automated Fact Checking: Task Formulations, Methods and Future
  Directions}.
\newblock In Proceedings of the 27th International Conference on Computational
  Linguistics, pages 3346--3359, 2018.

\bibitem{zhang2020overview}
X.~Zhang and A.~A. Ghorbani.
\newblock {\em An overview of online fake news: Characterization, detection,
  and discussion}.
\newblock Information Processing \& Management, 57(2):102025, 2020.

\bibitem{sun2017review}
S.~Sun, C.~Luo, and J.~Chen.
\newblock {\em A review of natural language processing techniques for opinion
  mining systems}.
\newblock Information fusion, 36:10--25, 2017.

\bibitem{cifariello2019wiser}
P.~Cifariello, P.~Ferragina, and M.~Ponza.
\newblock {\em Wiser: A semantic approach for expert finding in academia based
  on entity linking}.
\newblock Information Systems, 82:1--16, 2019.

\bibitem{roller2020recipes}
S.~Roller, E.~Dinan, N.~Goyal, D.~Ju, M.~Williamson, Y.~Liu, J.~Xu, M.~Ott,
  K.~Shuster, E.~M. Smith, et~al.
\newblock {\em Recipes for building an open-domain chatbot}.
\newblock arXiv preprint arXiv:2004.13637, 2020.

\bibitem{fei2020boundaries}
H.~Fei, Y.~Ren, and D.~Ji.
\newblock {\em Boundaries and edges rethinking: An end-to-end neural model for
  overlapping entity relation extraction}.
\newblock Information Processing \& Management, 57(6):102311, 2020.

\bibitem{augenstein2017semeval}
I.~Augenstein, M.~Das, S.~Riedel, L.~Vikraman, and A.~McCallum.
\newblock {\em SemEval 2017 Task 10: ScienceIE-Extracting Keyphrases and
  Relations from Scientific Publications}.
\newblock In Proceedings of the 11th International Workshop on Semantic
  Evaluation (SemEval-2017), pages 546--555, 2017.

\bibitem{han2018fewrel}
X.~Han, H.~Zhu, P.~Yu, Z.~Wang, Y.~Yao, Z.~Liu, and M.~Sun.
\newblock {\em {FewRel}: A Large-Scale Supervised Few-Shot Relation
  Classification Dataset with State-of-the-Art Evaluation}.
\newblock In Proceedings of the 2018 Conference on Empirical Methods in Natural
  Language Processing (EMNLP 2018), pages 4803--4809, 2018.
\newblock Available from: \url{https://aclanthology.org/D18-1514}.

\bibitem{riedel2010modeling}
S.~Riedel, L.~Yao, and A.~McCallum.
\newblock {\em Modeling relations and their mentions without labeled text}.
\newblock In Proceedings of the 2010 European Conference on Machine Learning
  and Knowledge Discovery in Databases, pages 148--163, 2010.

\bibitem{quirk2017distant}
C.~Quirk and H.~Poon.
\newblock {\em Distant Supervision for Relation Extraction beyond the Sentence
  Boundary}.
\newblock In Proceedings of the 2017 Conference of the European Chapter of the
  Association for Computational Linguistics, pages 1171--1182, 2017.

\bibitem{peng2017cross}
N.~Peng, H.~Poon, C.~Quirk, K.~Toutanova, and W.-t. Yih.
\newblock {\em Cross-sentence n-ary relation extraction with graph lstms}.
\newblock Transactions of the Association for Computational Linguistics,
  5:101--115, 2017.

\bibitem{li2015gated}
Y.~Li, D.~Tarlow, M.~Brockschmidt, and R.~Zemel.
\newblock {\em Gated graph sequence neural networks}.
\newblock In Proceedings of the 2016 International Conference on Learning
  Representations, 2016.

\bibitem{xu2018powerful}
K.~Xu, W.~Hu, J.~Leskovec, and S.~Jegelka.
\newblock {\em How Powerful are Graph Neural Networks?}
\newblock In Proceedings of the 2018 International Conference on Learning
  Representations, 2018.

\bibitem{kantor2019coreference}
B.~Kantor and A.~Globerson.
\newblock {\em Coreference resolution with entity equalization}.
\newblock In Proceedings of the 2019 Annual Meeting of the Association for
  Computational Linguistics, pages 673--677, 2019.

\bibitem{fu2019graphrel}
T.-J. Fu, P.-H. Li, and W.-Y. Ma.
\newblock {\em GraphRel: Modeling text as relational graphs for joint entity
  and relation extraction}.
\newblock In Proceedings of the 2019 Annual Meeting of the Association for
  Computational Linguistics, pages 1409--1418, 2019.

\bibitem{chinchor1998muc}
N.~Chinchor and E.~Marsh.
\newblock {\em Muc-7 information extraction task definition}.
\newblock In Proceeding of the 1998 Message Understanding Conference (MUC-7),
  pages 359--367, 1998.

\bibitem{aguilar2014comparison}
J.~Aguilar, C.~Beller, P.~McNamee, B.~Van~Durme, S.~Strassel, Z.~Song, and
  J.~Ellis.
\newblock {\em A comparison of the events and relations across ace, ere,
  tac-kbp, and framenet annotation standards}.
\newblock In Proceedings of the 2nd Workshop on EVENTS: Definition, Detection,
  Coreference, and Representation, pages 45--53, 2014.

\bibitem{hovy2006ontonotes}
E.~Hovy, M.~Marcus, M.~Palmer, L.~Ramshaw, and R.~Weischedel.
\newblock {\em OntoNotes: the 90\% solution}.
\newblock In Proceedings of the 2006 Conference of the North American Chapter
  of the Association for Computational Linguistics: Human Language
  Technologies, pages 57--60, 2006.

\bibitem{kim2003genia}
J.-D. Kim, T.~Ohta, Y.~Tateisi, and J.~Tsujii.
\newblock {\em GENIA corpus - a semantically annotated corpus for
  bio-textmining}.
\newblock Bioinformatics, 19(suppl\_1):180--182, 2003.

\bibitem{bentivogli2010extending}
L.~Bentivogli, P.~Forner, C.~Giuliano, A.~Marchetti, E.~Pianta, and
  K.~Tymoshenko.
\newblock {\em Extending English ACE 2005 Corpus Annotation with Ground-truth
  Links to Wikipedia}.
\newblock In Proceedings of the 2nd Workshop on The People's Web Meets NLP:
  Collaboratively Constructed Semantic Resources, pages 19--27, 2010.

\bibitem{lample2016neural}
G.~Lample, M.~Ballesteros, S.~Subramanian, K.~Kawakami, and C.~Dyer.
\newblock {\em Neural Architectures for Named Entity Recognition}.
\newblock In Proceedings of the 2016 Conference of the North American Chapter
  of the Association for Computational Linguistics: Human Language
  Technologies, pages 260--270, 2016.

\bibitem{chiu2016named}
J.~P. Chiu and E.~Nichols.
\newblock {\em Named Entity Recognition with Bidirectional LSTM-CNNs}.
\newblock Transactions of the Association for Computational Linguistics,
  4:357--370, 2016.

\bibitem{baevski2019cloze}
A.~Baevski, S.~Edunov, Y.~Liu, L.~Zettlemoyer, and M.~Auli.
\newblock {\em Cloze-driven Pretraining of Self-attention Networks}.
\newblock In Proceedings of the 2019 Conference on Empirical Methods in Natural
  Language Processing and International Joint Conference on Natural Language
  Processing, pages 5363--5372, 2019.

\bibitem{akbik2019pooled}
A.~Akbik, T.~Bergmann, and R.~Vollgraf.
\newblock {\em Pooled contextualized embeddings for named entity recognition}.
\newblock In Proceedings of the 2019 Conference of the North American Chapter
  of the Association for Computational Linguistics: Human Language
  Technologies, pages 724--728, 2019.

\bibitem{akbik2018contextual}
A.~Akbik, D.~Blythe, and R.~Vollgraf.
\newblock {\em Contextual string embeddings for sequence labeling}.
\newblock In Proceedings of the 2018 International Conference on Computational
  Linguistics, pages 1638--1649, 2018.

\bibitem{clark2018semi}
K.~Clark, M.-T. Luong, C.~D. Manning, and Q.~Le.
\newblock {\em Semi-Supervised Sequence Modeling with Cross-View Training}.
\newblock In Proceedings of the 2018 Conference on Empirical Methods in Natural
  Language Processing, pages 1914--1925, 2018.

\bibitem{strubell2017fast}
E.~Strubell, P.~Verga, D.~Belanger, and A.~McCallum.
\newblock {\em Fast and Accurate Entity Recognition with Iterated Dilated
  Convolutions}.
\newblock In Proceedings of the 2017 Conference on Empirical Methods in Natural
  Language Processing, pages 2670--2680, 2017.

\bibitem{li2014incremental}
Q.~Li and H.~Ji.
\newblock {\em Incremental joint extraction of entity mentions and relations}.
\newblock In Proceedings of the 2014 Annual Meeting of the Association for
  Computational Linguistics, pages 402--412, 2014.

\bibitem{zhang2017end}
M.~Zhang, Y.~Zhang, and G.~Fu.
\newblock {\em End-to-end neural relation extraction with global optimization}.
\newblock In Proceedings of the 2017 Conference on Empirical Methods in Natural
  Language Processing, pages 1730--1740, 2017.

\bibitem{soares2019matching}
L.~B. Soares, N.~FitzGerald, J.~Ling, and T.~Kwiatkowski.
\newblock {\em Matching the Blanks: Distributional Similarity for Relation
  Learning}.
\newblock In Proceedings of the 2019 Annual Meeting of the Association for
  Computational Linguistics, pages 2895--2905, 2019.

\bibitem{zhang2018graph}
Y.~Zhang, P.~Qi, and C.~D. Manning.
\newblock {\em Graph Convolution over Pruned Dependency Trees Improves Relation
  Extraction}.
\newblock In Proceedings of the 2018 Conference on Empirical Methods in Natural
  Language Processing, pages 2205--2215, 2018.

\bibitem{hu2020cross}
W.~Hu, B.~Ma, Z.~Li, Y.~Li, and Y.~Wang.
\newblock {\em A Cross-Media Deep Relationship Classification Method Using
  Discrimination Information}.
\newblock Information Processing \& Management, 57(6):102344, 2020.

\bibitem{peters2019knowledge}
M.~E. Peters, M.~Neumann, R.~Logan, R.~Schwartz, V.~Joshi, S.~Singh, and N.~A.
  Smith.
\newblock {\em Knowledge Enhanced Contextual Word Representations}.
\newblock In Proceedings of the 2019 Conference on Empirical Methods in Natural
  Language Processing and the International Joint Conference on Natural
  Language Processing (EMNLP-IJCNLP 2019), pages 43--54, 2019.
\newblock Available from: \url{https://doi.org/10.18653/v1/D19-1005}.

\bibitem{guo2019attention}
Z.~Guo, Y.~Zhang, and W.~Lu.
\newblock {\em Attention Guided Graph Convolutional Networks for Relation
  Extraction}.
\newblock In Proceedings of the 2019 Annual Meeting of the Association for
  Computational Linguistics, pages 241--251, 2019.

\bibitem{mchugh2012interrater}
M.~L. McHugh.
\newblock {\em Interrater reliability: the kappa statistic}.
\newblock Biochemia medica: Biochemia medica, 22(3):276--282, 2012.

\bibitem{landis1977measurement}
J.~Landis and G.~Koch.
\newblock {\em The measurement of observer agreement for categorical data.}
\newblock Biometrics, 33(1):159--174, 1977.

\bibitem{scarselli2008graph}
F.~Scarselli, M.~Gori, A.~C. Tsoi, M.~Hagenbuchner, and G.~Monfardini.
\newblock {\em The graph neural network model}.
\newblock IEEE Transactions on Neural Networks, 20(1):61--80, 2008.

\bibitem{dixit2019span}
K.~Dixit and Y.~Al-Onaizan.
\newblock {\em Span-level model for relation extraction}.
\newblock In Proceedings of the 2019 Annual Meeting of the Association for
  Computational Linguistics, pages 5308--5314, 2019.

\bibitem{ma2016end}
X.~Ma and E.~Hovy.
\newblock {\em End-to-end Sequence Labeling via Bi-directional LSTM-CNNs-CRF}.
\newblock In Proceedings of the 2016 Annual Meeting of the Association for
  Computational Linguistics, pages 1064--1074, 2016.

\bibitem{luan2017scientific}
Y.~Luan, M.~Ostendorf, and H.~Hajishirzi.
\newblock {\em Scientific Information Extraction with Semi-supervised Neural
  Tagging}.
\newblock In Proceedings of the 2017 Conference on Empirical Methods in Natural
  Language Processing, pages 2641--2651, 2017.

\bibitem{katiyar2018nested}
A.~Katiyar and C.~Cardie.
\newblock {\em Nested named entity recognition revisited}.
\newblock In Proceedings of the 2018 Conference of the North American Chapter
  of the Association for Computational Linguistics: Human Language
  Technologies, pages 861--871, 2018.

\bibitem{pennington2014glove}
J.~Pennington, R.~Socher, and C.~Manning.
\newblock {\em Glove: {G}lobal vectors for word representation}.
\newblock In Proceedings of the 2014 conference on empirical methods in natural
  language processing (EMNLP), pages 1532--1543, 2014.

\bibitem{devlin2019bert}
J.~Devlin, M.-W. Chang, K.~Lee, and K.~Toutanova.
\newblock {\em {BERT}: Pre-training of Deep Bidirectional Transformers for
  Language Understanding}.
\newblock In Proceedings of the 2019 Conference of the North American Chapter
  of the Association for Computational Linguistics: Human Language Technologies
  (NAACL-HLT 2019), pages 4171--4186, 2019.
\newblock Available from: \url{https://www.aclweb.org/anthology/N19-1423}.

\bibitem{pradhan2014scoring}
S.~Pradhan, X.~Luo, M.~Recasens, E.~Hovy, V.~Ng, and M.~Strube.
\newblock {\em Scoring Coreference Partitions of Predicted Mentions: A
  Reference Implementation}.
\newblock In Proceedings of the 2014 Annual Meeting of the Association for
  Computational Linguistics (ACL 2014), pages 30--35, 2014.
\newblock Available from: \url{https://doi.org/10.3115/v1/p14-2006}.

\bibitem{bagga1998algorithms}
A.~Bagga and B.~Baldwin.
\newblock {\em Algorithms for scoring coreference chains}.
\newblock In Proceedings of the 1998 International Conference on Language
  Resources and Evaluation Workshop on Linguistics Coreference (LREC 1998),
  pages 563--566, 1998.
\newblock Available from:
  \url{https://citeseerx.ist.psu.edu/viewdoc/download?doi=10.1.1.47.5848&rep=rep1&type=pdf}.

\bibitem{luo2005coreference}
X.~Luo.
\newblock {\em On coreference resolution performance metrics}.
\newblock In Proceedings of the 2005 Conference on Empirical Methods in Natural
  Language Processing (EMNLP 2005), pages 25--32, 2005.
\newblock Available from: \url{https://www.aclweb.org/anthology/H05-1004/}.

\bibitem{durrett2013easy}
G.~Durrett and D.~Klein.
\newblock {\em Easy victories and uphill battles in coreference resolution}.
\newblock In Proceedings of the 2013 Conference on Empirical Methods in Natural
  Language Processing, pages 1971--1982, 2013.

\bibitem{wiseman2015learning}
S.~Wiseman, A.~M. Rush, S.~M. Shieber, and J.~Weston.
\newblock {\em Learning Anaphoricity and Antecedent Ranking Features for
  Coreference Resolution}.
\newblock In Proceedings of the 2015 Annual Meeting of the Association for
  Computational Linguistics and International Joint Conference on Natural
  Language Processing, pages 1416--1426, 2015.

\bibitem{kulkarni2018annotated}
C.~Kulkarni, W.~Xu, A.~Ritter, and R.~Machiraju.
\newblock {\em An Annotated Corpus for Machine Reading of Instructions in Wet
  Lab Protocols}.
\newblock In Proceedings of the 2018 Conference of the North American Chapter
  of the Association for Computational Linguistics: Human Language
  Technologies, pages 97--106, 2018.

\bibitem{weischedel2011ontonotes}
R.~Weischedel, E.~Hovy, M.~Marcus, M.~Palmer, R.~Belvin, S.~Pradhan,
  L.~Ramshaw, and N.~Xue.
\newblock {\em OntoNotes: A large training corpus for enhanced processing}.
\newblock Handbook of Natural Language Processing and Machine Translation.
  Springer, page~59, 2011.

\bibitem{bekoulis2017reconstructing}
G.~Bekoulis, J.~Deleu, T.~Demeester, and C.~Develder.
\newblock {\em Reconstructing the house from the ad: Structured prediction on
  real estate classifieds}.
\newblock In Proceedings of the 15th Conference of the European Chapter of the
  Association for Computational Linguistics: Volume 2, Short Papers, pages
  274--279, 2017.

\bibitem{han2020novel}
X.~Han and L.~Wang.
\newblock {\em A Novel Document-Level Relation Extraction Method Based on BERT
  and Entity Information}.
\newblock IEEE Access, 2020.

\bibitem{wu2019enriching}
S.~Wu and Y.~He.
\newblock {\em Enriching pre-trained language model with entity information for
  relation classification}.
\newblock In Proceedings of the 2019 ACM International Conference on
  Information and Knowledge Management, pages 2361--2364, 2019.

\bibitem{cohen1960coefficient}
J.~Cohen.
\newblock {\em A coefficient of agreement for nominal scales}.
\newblock Educational and psychological measurement, 20(1):37--46, 1960.

\bibitem{scott1955reliability}
W.~A. Scott.
\newblock {\em Reliability of content analysis: The case of nominal scale
  coding}.
\newblock Public opinion quarterly, pages 321--325, 1955.

\bibitem{koo2007structured}
T.~Koo, A.~Globerson, X.~Carreras, and M.~Collins.
\newblock {\em Structured prediction models via the matrix-tree theorem}.
\newblock In Proceedings of the 2007 Joint Conference on Empirical Methods in
  Natural Language Processing and Computational Natural Language Learning
  (EMNLP-CoNLL 2007), pages 141--150, 2007.
\newblock Available from: \url{https://www.aclweb.org/anthology/D07-1015/}.

\bibitem{william1984tutte}
W.~Tutte.
\newblock {\em Graph Theory}.
\newblock Encyclopedia of Mathematics and its Applications, 21, 1984.
\newblock Available from: \url{https://doi.org/10.1002/net.3230160110}.

\bibitem{edmonds1967optimum}
J.~Edmonds.
\newblock {\em Optimum branchings}.
\newblock Journal of Research of the National Bureau of Standards B,
  71(4):233--240, 1967.
\newblock Available from:
  \url{https://nvlpubs.nist.gov/nistpubs/jres/71b/jresv71bn4p233_a1b.pdf}.

\bibitem{yamada2016joint}
I.~Yamada, H.~Shindo, H.~Takeda, and Y.~Takefuji.
\newblock {\em Joint Learning of the Embedding of Words and Entities for Named
  Entity Disambiguation}.
\newblock In Proceedings of The 2016 SIGNLL Conference on Computational Natural
  Language Learning (CoNLL 2016), pages 250--259, 2016.
\newblock Available from: \url{https://doi.org/10.18653/v1/k16-1025}.

\bibitem{vilain1995model}
M.~Vilain, J.~Burger, J.~Aberdeen, D.~Connolly, and L.~Hirschman.
\newblock {\em A model-theoretic coreference scoring scheme}.
\newblock In Proceedings of the 1995 Conference on Message understanding (MUC6,
  1995), pages 45--52, 1995.
\newblock Available from: \url{https://doi.org/10.3115/1072399.1072405}.

\bibitem{yang2019learning}
X.~Yang, X.~Gu, S.~Lin, S.~Tang, Y.~Zhuang, F.~Wu, Z.~Chen, G.~Hu, and X.~Ren.
\newblock {\em Learning Dynamic Context Augmentation for Global Entity
  Linking}.
\newblock In Proceedings of the 2019 Conference on Empirical Methods in Natural
  Language Processing and the International Joint Conference on Natural
  Language Processing (EMNLP-IJCNLP 2019), pages 271--281, 2019.
\newblock Available from: \url{https://doi.org/10.18653/v1/D19-1026}.

\bibitem{le2019boosting}
P.~Le and I.~Titov.
\newblock {\em Boosting Entity Linking Performance by Leveraging Unlabeled
  Documents}.
\newblock In Proceedings of the 2019 Annual Meeting of the Association for
  Computational Linguistics (ACL 2019), pages 1935--1945, 2019.
\newblock Available from: \url{https://www.aclweb.org/anthology/P19-1187}.

\bibitem{le2018improving}
P.~Le and I.~Titov.
\newblock {\em Improving Entity Linking by Modeling Latent Relations between
  Mentions}.
\newblock In Proceedings of the 2018 Annual Meeting of the Association for
  Computational Linguistics (ACL 2018), pages 1595--1604, 2018.
\newblock Available from: \url{https://www.aclweb.org/anthology/P18-1148}.

\bibitem{fahrni2012jointly}
A.~Fahrni and M.~Strube.
\newblock {\em Jointly disambiguating and clustering concepts and entities with
  Markov logic}.
\newblock In Proceedings of the 2012 International Conference on Computational
  Linguistics ({COLING} 2012), pages 815--832, 2012.
\newblock Available from: \url{https://aclanthology.org/C12-1050/}.

\bibitem{durrett2014joint}
G.~Durrett and D.~Klein.
\newblock {\em A joint model for entity analysis: Coreference, typing, and
  linking}.
\newblock Transactions of the Association for Computational Linguistics (TACL
  2014), 2:477--490, 2014.
\newblock Available from:
  \url{https://tacl2013.cs.columbia.edu/ojs/index.php/tacl/article/view/412}.

\bibitem{agarwal2021entity}
D.~Agarwal, R.~Angell, N.~Monath, and A.~McCallum.
\newblock {\em Entity Linking and Discovery via Arborescence-based Supervised
  Clustering}.
\newblock arXiv preprint arXiv:2109.01242, 2021.
\newblock Available from: \url{https://arxiv.org/abs/2109.01242}.

\bibitem{liu2020k}
W.~Liu, P.~Zhou, Z.~Zhao, Z.~Wang, Q.~Ju, H.~Deng, and P.~Wang.
\newblock {\em K-bert: Enabling language representation with knowledge graph}.
\newblock In Proceedings of the AAAI Conference on Artificial Intelligence,
  volume~34, pages 2901--2908, 2020.

\bibitem{yang2017leveraging}
B.~Yang and T.~Mitchell.
\newblock {\em Leveraging Knowledge Bases in {LSTM}s for Improving Machine
  Reading}.
\newblock In Proceedings of the 2017 Annual Meeting of the Association for
  Computational Linguistics (ACL 2017), pages 1436--1446, 2017.
\newblock Available from: \url{https://doi.org/10.18653/v1/P17-1132}.

\bibitem{han2018neural}
X.~Han, Z.~Liu, and M.~Sun.
\newblock {\em Neural knowledge acquisition via mutual attention between
  knowledge graph and text}.
\newblock In Proceedings of the AAAI Conference on Artificial Intelligence,
  volume~32, 2018.

\bibitem{zhang2019long}
N.~Zhang, S.~Deng, Z.~Sun, G.~Wang, X.~Chen, W.~Zhang, and H.~Chen.
\newblock {\em Long-tail relation extraction via knowledge graph embeddings and
  graph convolution networks}.
\newblock arXiv preprint arXiv:1903.01306, 2019.

\bibitem{yamada2020luke}
I.~Yamada, A.~Asai, H.~Shindo, H.~Takeda, and Y.~Matsumoto.
\newblock {\em {LUKE}: Deep Contextualized Entity Representations with
  Entity-aware Self-attention}.
\newblock In Proceedings of the 2020 Conference on Empirical Methods in Natural
  Language Processing (EMNLP 2020), pages 6442--6454, 2020.
\newblock Available from: \url{https://aclanthology.org/2020.emnlp-main.523}.

\bibitem{pennington2014}
J.~Pennington, R.~Socher, and C.~Manning.
\newblock {\em {GloVe}: Global vectors for word representation}.
\newblock In Proceedings of the 2014 Conference on Empirical Methods in Natural
  Language Processing (EMNLP 2014), pages 1532--1543, 2014.
\newblock Available from: \url{https://www.aclweb.org/anthology/D14-1162/}.

\bibitem{ma2016}
X.~Ma and E.~Hovy.
\newblock {\em End-to-end Sequence Labeling via Bi-directional
  {LSTM}-{CNN}s-{CRF}}.
\newblock In Proceedings of the 2016 Annual Meeting of the Association for
  Computational Linguistics (ACL 2016), pages 1064--1074, 2016.
\newblock Available from: \url{https://www.aclweb.org/anthology/P16-1101/}.

\bibitem{mikolov2013efficient}
T.~Mikolov, K.~Chen, G.~Corrado, and J.~Dean.
\newblock {\em Efficient estimation of word representations in vector space}.
\newblock In Proceedings of the 2013 International Conference on Learning
  Representations (ICLR 2013), pages 1--12, 2013.
\newblock Available from: \url{http://arxiv.org/abs/1301.3781}.

\bibitem{mikolov2013distributed}
T.~Mikolov, I.~Sutskever, K.~Chen, G.~S. Corrado, and J.~Dean.
\newblock {\em Distributed representations of words and phrases and their
  compositionality}.
\newblock In Advances in neural information processing systems, pages
  3111--3119, 2013.

\bibitem{joulin2017fast}
A.~Joulin, E.~Grave, P.~Bojanowski, M.~Nickel, and T.~Mikolov.
\newblock {\em Fast linear model for knowledge graph embeddings}.
\newblock arXiv:1710.10881, 2017.
\newblock Available from: \url{http://arxiv.org/abs/1710.10881}.

\bibitem{poerner2020bert}
N.~Poerner, U.~Waltinger, and H.~Sch{\"u}tze.
\newblock {\em E-bert: Efficient-yet-effective entity embeddings for bert}.
\newblock In Proceedings of the 2020 Conference on Empirical Methods in Natural
  Language Processing: Findings, pages 803--818, 2020.

\bibitem{jia2019document}
R.~Jia, C.~Wong, and H.~Poon.
\newblock {\em Document-Level N-ary Relation Extraction with Multiscale
  Representation Learning}.
\newblock In Proceedings of the 2019 Conference of the North American Chapter
  of the Association for Computational Linguistics: Human Language
  Technologies, pages 3693--3704, 2019.

\bibitem{yamada2019neural}
I.~Yamada and H.~Shindo.
\newblock {\em Neural Attentive Bag-of-Entities Model for Text Classification}.
\newblock In Proceedings of the 2019 Conference on Computational Natural
  Language Learning (CoNLL 2019), pages 563--573, 2019.
\newblock Available from: \url{https://www.aclweb.org/anthology/K19-1052}.

\bibitem{johnson2019billion}
J.~Johnson, M.~Douze, and H.~J{\'e}gou.
\newblock {\em Billion-scale similarity search with gpus}.
\newblock IEEE Transactions on Big Data, pages 535--547, 2021.
\newblock Available from: \url{https://doi.org/10.1109/TBDATA.2019.2921572}.

\bibitem{provatorova2021robustness}
V.~Provatorova, S.~Bhargav, S.~Vakulenko, and E.~Kanoulas.
\newblock {\em Robustness Evaluation of Entity Disambiguation Using Prior
  Probes: the Case of Entity Overshadowing}.
\newblock In Proceedings of the 2021 Conference on Empirical Methods in Natural
  Language Processing (EMNLP 2021), pages 10501--10510, 2021.
\newblock Available from: \url{https://aclanthology.org/2021.emnlp-main.820}.

\bibitem{lazaridou2021mind}
A.~Lazaridou, A.~Kuncoro, E.~Gribovskaya, D.~Agrawal, A.~Liska, T.~Terzi,
  M.~Gimenez, C.~de~Masson~d'Autume, T.~Kocisky, S.~Ruder, et~al.
\newblock {\em Mind the Gap: Assessing Temporal Generalization in Neural
  Language Models}.
\newblock In Proceedings of the 2021 Advances in Neural Information Processing
  Systems (NeurIPS 2021), pages 29348--29363, 2021.
\newblock Available from:
  \url{https://proceedings.neurips.cc/paper/2021/hash/f5bf0ba0a17ef18f9607774722f5698c-Abstract.html}.

\bibitem{dhingra2022time}
B.~Dhingra, J.~R. Cole, J.~M. Eisenschlos, D.~Gillick, J.~Eisenstein, and W.~W.
  Cohen.
\newblock {\em Time-Aware Language Models as Temporal Knowledge Bases}.
\newblock Transactions of the Association for Computational Linguistics,
  10:257--273, 2022.
\newblock Available from: \url{https://arxiv.org/abs/2106.15110}.

\bibitem{jang2022temporalwiki}
J.~Jang, S.~Ye, C.~Lee, S.~Yang, J.~Shin, J.~Han, G.~Kim, and M.~Seo.
\newblock {\em {TemporalWiki}: {A} Lifelong Benchmark for Training and
  Evaluating Ever-Evolving Language Models}.
\newblock CoRR, 2022.
\newblock Available from: \url{https://doi.org/10.48550/arXiv.2204.14211}.

\bibitem{loureiro2022timelms}
D.~Loureiro, F.~Barbieri, L.~Neves, L.~E. Anke, and J.~Camacho-Collados.
\newblock {\em {TimeLMs}: Diachronic Language Models from Twitter}.
\newblock In Proceedings of the 2022 Annual Meeting of the Association for
  Computational Linguistics (ACL 2022), pages 251--260, 2022.
\newblock Available from: \url{https://aclanthology.org/2022.acl-demo.25}.

\bibitem{lukes2018sentiment}
J.~Lukes and A.~S{\o}gaard.
\newblock {\em Sentiment analysis under temporal shift}.
\newblock In Proceedings of the 2018 Workshop on Computational Approaches to
  Subjectivity, Sentiment and Social Media Analysis (WASSA@EMNLP 2018), pages
  65--71, 2018.
\newblock Available from: \url{https://doi.org/10.18653/v1/w18-6210}.

\bibitem{ni2019justifying}
J.~Ni, J.~Li, and J.~McAuley.
\newblock {\em Justifying recommendations using distantly-labeled reviews and
  fine-grained aspects}.
\newblock In Proceedings of the 2019 Conference on Empirical Methods in Natural
  Language Processing and the International Joint Conference on Natural
  Language Processing (EMNLP-IJCNLP 2019), pages 188--197, 2019.
\newblock Available from: \url{https://aclanthology.org/D19-1018}.

\bibitem{agarwal2022temporal}
O.~Agarwal and A.~Nenkova.
\newblock {\em Temporal effects on pre-trained models for language processing
  tasks}.
\newblock Transactions of the Association for Computational Linguistics (TACL
  2022), 10:904--921, 2022.

\bibitem{huang2018examining}
X.~Huang and M.~J. Paul.
\newblock {\em Examining temporality in document classification}.
\newblock In Proceedings of the 2018 Annual Meeting of the Association for
  Computational Linguistics (ACL 2018), pages 694--699, 2018.
\newblock Available from: \url{https://aclanthology.org/P18-2110}.

\bibitem{he2018time}
Y.~He, J.~Li, Y.~Song, M.~He, H.~Peng, et~al.
\newblock {\em Time-evolving Text Classification with Deep Neural Networks}.
\newblock In Proceedings of the 2018 International Joint Conference on
  Artificial Intelligence (IJCAI 2018), pages 2241--2247, 2018.
\newblock Available from: \url{https://doi.org/10.24963/ijcai.2018/310}.

\bibitem{derczynski2016broad}
L.~Derczynski, K.~Bontcheva, and I.~Roberts.
\newblock {\em Broad twitter corpus: A diverse named entity recognition
  resource}.
\newblock In Proceedings of the 2016 International Conference on Computational
  Linguistics (COLING 2016), pages 1169--1179, 2016.
\newblock Available from: \url{https://aclanthology.org/C16-1111}.

\bibitem{rijhwani2020temporally}
S.~Rijhwani and D.~Preo{\c{t}}iuc-Pietro.
\newblock {\em Temporally-informed analysis of named entity recognition}.
\newblock In Proceedings of the 2020 Annual Meeting of the Association for
  Computational Linguistics (ACL 2020), pages 7605--7617, 2020.
\newblock Available from: \url{https://aclanthology.org/2020.acl-main.680}.

\bibitem{luu2021time}
K.~Luu, D.~Khashabi, S.~Gururangan, K.~Mandyam, and N.~A. Smith.
\newblock {\em Time Waits for No One! Analysis and Challenges of Temporal
  Misalignment}.
\newblock CoRR, 2021.
\newblock Available from: \url{https://arxiv.org/abs/2111.07408}.

\bibitem{agarwal2018dianed}
P.~Agarwal, J.~Str{\"o}tgen, L.~Del~Corro, J.~Hoffart, and G.~Weikum.
\newblock {\em {diaNED}: Time-aware named entity disambiguation for diachronic
  corpora}.
\newblock In Proceedings of the 2018 Annual Meeting of the Association for
  Computational Linguistics (ACL 2018), pages 686--693, 2018.
\newblock Available from: \url{https://aclanthology.org/P18-2109}.

\bibitem{guu2020realm}
K.~Guu, K.~Lee, Z.~Tung, P.~Pasupat, and M.-W. Chang.
\newblock {\em {REALM}: Retrieval-augmented language model pre-training}.
\newblock CoRR, 2020.
\newblock Available from: \url{https://arxiv.org/abs/2002.08909}.

\bibitem{verga2021adaptable}
P.~Verga, H.~Sun, L.~B. Soares, and W.~Cohen.
\newblock {\em Adaptable and interpretable neural memoryover symbolic
  knowledge}.
\newblock In Proceedings of the 2021 Conference of the North American Chapter
  of the Association for Computational Linguistics: Human Language Technologies
  (NAACL-HLT 2021), pages 3678--3691, 2021.
\newblock Available from: \url{https://aclanthology.org/2021.naacl-main.288}.

\bibitem{liu2022knowledge}
R.~Liu, G.~Zheng, S.~Gupta, R.~Gaonkar, C.~Gao, S.~Vosoughi, M.~Shokouhi, and
  A.~H. Awadallah.
\newblock {\em Knowledge infused decoding}.
\newblock In Proceedings of the 2022 International Conference on Learning
  Representations (ICLR 2022), 2022.
\newblock Available from: \url{https://openreview.net/forum?id=upnDJ7itech}.

\bibitem{petroni2019language}
F.~Petroni, T.~Rockt{\"a}schel, S.~Riedel, P.~Lewis, A.~Bakhtin, Y.~Wu, and
  A.~Miller.
\newblock {\em Language Models as Knowledge Bases?}
\newblock In Proceedings of the 2019 Conference on Empirical Methods in Natural
  Language Processing and the International Joint Conference on Natural
  Language Processing (EMNLP-IJCNLP 2019), pages 2463--2473, 2019.
\newblock Available from: \url{https://aclanthology.org/D19-1250}.

\bibitem{agarwal2021knowledge}
O.~Agarwal, H.~Ge, S.~Shakeri, and R.~Al-Rfou.
\newblock {\em Knowledge Graph Based Synthetic Corpus Generation for
  Knowledge-Enhanced Language Model Pre-training}.
\newblock In Proceedings of the 2021 Conference of the North American Chapter
  of the Association for Computational Linguistics: Human Language Technologies
  (NAACL-HLT 2021), pages 3554--3565, 2021.
\newblock Available from: \url{https://aclanthology.org/2021.naacl-main.278}.

\bibitem{kassner2021multilingual}
N.~Kassner, P.~Dufter, and H.~Sch{\"u}tze.
\newblock {\em Multilingual {LAMA}: Investigating Knowledge in Multilingual
  Pretrained Language Models}.
\newblock In Proceedings of the 2021 Conference of the European Chapter of the
  Association for Computational Linguistics (EACL 2021), pages 3250--3258,
  2021.
\newblock Available from: \url{https://aclanthology.org/2021.eacl-main.284}.

\bibitem{yih2015semantic}
S.~W.-t. Yih, M.-W. Chang, X.~He, and J.~Gao.
\newblock {\em Semantic parsing via staged query graph generation: Question
  answering with knowledge base}.
\newblock In Proceedings of the 2015 Conference of the Annual Meeting of the
  Association for Computational Linguistics and the International Joint
  Conference on Natural Language Processing (ACL-IJCNLP 2015), 2015.
\newblock Available from: \url{https://doi.org/10.3115/v1/p15-1128}.

\bibitem{joshi2017triviaqa}
M.~Joshi, E.~Choi, D.~S. Weld, and L.~Zettlemoyer.
\newblock {\em {TriviaQA}: A Large Scale Distantly Supervised Challenge Dataset
  for Reading Comprehension}.
\newblock In Proceedings of the 2017 Annual Meeting of the Association for
  Computational Linguistics (ACL 2017), pages 1601--1611, 2017.
\newblock Available from: \url{https://aclanthology.org/P17-1147}.

\bibitem{jiang2019freebaseqa}
K.~Jiang, D.~Wu, and H.~Jiang.
\newblock {\em {FreebaseQA}: A New Factoid QA Data Set Matching Trivia-Style
  Question-Answer Pairs with Freebase.}
\newblock In Proceedings of the 2019 Conference of the North American Chapter
  of the Association for Computational Linguistics: Human Language Technologies
  ({NAACL-HLT} 2019), pages 318--323, 2019.
\newblock Available from: \url{https://doi.org/10.18653/v1/n19-1028}.

\bibitem{lewis2021paq}
P.~Lewis, Y.~Wu, L.~Liu, P.~Minervini, H.~K{\"u}ttler, A.~Piktus, P.~Stenetorp,
  and S.~Riedel.
\newblock {\em Paq: 65 million probably-asked questions and what you can do
  with them}.
\newblock Transactions of the Association for Computational Linguistics,
  9:1098--1115, 2021.
\newblock Available from: \url{https://arxiv.org/abs/2102.07033}.

\bibitem{saxena2021question}
A.~Saxena, S.~Chakrabarti, and P.~Talukdar.
\newblock {\em Question Answering Over Temporal Knowledge Graphs}.
\newblock In Proceedings of the 2021 Annual Meeting of the Association for
  Computational Linguistics and the International Joint Conference on Natural
  Language Processing (ACL-IJCNLP 2021), pages 6663--6676, 2021.
\newblock Available from: \url{https://aclanthology.org/2021.acl-long.520}.

\bibitem{onoe2021creak}
Y.~Onoe, M.~J. Zhang, E.~Choi, and G.~Durrett.
\newblock {\em {CREAK}: A Dataset for Commonsense Reasoning over Entity
  Knowledge}.
\newblock In Proceedings of the 2021 Conference on Neural Information
  Processing Systems Datasets and Benchmarks Track (NeurIPS 2021), 2021.
\newblock Available from:
  \url{https://datasets-benchmarks-proceedings.neurips.cc/paper/2021/hash/5737c6ec2e0716f3d8a7a5c4e0de0d9a-Abstract-round2.html}.

\bibitem{runge2020exploring}
A.~Runge and E.~Hovy.
\newblock {\em Exploring Neural Entity Representations for Semantic
  Information}.
\newblock In Proceedings of the 2020 BlackboxNLP Workshop on Analyzing and
  Interpreting Neural Networks for NLP (BlackboxNLP@EMNLP 2020), pages
  204--216, 2020.
\newblock Available from: \url{https://aclanthology.org/2020.blackboxnlp-1.20}.

\bibitem{fevry2020entities}
T.~F{\'e}vry, L.~B. Soares, N.~FitzGerald, E.~Choi, and T.~Kwiatkowski.
\newblock {\em Entities as Experts: Sparse Memory Access with Entity
  Supervision}.
\newblock In Proceedings of the 2020 Conference on Empirical Methods in Natural
  Language Processing (EMNLP 2020), pages 4937--4951, 2020.
\newblock Available from: \url{https://aclanthology.org/2020.emnlp-main.400}.

\bibitem{lewis2020retrieval}
P.~Lewis, E.~Perez, A.~Piktus, F.~Petroni, V.~Karpukhin, N.~Goyal,
  H.~K{\"u}ttler, M.~Lewis, W.-t. Yih, T.~Rockt{\"a}schel, et~al.
\newblock {\em Retrieval-augmented generation for knowledge-intensive nlp
  tasks}.
\newblock In Proceedings of the 2020 Advances in Neural Information Processing
  Systems (NeurIPS 2020), pages 9459--9474, 2020.
\newblock Available from:
  \url{https://proceedings.neurips.cc/paper/2020/hash/6b493230205f780e1bc26945df7481e5-Abstract.html}.

\bibitem{heinzerling2021language}
B.~Heinzerling and K.~Inui.
\newblock {\em Language Models as Knowledge Bases: On Entity Representations,
  Storage Capacity, and Paraphrased Queries}.
\newblock In Proceedings of the 2021 Conference of the European Chapter of the
  Association for Computational Linguistics (EACL 2021), pages 1772--1791,
  2021.
\newblock Available from: \url{https://aclanthology.org/2021.eacl-main.153}.

\bibitem{ri2021mluke}
R.~Ri, I.~Yamada, and Y.~Tsuruoka.
\newblock {\em {mLUKE}: The Power of Entity Representations in Multilingual
  Pretrained Language Models}.
\newblock In Proceedings of the 2022 Annual Meeting of the Association for
  Computational Linguistics (ACL 2022), pages 7316--7330, 2022.
\newblock Available from: \url{https://aclanthology.org/2022.acl-long.505}.

\bibitem{west2010detecting}
A.~G. West, S.~Kannan, and I.~Lee.
\newblock {\em Detecting {W}ikipedia vandalism via spatio-temporal analysis of
  revision metadata?}
\newblock In Proceedings of the Third European Workshop on System Security,
  pages 22--28, 2010.
\newblock Available from: \url{https://doi.org/10.1145/1752046.1752050}.

\bibitem{dang2016quality}
Q.~V. Dang and C.-L. Ignat.
\newblock {\em Quality assessment of wikipedia articles without feature
  engineering}.
\newblock In Proceedings of the 16th ACM/IEEE-CS on Joint Conference on Digital
  Libraries, pages 27--30, 2016.
\newblock Available from: \url{https://doi.org/10.1145/2910896.2910917}.

\bibitem{wang2020assessing}
P.~Wang and X.~Li.
\newblock {\em Assessing the quality of information on {W}ikipedia: A
  deep-learning approach}.
\newblock Journal of the Association for Information Science and Technology,
  71(1):16--28, 2020.
\newblock Available from: \url{https://doi.org/10.1002/asi.24210}.

\bibitem{zheng2019roles}
L.~Zheng, C.~M. Albano, N.~M. Vora, F.~Mai, and J.~V. Nickerson.
\newblock {\em The roles bots play in {W}ikipedia}.
\newblock In Proceedings of the 2019 ACM on Human-Computer Interaction (ACM
  SIGCHI 2019), pages 1--20, 2019.
\newblock Available from: \url{https://doi.org/10.1145/3359317}.

\bibitem{jiang2020good}
J.~Jiang and M.~A. Vetter.
\newblock {\em The good, the bot, and the ugly: Problematic information and
  critical media literacy in the postdigital era}.
\newblock Postdigital Science and Education, 2(1):78--94, 2020.
\newblock Available from: \url{https://doi.org/10.1007/s42438-019-00069-4}.

\bibitem{chen2021evaluating}
A.~Chen, P.~Gudipati, S.~Longpre, X.~Ling, and S.~Singh.
\newblock {\em Evaluating Entity Disambiguation and the Role of Popularity in
  Retrieval-Based {NLP}}.
\newblock In Proceedings of the 2021 Annual Meeting of the Association for
  Computational Linguistics and the International Joint Conference on Natural
  Language Processing (ACL-IJCNLP 2021), pages 4472--4485, 2021.
\newblock Available from: \url{https://aclanthology.org/2021.acl-long.345}.

\bibitem{mazare2018training}
P.-E. Mazare, S.~Humeau, M.~Raison, and A.~Bordes.
\newblock {\em Training Millions of Personalized Dialogue Agents}.
\newblock In Proceedings of the 2018 Conference on Empirical Methods in Natural
  Language Processing (EMNLP 2018), pages 2775--2779, 2018.
\newblock Available from: \url{https://aclanthology.org/D18-1298}.

\bibitem{dinan2018wizard}
E.~Dinan, S.~Roller, K.~Shuster, A.~Fan, M.~Auli, and J.~Weston.
\newblock {\em Wizard of {W}ikipedia: Knowledge-powered conversational agents}.
\newblock In Proceedings of the 2018 International Conference on Learning
  Representations (ICLR 2018), 2018.
\newblock Available from: \url{https://openreview.net/forum?id=r1l73iRqKm}.

\bibitem{botha2020entity}
J.~A. Botha, Z.~Shan, and D.~Gillick.
\newblock {\em {E}ntity Linking in 100 Languages}.
\newblock In Proceedings of the 2020 Conference on Empirical Methods in Natural
  Language Processing (EMNLP 2020), pages 7833--7845, 2020.
\newblock Available from:
  \url{https://www.aclweb.org/anthology/2020.emnlp-main.630}.

\bibitem{de2021multilingual}
N.~De~Cao, L.~Wu, K.~Popat, M.~Artetxe, N.~Goyal, M.~Plekhanov, L.~Zettlemoyer,
  N.~Cancedda, S.~Riedel, and F.~Petroni.
\newblock {\em Multilingual Autoregressive Entity Linking}.
\newblock Transactions of the Association for Computational Linguistics,
  10:274--290, 2022.
\newblock Available from: \url{https://arxiv.org/abs/2103.12528}.

\bibitem{hennig2021mobie}
L.~Hennig, P.~T. Truong, and A.~Gabryszak.
\newblock {\em MobIE: A German Dataset for Named Entity Recognition, Entity
  Linking and Relation Extraction in the Mobility Domain}.
\newblock In Proceedings of the 2021 Conference on Natural Language Processing
  (KONVENS 2021), pages 223--227, 2021.
\newblock Available from: \url{https://aclanthology.org/2021.konvens-1.22.pdf}.

\bibitem{ogrodniczuk2020wikipedia}
M.~Ogrodniczuk and W.~Gruszczy{\'n}ski.
\newblock {\em Wikipedia-Based Entity Linking for the Digital Library of Polish
  and Poland-Related News Pamphlets}.
\newblock In Proceedings of the 2020 International Conference on Asian Digital
  Libraries (ICADL 2020), pages 81--88, 2020.
\newblock Available from: \url{https://doi.org/10.1007/978-3-030-64452-9\_7}.

\bibitem{caillaut2022automated}
G.~Caillaut, C.~Gracianne, N.~Abadie, G.~Touya, and S.~Auclair.
\newblock {\em Automated construction of a French Entity Linking dataset to
  geolocate social network posts in the context of natural disasters}.
\newblock In Proceedings of the 2022 International Conference on Information
  Systems for Crisis Response and Management (ISCRAM 2022), 2022.
\newblock Available from:
  \url{https://hal.archives-ouvertes.fr/hal-03631387/document}.

\bibitem{rosales2021towards}
H.~Rosales~M{\'e}ndez.
\newblock {\em Towards a fine-grained entity linking approach}.
\newblock PhD thesis, Universidad de Chile, 2021.
\newblock Available from:
  \url{https://repositorio.uchile.cl/bitstream/handle/2250/181834/Towards-a-fine-grained-entity-linking-approach.pdf?sequence=1}.

\bibitem{gebru2021datasheets}
T.~Gebru, J.~Morgenstern, B.~Vecchione, J.~W. Vaughan, H.~Wallach, H.~D. Iii,
  and K.~Crawford.
\newblock {\em Datasheets for datasets}.
\newblock Communications of the ACM, 64(12):86--92, 2021.
\newblock Available from: \url{https://doi.org/10.1145/3458723}.

\bibitem{wolf2020transformers}
T.~Wolf, L.~Debut, V.~Sanh, J.~Chaumond, C.~Delangue, A.~Moi, P.~Cistac,
  T.~Rault, R.~Louf, M.~Funtowicz, et~al.
\newblock {\em Transformers: State-of-the-art natural language processing}.
\newblock In Proceedings of the 2020 Conference on Empirical Methods in Natural
  Language Processing: System Demonstrations (EMNLP 2020), pages 38--45, 2020.
\newblock Available from: \url{https://aclanthology.org/2020.emnlp-demos.6}.

\bibitem{chen2018gradnorm}
Z.~Chen, V.~Badrinarayanan, C.-Y. Lee, and A.~Rabinovich.
\newblock {\em Gradnorm: Gradient normalization for adaptive loss balancing in
  deep multitask networks}.
\newblock In International conference on machine learning, pages 794--803.
  PMLR, 2018.

\bibitem{vandenhende2021multi}
S.~Vandenhende, S.~Georgoulis, W.~Van~Gansbeke, M.~Proesmans, D.~Dai, and
  L.~Van~Gool.
\newblock {\em Multi-task learning for dense prediction tasks: A survey}.
\newblock IEEE transactions on pattern analysis and machine intelligence, 2021.

\bibitem{zhang2021survey}
Y.~Zhang and Q.~Yang.
\newblock {\em A survey on multi-task learning}.
\newblock IEEE Transactions on Knowledge and Data Engineering, 2021.

\bibitem{yu2020gradient}
T.~Yu, S.~Kumar, A.~Gupta, S.~Levine, K.~Hausman, and C.~Finn.
\newblock {\em Gradient surgery for multi-task learning}.
\newblock Advances in Neural Information Processing Systems, 33:5824--5836,
  2020.

\bibitem{chen2020just}
Z.~Chen, J.~Ngiam, Y.~Huang, T.~Luong, H.~Kretzschmar, Y.~Chai, and
  D.~Anguelov.
\newblock {\em Just pick a sign: Optimizing deep multitask models with gradient
  sign dropout}.
\newblock Advances in Neural Information Processing Systems, 33:2039--2050,
  2020.

\bibitem{liu2021conflict}
B.~Liu, X.~Liu, X.~Jin, P.~Stone, and Q.~Liu.
\newblock {\em Conflict-averse gradient descent for multi-task learning}.
\newblock Advances in Neural Information Processing Systems, 34:18878--18890,
  2021.

\bibitem{agarwal2021temporal}
O.~Agarwal and A.~Nenkova.
\newblock {\em Temporal Effects on Pre-trained Models for Language Processing
  Tasks}.
\newblock CoRR, 2021.
\newblock Available from: \url{https://arxiv.org/abs/2111.12790}.

\bibitem{bodenreider2004unified}
O.~Bodenreider.
\newblock {\em The unified medical language system (UMLS): integrating
  biomedical terminology}.
\newblock Nucleic acids research, 32(suppl\_1):D267--D270, 2004.

\bibitem{rosales2020fine}
H.~Rosales-M{\'e}ndez, A.~Hogan, and B.~Poblete.
\newblock {\em Fine-Grained Entity Linking}.
\newblock Journal of Web Semantics, 65:100600, 2020.

\bibitem{bagga1998entity}
A.~Bagga and B.~Baldwin.
\newblock {\em Entity-Based Cross-Document Core f erencing Using the Vector
  Space Model}.
\newblock In 36th Annual Meeting of the Association for Computational
  Linguistics and 17th International Conference on Computational Linguistics,
  Volume 1, pages 79--85, 1998.

\bibitem{logan2021benchmarking}
R.~L. Logan~IV, A.~McCallum, S.~Singh, and D.~Bikel.
\newblock {\em Benchmarking scalable methods for streaming cross document
  entity coreference}.
\newblock In Proceedings of the 59th Annual Meeting of the Association for
  Computational Linguistics and the 11th International Joint Conference on
  Natural Language Processing (Volume 1: Long Papers), pages 4717--4731, 2021.

\bibitem{hsu2022contrastive}
B.~Hsu and G.~Horwood.
\newblock {\em Contrastive Representation Learning for Cross-Document
  Coreference Resolution of Events and Entities}.
\newblock arXiv preprint arXiv:2205.11438, 2022.

\bibitem{cattan2021realistic}
A.~Cattan, A.~Eirew, G.~Stanovsky, M.~Joshi, and I.~Dagan.
\newblock {\em Realistic Evaluation Principles for Cross-document Coreference
  Resolution}.
\newblock In Proceedings of* SEM 2021: The Tenth Joint Conference on Lexical
  and Computational Semantics, pages 143--151, 2021.

\bibitem{cattan2021cross}
A.~Cattan, A.~Eirew, G.~Stanovsky, M.~Joshi, and I.~Dagan.
\newblock {\em Cross-document Coreference Resolution over Predicted Mentions}.
\newblock In Findings of the Association for Computational Linguistics:
  ACL-IJCNLP 2021, pages 5100--5107, 2021.

\bibitem{cattan2021scico}
A.~Cattan, S.~Johnson, D.~Weld, I.~Dagan, I.~Beltagy, D.~Downey, and T.~Hope.
\newblock {\em Scico: Hierarchical cross-document coreference for scientific
  concepts}.
\newblock arXiv preprint arXiv:2104.08809, 2021.

\bibitem{cattan2020streamlining}
A.~Cattan, A.~Eirew, G.~Stanovsky, M.~Joshi, and I.~Dagan.
\newblock {\em Streamlining Cross-Document Coreference Resolution: Evaluation
  and Modeling}.
\newblock arXiv preprint arXiv:2009.11032, 2020.

\bibitem{gooi2004cross}
C.~H. Gooi and J.~Allan.
\newblock {\em Cross-document coreference on a large scale corpus}.
\newblock Technical report, MASSACHUSETTS UNIV AMHERST CENTER FOR INTELLIGENT
  INFORMATION RETRIEVAL, 2004.

\bibitem{singh2011large}
S.~Singh, A.~Subramanya, F.~Pereira, and A.~McCallum.
\newblock {\em Large-Scale Cross-Document Coreference Using Distributed
  Inference and Hierarchical Models}.
\newblock In Proceedings of the 49th Annual Meeting of the Association for
  Computational Linguistics: Human Language Technologies, pages 793--803, 2011.

\bibitem{barhom2019revisiting}
S.~Barhom, V.~Shwartz, A.~Eirew, M.~Bugert, N.~Reimers, and I.~Dagan.
\newblock {\em Revisiting Joint Modeling of Cross-document Entity and Event
  Coreference Resolution}.
\newblock In Proceedings of the 57th Annual Meeting of the Association for
  Computational Linguistics, pages 4179--4189, 2019.

\bibitem{caciularu2021cross}
A.~Caciularu, A.~Cohan, I.~Beltagy, M.~E. Peters, A.~Cattan, and I.~Dagan.
\newblock {\em Cross-document language modeling}.
\newblock arXiv preprint arXiv:2101.00406, 2021.

\bibitem{ravenscroft2021cd2cr}
J.~Ravenscroft, A.~Cattan, A.~Clare, I.~Dagan, and M.~Liakata.
\newblock {\em CD2CR: Co-reference Resolution Across Documents and Domains}.
\newblock arXiv preprint arXiv:2101.12637, 2021.

\bibitem{yao2021codred}
Y.~Yao, J.~Du, Y.~Lin, P.~Li, Z.~Liu, J.~Zhou, and M.~Sun.
\newblock {\em CodRED: A Cross-Document Relation Extraction Dataset for
  Acquiring Knowledge in the Wild}.
\newblock In Proceedings of the 2021 Conference on Empirical Methods in Natural
  Language Processing, pages 4452--4472, 2021.

\bibitem{beltagy2020longformer}
I.~Beltagy, M.~E. Peters, and A.~Cohan.
\newblock {\em Longformer: The long-document transformer}.
\newblock arXiv preprint arXiv:2004.05150, 2020.

\bibitem{zaheer2020big}
M.~Zaheer, G.~Guruganesh, K.~A. Dubey, J.~Ainslie, C.~Alberti, S.~Ontanon,
  P.~Pham, A.~Ravula, Q.~Wang, L.~Yang, et~al.
\newblock {\em Big Bird: Transformers for Longer Sequences.}
\newblock In NeurIPS, 2020.

\bibitem{treviso2022efficient}
M.~Treviso, T.~Ji, J.-U. Lee, B.~van Aken, Q.~Cao, M.~R. Ciosici, M.~Hassid,
  K.~Heafield, S.~Hooker, P.~H. Martins, et~al.
\newblock {\em Efficient Methods for Natural Language Processing: A Survey}.
\newblock arXiv preprint arXiv:2209.00099, 2022.

\bibitem{caciularu2021cdlm}
A.~Caciularu, A.~Cohan, I.~Beltagy, M.~E. Peters, A.~Cattan, and I.~Dagan.
\newblock {\em CDLM: Cross-Document Language Modeling}.
\newblock In Findings of the Association for Computational Linguistics: EMNLP
  2021, pages 2648--2662, 2021.

\bibitem{lewis2020bart}
M.~Lewis, Y.~Liu, N.~Goyal, M.~Ghazvininejad, A.~Mohamed, O.~Levy, V.~Stoyanov,
  and L.~Zettlemoyer.
\newblock {\em BART: Denoising Sequence-to-Sequence Pre-training for Natural
  Language Generation, Translation, and Comprehension}.
\newblock In Proceedings of the 58th Annual Meeting of the Association for
  Computational Linguistics, pages 7871--7880, 2020.

\bibitem{brown2020language}
T.~B. Brown, B.~Mann, N.~Ryder, M.~Subbiah, J.~Kaplan, P.~Dhariwal,
  A.~Neelakantan, P.~Shyam, G.~Sastry, A.~Askell, et~al.
\newblock {\em Language models are few-shot learners}.
\newblock arXiv preprint arXiv:2005.14165, 2020.

\bibitem{eberts2020span}
M.~Eberts and A.~Ulges.
\newblock {\em Span-Based Joint Entity and Relation Extraction with Transformer
  Pre-Training}.
\newblock In ECAI 2020, pages 2006--2013. IOS Press, 2020.

\bibitem{zhong2021frustratingly}
Z.~Zhong and D.~Chen.
\newblock {\em A Frustratingly Easy Approach for Entity and Relation
  Extraction}.
\newblock In Proceedings of the 2021 Conference of the North American Chapter
  of the Association for Computational Linguistics: Human Language
  Technologies, pages 50--61, 2021.

\bibitem{lin2020joint}
Y.~Lin, H.~Ji, F.~Huang, and L.~Wu.
\newblock {\em A Joint Neural Model for Information Extraction with Global
  Features}.
\newblock In Proceedings of the 58th Annual Meeting of the Association for
  Computational Linguistics, pages 7999--8009, 2020.

\bibitem{wang2020two}
J.~Wang and W.~Lu.
\newblock {\em Two are Better than One: Joint Entity and Relation Extraction
  with Table-Sequence Encoders}.
\newblock In Proceedings of the 2020 Conference on Empirical Methods in Natural
  Language Processing (EMNLP), pages 1706--1721, 2020.

\bibitem{li2020efficient}
B.~Z. Li, S.~Min, S.~Iyer, Y.~Mehdad, and W.-t. Yih.
\newblock {\em Efficient One-Pass End-to-End Entity Linking for Questions}.
\newblock In Proceedings of the 2020 Conference on Empirical Methods in Natural
  Language Processing (EMNLP), pages 6433--6441, 2020.

\bibitem{petroni2020context}
F.~Petroni, P.~Lewis, A.~Piktus, T.~Rockt{\"a}schel, Y.~Wu, A.~H. Miller, and
  S.~Riedel.
\newblock {\em How Context Affects Language Models' Factual Predictions}.
\newblock In Automated Knowledge Base Construction, 2020.

\bibitem{rongali2020don}
S.~Rongali, L.~Soldaini, E.~Monti, and W.~Hamza.
\newblock {\em Don’t parse, generate! a sequence to sequence architecture for
  task-oriented semantic parsing}.
\newblock In Proceedings of The Web Conference 2020, pages 2962--2968, 2020.

\bibitem{nogueira2020document}
R.~Nogueira, Z.~Jiang, R.~Pradeep, and J.~Lin.
\newblock {\em Document Ranking with a Pretrained Sequence-to-Sequence Model}.
\newblock In Proceedings of the 2020 Conference on Empirical Methods in Natural
  Language Processing: Findings, pages 708--718, 2020.

\bibitem{mrini2022detection}
K.~Mrini, S.~Nie, J.~Gu, S.~Wang, M.~Sanjabi, and H.~Firooz.
\newblock {\em Detection, Disambiguation, Re-ranking: Autoregressive Entity
  Linking as a Multi-Task Problem}.
\newblock arXiv preprint arXiv:2204.05990, 2022.

\bibitem{yuan2022generative}
H.~Yuan, Z.~Yuan, and S.~Yu.
\newblock {\em Generative Biomedical Entity Linking via Knowledge Base-Guided
  Pre-training and Synonyms-Aware Fine-tuning}.
\newblock arXiv preprint arXiv:2204.05164, 2022.

\bibitem{cabot2021rebel}
P.-L.~H. Cabot and R.~Navigli.
\newblock {\em REBEL: Relation extraction by end-to-end language generation}.
\newblock In Findings of the Association for Computational Linguistics: EMNLP
  2021, pages 2370--2381, 2021.

\bibitem{josifoski2021genie}
M.~Josifoski, N.~De~Cao, M.~Peyrard, and R.~West.
\newblock {\em GenIE: generative information extraction}.
\newblock arXiv preprint arXiv:2112.08340, 2021.

\bibitem{saxena2022sequence}
A.~Saxena, A.~Kochsiek, and R.~Gemulla.
\newblock {\em Sequence-to-Sequence Knowledge Graph Completion and Question
  Answering}.
\newblock In Proceedings of the 60th Annual Meeting of the Association for
  Computational Linguistics (Volume 1: Long Papers), pages 2814--2828, 2022.

\bibitem{jin2019recurrent}
W.~Jin, M.~Qu, X.~Jin, and X.~Ren.
\newblock {\em Recurrent event network: Autoregressive structure inference over
  temporal knowledge graphs}.
\newblock arXiv preprint arXiv:1904.05530, 2019.

\bibitem{huang2022multilingual}
K.-H. Huang, I.~Hsu, P.~Natarajan, K.-W. Chang, N.~Peng, et~al.
\newblock {\em Multilingual Generative Language Models for Zero-Shot
  Cross-Lingual Event Argument Extraction}.
\newblock arXiv preprint arXiv:2203.08308, 2022.

\bibitem{glass2022re2g}
M.~Glass, G.~Rossiello, M.~F.~M. Chowdhury, A.~R. Naik, P.~Cai, and A.~Gliozzo.
\newblock {\em Re2G: Retrieve, Rerank, Generate}.
\newblock arXiv preprint arXiv:2207.06300, 2022.

\bibitem{lu2022unified}
Y.~Lu, Q.~Liu, D.~Dai, X.~Xiao, H.~Lin, X.~Han, L.~Sun, and H.~Wu.
\newblock {\em Unified Structure Generation for Universal Information
  Extraction}.
\newblock arXiv preprint arXiv:2203.12277, 2022.

\bibitem{liu2021pre}
P.~Liu, W.~Yuan, J.~Fu, Z.~Jiang, H.~Hayashi, and G.~Neubig.
\newblock {\em Pre-train, prompt, and predict: A systematic survey of prompting
  methods in natural language processing}.
\newblock arXiv preprint arXiv:2107.13586, 2021.

\bibitem{ma2021template}
R.~Ma, X.~Zhou, T.~Gui, Y.~Tan, Q.~Zhang, and X.~Huang.
\newblock {\em Template-free prompt tuning for few-shot NER}.
\newblock arXiv preprint arXiv:2109.13532, 2021.

\bibitem{liu2022qaner}
A.~T. Liu, W.~Xiao, H.~Zhu, D.~Zhang, S.-W. Li, and A.~Arnold.
\newblock {\em QaNER: Prompting question answering models for few-shot named
  entity recognition}.
\newblock arXiv preprint arXiv:2203.01543, 2022.

\bibitem{zhang2022prompt}
H.~Zhang, B.~Liang, M.~Yang, H.~Wang, and R.~Xu.
\newblock {\em Prompt-Based Prototypical Framework for Continual Relation
  Extraction}.
\newblock IEEE/ACM Transactions on Audio, Speech, and Language Processing,
  2022.

\bibitem{chen2022knowprompt}
X.~Chen, N.~Zhang, X.~Xie, S.~Deng, Y.~Yao, C.~Tan, F.~Huang, L.~Si, and
  H.~Chen.
\newblock {\em Knowprompt: Knowledge-aware prompt-tuning with synergistic
  optimization for relation extraction}.
\newblock In Proceedings of the ACM Web Conference 2022, pages 2778--2788,
  2022.

\bibitem{yeh2022decorate}
H.-S. Yeh, T.~Lavergne, and P.~Zweigenbaum.
\newblock {\em Decorate the Examples: A Simple Method of Prompt Design for
  Biomedical Relation Extraction}.
\newblock arXiv preprint arXiv:2204.10360, 2022.

\bibitem{gong2021prompt}
J.~Gong and H.~Eldardiry.
\newblock {\em Prompt-based Zero-shot Relation Classification with Semantic
  Knowledge Augmentation}.
\newblock arXiv preprint arXiv:2112.04539, 2021.

\bibitem{chia2022relationprompt}
Y.~K. Chia, L.~Bing, S.~Poria, and L.~Si.
\newblock {\em RelationPrompt: Leveraging Prompts to Generate Synthetic Data
  for Zero-Shot Relation Triplet Extraction}.
\newblock In Findings of the Association for Computational Linguistics: ACL
  2022, pages 45--57, 2022.

\bibitem{liu2022dynamic}
X.~Liu, H.-Y. Huang, G.~Shi, and B.~Wang.
\newblock {\em Dynamic Prefix-Tuning for Generative Template-based Event
  Extraction}.
\newblock In Proceedings of the 60th Annual Meeting of the Association for
  Computational Linguistics (Volume 1: Long Papers), pages 5216--5228, 2022.

\bibitem{ma2022prompt}
Y.~Ma, Z.~Wang, Y.~Cao, M.~Li, M.~Chen, K.~Wang, and J.~Shao.
\newblock {\em Prompt for Extraction? PAIE: Prompting Argument Interaction for
  Event Argument Extraction}.
\newblock In Proceedings of the 60th Annual Meeting of the Association for
  Computational Linguistics (Volume 1: Long Papers), pages 6759--6774, 2022.

\bibitem{alkhamissi2022review}
B.~AlKhamissi, M.~Li, A.~Celikyilmaz, M.~Diab, and M.~Ghazvininejad.
\newblock {\em A review on language models as knowledge bases}.
\newblock arXiv preprint arXiv:2204.06031, 2022.

\bibitem{yin2022survey}
D.~Yin, L.~Dong, H.~Cheng, X.~Liu, K.-W. Chang, F.~Wei, and J.~Gao.
\newblock {\em A survey of knowledge-intensive nlp with pre-trained language
  models}.
\newblock arXiv preprint arXiv:2202.08772, 2022.

\bibitem{wang2020k}
R.~Wang, D.~Tang, N.~Duan, Z.~Wei, X.~Huang, G.~Cao, D.~Jiang, M.~Zhou, et~al.
\newblock {\em K-adapter: Infusing knowledge into pre-trained models with
  adapters}.
\newblock arXiv preprint arXiv:2002.01808, 2020.

\bibitem{de2021editing}
N.~De~Cao, W.~Aziz, and I.~Titov.
\newblock {\em Editing Factual Knowledge in Language Models}.
\newblock arXiv preprint arXiv:2104.08164, 2021.

\bibitem{sinitsin2020editable}
A.~Sinitsin, V.~Plokhotnyuk, D.~Pyrkin, S.~Popov, and A.~Babenko.
\newblock {\em Editable Neural Networks}.
\newblock arXiv preprint arXiv:2004.00345, 2020.

\bibitem{cossu2022continual}
A.~Cossu, T.~Tuytelaars, A.~Carta, L.~Passaro, V.~Lomonaco, and D.~Bacciu.
\newblock {\em Continual Pre-Training Mitigates Forgetting in Language and
  Vision}.
\newblock arXiv preprint arXiv:2205.09357, 2022.

\bibitem{jin2022lifelong}
X.~Jin, D.~Zhang, H.~Zhu, W.~Xiao, S.-W. Li, X.~Wei, A.~Arnold, and X.~Ren.
\newblock {\em Lifelong Pretraining: Continually Adapting Language Models to
  Emerging Corpora}.
\newblock In Proceedings of BigScience Episode$\backslash$\# 5--Workshop on
  Challenges \& Perspectives in Creating Large Language Models, pages 1--16,
  2022.

\bibitem{zhu2020modifying}
C.~Zhu, A.~S. Rawat, M.~Zaheer, S.~Bhojanapalli, D.~Li, F.~Yu, and S.~Kumar.
\newblock {\em Modifying Memories in Transformer Models}.
\newblock arXiv preprint arXiv:2012.00363, 2020.

\bibitem{hospedales2021meta}
T.~M. Hospedales, A.~Antoniou, P.~Micaelli, and A.~J. Storkey.
\newblock {\em Meta-learning in neural networks: A survey}.
\newblock IEEE transactions on pattern analysis and machine intelligence, 2021.

\bibitem{baker1998berkeley}
C.~F. Baker, C.~J. Fillmore, and J.~B. Lowe.
\newblock {\em The berkeley framenet project}.
\newblock In Proceedings of the 1998 International Conference on Computational
  Linguistics, pages 86--90, 1998.

\bibitem{song2018n}
L.~Song, Y.~Zhang, Z.~Wang, and D.~Gildea.
\newblock {\em N-ary Relation Extraction using Graph-State LSTM}.
\newblock In Proceedings of the 2018 Conference on Empirical Methods in Natural
  Language Processing, pages 2226--2235, 2018.

\bibitem{giunti2021representing}
M.~Giunti, G.~Sergioli, G.~Vivanet, and S.~Pinna.
\newblock {\em Representing n-ary relations in the Semantic Web}.
\newblock Logic Journal of the IGPL, 29(4):697--717, 2021.

\bibitem{lentschat2022new}
M.~Lentschat, P.~Buche, J.~Dibie-Barthelemy, and M.~Roche.
\newblock {\em A new method to extract n-Ary relation instances from scientific
  documents}.
\newblock Expert Systems with Applications, page 118332, 2022.

\bibitem{lai2020bert}
P.-T. Lai and Z.~Lu.
\newblock {\em BERT-GT: cross-sentence n-ary relation extraction with BERT and
  Graph Transformer}.
\newblock Bioinformatics, 36(24):5678--5685, 2020.

\bibitem{lehmberg2019synthesizing}
O.~Lehmberg and C.~Bizer.
\newblock {\em Synthesizing n-ary relations from web tables}.
\newblock In Proceedings of the 9th International Conference on Web
  Intelligence, Mining and Semantics, pages 1--12, 2019.

\bibitem{xiang2019survey}
W.~Xiang and B.~Wang.
\newblock {\em A survey of event extraction from text}.
\newblock IEEE Access, 7:173111--173137, 2019.

\bibitem{hogenboom2016survey}
F.~Hogenboom, F.~Frasincar, U.~Kaymak, F.~De~Jong, and E.~Caron.
\newblock {\em A survey of event extraction methods from text for decision
  support systems}.
\newblock Decision Support Systems, 85:12--22, 2016.

\bibitem{zhan2019survey}
L.~Zhan and X.~Jiang.
\newblock {\em Survey on event extraction technology in information extraction
  research area}.
\newblock In 2019 IEEE 3rd Information Technology, Networking, Electronic and
  Automation Control Conference (ITNEC), pages 2121--2126. IEEE, 2019.

\bibitem{kushman2014learning}
N.~Kushman, Y.~Artzi, L.~Zettlemoyer, and R.~Barzilay.
\newblock {\em Learning to automatically solve algebra word problems}.
\newblock In Proceedings of the 52nd Annual Meeting of the Association for
  Computational Linguistics (Volume 1: Long Papers), volume~1, pages 271--281,
  2014.

\bibitem{mitra2016learning}
A.~Mitra and C.~Baral.
\newblock {\em Learning to use formulas to solve simple arithmetic problems}.
\newblock In Proceedings of the 54th Annual Meeting of the Association for
  Computational Linguistics (Volume 1: Long Papers), volume~1, pages
  2144--2153, 2016.

\bibitem{wang2018mathdqn}
L.~Wang, D.~Zhang, L.~Gao, J.~Song, L.~Guo, and H.~T. Shen.
\newblock {\em Mathdqn: Solving arithmetic word problems via deep reinforcement
  learning}.
\newblock In Thirty-Second AAAI Conference on Artificial Intelligence, 2018.

\bibitem{zhang2019gap}
D.~Zhang, L.~Wang, L.~Zhang, B.~T. Dai, and H.~T. Shen.
\newblock {\em The gap of semantic parsing: A survey on automatic math word
  problem solvers}.
\newblock IEEE, 2019.

\bibitem{hosseini2014learning}
M.~J. Hosseini, H.~Hajishirzi, O.~Etzioni, and N.~Kushman.
\newblock {\em Learning to solve arithmetic word problems with verb
  categorization}.
\newblock In Proceedings of the 2014 Conference on Empirical Methods in Natural
  Language Processing (EMNLP), pages 523--533, 2014.

\bibitem{wang2017deep}
Y.~Wang, X.~Liu, and S.~Shi.
\newblock {\em Deep neural solver for math word problems}.
\newblock In Proceedings of the 2017 Conference on Empirical Methods in Natural
  Language Processing, pages 845--854, 2017.

\bibitem{li2019modeling}
J.~Li, L.~Wang, J.~Zhang, Y.~Wang, B.~T. Dai, and D.~Zhang.
\newblock {\em Modeling Intra-Relation in Math Word Problems with Different
  Functional Multi-Head Attentions}.
\newblock In Proceedings of the 57th Conference of the Association for
  Computational Linguistics, pages 6162--6167, 2019.

\bibitem{chiang2019semantically}
T.-R. Chiang and Y.-N. Chen.
\newblock {\em Semantically-Aligned Equation Generation for Solving and
  Reasoning Math Word Problems}.
\newblock In Proceedings of the 2019 Conference of the North American Chapter
  of the Association for Computational Linguistics: Human Language
  Technologies, Volume 1 (Long and Short Papers), pages 2656--2668, 2019.

\bibitem{shi2015automatically}
S.~Shi, Y.~Wang, C.-Y. Lin, X.~Liu, and Y.~Rui.
\newblock {\em Automatically solving number word problems by semantic parsing
  and reasoning}.
\newblock In Proceedings of the 2015 Conference on Empirical Methods in Natural
  Language Processing, pages 1132--1142, 2015.

\bibitem{ling2017program}
W.~Ling, D.~Yogatama, C.~Dyer, and P.~Blunsom.
\newblock {\em Program Induction by Rationale Generation: Learning to Solve and
  Explain Algebraic Word Problems}.
\newblock In Proceedings of the 55th Annual Meeting of the Association for
  Computational Linguistics (Volume 1: Long Papers), pages 158--167, 2017.

\bibitem{amini2019mathqa}
A.~Amini, S.~Gabriel, S.~Lin, R.~Koncel-Kedziorski, Y.~Choi, and H.~Hajishirzi.
\newblock {\em MathQA: Towards Interpretable Math Word Problem Solving with
  Operation-Based Formalisms}.
\newblock In Proceedings of the 2019 Conference of the North American Chapter
  of the Association for Computational Linguistics: Human Language
  Technologies, Volume 1 (Long and Short Papers), pages 2357--2367, 2019.

\bibitem{wang2018translating}
L.~Wang, Y.~Wang, D.~Cai, D.~Zhang, and X.~Liu.
\newblock {\em Translating a Math Word Problem to a Expression Tree}.
\newblock In Proceedings of the 2018 Conference on Empirical Methods in Natural
  Language Processing, pages 1064--1069, 2018.

\bibitem{tai2015improved}
K.~S. Tai, R.~Socher, and C.~D. Manning.
\newblock {\em Improved Semantic Representations From Tree-Structured Long
  Short-Term Memory Networks}.
\newblock In Proceedings of the 53rd Annual Meeting of the Association for
  Computational Linguistics and the 7th International Joint Conference on
  Natural Language Processing (Volume 1: Long Papers), volume~1, pages
  1556--1566, 2015.

\bibitem{bobrow1964natural}
D.~G. Bobrow.
\newblock {\em Natural language input for a computer problem solving system}.
\newblock 1964.

\bibitem{charniak1968carps}
E.~Charniak.
\newblock {\em CARPS: a program which solves calculus word problems}.
\newblock 1968.

\bibitem{charniak1969computer}
E.~Charniak.
\newblock {\em Computer solution of calculus word problems}.
\newblock In Proceedings of the 1st international joint conference on
  Artificial intelligence, pages 303--316. Morgan Kaufmann Publishers Inc.,
  1969.

\bibitem{fletcher1985understanding}
C.~R. Fletcher.
\newblock {\em Understanding and solving arithmetic word problems: A computer
  simulation}.
\newblock Behavior Research Methods, Instruments, \& Computers, 17(5):565--571,
  1985.

\bibitem{bakman2007robust}
Y.~Bakman.
\newblock {\em Robust understanding of word problems with extraneous
  information}.
\newblock 2007.

\bibitem{roy2015solving}
S.~Roy and D.~Roth.
\newblock {\em Solving General Arithmetic Word Problems}.
\newblock In Proceedings of the 2015 Conference on Empirical Methods in Natural
  Language Processing, pages 1743--1752, 2015.

\bibitem{roy2017unit}
S.~Roy and D.~Roth.
\newblock {\em Unit dependency graph and its application to arithmetic word
  problem solving}.
\newblock In Thirty-First AAAI Conference on Artificial Intelligence, 2017.

\bibitem{roy2018mapping}
S.~Roy and D.~Roth.
\newblock {\em Mapping to declarative knowledge for word problem solving}.
\newblock Transactions of the Association of Computational Linguistics,
  6:159--172, 2018.

\bibitem{huang2018neural}
D.~Huang, J.~Liu, C.-Y. Lin, and J.~Yin.
\newblock {\em Neural Math Word Problem Solver with Reinforcement Learning}.
\newblock In Proceedings of the 27th International Conference on Computational
  Linguistics, pages 213--223, 2018.

\bibitem{socher2011parsing}
R.~Socher, C.~C. Lin, C.~Manning, and A.~Y. Ng.
\newblock {\em Parsing natural scenes and natural language with recursive
  neural networks}.
\newblock In Proceedings of the 28th international conference on machine
  learning (ICML-11), pages 129--136, 2011.

\bibitem{socher2013parsing}
R.~Socher, J.~Bauer, C.~D. Manning, et~al.
\newblock {\em Parsing with compositional vector grammars}.
\newblock In Proceedings of the 51st Annual Meeting of the Association for
  Computational Linguistics (Volume 1: Long Papers), volume~1, pages 455--465,
  2013.

\bibitem{chen2017enhanced}
Q.~Chen, X.~Zhu, Z.-H. Ling, S.~Wei, H.~Jiang, and D.~Inkpen.
\newblock {\em Enhanced {LSTM} for Natural Language Inference}.
\newblock In Proceedings of the 55th Annual Meeting of the Association for
  Computational Linguistics (Volume 1: Long Papers), pages 1657--1668, 2017.

\bibitem{bengio1994learning}
Y.~Bengio, P.~Simard, and P.~Frasconi.
\newblock {\em Learning long-term dependencies with gradient descent is
  difficult}.
\newblock IEEE transactions on neural networks, 5(2):157--166, 1994.

\bibitem{pennington:14}
J.~Pennington, R.~Socher, and C.~Manning.
\newblock {\em Glove: Global vectors for word representation}.
\newblock In Proceedings of the 2014 conference on empirical methods in natural
  language processing (EMNLP), pages 1532--1543, 2014.

\bibitem{kingma:14}
D.~Kingma and J.~Ba.
\newblock {\em {A}dam: {A} method for stochastic optimization}.
\newblock In Proceedings of the International Conference on Learning
  Representations, San Diego, USA, 2015.

\bibitem{efron1994introduction}
B.~Efron and R.~J. Tibshirani.
\newblock {\em An introduction to the bootstrap}.
\newblock CRC press, 1994.

\bibitem{power2005cohort}
C.~Power and J.~Elliott.
\newblock {\em Cohort profile: 1958 {B}ritish birth cohort (national child
  development study)}.
\newblock International journal of epidemiology, 35(1):34--41, 2005.

\bibitem{shepherd2013bristol}
P.~Shepherd.
\newblock {\em Bristol social adjustment guides at 7 and 11 years}.
\newblock Centre for Longitudinal Studies, 2013.

\bibitem{pennebaker2015development}
J.~W. Pennebaker, R.~L. Boyd, K.~Jordan, and K.~Blackburn.
\newblock {\em The development and psychometric properties of {LIWC2015}}.
\newblock Technical report, 2015.

\bibitem{staiano2014depechemood}
J.~Staiano and M.~Guerini.
\newblock {\em Depeche{M}ood: a lexicon for emotion analysis from
  crowd-annotated news}.
\newblock arXiv preprint arXiv:1405.1605, 2014.

\bibitem{cagan2014generating}
T.~Cagan, S.~L. Frank, and R.~Tsarfaty.
\newblock {\em Generating subjective responses to opinionated articles in
  social media: an agenda-driven architecture and a Turing-like test}.
\newblock In Proceedings of the Joint Workshop on Social Dynamics and Personal
  Attributes in Social Media, pages 58--67, 2014.

\bibitem{merity2017}
S.~Merity, N.~S. Keskar, and R.~Socher.
\newblock {\em {Regularizing and Optimizing {LSTM} Language Models}}.
\newblock arXiv preprint arXiv:1708.02182, 2017.

\bibitem{chen2016xgboost}
T.~Chen and C.~Guestrin.
\newblock {\em {XGBoost}: A Scalable Tree Boosting System}.
\newblock In Proceedings of the 22nd ACM SIGKDD International Conference on
  Knowledge Discovery and Data Mining, KDD '16, pages 785--794, New York, NY,
  USA, 2016. ACM.
\newblock Available from: \url{http://doi.acm.org/10.1145/2939672.2939785},
  doi:10.1145/2939672.2939785.

\bibitem{kim2014convolutional}
Y.~Kim.
\newblock {\em Convolutional Neural Networks for Sentence Classification}.
\newblock In Proceedings of the 2014 Conference on Empirical Methods in Natural
  Language Processing (EMNLP), pages 1746--1751, 2014.

\bibitem{cho2014learning}
K.~Cho, B.~Van~Merri{\"e}nboer, C.~Gulcehre, D.~Bahdanau, F.~Bougares,
  H.~Schwenk, and Y.~Bengio.
\newblock {\em Learning phrase representations using {RNN} encoder-decoder for
  statistical machine translation}.
\newblock arXiv preprint arXiv:1406.1078, 2014.

\bibitem{lin2017structured}
Z.~Lin, M.~Feng, C.~N.~d. Santos, M.~Yu, B.~Xiang, B.~Zhou, and Y.~Bengio.
\newblock {\em A structured self-attentive sentence embedding}.
\newblock arXiv preprint arXiv:1703.03130, 2017.

\bibitem{resnik2015university}
P.~Resnik, W.~Armstrong, L.~Claudino, and T.~Nguyen.
\newblock {\em The {U}niversity of {M}aryland {CLPsych} 2015 shared task
  system}.
\newblock In Proceedings of the 2nd Workshop on Computational Linguistics and
  Clinical Psychology: From Linguistic Signal to Clinical Reality, pages
  54--60, 2015.

\bibitem{cohan2016triaging}
A.~Cohan, S.~Young, and N.~Goharian.
\newblock {\em Triaging mental health forum posts}.
\newblock In Proceedings of the Third Workshop on Computational Lingusitics and
  Clinical Psychology, pages 143--147, 2016.

\bibitem{yates2017depression}
A.~Yates, A.~Cohan, and N.~Goharian.
\newblock {\em Depression and Self-Harm Risk Assessment in Online Forums}.
\newblock arXiv preprint arXiv:1709.01848, 2017.

\bibitem{coppersmith2015clpsych}
G.~Coppersmith, M.~Dredze, C.~Harman, K.~Hollingshead, and M.~Mitchell.
\newblock {\em {CLPsych} 2015 shared task: Depression and {PTSD} on {T}witter}.
\newblock In Proceedings of the 2nd Workshop on Computational Linguistics and
  Clinical Psychology: From Linguistic Signal to Clinical Reality, pages
  31--39, 2015.

\end{thebibliography}

\newpage\null\thispagestyle{empty}\newpage
\newpage\null\thispagestyle{empty}\newpage

\end{document}